\documentclass{article}
\usepackage[main,preprint]{neurips_2026}
\usepackage{times}

\usepackage{amsfonts}

\usepackage{flafter}
\usepackage{float}
\usepackage{placeins}
\usepackage{hyperref}
\usepackage[strings]{underscore}
\usepackage{url}
\usepackage{xspace}
\usepackage{amsmath,amsthm}
\usepackage{booktabs}
\usepackage{longtable}
\usepackage{graphicx}

\newcommand{\nActDtype}{bfloat16\xspace}
\newcommand{\nActivationsF}{199,999,488\xspace}
\newcommand{\nAdamBetaOne}{0.9\xspace}
\newcommand{\nAdamBetaTwo}{0.999\xspace}
\newcommand{\nAdamEps}{$1.0\times 10^{-8}$\xspace}
\newcommand{\nAnatBetweenDecFew}{0.15\xspace}
\newcommand{\nAnatBetweenDecMany}{0.36\xspace}
\newcommand{\nAnatBetweenDecNone}{0.29\xspace}
\newcommand{\nAnatBetweenDecOne}{0.20\xspace}
\newcommand{\nAnatBetweenEncFew}{0.05\xspace}
\newcommand{\nAnatBetweenEncMany}{0.08\xspace}
\newcommand{\nAnatBetweenEncNone}{0.76\xspace}
\newcommand{\nAnatBetweenEncOne}{0.12\xspace}
\newcommand{\nAnatControlDecNone}{1.00\xspace}
\newcommand{\nAnatRetrainDecNone}{0.21\xspace}
\newcommand{\nAnatRetrainDecOne}{0.30\xspace}
\newcommand{\nAnatRetrainEncNone}{0.58\xspace}
\newcommand{\nAnatRetrainEncOne}{0.26\xspace}
\newcommand{\nAuditPerDoseZeroA}{8,192\xspace}
\newcommand{\nAuditPerDoseZeroB}{8,192\xspace}
\newcommand{\nAuditPerDoseZeroC}{8,192\xspace}
\newcommand{\nAuditPerDoseZeroD}{16,384\xspace}
\newcommand{\nAuditPerDoseZeroE}{8,192\xspace}
\newcommand{\nAuditPerDoseZeroF}{131,072\xspace}
\newcommand{\nAuditPerDoseZeroR}{44,343\xspace}
\newcommand{\nAuditRowsAll}{673,701\xspace}
\newcommand{\nAuxAtFourEnd}{199,999,488\xspace}
\newcommand{\nAuxAtFourMid}{100,941,824\xspace}
\newcommand{\nAuxAtThreeEnd}{199,999,488\xspace}
\newcommand{\nAuxAtThreeMid}{100,941,824\xspace}
\newcommand{\nAuxCheckpoints}{14\xspace}
\newcommand{\nAuxCoefFourEnd}{0.3750\xspace}
\newcommand{\nAuxCoefFourMid}{0.2500\xspace}
\newcommand{\nAuxCoefMax}{0.44\xspace}
\newcommand{\nAuxCoefMin}{0.25\xspace}
\newcommand{\nAuxCoefThreeEnd}{0.4375\xspace}
\newcommand{\nAuxCoefThreeMid}{0.3125\xspace}
\newcommand{\nAuxDeadFourEndFive}{2\xspace}
\newcommand{\nAuxDeadFourEndFour}{1\xspace}
\newcommand{\nAuxDeadFourEndOne}{1\xspace}
\newcommand{\nAuxDeadFourEndSeven}{1\xspace}
\newcommand{\nAuxDeadFourEndSix}{1\xspace}
\newcommand{\nAuxDeadFourEndThree}{0\xspace}
\newcommand{\nAuxDeadFourEndTwo}{0\xspace}
\newcommand{\nAuxDeadFourEndZero}{0\xspace}
\newcommand{\nAuxDeadFourMidFive}{0\xspace}
\newcommand{\nAuxDeadFourMidFour}{0\xspace}
\newcommand{\nAuxDeadFourMidOne}{0\xspace}
\newcommand{\nAuxDeadFourMidSeven}{1\xspace}
\newcommand{\nAuxDeadFourMidSix}{0\xspace}
\newcommand{\nAuxDeadFourMidThree}{3\xspace}
\newcommand{\nAuxDeadFourMidTwo}{2\xspace}
\newcommand{\nAuxDeadFourMidZero}{0\xspace}
\newcommand{\nAuxDeadMax}{3\xspace}
\newcommand{\nAuxDeadThreeEndFive}{2\xspace}
\newcommand{\nAuxDeadThreeEndFour}{1\xspace}
\newcommand{\nAuxDeadThreeEndOne}{0\xspace}
\newcommand{\nAuxDeadThreeEndSeven}{1\xspace}
\newcommand{\nAuxDeadThreeEndSix}{1\xspace}
\newcommand{\nAuxDeadThreeEndThree}{1\xspace}
\newcommand{\nAuxDeadThreeEndTwo}{0\xspace}
\newcommand{\nAuxDeadThreeEndZero}{1\xspace}
\newcommand{\nAuxDeadThreeMidFive}{0\xspace}
\newcommand{\nAuxDeadThreeMidFour}{0\xspace}
\newcommand{\nAuxDeadThreeMidOne}{0\xspace}
\newcommand{\nAuxDeadThreeMidSeven}{2\xspace}
\newcommand{\nAuxDeadThreeMidSix}{2\xspace}
\newcommand{\nAuxDeadThreeMidThree}{1\xspace}
\newcommand{\nAuxDeadThreeMidTwo}{0\xspace}
\newcommand{\nAuxDeadThreeMidZero}{0\xspace}
\newcommand{\nAuxOffCount}{12\xspace}
\newcommand{\nAuxOffThrough}{50,970,624\xspace}
\newcommand{\nAuxOverThreshold}{2\xspace}
\newcommand{\nAuxThreshold}{0.01\xspace}
\newcommand{\nBandNeeded}{0.5\xspace}
\newcommand{\nBandSmall}{0.1\xspace}
\newcommand{\nBandTokens}{1,024\xspace}
\newcommand{\nBandWritesPerSize}{1,024\xspace}
\newcommand{\nBatchR}{16,384\xspace}
\newcommand{\nBinHigh}{64\xspace}
\newcommand{\nBinLow}{0.0625\xspace}
\newcommand{\nBinsMinRows}{20\xspace}
\newcommand{\nBinsPerDoubling}{2\xspace}
\newcommand{\nBlocksA}{6\xspace}
\newcommand{\nBlocksD}{12\xspace}
\newcommand{\nBootAtOrUnderZeroA}{0.0\xspace}
\newcommand{\nBootAtOrUnderZeroB}{0.0\xspace}
\newcommand{\nBootAtOrUnderZeroC}{0.0\xspace}
\newcommand{\nBootAtOrUnderZeroD}{0.0\xspace}
\newcommand{\nBootAtOrUnderZeroE}{0.0\xspace}
\newcommand{\nBootAtOrUnderZeroF}{0.0\xspace}
\newcommand{\nBootAtOrUnderZeroR}{0.0\xspace}
\newcommand{\nBootDiffHighA}{0.564\xspace}
\newcommand{\nBootDiffHighB}{0.410\xspace}
\newcommand{\nBootDiffHighC}{0.654\xspace}
\newcommand{\nBootDiffHighD}{0.630\xspace}
\newcommand{\nBootDiffHighE}{0.663\xspace}
\newcommand{\nBootDiffHighF}{0.777\xspace}
\newcommand{\nBootDiffHighR}{0.0107\xspace}
\newcommand{\nBootDiffLowA}{0.496\xspace}
\newcommand{\nBootDiffLowB}{0.138\xspace}
\newcommand{\nBootDiffLowC}{0.606\xspace}
\newcommand{\nBootDiffLowD}{0.577\xspace}
\newcommand{\nBootDiffLowE}{0.612\xspace}
\newcommand{\nBootDiffLowF}{0.760\xspace}
\newcommand{\nBootDiffLowR}{0.0076\xspace}
\newcommand{\nBootDraws}{2,000\xspace}
\newcommand{\nBootFixedHighA}{0.596\xspace}
\newcommand{\nBootFixedHighB}{0.433\xspace}
\newcommand{\nBootFixedHighC}{0.686\xspace}
\newcommand{\nBootFixedHighD}{0.643\xspace}
\newcommand{\nBootFixedHighE}{0.693\xspace}
\newcommand{\nBootFixedHighF}{0.794\xspace}
\newcommand{\nBootFixedHighR}{0.0185\xspace}
\newcommand{\nBootFixedLowA}{0.529\xspace}
\newcommand{\nBootFixedLowB}{0.162\xspace}
\newcommand{\nBootFixedLowC}{0.639\xspace}
\newcommand{\nBootFixedLowD}{0.590\xspace}
\newcommand{\nBootFixedLowE}{0.644\xspace}
\newcommand{\nBootFixedLowF}{0.777\xspace}
\newcommand{\nBootFixedLowR}{0.0154\xspace}
\newcommand{\nBootInterval}{the 2.5th and 97.5th percentiles\xspace}
\newcommand{\nBootMarginHighA}{0.0342\xspace}
\newcommand{\nBootMarginHighB}{0.0240\xspace}
\newcommand{\nBootMarginHighC}{0.0340\xspace}
\newcommand{\nBootMarginHighD}{0.0135\xspace}
\newcommand{\nBootMarginHighE}{0.0317\xspace}
\newcommand{\nBootMarginHighF}{0.0173\xspace}
\newcommand{\nBootMarginHighR}{0.0080\xspace}
\newcommand{\nBootMarginLowA}{0.0316\xspace}
\newcommand{\nBootMarginLowB}{0.0219\xspace}
\newcommand{\nBootMarginLowC}{0.0315\xspace}
\newcommand{\nBootMarginLowD}{0.0125\xspace}
\newcommand{\nBootMarginLowE}{0.0294\xspace}
\newcommand{\nBootMarginLowF}{0.0164\xspace}
\newcommand{\nBootMarginLowR}{0.0075\xspace}
\newcommand{\nBootSeed}{20260924\xspace}
\newcommand{\nBoundaryF}{7.26\xspace}
\newcommand{\nBoundaryFactor}{2.0\xspace}
\newcommand{\nBoundaryPubSingleMax}{9.60\xspace}
\newcommand{\nBoundaryPubSingleMin}{3.22\xspace}
\newcommand{\nBoundaryRatio}{1.66\xspace}
\newcommand{\nBoundaryRel}{4.36\xspace}
\newcommand{\nBoundarySingleF}{5.55\xspace}
\newcommand{\nBoundarySingleRel}{3.90\xspace}
\newcommand{\nBoundarySubsetF}{7.54\xspace}
\newcommand{\nBoundarySubsetRel}{4.48\xspace}
\newcommand{\nBudgetCount}{9\xspace}
\newcommand{\nBudgetEnd}{199,999,488\xspace}
\newcommand{\nBudgetFiledEnd}{100,941,824\xspace}
\newcommand{\nBudgetPairs}{15\xspace}
\newcommand{\nBudgetTokensA}{8.2\xspace}
\newcommand{\nBudgetTokensR}{2\xspace}
\newcommand{\nCap}{32\xspace}
\newcommand{\nCarryMarginal}{1.02\xspace}
\newcommand{\nCarryRemainder}{1.06\xspace}
\newcommand{\nCarryStable}{0.92\xspace}
\newcommand{\nCertFlips}{0\xspace}
\newcommand{\nCertGlobalActive}{267\xspace}
\newcommand{\nCertGlobalPredicted}{524,288\xspace}
\newcommand{\nCertOwnSizeEightFSci}{$1.2\times 10^{-5}$\xspace}
\newcommand{\nCertOwnSizeSixteenFSci}{$6.0\times 10^{-8}$\xspace}
\newcommand{\nCertOwnSizeTwoF}{0.230\xspace}
\newcommand{\nCfAuxAlphaDen}{32\xspace}
\newcommand{\nCfBatch}{4,096\xspace}
\newcommand{\nCfCheckpointCount}{14\xspace}
\newcommand{\nCfCheckpointFirst}{10,000\xspace}
\newcommand{\nCfCheckpointLast}{200,000,000\xspace}
\newcommand{\nCfChunk}{1,000,000\xspace}
\newcommand{\nCfChunkTrained}{999,424\xspace}
\newcommand{\nCfDModel}{512\xspace}
\newcommand{\nCfDeadTokens}{10,000,000\xspace}
\newcommand{\nCfFvuSample}{65,536\xspace}
\newcommand{\nCfHeldout}{1,048,576\xspace}
\newcommand{\nCfKAux}{256\xspace}
\newcommand{\nCfLr}{$1.0\times 10^{-4}$\xspace}
\newcommand{\nCfSeedInitFirst}{100\xspace}
\newcommand{\nCfSeedInitLast}{105\xspace}
\newcommand{\nCfSeedInitRetrain}{106\xspace}
\newcommand{\nCfShuffleFirst}{200\xspace}
\newcommand{\nCfShuffleLastIndependent}{205\xspace}
\newcommand{\nCfShuffleRetrainFirst}{206\xspace}
\newcommand{\nCfShuffleRetrainSecond}{207\xspace}
\newcommand{\nCfTargetsFirstChunk}{6\xspace}
\newcommand{\nCfWarmup}{1,000\xspace}
\newcommand{\nChainBoundFixed}{0.663\xspace}
\newcommand{\nChainBoundOracle}{0.114\xspace}
\newcommand{\nChainExactInput}{0.029\xspace}
\newcommand{\nChainFixed}{0.884\xspace}
\newcommand{\nChainMargin}{0.999\xspace}
\newcommand{\nChainOracle}{0.996\xspace}
\newcommand{\nChainOracleMultiple}{4.0\xspace}
\newcommand{\nChainResidual}{0.053\xspace}
\newcommand{\nCheckAgainstPairingsF}{$1.8\times 10^{-4}$\xspace}
\newcommand{\nCheckAgainstPairingsTolF}{$2.0\times 10^{-3}$\xspace}
\newcommand{\nCheckCentralA}{$1.2\times 10^{-7}$\xspace}
\newcommand{\nCheckCentralB}{$2.1\times 10^{-10}$\xspace}
\newcommand{\nCheckCentralC}{$1.2\times 10^{-7}$\xspace}
\newcommand{\nCheckCentralD}{$2.2\times 10^{-8}$\xspace}
\newcommand{\nCheckCentralE}{$8.4\times 10^{-8}$\xspace}
\newcommand{\nCheckCentralF}{$1.2\times 10^{-7}$\xspace}
\newcommand{\nCheckCentralR}{$8.4\times 10^{-8}$\xspace}
\newcommand{\nCheckCentralTol}{$1.0\times 10^{-5}$\xspace}
\newcommand{\nCheckDirs}{3\xspace}
\newcommand{\nCheckEveryTokenF}{$2.1\times 10^{-4}$\xspace}
\newcommand{\nCheckEveryTokenTolF}{$1.0\times 10^{-3}$\xspace}
\newcommand{\nCheckJvpA}{$2.7\times 10^{-4}$\xspace}
\newcommand{\nCheckJvpB}{$1.8\times 10^{-6}$\xspace}
\newcommand{\nCheckJvpC}{$8.1\times 10^{-4}$\xspace}
\newcommand{\nCheckJvpD}{$1.5\times 10^{-5}$\xspace}
\newcommand{\nCheckJvpE}{$1.7\times 10^{-4}$\xspace}
\newcommand{\nCheckJvpF}{$2.7\times 10^{-4}$\xspace}
\newcommand{\nCheckJvpR}{$1.7\times 10^{-4}$\xspace}
\newcommand{\nCheckJvpTol}{$2.0\times 10^{-3}$\xspace}
\newcommand{\nCheckOutputA}{$4.8\times 10^{-5}$\xspace}
\newcommand{\nCheckOutputB}{$1.0\times 10^{-7}$\xspace}
\newcommand{\nCheckOutputC}{$1.9\times 10^{-4}$\xspace}
\newcommand{\nCheckOutputD}{$5.0\times 10^{-6}$\xspace}
\newcommand{\nCheckOutputE}{$4.9\times 10^{-5}$\xspace}
\newcommand{\nCheckOutputF}{$4.8\times 10^{-5}$\xspace}
\newcommand{\nCheckOutputR}{$4.9\times 10^{-5}$\xspace}
\newcommand{\nCheckOutputTol}{$1.0\times 10^{-3}$\xspace}
\newcommand{\nCheckPairingsF}{$1.1\times 10^{-4}$\xspace}
\newcommand{\nCheckPairingsTolF}{$1.0\times 10^{-4}$\xspace}
\newcommand{\nCheckSeedDouble}{2\xspace}
\newcommand{\nCheckSeedRelease}{1\xspace}
\newcommand{\nCheckSoftmaxA}{$1.4\times 10^{-15}$\xspace}
\newcommand{\nCheckSoftmaxB}{$9.3\times 10^{-16}$\xspace}
\newcommand{\nCheckSoftmaxC}{$1.2\times 10^{-15}$\xspace}
\newcommand{\nCheckSoftmaxD}{$1.1\times 10^{-15}$\xspace}
\newcommand{\nCheckSoftmaxE}{$1.5\times 10^{-15}$\xspace}
\newcommand{\nCheckSoftmaxF}{$1.4\times 10^{-15}$\xspace}
\newcommand{\nCheckSoftmaxR}{$1.5\times 10^{-15}$\xspace}
\newcommand{\nCheckSoftmaxTol}{$1.0\times 10^{-10}$\xspace}
\newcommand{\nCheckStep}{$1.0\times 10^{-6}$\xspace}
\newcommand{\nCheckTokensCount}{4\xspace}
\newcommand{\nConstGlobalMax}{552\xspace}
\newcommand{\nConstGlobalMin}{283\xspace}
\newcommand{\nConstMedianMax}{15.8\xspace}
\newcommand{\nConstMedianMin}{15.0\xspace}
\newcommand{\nContextA}{2,049\xspace}
\newcommand{\nControlDraws}{3\xspace}
\newcommand{\nControlExplained}{0.13\xspace}
\newcommand{\nControlPairs}{6\xspace}
\newcommand{\nControlRepresented}{0.00\xspace}
\newcommand{\nControlSeedBase}{20,260,923\xspace}
\newcommand{\nControlTenth}{0.11\xspace}
\newcommand{\nCorpusRemoved}{Books3, BookCorpus2, OpenSubtitles, YTSubtitles and OWT2\xspace}
\newcommand{\nCorpusRepo}{monology/pile-uncopyrighted\xspace}
\newcommand{\nCountChangeZeroR}{0.59\xspace}
\newcommand{\nCrossHalfCountA}{5,347\xspace}
\newcommand{\nCrossHalfCountB}{4,553\xspace}
\newcommand{\nCrossHalfCountC}{5,799\xspace}
\newcommand{\nCrossHalfCountD}{11,467\xspace}
\newcommand{\nCrossHalfCountE}{5,673\xspace}
\newcommand{\nCrossHalfCountF}{100,034\xspace}
\newcommand{\nCrossHalfCountR}{23,689\xspace}
\newcommand{\nCrossNineB}{0.19\xspace}
\newcommand{\nCrossNineCountA}{1,765\xspace}
\newcommand{\nCrossNineCountB}{1,528\xspace}
\newcommand{\nCrossNineCountC}{1,970\xspace}
\newcommand{\nCrossNineCountD}{3,853\xspace}
\newcommand{\nCrossNineCountE}{2,012\xspace}
\newcommand{\nCrossNineCountF}{38,829\xspace}
\newcommand{\nCrossNineCountR}{9,485\xspace}
\newcommand{\nCrossNineE}{0.25\xspace}
\newcommand{\nCrossQuadB}{0.95\xspace}
\newcommand{\nCrossQuadC}{0.99\xspace}
\newcommand{\nCrossZeroA}{0.84\xspace}
\newcommand{\nCrossZeroCountA}{6,867\xspace}
\newcommand{\nCrossZeroCountB}{6,078\xspace}
\newcommand{\nCrossZeroCountC}{7,251\xspace}
\newcommand{\nCrossZeroCountD}{14,519\xspace}
\newcommand{\nCrossZeroCountE}{7,171\xspace}
\newcommand{\nCrossZeroCountF}{120,462\xspace}
\newcommand{\nCrossZeroCountR}{30,404\xspace}
\newcommand{\nCrossZeroF}{0.92\xspace}
\newcommand{\nCrossZeroMax}{0.89\xspace}
\newcommand{\nCrossZeroMin}{0.74\xspace}
\newcommand{\nCrossZeroR}{0.69\xspace}
\newcommand{\nCrossingDoses}{10\xspace}
\newcommand{\nCrossingFloor}{0.1\xspace}
\newcommand{\nCtxR}{128\xspace}
\newcommand{\nCutoffOverLargestMax}{0.15\xspace}
\newcommand{\nCutoffOverLargestMin}{0.06\xspace}
\newcommand{\nDecodedCosF}{0.38\xspace}
\newcommand{\nDecodedCosMax}{0.50\xspace}
\newcommand{\nDecodedCosMin}{0.27\xspace}
\newcommand{\nDecodedCosR}{0.60\xspace}
\newcommand{\nDecodedRatioMax}{1.82\xspace}
\newcommand{\nDecodedRatioMin}{1.41\xspace}
\newcommand{\nDecodedRatioR}{0.64\xspace}
\newcommand{\nDecodedShareR}{0.38\xspace}
\newcommand{\nDecompGapMax}{$1.2\times 10^{-14}$\xspace}
\newcommand{\nDesignBoundA}{0.423\xspace}
\newcommand{\nDesignBoundB}{0.186\xspace}
\newcommand{\nDesignBoundC}{0.498\xspace}
\newcommand{\nDesignBoundD}{0.462\xspace}
\newcommand{\nDesignBoundE}{0.502\xspace}
\newcommand{\nDesignBoundF}{0.590\xspace}
\newcommand{\nDesignFactor}{0.75\xspace}
\newcommand{\nDevIgFiveFixedProducts}{5\xspace}
\newcommand{\nDevIgFiveRerank}{$8.8\times 10^{-6}$\xspace}
\newcommand{\nDevIgFiveRerankProducts}{5\xspace}
\newcommand{\nDevIgTenFixedProducts}{10\xspace}
\newcommand{\nDevMargin}{0.017\xspace}
\newcommand{\nDevSoftmaxRerank}{0.013\xspace}
\newcommand{\nDevTabBelowA}{0.77\xspace}
\newcommand{\nDevTabBelowB}{0.95\xspace}
\newcommand{\nDevTabBelowC}{0.64\xspace}
\newcommand{\nDevTabBelowD}{0.71\xspace}
\newcommand{\nDevTabBelowE}{0.82\xspace}
\newcommand{\nDevTabBelowF}{0.80\xspace}
\newcommand{\nDevTabFxHalfA}{0.702\xspace}
\newcommand{\nDevTabFxHalfB}{0.612\xspace}
\newcommand{\nDevTabFxHalfC}{0.780\xspace}
\newcommand{\nDevTabFxHalfD}{0.759\xspace}
\newcommand{\nDevTabFxHalfE}{0.781\xspace}
\newcommand{\nDevTabFxHalfF}{0.866\xspace}
\newcommand{\nDevTabFxNineA}{0.017\xspace}
\newcommand{\nDevTabFxNineB}{0.013\xspace}
\newcommand{\nDevTabFxNineC}{0.020\xspace}
\newcommand{\nDevTabFxNineD}{0.007\xspace}
\newcommand{\nDevTabFxNineE}{0.018\xspace}
\newcommand{\nDevTabFxNineF}{0.012\xspace}
\newcommand{\nDevTabFxPooledA}{0.564\xspace}
\newcommand{\nDevTabFxPooledB}{0.247\xspace}
\newcommand{\nDevTabFxPooledC}{0.663\xspace}
\newcommand{\nDevTabFxPooledD}{0.617\xspace}
\newcommand{\nDevTabFxPooledE}{0.669\xspace}
\newcommand{\nDevTabFxPooledF}{0.786\xspace}
\newcommand{\nDevTabFxZeroA}{0.739\xspace}
\newcommand{\nDevTabFxZeroB}{0.803\xspace}
\newcommand{\nDevTabFxZeroC}{0.784\xspace}
\newcommand{\nDevTabFxZeroD}{0.774\xspace}
\newcommand{\nDevTabFxZeroE}{0.794\xspace}
\newcommand{\nDevTabFxZeroF}{0.842\xspace}
\newcommand{\nDevTabMgHalfA}{0.042\xspace}
\newcommand{\nDevTabMgHalfB}{0.036\xspace}
\newcommand{\nDevTabMgHalfC}{0.036\xspace}
\newcommand{\nDevTabMgHalfD}{0.016\xspace}
\newcommand{\nDevTabMgHalfE}{0.039\xspace}
\newcommand{\nDevTabMgHalfF}{0.020\xspace}
\newcommand{\nDevTabMgNineA}{0.010\xspace}
\newcommand{\nDevTabMgNineB}{0.010\xspace}
\newcommand{\nDevTabMgNineC}{0.010\xspace}
\newcommand{\nDevTabMgNineD}{0.005\xspace}
\newcommand{\nDevTabMgNineE}{0.010\xspace}
\newcommand{\nDevTabMgNineF}{0.006\xspace}
\newcommand{\nDevTabMgPooledA}{0.033\xspace}
\newcommand{\nDevTabMgPooledB}{0.023\xspace}
\newcommand{\nDevTabMgPooledC}{0.033\xspace}
\newcommand{\nDevTabMgPooledD}{0.013\xspace}
\newcommand{\nDevTabMgPooledE}{0.031\xspace}
\newcommand{\nDevTabMgPooledF}{0.017\xspace}
\newcommand{\nDevTabMgZeroA}{0.121\xspace}
\newcommand{\nDevTabMgZeroB}{0.073\xspace}
\newcommand{\nDevTabMgZeroC}{0.106\xspace}
\newcommand{\nDevTabMgZeroD}{0.032\xspace}
\newcommand{\nDevTabMgZeroE}{0.128\xspace}
\newcommand{\nDevTabMgZeroF}{0.050\xspace}
\newcommand{\nDevTabPairedA}{-0.0017\xspace}
\newcommand{\nDevTabPairedB}{-0.0036\xspace}
\newcommand{\nDevTabPairedC}{-0.0005\xspace}
\newcommand{\nDevTabPairedD}{-0.0004\xspace}
\newcommand{\nDevTabPairedE}{-0.0023\xspace}
\newcommand{\nDevTabPairedF}{-0.0009\xspace}
\newcommand{\nDevTabSfHalfA}{0.699\xspace}
\newcommand{\nDevTabSfHalfB}{0.608\xspace}
\newcommand{\nDevTabSfHalfC}{0.775\xspace}
\newcommand{\nDevTabSfHalfD}{0.753\xspace}
\newcommand{\nDevTabSfHalfE}{0.777\xspace}
\newcommand{\nDevTabSfHalfF}{0.864\xspace}
\newcommand{\nDevTabSfNineA}{0.013\xspace}
\newcommand{\nDevTabSfNineB}{0.010\xspace}
\newcommand{\nDevTabSfNineC}{0.016\xspace}
\newcommand{\nDevTabSfNineD}{0.006\xspace}
\newcommand{\nDevTabSfNineE}{0.012\xspace}
\newcommand{\nDevTabSfNineF}{0.009\xspace}
\newcommand{\nDevTabSfPooledA}{0.552\xspace}
\newcommand{\nDevTabSfPooledB}{0.191\xspace}
\newcommand{\nDevTabSfPooledC}{0.654\xspace}
\newcommand{\nDevTabSfPooledD}{0.610\xspace}
\newcommand{\nDevTabSfPooledE}{0.660\xspace}
\newcommand{\nDevTabSfPooledF}{0.783\xspace}
\newcommand{\nDevTabSfZeroA}{0.731\xspace}
\newcommand{\nDevTabSfZeroB}{0.797\xspace}
\newcommand{\nDevTabSfZeroC}{0.777\xspace}
\newcommand{\nDevTabSfZeroD}{0.769\xspace}
\newcommand{\nDevTabSfZeroE}{0.787\xspace}
\newcommand{\nDevTabSfZeroF}{0.839\xspace}
\newcommand{\nDevTabSrHalfA}{0.031\xspace}
\newcommand{\nDevTabSrHalfB}{0.025\xspace}
\newcommand{\nDevTabSrHalfC}{0.028\xspace}
\newcommand{\nDevTabSrHalfD}{0.014\xspace}
\newcommand{\nDevTabSrHalfE}{0.026\xspace}
\newcommand{\nDevTabSrHalfF}{0.015\xspace}
\newcommand{\nDevTabSrNineA}{0.008\xspace}
\newcommand{\nDevTabSrNineB}{0.007\xspace}
\newcommand{\nDevTabSrNineC}{0.008\xspace}
\newcommand{\nDevTabSrNineD}{0.004\xspace}
\newcommand{\nDevTabSrNineE}{0.007\xspace}
\newcommand{\nDevTabSrNineF}{0.005\xspace}
\newcommand{\nDevTabSrPooledA}{0.024\xspace}
\newcommand{\nDevTabSrPooledB}{0.016\xspace}
\newcommand{\nDevTabSrPooledC}{0.026\xspace}
\newcommand{\nDevTabSrPooledD}{0.011\xspace}
\newcommand{\nDevTabSrPooledE}{0.021\xspace}
\newcommand{\nDevTabSrPooledF}{0.013\xspace}
\newcommand{\nDevTabSrZeroA}{0.072\xspace}
\newcommand{\nDevTabSrZeroB}{0.049\xspace}
\newcommand{\nDevTabSrZeroC}{0.075\xspace}
\newcommand{\nDevTabSrZeroD}{0.027\xspace}
\newcommand{\nDevTabSrZeroE}{0.058\xspace}
\newcommand{\nDevTabSrZeroF}{0.032\xspace}
\newcommand{\nDeviceHours}{31.8\xspace}
\newcommand{\nDeviceName}{NVIDIA A100 80GB\xspace}
\newcommand{\nDistinctActivationsBlock}{12,288\xspace}
\newcommand{\nDoseHalf}{0.5\xspace}
\newcommand{\nDoseNine}{0.9\xspace}
\newcommand{\nDoseQuad}{4\xspace}
\newcommand{\nDzDpMax}{10.1\xspace}
\newcommand{\nDzDpMin}{5.7\xspace}
\newcommand{\nEaEnterFNA}{25,839\xspace}
\newcommand{\nEaEnterFNB}{20,297\xspace}
\newcommand{\nEaEnterFNC}{27,710\xspace}
\newcommand{\nEaEnterFND}{55,137\xspace}
\newcommand{\nEaEnterFNE}{27,603\xspace}
\newcommand{\nEaEnterFNF}{562,336\xspace}
\newcommand{\nEaEnterFNR}{105,408\xspace}
\newcommand{\nEaEnterFPA}{25,839\xspace}
\newcommand{\nEaEnterFPB}{20,297\xspace}
\newcommand{\nEaEnterFPC}{27,710\xspace}
\newcommand{\nEaEnterFPD}{55,137\xspace}
\newcommand{\nEaEnterFPE}{27,603\xspace}
\newcommand{\nEaEnterFPF}{562,336\xspace}
\newcommand{\nEaEnterFPR}{105,408\xspace}
\newcommand{\nEaEnterPrecA}{0.026\xspace}
\newcommand{\nEaEnterPrecB}{0.013\xspace}
\newcommand{\nEaEnterPrecC}{0.020\xspace}
\newcommand{\nEaEnterPrecD}{0.031\xspace}
\newcommand{\nEaEnterPrecE}{0.019\xspace}
\newcommand{\nEaEnterPrecF}{0.027\xspace}
\newcommand{\nEaEnterPrecR}{0.002\xspace}
\newcommand{\nEaEnterRecA}{0.026\xspace}
\newcommand{\nEaEnterRecB}{0.013\xspace}
\newcommand{\nEaEnterRecC}{0.020\xspace}
\newcommand{\nEaEnterRecD}{0.031\xspace}
\newcommand{\nEaEnterRecE}{0.019\xspace}
\newcommand{\nEaEnterRecF}{0.027\xspace}
\newcommand{\nEaEnterRecR}{0.002\xspace}
\newcommand{\nEaEnterTPA}{681\xspace}
\newcommand{\nEaEnterTPB}{260\xspace}
\newcommand{\nEaEnterTPC}{572\xspace}
\newcommand{\nEaEnterTPD}{1,766\xspace}
\newcommand{\nEaEnterTPE}{530\xspace}
\newcommand{\nEaEnterTPF}{15,768\xspace}
\newcommand{\nEaEnterTPR}{245\xspace}
\newcommand{\nEaLeaveFNA}{16,893\xspace}
\newcommand{\nEaLeaveFNB}{14,389\xspace}
\newcommand{\nEaLeaveFNC}{18,383\xspace}
\newcommand{\nEaLeaveFND}{42,820\xspace}
\newcommand{\nEaLeaveFNE}{18,759\xspace}
\newcommand{\nEaLeaveFNF}{438,988\xspace}
\newcommand{\nEaLeaveFNR}{86,032\xspace}
\newcommand{\nEaLeaveFPA}{16,893\xspace}
\newcommand{\nEaLeaveFPB}{14,389\xspace}
\newcommand{\nEaLeaveFPC}{18,383\xspace}
\newcommand{\nEaLeaveFPD}{42,820\xspace}
\newcommand{\nEaLeaveFPE}{18,759\xspace}
\newcommand{\nEaLeaveFPF}{438,988\xspace}
\newcommand{\nEaLeaveFPR}{86,032\xspace}
\newcommand{\nEaLeavePrecA}{0.363\xspace}
\newcommand{\nEaLeavePrecB}{0.300\xspace}
\newcommand{\nEaLeavePrecC}{0.350\xspace}
\newcommand{\nEaLeavePrecD}{0.247\xspace}
\newcommand{\nEaLeavePrecE}{0.333\xspace}
\newcommand{\nEaLeavePrecF}{0.241\xspace}
\newcommand{\nEaLeavePrecR}{0.220\xspace}
\newcommand{\nEaLeaveRecA}{0.363\xspace}
\newcommand{\nEaLeaveRecB}{0.300\xspace}
\newcommand{\nEaLeaveRecC}{0.350\xspace}
\newcommand{\nEaLeaveRecD}{0.247\xspace}
\newcommand{\nEaLeaveRecE}{0.333\xspace}
\newcommand{\nEaLeaveRecF}{0.241\xspace}
\newcommand{\nEaLeaveRecR}{0.220\xspace}
\newcommand{\nEaLeaveTPA}{9,627\xspace}
\newcommand{\nEaLeaveTPB}{6,168\xspace}
\newcommand{\nEaLeaveTPC}{9,899\xspace}
\newcommand{\nEaLeaveTPD}{14,083\xspace}
\newcommand{\nEaLeaveTPE}{9,374\xspace}
\newcommand{\nEaLeaveTPF}{139,116\xspace}
\newcommand{\nEaLeaveTPR}{24,232\xspace}
\newcommand{\nEnergyFloor}{0.1\xspace}
\newcommand{\nEnergyZeroA}{0.77\xspace}
\newcommand{\nEnergyZeroB}{0.84\xspace}
\newcommand{\nEnergyZeroC}{0.75\xspace}
\newcommand{\nEnergyZeroD}{0.73\xspace}
\newcommand{\nEnergyZeroE}{0.80\xspace}
\newcommand{\nEnergyZeroF}{0.79\xspace}
\newcommand{\nEnergyZeroMax}{0.84\xspace}
\newcommand{\nEnergyZeroMin}{0.73\xspace}
\newcommand{\nEnergyZeroR}{0.05\xspace}
\newcommand{\nEnterShareA}{0.33\xspace}
\newcommand{\nEnterShareB}{0.39\xspace}
\newcommand{\nEnterShareC}{0.34\xspace}
\newcommand{\nEnterShareD}{0.33\xspace}
\newcommand{\nEnterShareE}{0.35\xspace}
\newcommand{\nEnterShareF}{0.37\xspace}
\newcommand{\nEnterShareMax}{0.39\xspace}
\newcommand{\nEnterShareMin}{0.33\xspace}
\newcommand{\nEnterShareR}{0.01\xspace}
\newcommand{\nExpansionF}{32\xspace}
\newcommand{\nFFourAcrossMax}{0.860\xspace}
\newcommand{\nFFourAcrossMean}{0.855\xspace}
\newcommand{\nFFourAcrossMin}{0.852\xspace}
\newcommand{\nFFourBetweenMax}{0.886\xspace}
\newcommand{\nFFourBetweenMean}{0.878\xspace}
\newcommand{\nFFourBetweenMin}{0.871\xspace}
\newcommand{\nFThreeBound}{4\xspace}
\newcommand{\nFThreeFloor}{0.5\xspace}
\newcommand{\nFVUDownA}{0.148\xspace}
\newcommand{\nFVUDownB}{0.165\xspace}
\newcommand{\nFVUDownC}{0.103\xspace}
\newcommand{\nFVUDownD}{0.063\xspace}
\newcommand{\nFVUDownE}{0.164\xspace}
\newcommand{\nFVUDownR}{0.217\xspace}
\newcommand{\nFVUFourMaxF}{0.156\xspace}
\newcommand{\nFVUFourMinF}{0.155\xspace}
\newcommand{\nFVUThreeMaxF}{0.1094\xspace}
\newcommand{\nFVUThreeMinF}{0.1090\xspace}
\newcommand{\nFVUUpA}{0.125\xspace}
\newcommand{\nFVUUpB}{0.110\xspace}
\newcommand{\nFVUUpC}{0.148\xspace}
\newcommand{\nFVUUpD}{0.071\xspace}
\newcommand{\nFVUUpE}{0.126\xspace}
\newcommand{\nFVUUpR}{0.174\xspace}
\newcommand{\nFixedErrA}{0.564\xspace}
\newcommand{\nFixedErrB}{0.247\xspace}
\newcommand{\nFixedErrC}{0.663\xspace}
\newcommand{\nFixedErrD}{0.617\xspace}
\newcommand{\nFixedErrDecA}{0.512\xspace}
\newcommand{\nFixedErrDecB}{0.287\xspace}
\newcommand{\nFixedErrDecC}{0.625\xspace}
\newcommand{\nFixedErrDecD}{0.577\xspace}
\newcommand{\nFixedErrDecE}{0.625\xspace}
\newcommand{\nFixedErrDecF}{0.749\xspace}
\newcommand{\nFixedErrDecR}{0.017\xspace}
\newcommand{\nFixedErrDoseFHalf}{0.866\xspace}
\newcommand{\nFixedErrDoseFNine}{0.012\xspace}
\newcommand{\nFixedErrDoseFZero}{0.842\xspace}
\newcommand{\nFixedErrE}{0.669\xspace}
\newcommand{\nFixedErrF}{0.786\xspace}
\newcommand{\nFixedErrHalfA}{0.702\xspace}
\newcommand{\nFixedErrHalfB}{0.612\xspace}
\newcommand{\nFixedErrHalfC}{0.780\xspace}
\newcommand{\nFixedErrHalfD}{0.759\xspace}
\newcommand{\nFixedErrHalfE}{0.781\xspace}
\newcommand{\nFixedErrHalfR}{0.027\xspace}
\newcommand{\nFixedErrMax}{0.669\xspace}
\newcommand{\nFixedErrMin}{0.247\xspace}
\newcommand{\nFixedErrNineA}{0.017\xspace}
\newcommand{\nFixedErrNineB}{0.013\xspace}
\newcommand{\nFixedErrNineC}{0.020\xspace}
\newcommand{\nFixedErrNineD}{0.007\xspace}
\newcommand{\nFixedErrNineE}{0.018\xspace}
\newcommand{\nFixedErrNineR}{0.003\xspace}
\newcommand{\nFixedErrR}{0.017\xspace}
\newcommand{\nFixedErrRunFive}{0.782\xspace}
\newcommand{\nFixedErrRunFiveHalf}{0.861\xspace}
\newcommand{\nFixedErrRunFiveNine}{0.012\xspace}
\newcommand{\nFixedErrRunFiveZero}{0.842\xspace}
\newcommand{\nFixedErrRunFour}{0.770\xspace}
\newcommand{\nFixedErrRunFourHalf}{0.855\xspace}
\newcommand{\nFixedErrRunFourNine}{0.012\xspace}
\newcommand{\nFixedErrRunFourZero}{0.838\xspace}
\newcommand{\nFixedErrRunOne}{0.782\xspace}
\newcommand{\nFixedErrRunOneHalf}{0.865\xspace}
\newcommand{\nFixedErrRunOneNine}{0.012\xspace}
\newcommand{\nFixedErrRunOneZero}{0.838\xspace}
\newcommand{\nFixedErrRunSeven}{0.790\xspace}
\newcommand{\nFixedErrRunSevenHalf}{0.869\xspace}
\newcommand{\nFixedErrRunSevenNine}{0.012\xspace}
\newcommand{\nFixedErrRunSevenZero}{0.848\xspace}
\newcommand{\nFixedErrRunSix}{0.785\xspace}
\newcommand{\nFixedErrRunSixHalf}{0.867\xspace}
\newcommand{\nFixedErrRunSixNine}{0.012\xspace}
\newcommand{\nFixedErrRunSixZero}{0.844\xspace}
\newcommand{\nFixedErrRunThree}{0.796\xspace}
\newcommand{\nFixedErrRunThreeHalf}{0.870\xspace}
\newcommand{\nFixedErrRunThreeNine}{0.013\xspace}
\newcommand{\nFixedErrRunThreeZero}{0.843\xspace}
\newcommand{\nFixedErrRunTwo}{0.794\xspace}
\newcommand{\nFixedErrRunTwoHalf}{0.872\xspace}
\newcommand{\nFixedErrRunTwoNine}{0.013\xspace}
\newcommand{\nFixedErrRunTwoZero}{0.843\xspace}
\newcommand{\nFixedErrRunZero}{0.790\xspace}
\newcommand{\nFixedErrRunZeroHalf}{0.866\xspace}
\newcommand{\nFixedErrRunZeroNine}{0.013\xspace}
\newcommand{\nFixedErrRunZeroZero}{0.839\xspace}
\newcommand{\nFixedErrZeroA}{0.739\xspace}
\newcommand{\nFixedErrZeroB}{0.803\xspace}
\newcommand{\nFixedErrZeroC}{0.784\xspace}
\newcommand{\nFixedErrZeroD}{0.774\xspace}
\newcommand{\nFixedErrZeroE}{0.794\xspace}
\newcommand{\nFixedErrZeroR}{0.075\xspace}
\newcommand{\nFixedSizeEightF}{0.590\xspace}
\newcommand{\nFixedSizeEightRel}{0.578\xspace}
\newcommand{\nFixedSizeFourF}{0.676\xspace}
\newcommand{\nFixedSizeFourRel}{0.613\xspace}
\newcommand{\nFixedSizeSixteenF}{0.551\xspace}
\newcommand{\nFixedSizeSixteenRel}{0.631\xspace}
\newcommand{\nFixedSizeTwoF}{0.770\xspace}
\newcommand{\nFixedSizeTwoRel}{0.685\xspace}
\newcommand{\nFvuTailFourMax}{0.1559\xspace}
\newcommand{\nFvuTailFourMin}{0.1543\xspace}
\newcommand{\nFvuTailThreeMax}{0.1131\xspace}
\newcommand{\nFvuTailThreeMin}{0.1128\xspace}
\newcommand{\nGainBound}{0.19\xspace}
\newcommand{\nGap}{0.38\xspace}
\newcommand{\nGapRetrain}{0.34\xspace}
\newcommand{\nGaugeFactorF}{0.949\xspace}
\newcommand{\nHeldout}{1,024\xspace}
\newcommand{\nHoursPublic}{1.7\xspace}
\newcommand{\nHoursReadout}{3.9\xspace}
\newcommand{\nHoursTraining}{20.8\xspace}
\newcommand{\nHoursTransport}{5.4\xspace}
\newcommand{\nInitLayerMult}{7,919\xspace}
\newcommand{\nInterventionsPerRunF}{16,384\xspace}
\newcommand{\nIvDraws}{2,000\xspace}
\newcommand{\nIvHighCrossTerm}{0.053\xspace}
\newcommand{\nIvHighCrossTermRetrain}{0.050\xspace}
\newcommand{\nIvHighGap}{0.390\xspace}
\newcommand{\nIvHighGapRetrain}{0.346\xspace}
\newcommand{\nIvHighMaskTerm}{0.343\xspace}
\newcommand{\nIvHighMaskTermRetrain}{0.297\xspace}
\newcommand{\nIvHighMatchMasked}{0.376\xspace}
\newcommand{\nIvHighMatchMaskedRetrain}{0.441\xspace}
\newcommand{\nIvHighMatchUnmasked}{0.222\xspace}
\newcommand{\nIvHighMatchUnmaskedRetrain}{0.350\xspace}
\newcommand{\nIvHighResid}{0.153\xspace}
\newcommand{\nIvHighResidRetrain}{0.154\xspace}
\newcommand{\nIvHighSparseMasked}{0.476\xspace}
\newcommand{\nIvHighSparseMaskedRetrain}{0.515\xspace}
\newcommand{\nIvHighSparseUnmasked}{0.571\xspace}
\newcommand{\nIvHighSparseUnmaskedRetrain}{0.640\xspace}
\newcommand{\nIvHighWholeTarget}{0.857\xspace}
\newcommand{\nIvHighWholeTargetRetrain}{0.851\xspace}
\newcommand{\nIvLowCrossTerm}{0.044\xspace}
\newcommand{\nIvLowCrossTermRetrain}{0.047\xspace}
\newcommand{\nIvLowGap}{0.379\xspace}
\newcommand{\nIvLowGapRetrain}{0.334\xspace}
\newcommand{\nIvLowMaskTerm}{0.330\xspace}
\newcommand{\nIvLowMaskTermRetrain}{0.286\xspace}
\newcommand{\nIvLowMatchMasked}{0.352\xspace}
\newcommand{\nIvLowMatchMaskedRetrain}{0.426\xspace}
\newcommand{\nIvLowMatchUnmasked}{0.220\xspace}
\newcommand{\nIvLowMatchUnmaskedRetrain}{0.349\xspace}
\newcommand{\nIvLowResid}{0.143\xspace}
\newcommand{\nIvLowResidRetrain}{0.149\xspace}
\newcommand{\nIvLowSparseMasked}{0.457\xspace}
\newcommand{\nIvLowSparseMaskedRetrain}{0.501\xspace}
\newcommand{\nIvLowSparseUnmasked}{0.568\xspace}
\newcommand{\nIvLowSparseUnmaskedRetrain}{0.638\xspace}
\newcommand{\nIvLowWholeTarget}{0.847\xspace}
\newcommand{\nIvLowWholeTargetRetrain}{0.846\xspace}
\newcommand{\nIvPctHigh}{97.5\xspace}
\newcommand{\nIvPctLow}{2.5\xspace}
\newcommand{\nIvSeed}{20260925\xspace}
\newcommand{\nKA}{16\xspace}
\newcommand{\nKB}{16\xspace}
\newcommand{\nKC}{16\xspace}
\newcommand{\nKD}{32\xspace}
\newcommand{\nKE}{16\xspace}
\newcommand{\nKF}{32\xspace}
\newcommand{\nKFiled}{25\xspace}
\newcommand{\nKR}{none\xspace}
\newcommand{\nLayerDownA}{4\xspace}
\newcommand{\nLayerDownB}{2\xspace}
\newcommand{\nLayerDownC}{5\xspace}
\newcommand{\nLayerDownD}{7\xspace}
\newcommand{\nLayerDownE}{4\xspace}
\newcommand{\nLayerDownR}{4\xspace}
\newcommand{\nLayerThird}{5\xspace}
\newcommand{\nLayerUpA}{3\xspace}
\newcommand{\nLayerUpB}{1\xspace}
\newcommand{\nLayerUpC}{4\xspace}
\newcommand{\nLayerUpD}{6\xspace}
\newcommand{\nLayerUpE}{3\xspace}
\newcommand{\nLayerUpR}{3\xspace}
\newcommand{\nLeavePrecMax}{0.93\xspace}
\newcommand{\nLeavePrecMin}{0.89\xspace}
\newcommand{\nLeaveRecMax}{0.95\xspace}
\newcommand{\nLeaveRecMin}{0.92\xspace}
\newcommand{\nLevelChosen}{32\xspace}
\newcommand{\nLevelEight}{0.463\xspace}
\newcommand{\nLevelFour}{0.403\xspace}
\newcommand{\nLevelOne}{0.246\xspace}
\newcommand{\nLevelSixteen}{0.521\xspace}
\newcommand{\nLevelThirtyTwo}{0.568\xspace}
\newcommand{\nLevelTwo}{0.336\xspace}
\newcommand{\nLevelWithin}{0.01\xspace}
\newcommand{\nLipFeaturesPerToken}{4\xspace}
\newcommand{\nLipSeed}{20260921\xspace}
\newcommand{\nLipWritesPerRun}{2,048\xspace}
\newcommand{\nMaEnterFNA}{15,151\xspace}
\newcommand{\nMaEnterFNB}{10,382\xspace}
\newcommand{\nMaEnterFNC}{15,831\xspace}
\newcommand{\nMaEnterFND}{29,782\xspace}
\newcommand{\nMaEnterFNE}{16,187\xspace}
\newcommand{\nMaEnterFNF}{323,358\xspace}
\newcommand{\nMaEnterFNR}{51,791\xspace}
\newcommand{\nMaEnterFPA}{15,151\xspace}
\newcommand{\nMaEnterFPB}{10,382\xspace}
\newcommand{\nMaEnterFPC}{15,831\xspace}
\newcommand{\nMaEnterFPD}{29,782\xspace}
\newcommand{\nMaEnterFPE}{16,187\xspace}
\newcommand{\nMaEnterFPF}{323,358\xspace}
\newcommand{\nMaEnterFPR}{51,791\xspace}
\newcommand{\nMaEnterPrecA}{0.429\xspace}
\newcommand{\nMaEnterPrecB}{0.495\xspace}
\newcommand{\nMaEnterPrecC}{0.440\xspace}
\newcommand{\nMaEnterPrecD}{0.477\xspace}
\newcommand{\nMaEnterPrecE}{0.425\xspace}
\newcommand{\nMaEnterPrecF}{0.441\xspace}
\newcommand{\nMaEnterPrecR}{0.510\xspace}
\newcommand{\nMaEnterRecA}{0.429\xspace}
\newcommand{\nMaEnterRecB}{0.495\xspace}
\newcommand{\nMaEnterRecC}{0.440\xspace}
\newcommand{\nMaEnterRecD}{0.477\xspace}
\newcommand{\nMaEnterRecE}{0.425\xspace}
\newcommand{\nMaEnterRecF}{0.441\xspace}
\newcommand{\nMaEnterRecR}{0.510\xspace}
\newcommand{\nMaEnterTPA}{11,369\xspace}
\newcommand{\nMaEnterTPB}{10,175\xspace}
\newcommand{\nMaEnterTPC}{12,451\xspace}
\newcommand{\nMaEnterTPD}{27,121\xspace}
\newcommand{\nMaEnterTPE}{11,946\xspace}
\newcommand{\nMaEnterTPF}{254,746\xspace}
\newcommand{\nMaEnterTPR}{53,862\xspace}
\newcommand{\nMaLeaveFNA}{11,994\xspace}
\newcommand{\nMaLeaveFNB}{9,011\xspace}
\newcommand{\nMaLeaveFNC}{12,834\xspace}
\newcommand{\nMaLeaveFND}{28,754\xspace}
\newcommand{\nMaLeaveFNE}{12,247\xspace}
\newcommand{\nMaLeaveFNF}{279,809\xspace}
\newcommand{\nMaLeaveFNR}{48,414\xspace}
\newcommand{\nMaLeaveFPA}{11,994\xspace}
\newcommand{\nMaLeaveFPB}{9,011\xspace}
\newcommand{\nMaLeaveFPC}{12,834\xspace}
\newcommand{\nMaLeaveFPD}{28,754\xspace}
\newcommand{\nMaLeaveFPE}{12,247\xspace}
\newcommand{\nMaLeaveFPF}{279,809\xspace}
\newcommand{\nMaLeaveFPR}{48,414\xspace}
\newcommand{\nMaLeavePrecA}{0.548\xspace}
\newcommand{\nMaLeavePrecB}{0.562\xspace}
\newcommand{\nMaLeavePrecC}{0.546\xspace}
\newcommand{\nMaLeavePrecD}{0.495\xspace}
\newcommand{\nMaLeavePrecE}{0.565\xspace}
\newcommand{\nMaLeavePrecF}{0.516\xspace}
\newcommand{\nMaLeavePrecR}{0.561\xspace}
\newcommand{\nMaLeaveRecA}{0.548\xspace}
\newcommand{\nMaLeaveRecB}{0.562\xspace}
\newcommand{\nMaLeaveRecC}{0.546\xspace}
\newcommand{\nMaLeaveRecD}{0.495\xspace}
\newcommand{\nMaLeaveRecE}{0.565\xspace}
\newcommand{\nMaLeaveRecF}{0.516\xspace}
\newcommand{\nMaLeaveRecR}{0.561\xspace}
\newcommand{\nMaLeaveTPA}{14,526\xspace}
\newcommand{\nMaLeaveTPB}{11,546\xspace}
\newcommand{\nMaLeaveTPC}{15,448\xspace}
\newcommand{\nMaLeaveTPD}{28,149\xspace}
\newcommand{\nMaLeaveTPE}{15,886\xspace}
\newcommand{\nMaLeaveTPF}{298,295\xspace}
\newcommand{\nMaLeaveTPR}{61,850\xspace}
\newcommand{\nMapSeedNull}{7,000\xspace}
\newcommand{\nMapSeedRandom}{6,000\xspace}
\newcommand{\nMarginErrA}{0.033\xspace}
\newcommand{\nMarginErrB}{0.023\xspace}
\newcommand{\nMarginErrC}{0.033\xspace}
\newcommand{\nMarginErrD}{0.013\xspace}
\newcommand{\nMarginErrDecA}{0.036\xspace}
\newcommand{\nMarginErrDecB}{0.029\xspace}
\newcommand{\nMarginErrDecC}{0.036\xspace}
\newcommand{\nMarginErrDecD}{0.014\xspace}
\newcommand{\nMarginErrDecE}{0.034\xspace}
\newcommand{\nMarginErrDecF}{0.018\xspace}
\newcommand{\nMarginErrDecR}{0.009\xspace}
\newcommand{\nMarginErrDoseFHalf}{0.020\xspace}
\newcommand{\nMarginErrDoseFNine}{0.006\xspace}
\newcommand{\nMarginErrDoseFZero}{0.050\xspace}
\newcommand{\nMarginErrE}{0.031\xspace}
\newcommand{\nMarginErrF}{0.017\xspace}
\newcommand{\nMarginErrHalfA}{0.042\xspace}
\newcommand{\nMarginErrHalfB}{0.036\xspace}
\newcommand{\nMarginErrHalfC}{0.036\xspace}
\newcommand{\nMarginErrHalfD}{0.016\xspace}
\newcommand{\nMarginErrHalfE}{0.039\xspace}
\newcommand{\nMarginErrHalfR}{0.011\xspace}
\newcommand{\nMarginErrMax}{0.033\xspace}
\newcommand{\nMarginErrMin}{0.013\xspace}
\newcommand{\nMarginErrNineA}{0.010\xspace}
\newcommand{\nMarginErrNineB}{0.010\xspace}
\newcommand{\nMarginErrNineC}{0.010\xspace}
\newcommand{\nMarginErrNineD}{0.005\xspace}
\newcommand{\nMarginErrNineE}{0.010\xspace}
\newcommand{\nMarginErrNineR}{0.002\xspace}
\newcommand{\nMarginErrR}{0.008\xspace}
\newcommand{\nMarginErrRunFive}{0.017\xspace}
\newcommand{\nMarginErrRunFiveHalf}{0.021\xspace}
\newcommand{\nMarginErrRunFiveNine}{0.006\xspace}
\newcommand{\nMarginErrRunFiveZero}{0.048\xspace}
\newcommand{\nMarginErrRunFour}{0.017\xspace}
\newcommand{\nMarginErrRunFourHalf}{0.021\xspace}
\newcommand{\nMarginErrRunFourNine}{0.006\xspace}
\newcommand{\nMarginErrRunFourZero}{0.051\xspace}
\newcommand{\nMarginErrRunOne}{0.017\xspace}
\newcommand{\nMarginErrRunOneHalf}{0.021\xspace}
\newcommand{\nMarginErrRunOneNine}{0.006\xspace}
\newcommand{\nMarginErrRunOneZero}{0.052\xspace}
\newcommand{\nMarginErrRunSeven}{0.017\xspace}
\newcommand{\nMarginErrRunSevenHalf}{0.020\xspace}
\newcommand{\nMarginErrRunSevenNine}{0.006\xspace}
\newcommand{\nMarginErrRunSevenZero}{0.049\xspace}
\newcommand{\nMarginErrRunSix}{0.016\xspace}
\newcommand{\nMarginErrRunSixHalf}{0.020\xspace}
\newcommand{\nMarginErrRunSixNine}{0.006\xspace}
\newcommand{\nMarginErrRunSixZero}{0.047\xspace}
\newcommand{\nMarginErrRunThree}{0.017\xspace}
\newcommand{\nMarginErrRunThreeHalf}{0.020\xspace}
\newcommand{\nMarginErrRunThreeNine}{0.006\xspace}
\newcommand{\nMarginErrRunThreeZero}{0.050\xspace}
\newcommand{\nMarginErrRunTwo}{0.017\xspace}
\newcommand{\nMarginErrRunTwoHalf}{0.020\xspace}
\newcommand{\nMarginErrRunTwoNine}{0.006\xspace}
\newcommand{\nMarginErrRunTwoZero}{0.051\xspace}
\newcommand{\nMarginErrRunZero}{0.017\xspace}
\newcommand{\nMarginErrRunZeroHalf}{0.020\xspace}
\newcommand{\nMarginErrRunZeroNine}{0.006\xspace}
\newcommand{\nMarginErrRunZeroZero}{0.050\xspace}
\newcommand{\nMarginErrZeroA}{0.121\xspace}
\newcommand{\nMarginErrZeroB}{0.073\xspace}
\newcommand{\nMarginErrZeroC}{0.106\xspace}
\newcommand{\nMarginErrZeroD}{0.032\xspace}
\newcommand{\nMarginErrZeroE}{0.128\xspace}
\newcommand{\nMarginErrZeroR}{0.022\xspace}
\newcommand{\nMarginSizeEightF}{0.405\xspace}
\newcommand{\nMarginSizeEightRel}{0.498\xspace}
\newcommand{\nMarginSizeFourF}{0.371\xspace}
\newcommand{\nMarginSizeFourRel}{0.425\xspace}
\newcommand{\nMarginSizeSixteenF}{0.458\xspace}
\newcommand{\nMarginSizeSixteenRel}{0.670\xspace}
\newcommand{\nMarginSizeTwoF}{0.287\xspace}
\newcommand{\nMarginSizeTwoRel}{0.347\xspace}
\newcommand{\nMasked}{0.47\xspace}
\newcommand{\nMaskedRetrain}{0.51\xspace}
\newcommand{\nMeasureSeqs}{256\xspace}
\newcommand{\nMedMarginalShare}{0.14\xspace}
\newcommand{\nMedStableShare}{0.30\xspace}
\newcommand{\nMedianAtomsDec}{5\xspace}
\newcommand{\nMedianAtomsEnc}{2\xspace}
\newcommand{\nMediationDiff}{0.16\xspace}
\newcommand{\nMediationMargin}{0.2\xspace}
\newcommand{\nMgFNA}{4,378\xspace}
\newcommand{\nMgFNB}{2,901\xspace}
\newcommand{\nMgFNC}{4,080\xspace}
\newcommand{\nMgFND}{4,752\xspace}
\newcommand{\nMgFNE}{4,976\xspace}
\newcommand{\nMgFNF}{65,554\xspace}
\newcommand{\nMgFNR}{2,651\xspace}
\newcommand{\nMgFPA}{5,007\xspace}
\newcommand{\nMgFPB}{2,945\xspace}
\newcommand{\nMgFPC}{3,935\xspace}
\newcommand{\nMgFPD}{5,559\xspace}
\newcommand{\nMgFPE}{5,751\xspace}
\newcommand{\nMgFPF}{75,675\xspace}
\newcommand{\nMgFPR}{37,725\xspace}
\newcommand{\nMgLeaveFNA}{1,984\xspace}
\newcommand{\nMgLeaveFNB}{1,733\xspace}
\newcommand{\nMgLeaveFNC}{2,366\xspace}
\newcommand{\nMgLeaveFND}{3,120\xspace}
\newcommand{\nMgLeaveFNE}{2,327\xspace}
\newcommand{\nMgLeaveFNF}{39,465\xspace}
\newcommand{\nMgLeaveFNR}{7,906\xspace}
\newcommand{\nMgLeaveFPA}{2,613\xspace}
\newcommand{\nMgLeaveFPB}{1,777\xspace}
\newcommand{\nMgLeaveFPC}{2,221\xspace}
\newcommand{\nMgLeaveFPD}{3,927\xspace}
\newcommand{\nMgLeaveFPE}{3,102\xspace}
\newcommand{\nMgLeaveFPF}{49,586\xspace}
\newcommand{\nMgLeaveFPR}{2,168\xspace}
\newcommand{\nMgLeavePrecA}{0.904\xspace}
\newcommand{\nMgLeavePrecB}{0.914\xspace}
\newcommand{\nMgLeavePrecC}{0.921\xspace}
\newcommand{\nMgLeavePrecD}{0.932\xspace}
\newcommand{\nMgLeavePrecE}{0.893\xspace}
\newcommand{\nMgLeavePrecF}{0.916\xspace}
\newcommand{\nMgLeavePrecR}{0.979\xspace}
\newcommand{\nMgLeaveRecA}{0.925\xspace}
\newcommand{\nMgLeaveRecB}{0.916\xspace}
\newcommand{\nMgLeaveRecC}{0.916\xspace}
\newcommand{\nMgLeaveRecD}{0.945\xspace}
\newcommand{\nMgLeaveRecE}{0.917\xspace}
\newcommand{\nMgLeaveRecF}{0.932\xspace}
\newcommand{\nMgLeaveRecR}{0.928\xspace}
\newcommand{\nMgLeaveTPA}{24,536\xspace}
\newcommand{\nMgLeaveTPB}{18,824\xspace}
\newcommand{\nMgLeaveTPC}{25,916\xspace}
\newcommand{\nMgLeaveTPD}{53,783\xspace}
\newcommand{\nMgLeaveTPE}{25,806\xspace}
\newcommand{\nMgLeaveTPF}{538,639\xspace}
\newcommand{\nMgLeaveTPR}{102,358\xspace}
\newcommand{\nMgPrecAuditA}{0.816\xspace}
\newcommand{\nMgPrecAuditB}{0.857\xspace}
\newcommand{\nMgPrecAuditC}{0.860\xspace}
\newcommand{\nMgPrecAuditD}{0.904\xspace}
\newcommand{\nMgPrecAuditE}{0.801\xspace}
\newcommand{\nMgPrecAuditF}{0.871\xspace}
\newcommand{\nMgPrecAuditR}{0.732\xspace}
\newcommand{\nMgRecAuditA}{0.835\xspace}
\newcommand{\nMgRecAuditB}{0.859\xspace}
\newcommand{\nMgRecAuditC}{0.856\xspace}
\newcommand{\nMgRecAuditD}{0.916\xspace}
\newcommand{\nMgRecAuditE}{0.823\xspace}
\newcommand{\nMgRecAuditF}{0.887\xspace}
\newcommand{\nMgRecAuditR}{0.975\xspace}
\newcommand{\nMgTPA}{22,142\xspace}
\newcommand{\nMgTPB}{17,656\xspace}
\newcommand{\nMgTPC}{24,202\xspace}
\newcommand{\nMgTPD}{52,151\xspace}
\newcommand{\nMgTPE}{23,157\xspace}
\newcommand{\nMgTPF}{512,550\xspace}
\newcommand{\nMgTPR}{103,002\xspace}
\newcommand{\nMinChars}{16\xspace}
\newcommand{\nMultiAtomAcross}{0.85\xspace}
\newcommand{\nMultiAtomBetweenMax}{0.89\xspace}
\newcommand{\nNearestInactiveMax}{-0.008\xspace}
\newcommand{\nNearestInactiveMin}{-0.018\xspace}
\newcommand{\nNlEnterFNA}{26,519\xspace}
\newcommand{\nNlEnterFNB}{20,555\xspace}
\newcommand{\nNlEnterFNC}{28,282\xspace}
\newcommand{\nNlEnterFND}{56,899\xspace}
\newcommand{\nNlEnterFNE}{28,131\xspace}
\newcommand{\nNlEnterFNF}{577,967\xspace}
\newcommand{\nNlEnterFNR}{105,623\xspace}
\newcommand{\nNlEnterFPA}{26,519\xspace}
\newcommand{\nNlEnterFPB}{20,555\xspace}
\newcommand{\nNlEnterFPC}{28,282\xspace}
\newcommand{\nNlEnterFPD}{56,899\xspace}
\newcommand{\nNlEnterFPE}{28,131\xspace}
\newcommand{\nNlEnterFPF}{577,967\xspace}
\newcommand{\nNlEnterFPR}{105,623\xspace}
\newcommand{\nNlEnterPrecA}{$3.8\times 10^{-5}$\xspace}
\newcommand{\nNlEnterPrecB}{$9.7\times 10^{-5}$\xspace}
\newcommand{\nNlEnterPrecC}{0.000\xspace}
\newcommand{\nNlEnterPrecD}{$7.0\times 10^{-5}$\xspace}
\newcommand{\nNlEnterPrecE}{$7.1\times 10^{-5}$\xspace}
\newcommand{\nNlEnterPrecF}{$2.4\times 10^{-4}$\xspace}
\newcommand{\nNlEnterPrecR}{$2.8\times 10^{-4}$\xspace}
\newcommand{\nNlEnterRecA}{$3.8\times 10^{-5}$\xspace}
\newcommand{\nNlEnterRecB}{$9.7\times 10^{-5}$\xspace}
\newcommand{\nNlEnterRecC}{0.000\xspace}
\newcommand{\nNlEnterRecD}{$7.0\times 10^{-5}$\xspace}
\newcommand{\nNlEnterRecE}{$7.1\times 10^{-5}$\xspace}
\newcommand{\nNlEnterRecF}{$2.4\times 10^{-4}$\xspace}
\newcommand{\nNlEnterRecR}{$2.8\times 10^{-4}$\xspace}
\newcommand{\nNlEnterTPA}{1\xspace}
\newcommand{\nNlEnterTPB}{2\xspace}
\newcommand{\nNlEnterTPC}{0\xspace}
\newcommand{\nNlEnterTPD}{4\xspace}
\newcommand{\nNlEnterTPE}{2\xspace}
\newcommand{\nNlEnterTPF}{137\xspace}
\newcommand{\nNlEnterTPR}{30\xspace}
\newcommand{\nNlLeaveFNA}{21,504\xspace}
\newcommand{\nNlLeaveFNB}{17,224\xspace}
\newcommand{\nNlLeaveFNC}{23,229\xspace}
\newcommand{\nNlLeaveFND}{51,266\xspace}
\newcommand{\nNlLeaveFNE}{22,835\xspace}
\newcommand{\nNlLeaveFNF}{506,803\xspace}
\newcommand{\nNlLeaveFNR}{100,665\xspace}
\newcommand{\nNlLeaveFPA}{21,504\xspace}
\newcommand{\nNlLeaveFPB}{17,224\xspace}
\newcommand{\nNlLeaveFPC}{23,229\xspace}
\newcommand{\nNlLeaveFPD}{51,266\xspace}
\newcommand{\nNlLeaveFPE}{22,835\xspace}
\newcommand{\nNlLeaveFPF}{506,803\xspace}
\newcommand{\nNlLeaveFPR}{100,665\xspace}
\newcommand{\nNlLeavePrecA}{0.189\xspace}
\newcommand{\nNlLeavePrecB}{0.162\xspace}
\newcommand{\nNlLeavePrecC}{0.179\xspace}
\newcommand{\nNlLeavePrecD}{0.099\xspace}
\newcommand{\nNlLeavePrecE}{0.188\xspace}
\newcommand{\nNlLeavePrecF}{0.123\xspace}
\newcommand{\nNlLeavePrecR}{0.087\xspace}
\newcommand{\nNlLeaveRecA}{0.189\xspace}
\newcommand{\nNlLeaveRecB}{0.162\xspace}
\newcommand{\nNlLeaveRecC}{0.179\xspace}
\newcommand{\nNlLeaveRecD}{0.099\xspace}
\newcommand{\nNlLeaveRecE}{0.188\xspace}
\newcommand{\nNlLeaveRecF}{0.123\xspace}
\newcommand{\nNlLeaveRecR}{0.087\xspace}
\newcommand{\nNlLeaveTPA}{5,016\xspace}
\newcommand{\nNlLeaveTPB}{3,333\xspace}
\newcommand{\nNlLeaveTPC}{5,053\xspace}
\newcommand{\nNlLeaveTPD}{5,637\xspace}
\newcommand{\nNlLeaveTPE}{5,298\xspace}
\newcommand{\nNlLeaveTPF}{71,301\xspace}
\newcommand{\nNlLeaveTPR}{9,599\xspace}
\newcommand{\nNnlsCheckEvery}{25\xspace}
\newcommand{\nNnlsIters}{1,000\xspace}
\newcommand{\nNnlsStepMultiple}{1.01\xspace}
\newcommand{\nNnlsStop}{$1.0\times 10^{-8}$\xspace}
\newcommand{\nNoRerankErrF}{0.558\xspace}
\newcommand{\nNonzero}{$1.0\times 10^{-4}$\xspace}
\newcommand{\nNormMedianF}{1.68\xspace}
\newcommand{\nNormMedianRel}{3.57\xspace}
\newcommand{\nNullDrawsPerPair}{200\xspace}
\newcommand{\nNullSeed}{20260921\xspace}
\newcommand{\nOneToOne}{0.22\xspace}
\newcommand{\nOneToOneExact}{0.221\xspace}
\newcommand{\nOneToOneMasked}{0.36\xspace}
\newcommand{\nOneToOneMaskedRetrain}{0.43\xspace}
\newcommand{\nOneToOneRetrain}{0.35\xspace}
\newcommand{\nOracleFNA}{0\xspace}
\newcommand{\nOracleFNB}{0\xspace}
\newcommand{\nOracleFNC}{0\xspace}
\newcommand{\nOracleFND}{0\xspace}
\newcommand{\nOracleFNE}{0\xspace}
\newcommand{\nOracleFNF}{0\xspace}
\newcommand{\nOracleFNR}{0\xspace}
\newcommand{\nOracleFPA}{0\xspace}
\newcommand{\nOracleFPB}{0\xspace}
\newcommand{\nOracleFPC}{0\xspace}
\newcommand{\nOracleFPD}{0\xspace}
\newcommand{\nOracleFPE}{0\xspace}
\newcommand{\nOracleFPF}{0\xspace}
\newcommand{\nOracleFPR}{0\xspace}
\newcommand{\nOracleLeaveFNA}{0\xspace}
\newcommand{\nOracleLeaveFNB}{0\xspace}
\newcommand{\nOracleLeaveFNC}{0\xspace}
\newcommand{\nOracleLeaveFND}{0\xspace}
\newcommand{\nOracleLeaveFNE}{0\xspace}
\newcommand{\nOracleLeaveFNF}{0\xspace}
\newcommand{\nOracleLeaveFNR}{0\xspace}
\newcommand{\nOracleLeaveFPA}{0\xspace}
\newcommand{\nOracleLeaveFPB}{0\xspace}
\newcommand{\nOracleLeaveFPC}{0\xspace}
\newcommand{\nOracleLeaveFPD}{0\xspace}
\newcommand{\nOracleLeaveFPE}{0\xspace}
\newcommand{\nOracleLeaveFPF}{0\xspace}
\newcommand{\nOracleLeaveFPR}{0\xspace}
\newcommand{\nOracleLeavePrecA}{1.000\xspace}
\newcommand{\nOracleLeavePrecB}{1.000\xspace}
\newcommand{\nOracleLeavePrecC}{1.000\xspace}
\newcommand{\nOracleLeavePrecD}{1.000\xspace}
\newcommand{\nOracleLeavePrecE}{1.000\xspace}
\newcommand{\nOracleLeavePrecF}{1.000\xspace}
\newcommand{\nOracleLeavePrecR}{1.000\xspace}
\newcommand{\nOracleLeaveRecA}{1.000\xspace}
\newcommand{\nOracleLeaveRecB}{1.000\xspace}
\newcommand{\nOracleLeaveRecC}{1.000\xspace}
\newcommand{\nOracleLeaveRecD}{1.000\xspace}
\newcommand{\nOracleLeaveRecE}{1.000\xspace}
\newcommand{\nOracleLeaveRecF}{1.000\xspace}
\newcommand{\nOracleLeaveRecR}{1.000\xspace}
\newcommand{\nOracleLeaveTPA}{26,520\xspace}
\newcommand{\nOracleLeaveTPB}{20,557\xspace}
\newcommand{\nOracleLeaveTPC}{28,282\xspace}
\newcommand{\nOracleLeaveTPD}{56,903\xspace}
\newcommand{\nOracleLeaveTPE}{28,133\xspace}
\newcommand{\nOracleLeaveTPF}{578,104\xspace}
\newcommand{\nOracleLeaveTPR}{110,264\xspace}
\newcommand{\nOraclePrecAll}{1.00\xspace}
\newcommand{\nOraclePrecAuditA}{1.000\xspace}
\newcommand{\nOraclePrecAuditB}{1.000\xspace}
\newcommand{\nOraclePrecAuditC}{1.000\xspace}
\newcommand{\nOraclePrecAuditD}{1.000\xspace}
\newcommand{\nOraclePrecAuditE}{1.000\xspace}
\newcommand{\nOraclePrecAuditF}{1.000\xspace}
\newcommand{\nOraclePrecAuditR}{1.000\xspace}
\newcommand{\nOracleRecAuditA}{1.000\xspace}
\newcommand{\nOracleRecAuditB}{1.000\xspace}
\newcommand{\nOracleRecAuditC}{1.000\xspace}
\newcommand{\nOracleRecAuditD}{1.000\xspace}
\newcommand{\nOracleRecAuditE}{1.000\xspace}
\newcommand{\nOracleRecAuditF}{1.000\xspace}
\newcommand{\nOracleRecAuditR}{1.000\xspace}
\newcommand{\nOracleTPA}{26,520\xspace}
\newcommand{\nOracleTPB}{20,557\xspace}
\newcommand{\nOracleTPC}{28,282\xspace}
\newcommand{\nOracleTPD}{56,903\xspace}
\newcommand{\nOracleTPE}{28,133\xspace}
\newcommand{\nOracleTPF}{578,104\xspace}
\newcommand{\nOracleTPR}{105,653\xspace}
\newcommand{\nOverlapFiledConverged}{0.0034\xspace}
\newcommand{\nOverlapFiledEnd}{0.307\xspace}
\newcommand{\nOverlapTestbedConverged}{$3.7\times 10^{-5}$\xspace}
\newcommand{\nOverlapTestbedEnd}{0.198\xspace}
\newcommand{\nOverlapThreshold}{0.9\xspace}
\newcommand{\nOvershootBoundMedianF}{1.28\xspace}
\newcommand{\nOvershootBoundMedianPublic}{1.21\xspace}
\newcommand{\nOvershootCrossingWrites}{345,178\xspace}
\newcommand{\nOvershootMeasuredMedianF}{2.53\xspace}
\newcommand{\nOvershootMeasuredMedianPublic}{2.61\xspace}
\newcommand{\nOvershootPartPositiveF}{0.70\xspace}
\newcommand{\nOvershootPartPositivePublic}{0.55\xspace}
\newcommand{\nPackTrainDocs}{144,467\xspace}
\newcommand{\nPackTrainSkipped}{83\xspace}
\newcommand{\nPackTrainSkippedTokens}{335\xspace}
\newcommand{\nPackValDocs}{1,666\xspace}
\newcommand{\nPackValSkipped}{2\xspace}
\newcommand{\nPackValSkippedTokens}{4\xspace}
\newcommand{\nPairsBeatingNull}{30\xspace}
\newcommand{\nPairsBetween}{30\xspace}
\newcommand{\nPartnerActiveDec}{0.84\xspace}
\newcommand{\nPartnerActiveDecNull}{0.08\xspace}
\newcommand{\nPartnerCosine}{0.9\xspace}
\newcommand{\nPathFiveRerankMaxF}{0.0043\xspace}
\newcommand{\nPathFiveRerankMaxRel}{0.124\xspace}
\newcommand{\nPathFiveRerankSizeEightF}{0.0012\xspace}
\newcommand{\nPathFiveRerankSizeEightRel}{0.0034\xspace}
\newcommand{\nPathFiveRerankSizeFourF}{0.0005\xspace}
\newcommand{\nPathFiveRerankSizeFourRel}{0.0011\xspace}
\newcommand{\nPathFiveRerankSizeSixteenF}{0.0043\xspace}
\newcommand{\nPathFiveRerankSizeSixteenRel}{0.1237\xspace}
\newcommand{\nPathFiveRerankSizeTwoF}{0.0002\xspace}
\newcommand{\nPathFiveRerankSizeTwoRel}{0.0004\xspace}
\newcommand{\nPathProducts}{5\xspace}
\newcommand{\nPathTenProducts}{10\xspace}
\newcommand{\nPathTenRerankSizeEightF}{0.0003\xspace}
\newcommand{\nPathTenRerankSizeEightRel}{0.0006\xspace}
\newcommand{\nPathTenRerankSizeFourF}{0.0001\xspace}
\newcommand{\nPathTenRerankSizeFourRel}{0.0003\xspace}
\newcommand{\nPathTenRerankSizeSixteenF}{0.0007\xspace}
\newcommand{\nPathTenRerankSizeSixteenRel}{0.0191\xspace}
\newcommand{\nPathTenRerankSizeTwoF}{$4.2\times 10^{-5}$\xspace}
\newcommand{\nPathTenRerankSizeTwoRel}{0.0001\xspace}
\newcommand{\nPositionLast}{2,047\xspace}
\newcommand{\nPositionSeed}{20260921\xspace}
\newcommand{\nPrecA}{0.82\xspace}
\newcommand{\nPrecB}{0.86\xspace}
\newcommand{\nPrecC}{0.86\xspace}
\newcommand{\nPrecD}{0.90\xspace}
\newcommand{\nPrecE}{0.80\xspace}
\newcommand{\nPrecEdgeAloneA}{0.026\xspace}
\newcommand{\nPrecEdgeAloneF}{0.027\xspace}
\newcommand{\nPrecF}{0.87\xspace}
\newcommand{\nPrecMarginAloneA}{0.43\xspace}
\newcommand{\nPrecMarginAloneF}{0.44\xspace}
\newcommand{\nPrecMax}{0.90\xspace}
\newcommand{\nPrecMin}{0.80\xspace}
\newcommand{\nPrecNullA}{0.0000\xspace}
\newcommand{\nPrecNullF}{0.0002\xspace}
\newcommand{\nPrecR}{0.73\xspace}
\newcommand{\nPrecTrueCountA}{0.85\xspace}
\newcommand{\nPrecTrueCountF}{0.90\xspace}
\newcommand{\nPredictorDoses}{3\xspace}
\newcommand{\nRandomDrawsPerPair}{20\xspace}
\newcommand{\nRecA}{0.83\xspace}
\newcommand{\nRecB}{0.86\xspace}
\newcommand{\nRecC}{0.86\xspace}
\newcommand{\nRecD}{0.92\xspace}
\newcommand{\nRecE}{0.82\xspace}
\newcommand{\nRecF}{0.89\xspace}
\newcommand{\nRecMax}{0.92\xspace}
\newcommand{\nRecMin}{0.82\xspace}
\newcommand{\nRecR}{0.97\xspace}
\newcommand{\nRecoveredTenth}{0.73\xspace}
\newcommand{\nRecoveredTenthRetrain}{0.76\xspace}
\newcommand{\nRegBatch}{65,536\xspace}
\newcommand{\nRegEpochs}{2\xspace}
\newcommand{\nRegInitSeed}{20260921\xspace}
\newcommand{\nRegLr}{$1.0\times 10^{-3}$\xspace}
\newcommand{\nRegMarginMedFive}{0.016\xspace}
\newcommand{\nRegMarginMedFour}{0.017\xspace}
\newcommand{\nRegMarginMedOne}{0.017\xspace}
\newcommand{\nRegMarginMedSeven}{0.016\xspace}
\newcommand{\nRegMarginMedSix}{0.016\xspace}
\newcommand{\nRegMarginMedThree}{0.016\xspace}
\newcommand{\nRegMarginMedTwo}{0.017\xspace}
\newcommand{\nRegMarginMedZero}{0.016\xspace}
\newcommand{\nRegMarginQOneFive}{0.006\xspace}
\newcommand{\nRegMarginQOneFour}{0.006\xspace}
\newcommand{\nRegMarginQOneOne}{0.006\xspace}
\newcommand{\nRegMarginQOneSeven}{0.006\xspace}
\newcommand{\nRegMarginQOneSix}{0.006\xspace}
\newcommand{\nRegMarginQOneThree}{0.006\xspace}
\newcommand{\nRegMarginQOneTwo}{0.007\xspace}
\newcommand{\nRegMarginQOneZero}{0.006\xspace}
\newcommand{\nRegMarginQThreeFive}{0.054\xspace}
\newcommand{\nRegMarginQThreeFour}{0.056\xspace}
\newcommand{\nRegMarginQThreeOne}{0.057\xspace}
\newcommand{\nRegMarginQThreeSeven}{0.056\xspace}
\newcommand{\nRegMarginQThreeSix}{0.053\xspace}
\newcommand{\nRegMarginQThreeThree}{0.056\xspace}
\newcommand{\nRegMarginQThreeTwo}{0.057\xspace}
\newcommand{\nRegMarginQThreeZero}{0.054\xspace}
\newcommand{\nRegMarginScoringMean}{0.016\xspace}
\newcommand{\nRegRows}{67,108,864\xspace}
\newcommand{\nRegScoreTokens}{256\xspace}
\newcommand{\nRegSeedBase}{20260921\xspace}
\newcommand{\nRegSixFive}{2.38\xspace}
\newcommand{\nRegSixFour}{2.42\xspace}
\newcommand{\nRegSixMax}{2.45\xspace}
\newcommand{\nRegSixMin}{2.23\xspace}
\newcommand{\nRegSixOne}{2.45\xspace}
\newcommand{\nRegSixSeven}{2.35\xspace}
\newcommand{\nRegSixSix}{2.23\xspace}
\newcommand{\nRegSixThree}{2.36\xspace}
\newcommand{\nRegSixTwo}{2.44\xspace}
\newcommand{\nRegSixZero}{2.37\xspace}
\newcommand{\nRegSteps}{2,048\xspace}
\newcommand{\nRegTrain}{4,096\xspace}
\newcommand{\nRegTrainTokens}{256\xspace}
\newcommand{\nRegTwoFive}{1.48\xspace}
\newcommand{\nRegTwoFour}{1.49\xspace}
\newcommand{\nRegTwoMax}{1.53\xspace}
\newcommand{\nRegTwoMin}{1.44\xspace}
\newcommand{\nRegTwoOne}{1.53\xspace}
\newcommand{\nRegTwoSeven}{1.51\xspace}
\newcommand{\nRegTwoSix}{1.44\xspace}
\newcommand{\nRegTwoThree}{1.45\xspace}
\newcommand{\nRegTwoTwo}{1.47\xspace}
\newcommand{\nRegTwoZero}{1.47\xspace}
\newcommand{\nRegWidth}{64\xspace}
\newcommand{\nRegimeLevel}{0.3\xspace}
\newcommand{\nRegressorRunsToFire}{5\xspace}
\newcommand{\nRegressorRunsWithin}{0\xspace}
\newcommand{\nReleaseA}{EleutherAI/sae-pythia-70m-32k\xspace}
\newcommand{\nReleaseD}{EleutherAI/sae-pythia-160m-32k\xspace}
\newcommand{\nReleaseE}{EleutherAI/sae-pythia-70m-deduped-32k\xspace}
\newcommand{\nReleaseR}{saprmarks/pythia-70m-deduped-saes\xspace}
\newcommand{\nReluEnergyBound}{0.5\xspace}
\newcommand{\nReluRatioHigh}{1.25\xspace}
\newcommand{\nReluRatioLow}{0.50\xspace}
\newcommand{\nReluShareFloor}{0.5\xspace}
\newcommand{\nRemainderSizesMin}{1.00\xspace}
\newcommand{\nRemedyFactor}{0.75\xspace}
\newcommand{\nRepresentedDec}{0.71\xspace}
\newcommand{\nRepresentedEnc}{0.24\xspace}
\newcommand{\nRestricted}{0.85\xspace}
\newcommand{\nRestrictedRetrain}{0.85\xspace}
\newcommand{\nRotationSeedDec}{3,000\xspace}
\newcommand{\nRotationSeedEnc}{4,000\xspace}
\newcommand{\nRowsTwoHop}{24,576\xspace}
\newcommand{\nRuleBoundASingle}{3.98\xspace}
\newcommand{\nRuleBoundASubset}{4.60\xspace}
\newcommand{\nRuleBoundBSingle}{5.14\xspace}
\newcommand{\nRuleBoundCSingle}{3.22\xspace}
\newcommand{\nRuleBoundDSingle}{9.60\xspace}
\newcommand{\nRuleBoundESingle}{3.68\xspace}
\newcommand{\nRuleBoundFFiveSingle}{5.77\xspace}
\newcommand{\nRuleBoundFFiveSubset}{7.51\xspace}
\newcommand{\nRuleBoundFFourSingle}{5.84\xspace}
\newcommand{\nRuleBoundFFourSubset}{7.63\xspace}
\newcommand{\nRuleBoundFOneSingle}{5.59\xspace}
\newcommand{\nRuleBoundFOneSubset}{7.46\xspace}
\newcommand{\nRuleBoundFSevenSingle}{5.70\xspace}
\newcommand{\nRuleBoundFSevenSubset}{7.52\xspace}
\newcommand{\nRuleBoundFSixSingle}{5.39\xspace}
\newcommand{\nRuleBoundFSixSubset}{7.50\xspace}
\newcommand{\nRuleBoundFThreeSingle}{5.57\xspace}
\newcommand{\nRuleBoundFThreeSubset}{7.65\xspace}
\newcommand{\nRuleBoundFTwoSingle}{5.75\xspace}
\newcommand{\nRuleBoundFTwoSubset}{7.64\xspace}
\newcommand{\nRuleBoundFZeroSingle}{5.56\xspace}
\newcommand{\nRuleBoundFZeroSubset}{7.73\xspace}
\newcommand{\nRuleCovASingle}{0.248\xspace}
\newcommand{\nRuleCovASubset}{0.012\xspace}
\newcommand{\nRuleCovBSingle}{0.710\xspace}
\newcommand{\nRuleCovCSingle}{0.321\xspace}
\newcommand{\nRuleCovDSingle}{0.706\xspace}
\newcommand{\nRuleCovESingle}{0.211\xspace}
\newcommand{\nRuleCovFFiveSingle}{0.524\xspace}
\newcommand{\nRuleCovFFiveSubset}{0.058\xspace}
\newcommand{\nRuleCovFFourSingle}{0.518\xspace}
\newcommand{\nRuleCovFFourSubset}{0.063\xspace}
\newcommand{\nRuleCovFOneSingle}{0.503\xspace}
\newcommand{\nRuleCovFOneSubset}{0.066\xspace}
\newcommand{\nRuleCovFSevenSingle}{0.519\xspace}
\newcommand{\nRuleCovFSevenSubset}{0.062\xspace}
\newcommand{\nRuleCovFSixSingle}{0.520\xspace}
\newcommand{\nRuleCovFSixSubset}{0.065\xspace}
\newcommand{\nRuleCovFThreeSingle}{0.541\xspace}
\newcommand{\nRuleCovFThreeSubset}{0.061\xspace}
\newcommand{\nRuleCovFTwoSingle}{0.525\xspace}
\newcommand{\nRuleCovFTwoSubset}{0.067\xspace}
\newcommand{\nRuleCovFZeroSingle}{0.512\xspace}
\newcommand{\nRuleCovFZeroSubset}{0.060\xspace}
\newcommand{\nRuleCovSubsetMax}{0.067\xspace}
\newcommand{\nRuleCovSubsetMin}{0.012\xspace}
\newcommand{\nRuleDoseCovAHalf}{0.613\xspace}
\newcommand{\nRuleDoseCovANine}{0.982\xspace}
\newcommand{\nRuleDoseCovAPooled}{0.614\xspace}
\newcommand{\nRuleDoseCovAZero}{0.248\xspace}
\newcommand{\nRuleDoseCovBHalf}{0.889\xspace}
\newcommand{\nRuleDoseCovBNine}{1.000\xspace}
\newcommand{\nRuleDoseCovBPooled}{0.866\xspace}
\newcommand{\nRuleDoseCovBZero}{0.710\xspace}
\newcommand{\nRuleDoseCovCHalf}{0.627\xspace}
\newcommand{\nRuleDoseCovCNine}{0.972\xspace}
\newcommand{\nRuleDoseCovCPooled}{0.640\xspace}
\newcommand{\nRuleDoseCovCZero}{0.321\xspace}
\newcommand{\nRuleDoseCovDHalf}{0.858\xspace}
\newcommand{\nRuleDoseCovDNine}{0.989\xspace}
\newcommand{\nRuleDoseCovDPooled}{0.851\xspace}
\newcommand{\nRuleDoseCovDZero}{0.706\xspace}
\newcommand{\nRuleDoseCovEHalf}{0.579\xspace}
\newcommand{\nRuleDoseCovENine}{0.984\xspace}
\newcommand{\nRuleDoseCovEPooled}{0.591\xspace}
\newcommand{\nRuleDoseCovEZero}{0.211\xspace}
\newcommand{\nRuleDoseCovFFiveHalf}{0.805\xspace}
\newcommand{\nRuleDoseCovFFiveNine}{0.995\xspace}
\newcommand{\nRuleDoseCovFFivePooled}{0.775\xspace}
\newcommand{\nRuleDoseCovFFiveZero}{0.524\xspace}
\newcommand{\nRuleDoseCovFFourHalf}{0.789\xspace}
\newcommand{\nRuleDoseCovFFourNine}{0.994\xspace}
\newcommand{\nRuleDoseCovFFourPooled}{0.767\xspace}
\newcommand{\nRuleDoseCovFFourZero}{0.518\xspace}
\newcommand{\nRuleDoseCovFOneHalf}{0.785\xspace}
\newcommand{\nRuleDoseCovFOneNine}{0.991\xspace}
\newcommand{\nRuleDoseCovFOnePooled}{0.760\xspace}
\newcommand{\nRuleDoseCovFOneZero}{0.503\xspace}
\newcommand{\nRuleDoseCovFSevenHalf}{0.793\xspace}
\newcommand{\nRuleDoseCovFSevenNine}{0.990\xspace}
\newcommand{\nRuleDoseCovFSevenPooled}{0.767\xspace}
\newcommand{\nRuleDoseCovFSevenZero}{0.519\xspace}
\newcommand{\nRuleDoseCovFSixHalf}{0.808\xspace}
\newcommand{\nRuleDoseCovFSixNine}{0.993\xspace}
\newcommand{\nRuleDoseCovFSixPooled}{0.774\xspace}
\newcommand{\nRuleDoseCovFSixZero}{0.520\xspace}
\newcommand{\nRuleDoseCovFThreeHalf}{0.793\xspace}
\newcommand{\nRuleDoseCovFThreeNine}{0.991\xspace}
\newcommand{\nRuleDoseCovFThreePooled}{0.775\xspace}
\newcommand{\nRuleDoseCovFThreeZero}{0.541\xspace}
\newcommand{\nRuleDoseCovFTwoHalf}{0.796\xspace}
\newcommand{\nRuleDoseCovFTwoNine}{0.995\xspace}
\newcommand{\nRuleDoseCovFTwoPooled}{0.772\xspace}
\newcommand{\nRuleDoseCovFTwoZero}{0.525\xspace}
\newcommand{\nRuleDoseCovFZeroHalf}{0.787\xspace}
\newcommand{\nRuleDoseCovFZeroNine}{0.995\xspace}
\newcommand{\nRuleDoseCovFZeroPooled}{0.765\xspace}
\newcommand{\nRuleDoseCovFZeroZero}{0.512\xspace}
\newcommand{\nRuleDoseInsideAPooled}{2,870\xspace}
\newcommand{\nRuleDoseInsideBPooled}{3,031\xspace}
\newcommand{\nRuleDoseInsideCPooled}{2,935\xspace}
\newcommand{\nRuleDoseInsideDPooled}{3,052\xspace}
\newcommand{\nRuleDoseInsideEPooled}{2,861\xspace}
\newcommand{\nRuleDoseInsideFFivePooled}{2,988\xspace}
\newcommand{\nRuleDoseInsideFFourPooled}{2,971\xspace}
\newcommand{\nRuleDoseInsideFOnePooled}{2,978\xspace}
\newcommand{\nRuleDoseInsideFSevenPooled}{2,968\xspace}
\newcommand{\nRuleDoseInsideFSixPooled}{2,983\xspace}
\newcommand{\nRuleDoseInsideFThreePooled}{2,961\xspace}
\newcommand{\nRuleDoseInsideFTwoPooled}{2,991\xspace}
\newcommand{\nRuleDoseInsideFZeroPooled}{2,996\xspace}
\newcommand{\nRuleDoseLiveA}{0.350\xspace}
\newcommand{\nRuleDoseLiveB}{0.537\xspace}
\newcommand{\nRuleDoseLiveC}{0.342\xspace}
\newcommand{\nRuleDoseLiveD}{0.575\xspace}
\newcommand{\nRuleDoseLiveE}{0.321\xspace}
\newcommand{\nRuleDoseLiveFFive}{0.477\xspace}
\newcommand{\nRuleDoseLiveFFour}{0.486\xspace}
\newcommand{\nRuleDoseLiveFOne}{0.469\xspace}
\newcommand{\nRuleDoseLiveFSeven}{0.474\xspace}
\newcommand{\nRuleDoseLiveFSix}{0.486\xspace}
\newcommand{\nRuleDoseLiveFThree}{0.471\xspace}
\newcommand{\nRuleDoseLiveFTwo}{0.487\xspace}
\newcommand{\nRuleDoseLiveFZero}{0.481\xspace}
\newcommand{\nRuleDoseNineCovMax}{1.00\xspace}
\newcommand{\nRuleDoseNineCovMin}{0.97\xspace}
\newcommand{\nRuleDosePooledAccepted}{668,393,327\xspace}
\newcommand{\nRuleDosePooledCovMax}{0.87\xspace}
\newcommand{\nRuleDosePooledCovMin}{0.59\xspace}
\newcommand{\nRuleDosePooledLiveMax}{0.57\xspace}
\newcommand{\nRuleDosePooledLiveMin}{0.32\xspace}
\newcommand{\nRuleDoseWrongAPooled}{0\xspace}
\newcommand{\nRuleDoseWrongBPooled}{0\xspace}
\newcommand{\nRuleDoseWrongCPooled}{0\xspace}
\newcommand{\nRuleDoseWrongDPooled}{0\xspace}
\newcommand{\nRuleDoseWrongEPooled}{0\xspace}
\newcommand{\nRuleDoseWrongFFivePooled}{0\xspace}
\newcommand{\nRuleDoseWrongFFourPooled}{0\xspace}
\newcommand{\nRuleDoseWrongFOnePooled}{0\xspace}
\newcommand{\nRuleDoseWrongFSevenPooled}{0\xspace}
\newcommand{\nRuleDoseWrongFSixPooled}{0\xspace}
\newcommand{\nRuleDoseWrongFThreePooled}{0\xspace}
\newcommand{\nRuleDoseWrongFTwoPooled}{0\xspace}
\newcommand{\nRuleDoseWrongFZeroPooled}{0\xspace}
\newcommand{\nRuleDoseZeroCovMax}{0.71\xspace}
\newcommand{\nRuleDoseZeroCovMin}{0.21\xspace}
\newcommand{\nRuleErrASingle}{0.064\xspace}
\newcommand{\nRuleErrASubset}{0.221\xspace}
\newcommand{\nRuleErrBSingle}{0.042\xspace}
\newcommand{\nRuleErrCSingle}{0.054\xspace}
\newcommand{\nRuleErrDSingle}{0.026\xspace}
\newcommand{\nRuleErrESingle}{0.064\xspace}
\newcommand{\nRuleErrFFiveSingle}{0.038\xspace}
\newcommand{\nRuleErrFFiveSubset}{0.182\xspace}
\newcommand{\nRuleErrFFourSingle}{0.038\xspace}
\newcommand{\nRuleErrFFourSubset}{0.186\xspace}
\newcommand{\nRuleErrFOneSingle}{0.038\xspace}
\newcommand{\nRuleErrFOneSubset}{0.185\xspace}
\newcommand{\nRuleErrFSevenSingle}{0.038\xspace}
\newcommand{\nRuleErrFSevenSubset}{0.186\xspace}
\newcommand{\nRuleErrFSixSingle}{0.038\xspace}
\newcommand{\nRuleErrFSixSubset}{0.183\xspace}
\newcommand{\nRuleErrFThreeSingle}{0.038\xspace}
\newcommand{\nRuleErrFThreeSubset}{0.183\xspace}
\newcommand{\nRuleErrFTwoSingle}{0.039\xspace}
\newcommand{\nRuleErrFTwoSubset}{0.187\xspace}
\newcommand{\nRuleErrFZeroSingle}{0.039\xspace}
\newcommand{\nRuleErrFZeroSubset}{0.189\xspace}
\newcommand{\nRuleInsideASingle}{852\xspace}
\newcommand{\nRuleInsideASubset}{555\xspace}
\newcommand{\nRuleInsideBSingle}{983\xspace}
\newcommand{\nRuleInsideCSingle}{908\xspace}
\newcommand{\nRuleInsideDSingle}{1,005\xspace}
\newcommand{\nRuleInsideESingle}{836\xspace}
\newcommand{\nRuleInsideFFiveSingle}{952\xspace}
\newcommand{\nRuleInsideFFiveSubset}{1,086\xspace}
\newcommand{\nRuleInsideFFourSingle}{938\xspace}
\newcommand{\nRuleInsideFFourSubset}{1,135\xspace}
\newcommand{\nRuleInsideFOneSingle}{947\xspace}
\newcommand{\nRuleInsideFOneSubset}{1,083\xspace}
\newcommand{\nRuleInsideFSevenSingle}{935\xspace}
\newcommand{\nRuleInsideFSevenSubset}{1,099\xspace}
\newcommand{\nRuleInsideFSixSingle}{945\xspace}
\newcommand{\nRuleInsideFSixSubset}{1,110\xspace}
\newcommand{\nRuleInsideFThreeSingle}{933\xspace}
\newcommand{\nRuleInsideFThreeSubset}{1,102\xspace}
\newcommand{\nRuleInsideFTwoSingle}{956\xspace}
\newcommand{\nRuleInsideFTwoSubset}{1,136\xspace}
\newcommand{\nRuleInsideFZeroSingle}{957\xspace}
\newcommand{\nRuleInsideFZeroSubset}{1,131\xspace}
\newcommand{\nRuleLevel}{0.3\xspace}
\newcommand{\nRuleLiveASingle}{0.068\xspace}
\newcommand{\nRuleLiveASubset}{0.004\xspace}
\newcommand{\nRuleLiveBSingle}{0.242\xspace}
\newcommand{\nRuleLiveCSingle}{0.076\xspace}
\newcommand{\nRuleLiveDSingle}{0.287\xspace}
\newcommand{\nRuleLiveESingle}{0.055\xspace}
\newcommand{\nRuleLiveFFiveSingle}{0.174\xspace}
\newcommand{\nRuleLiveFFiveSubset}{0.018\xspace}
\newcommand{\nRuleLiveFFourSingle}{0.187\xspace}
\newcommand{\nRuleLiveFFourSubset}{0.022\xspace}
\newcommand{\nRuleLiveFOneSingle}{0.170\xspace}
\newcommand{\nRuleLiveFOneSubset}{0.020\xspace}
\newcommand{\nRuleLiveFSevenSingle}{0.174\xspace}
\newcommand{\nRuleLiveFSevenSubset}{0.019\xspace}
\newcommand{\nRuleLiveFSixSingle}{0.181\xspace}
\newcommand{\nRuleLiveFSixSubset}{0.020\xspace}
\newcommand{\nRuleLiveFThreeSingle}{0.173\xspace}
\newcommand{\nRuleLiveFThreeSubset}{0.018\xspace}
\newcommand{\nRuleLiveFTwoSingle}{0.186\xspace}
\newcommand{\nRuleLiveFTwoSubset}{0.023\xspace}
\newcommand{\nRuleLiveFZeroSingle}{0.181\xspace}
\newcommand{\nRuleLiveFZeroSubset}{0.020\xspace}
\newcommand{\nRuleScoring}{256\xspace}
\newcommand{\nRuleWritesASingle}{1,024\xspace}
\newcommand{\nRuleWritesASubset}{2,048\xspace}
\newcommand{\nRuleWritesBSingle}{1,024\xspace}
\newcommand{\nRuleWritesCSingle}{1,024\xspace}
\newcommand{\nRuleWritesDSingle}{1,024\xspace}
\newcommand{\nRuleWritesESingle}{1,024\xspace}
\newcommand{\nRuleWritesFFiveSingle}{1,024\xspace}
\newcommand{\nRuleWritesFFiveSubset}{2,048\xspace}
\newcommand{\nRuleWritesFFourSingle}{1,024\xspace}
\newcommand{\nRuleWritesFFourSubset}{2,048\xspace}
\newcommand{\nRuleWritesFOneSingle}{1,024\xspace}
\newcommand{\nRuleWritesFOneSubset}{2,048\xspace}
\newcommand{\nRuleWritesFSevenSingle}{1,024\xspace}
\newcommand{\nRuleWritesFSevenSubset}{2,048\xspace}
\newcommand{\nRuleWritesFSixSingle}{1,024\xspace}
\newcommand{\nRuleWritesFSixSubset}{2,048\xspace}
\newcommand{\nRuleWritesFThreeSingle}{1,024\xspace}
\newcommand{\nRuleWritesFThreeSubset}{2,048\xspace}
\newcommand{\nRuleWritesFTwoSingle}{1,024\xspace}
\newcommand{\nRuleWritesFTwoSubset}{2,048\xspace}
\newcommand{\nRuleWritesFZeroSingle}{1,024\xspace}
\newcommand{\nRuleWritesFZeroSubset}{2,048\xspace}
\newcommand{\nRuleWrong}{0\xspace}
\newcommand{\nRuleWrongASingle}{0\xspace}
\newcommand{\nRuleWrongASubset}{0\xspace}
\newcommand{\nRuleWrongBSingle}{0\xspace}
\newcommand{\nRuleWrongCSingle}{0\xspace}
\newcommand{\nRuleWrongDSingle}{0\xspace}
\newcommand{\nRuleWrongESingle}{0\xspace}
\newcommand{\nRuleWrongFFiveSingle}{0\xspace}
\newcommand{\nRuleWrongFFiveSubset}{0\xspace}
\newcommand{\nRuleWrongFFourSingle}{0\xspace}
\newcommand{\nRuleWrongFFourSubset}{0\xspace}
\newcommand{\nRuleWrongFOneSingle}{0\xspace}
\newcommand{\nRuleWrongFOneSubset}{0\xspace}
\newcommand{\nRuleWrongFSevenSingle}{0\xspace}
\newcommand{\nRuleWrongFSevenSubset}{0\xspace}
\newcommand{\nRuleWrongFSixSingle}{0\xspace}
\newcommand{\nRuleWrongFSixSubset}{0\xspace}
\newcommand{\nRuleWrongFThreeSingle}{0\xspace}
\newcommand{\nRuleWrongFThreeSubset}{0\xspace}
\newcommand{\nRuleWrongFTwoSingle}{0\xspace}
\newcommand{\nRuleWrongFTwoSubset}{0\xspace}
\newcommand{\nRuleWrongFZeroSingle}{0\xspace}
\newcommand{\nRuleWrongFZeroSubset}{0\xspace}
\newcommand{\nRulesFiredOther}{5\xspace}
\newcommand{\nRulesFiredRelu}{2\xspace}
\newcommand{\nRunIndexFive}{5\xspace}
\newcommand{\nRunIndexFour}{4\xspace}
\newcommand{\nRunIndexOne}{1\xspace}
\newcommand{\nRunIndexSeven}{7\xspace}
\newcommand{\nRunIndexSix}{6\xspace}
\newcommand{\nRunIndexThree}{3\xspace}
\newcommand{\nRunIndexTwo}{2\xspace}
\newcommand{\nRunIndexZero}{0\xspace}
\newcommand{\nRunsF}{8\xspace}
\newcommand{\nRunsTogether}{8\xspace}
\newcommand{\nScaleDownF}{13.093\xspace}
\newcommand{\nScaleUpF}{12.424\xspace}
\newcommand{\nSeqLen}{2,048\xspace}
\newcommand{\nShuffleLayerMult}{10,007\xspace}
\newcommand{\nShuffleRunMult}{1,000,003\xspace}
\newcommand{\nSizeEight}{8\xspace}
\newcommand{\nSizeFour}{4\xspace}
\newcommand{\nSizeSixteen}{16\xspace}
\newcommand{\nSizeTwo}{2\xspace}
\newcommand{\nSmPairsBelow}{5\xspace}
\newcommand{\nSmRerankPubMax}{0.026\xspace}
\newcommand{\nSmRerankPubMin}{0.011\xspace}
\newcommand{\nSmallBlockDistinct}{12,288\xspace}
\newcommand{\nSmallBlockPresentations}{32,768,000\xspace}
\newcommand{\nSparseGain}{0.35\xspace}
\newcommand{\nSparseRetrain}{0.64\xspace}
\newcommand{\nSparseUnmasked}{0.57\xspace}
\newcommand{\nSparsityPenaltyR}{0.1\xspace}
\newcommand{\nSplitHalf}{256\xspace}
\newcommand{\nSplitSeed}{20260925\xspace}
\newcommand{\nStepsR}{120,000\xspace}
\newcommand{\nSubsetSeed}{20260921\xspace}
\newcommand{\nSubsetsPerSize}{16\xspace}
\newcommand{\nSupportMargin}{0.1\xspace}
\newcommand{\nSwapRatioF}{1.56\xspace}
\newcommand{\nSwapRatioFloor}{0.5\xspace}
\newcommand{\nSwapRatioMax}{1.72\xspace}
\newcommand{\nSwapRatioMin}{1.25\xspace}
\newcommand{\nSwapRatioR}{0.045\xspace}
\newcommand{\nSwapShareCeiling}{0.5\xspace}
\newcommand{\nSwapShareF}{0.24\xspace}
\newcommand{\nSwapShareMax}{0.29\xspace}
\newcommand{\nSwapShareMin}{0.21\xspace}
\newcommand{\nTcFNA}{4,027\xspace}
\newcommand{\nTcFNB}{2,418\xspace}
\newcommand{\nTcFNC}{3,283\xspace}
\newcommand{\nTcFND}{4,110\xspace}
\newcommand{\nTcFNE}{4,555\xspace}
\newcommand{\nTcFNF}{59,922\xspace}
\newcommand{\nTcFNR}{5,272\xspace}
\newcommand{\nTcFPA}{4,027\xspace}
\newcommand{\nTcFPB}{2,418\xspace}
\newcommand{\nTcFPC}{3,283\xspace}
\newcommand{\nTcFPD}{4,110\xspace}
\newcommand{\nTcFPE}{4,555\xspace}
\newcommand{\nTcFPF}{59,922\xspace}
\newcommand{\nTcFPR}{5,272\xspace}
\newcommand{\nTcLeaveFNA}{1,429\xspace}
\newcommand{\nTcLeaveFNB}{1,182\xspace}
\newcommand{\nTcLeaveFNC}{1,460\xspace}
\newcommand{\nTcLeaveFND}{2,427\xspace}
\newcommand{\nTcLeaveFNE}{1,737\xspace}
\newcommand{\nTcLeaveFNF}{32,142\xspace}
\newcommand{\nTcLeaveFNR}{2,722\xspace}
\newcommand{\nTcLeaveFPA}{1,429\xspace}
\newcommand{\nTcLeaveFPB}{1,182\xspace}
\newcommand{\nTcLeaveFPC}{1,460\xspace}
\newcommand{\nTcLeaveFPD}{2,427\xspace}
\newcommand{\nTcLeaveFPE}{1,737\xspace}
\newcommand{\nTcLeaveFPF}{32,142\xspace}
\newcommand{\nTcLeaveFPR}{2,722\xspace}
\newcommand{\nTcLeavePrecA}{0.946\xspace}
\newcommand{\nTcLeavePrecB}{0.943\xspace}
\newcommand{\nTcLeavePrecC}{0.948\xspace}
\newcommand{\nTcLeavePrecD}{0.957\xspace}
\newcommand{\nTcLeavePrecE}{0.938\xspace}
\newcommand{\nTcLeavePrecF}{0.944\xspace}
\newcommand{\nTcLeavePrecR}{0.975\xspace}
\newcommand{\nTcLeaveRecA}{0.946\xspace}
\newcommand{\nTcLeaveRecB}{0.943\xspace}
\newcommand{\nTcLeaveRecC}{0.948\xspace}
\newcommand{\nTcLeaveRecD}{0.957\xspace}
\newcommand{\nTcLeaveRecE}{0.938\xspace}
\newcommand{\nTcLeaveRecF}{0.944\xspace}
\newcommand{\nTcLeaveRecR}{0.975\xspace}
\newcommand{\nTcLeaveTPA}{25,091\xspace}
\newcommand{\nTcLeaveTPB}{19,375\xspace}
\newcommand{\nTcLeaveTPC}{26,822\xspace}
\newcommand{\nTcLeaveTPD}{54,476\xspace}
\newcommand{\nTcLeaveTPE}{26,396\xspace}
\newcommand{\nTcLeaveTPF}{545,962\xspace}
\newcommand{\nTcLeaveTPR}{107,542\xspace}
\newcommand{\nTcPrecAuditA}{0.848\xspace}
\newcommand{\nTcPrecAuditB}{0.882\xspace}
\newcommand{\nTcPrecAuditC}{0.884\xspace}
\newcommand{\nTcPrecAuditD}{0.928\xspace}
\newcommand{\nTcPrecAuditE}{0.838\xspace}
\newcommand{\nTcPrecAuditF}{0.896\xspace}
\newcommand{\nTcPrecAuditR}{0.950\xspace}
\newcommand{\nTcRecAuditA}{0.848\xspace}
\newcommand{\nTcRecAuditB}{0.882\xspace}
\newcommand{\nTcRecAuditC}{0.884\xspace}
\newcommand{\nTcRecAuditD}{0.928\xspace}
\newcommand{\nTcRecAuditE}{0.838\xspace}
\newcommand{\nTcRecAuditF}{0.896\xspace}
\newcommand{\nTcRecAuditR}{0.950\xspace}
\newcommand{\nTcTPA}{22,493\xspace}
\newcommand{\nTcTPB}{18,139\xspace}
\newcommand{\nTcTPC}{24,999\xspace}
\newcommand{\nTcTPD}{52,793\xspace}
\newcommand{\nTcTPE}{23,578\xspace}
\newcommand{\nTcTPF}{518,182\xspace}
\newcommand{\nTcTPR}{100,381\xspace}
\newcommand{\nTermCross}{0.05\xspace}
\newcommand{\nTermCrossRetrain}{0.05\xspace}
\newcommand{\nTermMask}{0.34\xspace}
\newcommand{\nTermMaskRetrain}{0.29\xspace}
\newcommand{\nTermResid}{0.15\xspace}
\newcommand{\nTermResidRetrain}{0.15\xspace}
\newcommand{\nTheoryChecksMargin}{8\xspace}
\newcommand{\nTheoryChecksOvershoot}{6\xspace}
\newcommand{\nTheoryChecksReadout}{12\xspace}
\newcommand{\nTheoryChecksRestated}{12\xspace}
\newcommand{\nTheoryChecksTransport}{6\xspace}
\newcommand{\nTheoryD}{24\xspace}
\newcommand{\nTheoryK}{5\xspace}
\newcommand{\nTheoryM}{60\xspace}
\newcommand{\nTheorySeedMargin}{9200\xspace}
\newcommand{\nTheorySeedOvershoot}{9800\xspace}
\newcommand{\nTheorySeedReadout}{9500\xspace}
\newcommand{\nTheorySeedRestated}{9700\xspace}
\newcommand{\nTheorySeedTransport}{9000\xspace}
\newcommand{\nTheoryTol}{$1.0\times 10^{-9}$\xspace}
\newcommand{\nTjSequences}{1,280\xspace}
\newcommand{\nTokens}{512\xspace}
\newcommand{\nTolerance}{0.436\xspace}
\newcommand{\nToleranceCosine}{0.9\xspace}
\newcommand{\nTorchPin}{2.8.0\xspace}
\newcommand{\nTrFiveCrossTerm}{0.053\xspace}
\newcommand{\nTrFiveGap}{0.380\xspace}
\newcommand{\nTrFiveMaskTerm}{0.328\xspace}
\newcommand{\nTrFiveMatchMasked}{0.381\xspace}
\newcommand{\nTrFiveMatchUnmasked}{0.220\xspace}
\newcommand{\nTrFiveResid}{0.139\xspace}
\newcommand{\nTrFiveSparseMasked}{0.480\xspace}
\newcommand{\nTrFiveSparseUnmasked}{0.570\xspace}
\newcommand{\nTrFiveWholeTarget}{0.861\xspace}
\newcommand{\nTrFourCrossTerm}{0.048\xspace}
\newcommand{\nTrFourGap}{0.387\xspace}
\newcommand{\nTrFourMaskTerm}{0.339\xspace}
\newcommand{\nTrFourMatchMasked}{0.360\xspace}
\newcommand{\nTrFourMatchUnmasked}{0.221\xspace}
\newcommand{\nTrFourResid}{0.146\xspace}
\newcommand{\nTrFourSparseMasked}{0.467\xspace}
\newcommand{\nTrFourSparseUnmasked}{0.570\xspace}
\newcommand{\nTrFourWholeTarget}{0.854\xspace}
\newcommand{\nTrOneCrossTerm}{0.048\xspace}
\newcommand{\nTrOneGap}{0.384\xspace}
\newcommand{\nTrOneMaskTerm}{0.335\xspace}
\newcommand{\nTrOneMatchMasked}{0.371\xspace}
\newcommand{\nTrOneMatchUnmasked}{0.221\xspace}
\newcommand{\nTrOneResid}{0.151\xspace}
\newcommand{\nTrOneSparseMasked}{0.465\xspace}
\newcommand{\nTrOneSparseUnmasked}{0.569\xspace}
\newcommand{\nTrOneWholeTarget}{0.849\xspace}
\newcommand{\nTrPairsPerRun}{5\xspace}
\newcommand{\nTrRetrainSevenSixCrossTerm}{0.053\xspace}
\newcommand{\nTrRetrainSevenSixGap}{0.342\xspace}
\newcommand{\nTrRetrainSevenSixMaskTerm}{0.289\xspace}
\newcommand{\nTrRetrainSevenSixMatchMasked}{0.436\xspace}
\newcommand{\nTrRetrainSevenSixMatchUnmasked}{0.349\xspace}
\newcommand{\nTrRetrainSevenSixResid}{0.147\xspace}
\newcommand{\nTrRetrainSevenSixSparseMasked}{0.511\xspace}
\newcommand{\nTrRetrainSevenSixSparseUnmasked}{0.638\xspace}
\newcommand{\nTrRetrainSevenSixWholeTarget}{0.853\xspace}
\newcommand{\nTrRetrainSixSevenCrossTerm}{0.044\xspace}
\newcommand{\nTrRetrainSixSevenGap}{0.338\xspace}
\newcommand{\nTrRetrainSixSevenMaskTerm}{0.294\xspace}
\newcommand{\nTrRetrainSixSevenMatchMasked}{0.431\xspace}
\newcommand{\nTrRetrainSixSevenMatchUnmasked}{0.349\xspace}
\newcommand{\nTrRetrainSixSevenResid}{0.156\xspace}
\newcommand{\nTrRetrainSixSevenSparseMasked}{0.506\xspace}
\newcommand{\nTrRetrainSixSevenSparseUnmasked}{0.640\xspace}
\newcommand{\nTrRetrainSixSevenWholeTarget}{0.844\xspace}
\newcommand{\nTrThreeCrossTerm}{0.046\xspace}
\newcommand{\nTrThreeGap}{0.380\xspace}
\newcommand{\nTrThreeMaskTerm}{0.334\xspace}
\newcommand{\nTrThreeMatchMasked}{0.370\xspace}
\newcommand{\nTrThreeMatchUnmasked}{0.222\xspace}
\newcommand{\nTrThreeResid}{0.149\xspace}
\newcommand{\nTrThreeSparseMasked}{0.471\xspace}
\newcommand{\nTrThreeSparseUnmasked}{0.570\xspace}
\newcommand{\nTrThreeWholeTarget}{0.851\xspace}
\newcommand{\nTrTwoCrossTerm}{0.051\xspace}
\newcommand{\nTrTwoGap}{0.389\xspace}
\newcommand{\nTrTwoMaskTerm}{0.338\xspace}
\newcommand{\nTrTwoMatchMasked}{0.357\xspace}
\newcommand{\nTrTwoMatchUnmasked}{0.220\xspace}
\newcommand{\nTrTwoResid}{0.147\xspace}
\newcommand{\nTrTwoSparseMasked}{0.464\xspace}
\newcommand{\nTrTwoSparseUnmasked}{0.569\xspace}
\newcommand{\nTrTwoWholeTarget}{0.853\xspace}
\newcommand{\nTrZeroCrossTerm}{0.046\xspace}
\newcommand{\nTrZeroGap}{0.388\xspace}
\newcommand{\nTrZeroMaskTerm}{0.342\xspace}
\newcommand{\nTrZeroMatchMasked}{0.335\xspace}
\newcommand{\nTrZeroMatchUnmasked}{0.221\xspace}
\newcommand{\nTrZeroResid}{0.167\xspace}
\newcommand{\nTrZeroSparseMasked}{0.445\xspace}
\newcommand{\nTrZeroSparseUnmasked}{0.571\xspace}
\newcommand{\nTrZeroWholeTarget}{0.833\xspace}
\newcommand{\nWidthA}{32,768\xspace}
\newcommand{\nWidthB}{32,768\xspace}
\newcommand{\nWidthC}{32,768\xspace}
\newcommand{\nWidthD}{32,768\xspace}
\newcommand{\nWidthE}{32,768\xspace}
\newcommand{\nWidthF}{16,384\xspace}
\newcommand{\nWidthFiled}{2,048\xspace}
\newcommand{\nWidthR}{32,768\xspace}
\newcommand{\nWithinTestbedEnd}{0.300\xspace}
\newcommand{\nZeroResponseA}{0\xspace}
\newcommand{\nZeroResponseB}{0\xspace}
\newcommand{\nZeroResponseC}{0\xspace}
\newcommand{\nZeroResponseD}{0\xspace}
\newcommand{\nZeroResponseE}{0\xspace}
\newcommand{\nZeroResponseF}{0\xspace}
\newcommand{\nZeroResponseR}{0\xspace}
\newcommand{\nDsAbove}{4,893\xspace}
\newcommand{\nDsActive}{16\xspace}
\newcommand{\nDsAgreeShare}{0.93\xspace}
\newcommand{\nDsAttainShare}{0.01\xspace}
\newcommand{\nDsAttractors}{25\xspace}
\newcommand{\nDsCheckBound}{$1.0\times 10^{-6}$\xspace}
\newcommand{\nDsCheckUnits}{20\xspace}
\newcommand{\nDsContBound}{0.001\xspace}
\newcommand{\nDsContFirst}{0.0013\xspace}
\newcommand{\nDsContSecond}{$3.2\times 10^{-12}$\xspace}
\newcommand{\nDsDevPrompts}{100\xspace}
\newcommand{\nDsDraws}{2,000\xspace}
\newcommand{\nDsEntrantShare}{0.08\xspace}
\newcommand{\nDsErrControl}{0.185\xspace}
\newcommand{\nDsErrEntrants}{0.247\xspace}
\newcommand{\nDsErrFixed}{0.146\xspace}
\newcommand{\nDsErrFixedHalf}{0.113\xspace}
\newcommand{\nDsErrFixedNine}{0.054\xspace}
\newcommand{\nDsErrFixedZero}{0.195\xspace}
\newcommand{\nDsErrMargin}{0.285\xspace}
\newcommand{\nDsErrMarginHalf}{0.239\xspace}
\newcommand{\nDsErrMarginNine}{0.163\xspace}
\newcommand{\nDsErrMarginZero}{0.350\xspace}
\newcommand{\nDsErrNull}{0.237\xspace}
\newcommand{\nDsErrNullHalf}{0.190\xspace}
\newcommand{\nDsErrNullNine}{0.103\xspace}
\newcommand{\nDsErrNullZero}{0.308\xspace}
\newcommand{\nDsExactMean}{0.136\xspace}
\newcommand{\nDsExactMedian}{0.026\xspace}
\newcommand{\nDsExactTrue}{0.234\xspace}
\newcommand{\nDsExcess}{95\xspace}
\newcommand{\nDsFloor}{0.034\xspace}
\newcommand{\nDsIncludeShare}{0.70\xspace}
\newcommand{\nDsInductShare}{0.44\xspace}
\newcommand{\nDsIntervalLevel}{90\xspace}
\newcommand{\nDsMinUnits}{100\xspace}
\newcommand{\nDsNullLargest}{$3.2\times 10^{-12}$\xspace}
\newcommand{\nDsOneLargest}{$3.2\times 10^{-12}$\xspace}
\newcommand{\nDsPlantedLargest}{$2.3\times 10^{-13}$\xspace}
\newcommand{\nDsPoolSize}{13,269\xspace}
\newcommand{\nDsPrepositions}{10\xspace}
\newcommand{\nDsRatio}{1.95\xspace}
\newcommand{\nDsRatioBound}{0.80\xspace}
\newcommand{\nDsRatioHigh}{2.07\xspace}
\newcommand{\nDsRatioLow}{1.84\xspace}
\newcommand{\nDsReproBound}{$1.0\times 10^{-4}$\xspace}
\newcommand{\nDsReproGap}{$1.4\times 10^{-5}$\xspace}
\newcommand{\nDsReproMove}{$3.2\times 10^{-5}$\xspace}
\newcommand{\nDsReproRows}{20\xspace}
\newcommand{\nDsSeedDev}{20261108\xspace}
\newcommand{\nDsSeedTest}{20261107\xspace}
\newcommand{\nDsSequenceLength}{20\xspace}
\newcommand{\nDsShareSmaller}{0.22\xspace}
\newcommand{\nDsSubjects}{30\xspace}
\newcommand{\nDsTestPrompts}{200\xspace}
\newcommand{\nDsUnitsDev}{4,800\xspace}
\newcommand{\nDsUnitsHalf}{1,938\xspace}
\newcommand{\nDsUnitsNine}{566\xspace}
\newcommand{\nDsUnitsTest}{9,600\xspace}
\newcommand{\nDsUnitsZero}{2,389\xspace}
\newcommand{\nDsVerbPairs}{3\xspace}
\newcommand{\nCostProcessor}{Apple M4\xspace}
\newcommand{\nCostReleaseExactConstructionMaximumMs}{26.284\xspace}
\newcommand{\nCostReleaseExactConstructionMedianMs}{12.664\xspace}
\newcommand{\nCostReleaseExactConstructionMinimumMs}{6.148\xspace}
\newcommand{\nCostReleaseExactQueryMaximumUs}{1128.549\xspace}
\newcommand{\nCostReleaseExactQueryMedianUs}{335.514\xspace}
\newcommand{\nCostReleaseExactQueryMinimumUs}{265.872\xspace}
\newcommand{\nCostReleaseFixedConstructionMaximumMs}{52.290\xspace}
\newcommand{\nCostReleaseFixedConstructionMedianMs}{24.665\xspace}
\newcommand{\nCostReleaseFixedConstructionMinimumMs}{13.641\xspace}
\newcommand{\nCostReleaseFixedQueryMaximumUs}{0.576\xspace}
\newcommand{\nCostReleaseFixedQueryMedianUs}{0.161\xspace}
\newcommand{\nCostReleaseFixedQueryMinimumUs}{0.138\xspace}
\newcommand{\nCostReleaseIgFiveRerankConstructionMaximumMs}{31.016\xspace}
\newcommand{\nCostReleaseIgFiveRerankConstructionMedianMs}{12.709\xspace}
\newcommand{\nCostReleaseIgFiveRerankConstructionMinimumMs}{5.999\xspace}
\newcommand{\nCostReleaseIgFiveRerankQueryMaximumUs}{5305.643\xspace}
\newcommand{\nCostReleaseIgFiveRerankQueryMedianUs}{1277.181\xspace}
\newcommand{\nCostReleaseIgFiveRerankQueryMinimumUs}{786.326\xspace}
\newcommand{\nCostReleaseIgOneZeroRerankConstructionMaximumMs}{43.065\xspace}
\newcommand{\nCostReleaseIgOneZeroRerankConstructionMedianMs}{12.675\xspace}
\newcommand{\nCostReleaseIgOneZeroRerankConstructionMinimumMs}{6.146\xspace}
\newcommand{\nCostReleaseIgOneZeroRerankQueryMaximumUs}{6738.801\xspace}
\newcommand{\nCostReleaseIgOneZeroRerankQueryMedianUs}{2123.685\xspace}
\newcommand{\nCostReleaseIgOneZeroRerankQueryMinimumUs}{1192.417\xspace}
\newcommand{\nCostReleaseMarginConstructionMaximumMs}{60.791\xspace}
\newcommand{\nCostReleaseMarginConstructionMedianMs}{30.403\xspace}
\newcommand{\nCostReleaseMarginConstructionMinimumMs}{19.819\xspace}
\newcommand{\nCostReleaseMarginQueryMaximumUs}{122.154\xspace}
\newcommand{\nCostReleaseMarginQueryMedianUs}{50.303\xspace}
\newcommand{\nCostReleaseMarginQueryMinimumUs}{45.588\xspace}
\newcommand{\nCostReleaseSoftmaxFixedConstructionMaximumMs}{56.806\xspace}
\newcommand{\nCostReleaseSoftmaxFixedConstructionMedianMs}{25.747\xspace}
\newcommand{\nCostReleaseSoftmaxFixedConstructionMinimumMs}{11.784\xspace}
\newcommand{\nCostReleaseSoftmaxFixedQueryMaximumUs}{178.044\xspace}
\newcommand{\nCostReleaseSoftmaxFixedQueryMedianUs}{67.404\xspace}
\newcommand{\nCostReleaseSoftmaxFixedQueryMinimumUs}{17.688\xspace}
\newcommand{\nCostReleaseSoftmaxRerankConstructionMaximumMs}{57.682\xspace}
\newcommand{\nCostReleaseSoftmaxRerankConstructionMedianMs}{25.645\xspace}
\newcommand{\nCostReleaseSoftmaxRerankConstructionMinimumMs}{11.978\xspace}
\newcommand{\nCostReleaseSoftmaxRerankQueryMaximumUs}{698.449\xspace}
\newcommand{\nCostReleaseSoftmaxRerankQueryMedianUs}{299.785\xspace}
\newcommand{\nCostReleaseSoftmaxRerankQueryMinimumUs}{240.660\xspace}
\newcommand{\nCostRepetitions}{3\xspace}
\newcommand{\nCostTestbedExactConstructionMaximumMs}{37.723\xspace}
\newcommand{\nCostTestbedExactConstructionMedianMs}{14.897\xspace}
\newcommand{\nCostTestbedExactConstructionMinimumMs}{3.733\xspace}
\newcommand{\nCostTestbedExactQueryMaximumUs}{421.729\xspace}
\newcommand{\nCostTestbedExactQueryMedianUs}{262.075\xspace}
\newcommand{\nCostTestbedExactQueryMinimumUs}{186.600\xspace}
\newcommand{\nCostTestbedFixedConstructionMaximumMs}{81.657\xspace}
\newcommand{\nCostTestbedFixedConstructionMedianMs}{39.233\xspace}
\newcommand{\nCostTestbedFixedConstructionMinimumMs}{20.419\xspace}
\newcommand{\nCostTestbedFixedQueryMaximumUs}{0.773\xspace}
\newcommand{\nCostTestbedFixedQueryMedianUs}{0.375\xspace}
\newcommand{\nCostTestbedFixedQueryMinimumUs}{0.256\xspace}
\newcommand{\nCostTestbedIgFiveRerankConstructionMaximumMs}{27.172\xspace}
\newcommand{\nCostTestbedIgFiveRerankConstructionMedianMs}{14.852\xspace}
\newcommand{\nCostTestbedIgFiveRerankConstructionMinimumMs}{4.395\xspace}
\newcommand{\nCostTestbedIgFiveRerankQueryMaximumUs}{2985.473\xspace}
\newcommand{\nCostTestbedIgFiveRerankQueryMedianUs}{1602.543\xspace}
\newcommand{\nCostTestbedIgFiveRerankQueryMinimumUs}{849.068\xspace}
\newcommand{\nCostTestbedIgOneZeroRerankConstructionMaximumMs}{27.529\xspace}
\newcommand{\nCostTestbedIgOneZeroRerankConstructionMedianMs}{14.378\xspace}
\newcommand{\nCostTestbedIgOneZeroRerankConstructionMinimumMs}{3.831\xspace}
\newcommand{\nCostTestbedIgOneZeroRerankQueryMaximumUs}{5336.180\xspace}
\newcommand{\nCostTestbedIgOneZeroRerankQueryMedianUs}{2818.377\xspace}
\newcommand{\nCostTestbedIgOneZeroRerankQueryMinimumUs}{1329.763\xspace}
\newcommand{\nCostTestbedMarginConstructionMaximumMs}{90.750\xspace}
\newcommand{\nCostTestbedMarginConstructionMedianMs}{44.685\xspace}
\newcommand{\nCostTestbedMarginConstructionMinimumMs}{25.699\xspace}
\newcommand{\nCostTestbedMarginQueryMaximumUs}{331.112\xspace}
\newcommand{\nCostTestbedMarginQueryMedianUs}{44.228\xspace}
\newcommand{\nCostTestbedMarginQueryMinimumUs}{37.499\xspace}
\newcommand{\nCostTestbedSoftmaxFixedConstructionMaximumMs}{62.628\xspace}
\newcommand{\nCostTestbedSoftmaxFixedConstructionMedianMs}{38.048\xspace}
\newcommand{\nCostTestbedSoftmaxFixedConstructionMinimumMs}{17.269\xspace}
\newcommand{\nCostTestbedSoftmaxFixedQueryMaximumUs}{202.369\xspace}
\newcommand{\nCostTestbedSoftmaxFixedQueryMedianUs}{83.897\xspace}
\newcommand{\nCostTestbedSoftmaxFixedQueryMinimumUs}{26.476\xspace}
\newcommand{\nCostTestbedSoftmaxRerankConstructionMaximumMs}{82.558\xspace}
\newcommand{\nCostTestbedSoftmaxRerankConstructionMedianMs}{37.478\xspace}
\newcommand{\nCostTestbedSoftmaxRerankConstructionMinimumMs}{17.587\xspace}
\newcommand{\nCostTestbedSoftmaxRerankQueryMaximumUs}{364.609\xspace}
\newcommand{\nCostTestbedSoftmaxRerankQueryMedianUs}{223.701\xspace}
\newcommand{\nCostTestbedSoftmaxRerankQueryMinimumUs}{166.184\xspace}
\newcommand{\nCostThreads}{4\xspace}
\newcommand{\nCostWarmUps}{1\xspace}
\newcommand{\nCutoffFiveMargin}{0.0169\xspace}
\newcommand{\nCutoffFiveOldCutoff}{0.5577\xspace}
\newcommand{\nCutoffFiveRun}{5\xspace}
\newcommand{\nCutoffFourMargin}{0.0172\xspace}
\newcommand{\nCutoffFourOldCutoff}{0.5397\xspace}
\newcommand{\nCutoffFourRun}{4\xspace}
\newcommand{\nCutoffOneMargin}{0.0171\xspace}
\newcommand{\nCutoffOneOldCutoff}{0.5660\xspace}
\newcommand{\nCutoffOneRun}{1\xspace}
\newcommand{\nCutoffSevenMargin}{0.0167\xspace}
\newcommand{\nCutoffSevenOldCutoff}{0.5687\xspace}
\newcommand{\nCutoffSevenRun}{7\xspace}
\newcommand{\nCutoffSixMargin}{0.0163\xspace}
\newcommand{\nCutoffSixOldCutoff}{0.5568\xspace}
\newcommand{\nCutoffSixRun}{6\xspace}
\newcommand{\nCutoffThreeMargin}{0.0166\xspace}
\newcommand{\nCutoffThreeOldCutoff}{0.5568\xspace}
\newcommand{\nCutoffThreeRun}{3\xspace}
\newcommand{\nCutoffTwoMargin}{0.0169\xspace}
\newcommand{\nCutoffTwoOldCutoff}{0.5620\xspace}
\newcommand{\nCutoffTwoRun}{2\xspace}
\newcommand{\nCutoffZeroMargin}{0.0168\xspace}
\newcommand{\nCutoffZeroOldCutoff}{0.5587\xspace}
\newcommand{\nCutoffZeroRun}{0\xspace}
\newcommand{\nProofEight}{300\xspace}
\newcommand{\nProofEightSeven}{400\xspace}
\newcommand{\nProofEightZero}{0\xspace}
\newcommand{\nProofFiveThree}{0.333\xspace}
\newcommand{\nProofFiveTwo}{0.500\xspace}
\newcommand{\nProofFourFive}{0.071\xspace}
\newcommand{\nProofFourFour}{0.100\xspace}
\newcommand{\nProofFourOne}{0.416\xspace}
\newcommand{\nProofFourZero}{$2.8\times 10^{-16}$\xspace}
\newcommand{\nProofFourZeroFour}{1,710\xspace}
\newcommand{\nProofFourZeroOne}{$3.6\times 10^{-14}$\xspace}
\newcommand{\nProofFourZeroThree}{0\xspace}
\newcommand{\nProofFourZeroTwo}{592\xspace}
\newcommand{\nProofFourZeroZero}{$1.3\times 10^{-14}$\xspace}
\newcommand{\nProofNine}{300\xspace}
\newcommand{\nProofNineZero}{11999.571\xspace}
\newcommand{\nProofOneEightFour}{0.500\xspace}
\newcommand{\nProofOneFive}{10,098\xspace}
\newcommand{\nProofOneFiveEight}{240\xspace}
\newcommand{\nProofOneFiveFour}{2.548\xspace}
\newcommand{\nProofOneFiveSix}{12.397\xspace}
\newcommand{\nProofOneFiveThree}{2.548\xspace}
\newcommand{\nProofOneFour}{400\xspace}
\newcommand{\nProofOneFourNine}{2.548\xspace}
\newcommand{\nProofOneFourThree}{$1.5\times 10^{-5}$\xspace}
\newcommand{\nProofOneNine}{267\xspace}
\newcommand{\nProofOneOne}{300\xspace}
\newcommand{\nProofOneSeven}{10,098\xspace}
\newcommand{\nProofOneSix}{114\xspace}
\newcommand{\nProofOneSixFive}{1.511\xspace}
\newcommand{\nProofOneSixFour}{$4.4\times 10^{-15}$\xspace}
\newcommand{\nProofOneSixNine}{200\xspace}
\newcommand{\nProofOneSixSix}{$7.1\times 10^{-15}$\xspace}
\newcommand{\nProofOneSixThree}{300\xspace}
\newcommand{\nProofOneSixTwo}{36\xspace}
\newcommand{\nProofOneSixZero}{$4.4\times 10^{-16}$\xspace}
\newcommand{\nProofOneThree}{0.273\xspace}
\newcommand{\nProofOneTwo}{400\xspace}
\newcommand{\nProofOneZero}{300\xspace}
\newcommand{\nProofSeven}{268\xspace}
\newcommand{\nProofSevenFive}{2,416\xspace}
\newcommand{\nProofSevenNine}{0\xspace}
\newcommand{\nProofSevenSix}{20,700\xspace}
\newcommand{\nProofSixEight}{485.908\xspace}
\newcommand{\nProofSixOne}{0.391\xspace}
\newcommand{\nProofSixSeven}{2535.325\xspace}
\newcommand{\nProofSixSix}{$3.1\times 10^{-15}$\xspace}
\newcommand{\nProofSixZero}{0.600\xspace}
\newcommand{\nProofThreeEightZero}{0.500\xspace}
\newcommand{\nProofThreeFiveOne}{0.000\xspace}
\newcommand{\nProofThreeFiveTwo}{1.000\xspace}
\newcommand{\nProofThreeFiveZero}{0.000\xspace}
\newcommand{\nProofThreeFourFive}{0.650\xspace}
\newcommand{\nProofThreeFourFour}{0.200\xspace}
\newcommand{\nProofThreeFourNine}{$5.3\times 10^{-16}$\xspace}
\newcommand{\nProofThreeFourOne}{1.000\xspace}
\newcommand{\nProofThreeFourSeven}{0.000\xspace}
\newcommand{\nProofThreeFourSix}{1\xspace}
\newcommand{\nProofThreeFourThree}{0.800\xspace}
\newcommand{\nProofThreeFourTwo}{2.000\xspace}
\newcommand{\nProofThreeNine}{$3.5\times 10^{-16}$\xspace}
\newcommand{\nProofThreeNineNine}{0.500\xspace}
\newcommand{\nProofThreeOne}{0.115\xspace}
\newcommand{\nProofThreeSevenOne}{-3.000\xspace}
\newcommand{\nProofThreeSevenThree}{0.500\xspace}
\newcommand{\nProofThreeSevenZero}{1,710\xspace}
\newcommand{\nProofThreeSixNine}{1,710\xspace}
\newcommand{\nProofThreeSixTwo}{1.118\xspace}
\newcommand{\nProofThreeSixZero}{-1.586\xspace}
\newcommand{\nProofThreeThreeEight}{0.685\xspace}
\newcommand{\nProofThreeThreeFive}{0.500\xspace}
\newcommand{\nProofThreeThreeNine}{-2.000\xspace}
\newcommand{\nProofThreeThreeSeven}{0.448\xspace}
\newcommand{\nProofThreeThreeSix}{1.000\xspace}
\newcommand{\nProofThreeThreeThree}{0.000\xspace}
\newcommand{\nProofThreeTwoNine}{202635.092\xspace}
\newcommand{\nProofThreeZero}{47.055\xspace}
\newcommand{\nProofTwoEightFive}{16.730\xspace}
\newcommand{\nProofTwoNine}{0.000\xspace}
\newcommand{\nProofTwoSevenFour}{1.118\xspace}
\newcommand{\nProofTwoThree}{0.000\xspace}
\newcommand{\nProofTwoThreeFour}{1.118\xspace}
\newcommand{\nProofTwoTwo}{$3.9\times 10^{-17}$\xspace}
\newcommand{\nProofTwoTwoFour}{0.500\xspace}
\newcommand{\nProofTwoZero}{33\xspace}
\newcommand{\nReadAllControlMeanError}{0.1142\xspace}
\newcommand{\nReadAllControlMedianError}{0.0439\xspace}
\newcommand{\nReadAllExactCleanResidualMeanError}{0.1689\xspace}
\newcommand{\nReadAllExactCleanResidualMeanShort}{0.169\xspace}
\newcommand{\nReadAllExactCleanResidualMedianError}{0.0588\xspace}
\newcommand{\nReadAllFixedMeanError}{0.0843\xspace}
\newcommand{\nReadAllFixedMeanShort}{0.0843\xspace}
\newcommand{\nReadAllFixedMedianError}{0.0264\xspace}
\newcommand{\nReadAllLowerShort}{1.99\xspace}
\newcommand{\nReadAllMarginMeanError}{0.1783\xspace}
\newcommand{\nReadAllMarginMeanShort}{0.178\xspace}
\newcommand{\nReadAllMarginMedianError}{0.0606\xspace}
\newcommand{\nReadAllNullMeanError}{0.1267\xspace}
\newcommand{\nReadAllNullMedianError}{0.0362\xspace}
\newcommand{\nReadAllRatioShort}{2.11\xspace}
\newcommand{\nReadAllUnits}{9,600\xspace}
\newcommand{\nReadAllUpperShort}{2.24\xspace}
\newcommand{\nReadKeptControlMeanError}{0.1847\xspace}
\newcommand{\nReadKeptControlMedianError}{0.0989\xspace}
\newcommand{\nReadKeptExactCleanResidualMeanError}{0.2694\xspace}
\newcommand{\nReadKeptExactCleanResidualMeanShort}{0.269\xspace}
\newcommand{\nReadKeptExactCleanResidualMedianError}{0.1400\xspace}
\newcommand{\nReadKeptFixedMeanError}{0.1460\xspace}
\newcommand{\nReadKeptFixedMedianError}{0.0713\xspace}
\newcommand{\nReadKeptMarginMeanError}{0.2846\xspace}
\newcommand{\nReadKeptMarginMedianError}{0.1489\xspace}
\newcommand{\nReadKeptNullMeanError}{0.2374\xspace}
\newcommand{\nReadKeptNullMedianError}{0.1158\xspace}
\newcommand{\nReadKeptUnits}{4,893\xspace}
\newcommand{\nExtraFactorOneControlError}{0.1465\xspace}
\newcommand{\nExtraFactorTwoControlError}{0.0832\xspace}
\newcommand{\nExtraFactorZeroControlError}{0.2398\xspace}
\newcommand{\nExtraMapEightFitted}{0.5708\xspace}
\newcommand{\nExtraMapEightNullMax}{-0.5222\xspace}
\newcommand{\nExtraMapEightRandomMax}{0.0027\xspace}
\newcommand{\nExtraMapEightSourceRun}{1\xspace}
\newcommand{\nExtraMapEightTargetRun}{4\xspace}
\newcommand{\nExtraMapFiveFitted}{0.5673\xspace}
\newcommand{\nExtraMapFiveNullMax}{-0.5258\xspace}
\newcommand{\nExtraMapFiveRandomMax}{0.0026\xspace}
\newcommand{\nExtraMapFiveSourceRun}{1\xspace}
\newcommand{\nExtraMapFiveTargetRun}{0\xspace}
\newcommand{\nExtraMapFourFitted}{0.5674\xspace}
\newcommand{\nExtraMapFourNullMax}{-0.5302\xspace}
\newcommand{\nExtraMapFourRandomMax}{0.0026\xspace}
\newcommand{\nExtraMapFourSourceRun}{0\xspace}
\newcommand{\nExtraMapFourTargetRun}{5\xspace}
\newcommand{\nExtraMapNineFitted}{0.5657\xspace}
\newcommand{\nExtraMapNineNullMax}{-0.5323\xspace}
\newcommand{\nExtraMapNineRandomMax}{0.0026\xspace}
\newcommand{\nExtraMapNineSourceRun}{1\xspace}
\newcommand{\nExtraMapNineTargetRun}{5\xspace}
\newcommand{\nExtraMapOneEightFitted}{0.5724\xspace}
\newcommand{\nExtraMapOneEightNullMax}{-0.5211\xspace}
\newcommand{\nExtraMapOneEightRandomMax}{0.0026\xspace}
\newcommand{\nExtraMapOneEightSourceRun}{3\xspace}
\newcommand{\nExtraMapOneEightTargetRun}{4\xspace}
\newcommand{\nExtraMapOneFitted}{0.5714\xspace}
\newcommand{\nExtraMapOneFiveFitted}{0.5680\xspace}
\newcommand{\nExtraMapOneFiveNullMax}{-0.5237\xspace}
\newcommand{\nExtraMapOneFiveRandomMax}{0.0026\xspace}
\newcommand{\nExtraMapOneFiveSourceRun}{3\xspace}
\newcommand{\nExtraMapOneFiveTargetRun}{0\xspace}
\newcommand{\nExtraMapOneFourFitted}{0.5661\xspace}
\newcommand{\nExtraMapOneFourNullMax}{-0.5302\xspace}
\newcommand{\nExtraMapOneFourRandomMax}{0.0026\xspace}
\newcommand{\nExtraMapOneFourSourceRun}{2\xspace}
\newcommand{\nExtraMapOneFourTargetRun}{5\xspace}
\newcommand{\nExtraMapOneNineFitted}{0.5672\xspace}
\newcommand{\nExtraMapOneNineNullMax}{-0.5317\xspace}
\newcommand{\nExtraMapOneNineRandomMax}{0.0027\xspace}
\newcommand{\nExtraMapOneNineSourceRun}{3\xspace}
\newcommand{\nExtraMapOneNineTargetRun}{5\xspace}
\newcommand{\nExtraMapOneNullMax}{-0.5146\xspace}
\newcommand{\nExtraMapOneOneFitted}{0.5699\xspace}
\newcommand{\nExtraMapOneOneNullMax}{-0.5223\xspace}
\newcommand{\nExtraMapOneOneRandomMax}{0.0026\xspace}
\newcommand{\nExtraMapOneOneSourceRun}{2\xspace}
\newcommand{\nExtraMapOneOneTargetRun}{1\xspace}
\newcommand{\nExtraMapOneRandomMax}{0.0026\xspace}
\newcommand{\nExtraMapOneSevenFitted}{0.5716\xspace}
\newcommand{\nExtraMapOneSevenNullMax}{-0.5141\xspace}
\newcommand{\nExtraMapOneSevenRandomMax}{0.0026\xspace}
\newcommand{\nExtraMapOneSevenSourceRun}{3\xspace}
\newcommand{\nExtraMapOneSevenTargetRun}{2\xspace}
\newcommand{\nExtraMapOneSixFitted}{0.5720\xspace}
\newcommand{\nExtraMapOneSixNullMax}{-0.5227\xspace}
\newcommand{\nExtraMapOneSixRandomMax}{0.0026\xspace}
\newcommand{\nExtraMapOneSixSourceRun}{3\xspace}
\newcommand{\nExtraMapOneSixTargetRun}{1\xspace}
\newcommand{\nExtraMapOneSourceRun}{0\xspace}
\newcommand{\nExtraMapOneTargetRun}{2\xspace}
\newcommand{\nExtraMapOneThreeFitted}{0.5704\xspace}
\newcommand{\nExtraMapOneThreeNullMax}{-0.5172\xspace}
\newcommand{\nExtraMapOneThreeRandomMax}{0.0027\xspace}
\newcommand{\nExtraMapOneThreeSourceRun}{2\xspace}
\newcommand{\nExtraMapOneThreeTargetRun}{4\xspace}
\newcommand{\nExtraMapOneTwoFitted}{0.5705\xspace}
\newcommand{\nExtraMapOneTwoNullMax}{-0.5202\xspace}
\newcommand{\nExtraMapOneTwoRandomMax}{0.0026\xspace}
\newcommand{\nExtraMapOneTwoSourceRun}{2\xspace}
\newcommand{\nExtraMapOneTwoTargetRun}{3\xspace}
\newcommand{\nExtraMapOneZeroFitted}{0.5673\xspace}
\newcommand{\nExtraMapOneZeroNullMax}{-0.5224\xspace}
\newcommand{\nExtraMapOneZeroRandomMax}{0.0026\xspace}
\newcommand{\nExtraMapOneZeroSourceRun}{2\xspace}
\newcommand{\nExtraMapOneZeroTargetRun}{0\xspace}
\newcommand{\nExtraMapSevenFitted}{0.5710\xspace}
\newcommand{\nExtraMapSevenNullMax}{-0.5239\xspace}
\newcommand{\nExtraMapSevenRandomMax}{0.0026\xspace}
\newcommand{\nExtraMapSevenSourceRun}{1\xspace}
\newcommand{\nExtraMapSevenTargetRun}{3\xspace}
\newcommand{\nExtraMapSixFitted}{0.5697\xspace}
\newcommand{\nExtraMapSixNullMax}{-0.5155\xspace}
\newcommand{\nExtraMapSixRandomMax}{0.0026\xspace}
\newcommand{\nExtraMapSixSourceRun}{1\xspace}
\newcommand{\nExtraMapSixTargetRun}{2\xspace}
\newcommand{\nExtraMapThreeFitted}{0.5720\xspace}
\newcommand{\nExtraMapThreeNullMax}{-0.5200\xspace}
\newcommand{\nExtraMapThreeRandomMax}{0.0026\xspace}
\newcommand{\nExtraMapThreeSourceRun}{0\xspace}
\newcommand{\nExtraMapThreeTargetRun}{4\xspace}
\newcommand{\nExtraMapTwoEightFitted}{0.5698\xspace}
\newcommand{\nExtraMapTwoEightNullMax}{-0.5182\xspace}
\newcommand{\nExtraMapTwoEightRandomMax}{0.0026\xspace}
\newcommand{\nExtraMapTwoEightSourceRun}{5\xspace}
\newcommand{\nExtraMapTwoEightTargetRun}{3\xspace}
\newcommand{\nExtraMapTwoFitted}{0.5717\xspace}
\newcommand{\nExtraMapTwoFiveFitted}{0.5661\xspace}
\newcommand{\nExtraMapTwoFiveNullMax}{-0.5191\xspace}
\newcommand{\nExtraMapTwoFiveRandomMax}{0.0026\xspace}
\newcommand{\nExtraMapTwoFiveSourceRun}{5\xspace}
\newcommand{\nExtraMapTwoFiveTargetRun}{0\xspace}
\newcommand{\nExtraMapTwoFourFitted}{0.5663\xspace}
\newcommand{\nExtraMapTwoFourNullMax}{-0.5316\xspace}
\newcommand{\nExtraMapTwoFourRandomMax}{0.0026\xspace}
\newcommand{\nExtraMapTwoFourSourceRun}{4\xspace}
\newcommand{\nExtraMapTwoFourTargetRun}{5\xspace}
\newcommand{\nExtraMapTwoNineFitted}{0.5703\xspace}
\newcommand{\nExtraMapTwoNineNullMax}{-0.5156\xspace}
\newcommand{\nExtraMapTwoNineRandomMax}{0.0026\xspace}
\newcommand{\nExtraMapTwoNineSourceRun}{5\xspace}
\newcommand{\nExtraMapTwoNineTargetRun}{4\xspace}
\newcommand{\nExtraMapTwoNullMax}{-0.5216\xspace}
\newcommand{\nExtraMapTwoOneFitted}{0.5714\xspace}
\newcommand{\nExtraMapTwoOneNullMax}{-0.5244\xspace}
\newcommand{\nExtraMapTwoOneRandomMax}{0.0026\xspace}
\newcommand{\nExtraMapTwoOneSourceRun}{4\xspace}
\newcommand{\nExtraMapTwoOneTargetRun}{1\xspace}
\newcommand{\nExtraMapTwoRandomMax}{0.0026\xspace}
\newcommand{\nExtraMapTwoSevenFitted}{0.5724\xspace}
\newcommand{\nExtraMapTwoSevenNullMax}{-0.5099\xspace}
\newcommand{\nExtraMapTwoSevenRandomMax}{0.0026\xspace}
\newcommand{\nExtraMapTwoSevenSourceRun}{5\xspace}
\newcommand{\nExtraMapTwoSevenTargetRun}{2\xspace}
\newcommand{\nExtraMapTwoSixFitted}{0.5700\xspace}
\newcommand{\nExtraMapTwoSixNullMax}{-0.5184\xspace}
\newcommand{\nExtraMapTwoSixRandomMax}{0.0026\xspace}
\newcommand{\nExtraMapTwoSixSourceRun}{5\xspace}
\newcommand{\nExtraMapTwoSixTargetRun}{1\xspace}
\newcommand{\nExtraMapTwoSourceRun}{0\xspace}
\newcommand{\nExtraMapTwoTargetRun}{3\xspace}
\newcommand{\nExtraMapTwoThreeFitted}{0.5717\xspace}
\newcommand{\nExtraMapTwoThreeNullMax}{-0.5243\xspace}
\newcommand{\nExtraMapTwoThreeRandomMax}{0.0026\xspace}
\newcommand{\nExtraMapTwoThreeSourceRun}{4\xspace}
\newcommand{\nExtraMapTwoThreeTargetRun}{3\xspace}
\newcommand{\nExtraMapTwoTwoFitted}{0.5712\xspace}
\newcommand{\nExtraMapTwoTwoNullMax}{-0.5164\xspace}
\newcommand{\nExtraMapTwoTwoRandomMax}{0.0026\xspace}
\newcommand{\nExtraMapTwoTwoSourceRun}{4\xspace}
\newcommand{\nExtraMapTwoTwoTargetRun}{2\xspace}
\newcommand{\nExtraMapTwoZeroFitted}{0.5685\xspace}
\newcommand{\nExtraMapTwoZeroNullMax}{-0.5233\xspace}
\newcommand{\nExtraMapTwoZeroRandomMax}{0.0026\xspace}
\newcommand{\nExtraMapTwoZeroSourceRun}{4\xspace}
\newcommand{\nExtraMapTwoZeroTargetRun}{0\xspace}
\newcommand{\nExtraMapZeroFitted}{0.5715\xspace}
\newcommand{\nExtraMapZeroNullMax}{-0.5227\xspace}
\newcommand{\nExtraMapZeroRandomMax}{0.0027\xspace}
\newcommand{\nExtraMapZeroSourceRun}{0\xspace}
\newcommand{\nExtraMapZeroTargetRun}{1\xspace}
\newcommand{\nExtraReleaseErrMargin}{0.033\xspace}
\newcommand{\nExtraReleaseErrThreshold}{0.516\xspace}
\newcommand{\nExtraScaleFloor}{$1.0\times 10^{-6}$\xspace}
\newcommand{\nWorkCountFive}{768\xspace}
\newcommand{\nWorkCountFour}{384\xspace}
\newcommand{\nWorkCountOne}{48\xspace}
\newcommand{\nWorkCountThree}{192\xspace}
\newcommand{\nWorkCountTwo}{96\xspace}
\newcommand{\nWorkCountZero}{16\xspace}
\newcommand{\nWorkCpuReleaseExactCountFourEightMedianMs}{29.068\xspace}
\newcommand{\nWorkCpuReleaseExactCountOneNineTwoMedianMs}{58.306\xspace}
\newcommand{\nWorkCpuReleaseExactCountOneSixMedianMs}{23.283\xspace}
\newcommand{\nWorkCpuReleaseExactCountSevenSixEightMedianMs}{165.080\xspace}
\newcommand{\nWorkCpuReleaseExactCountThreeEightFourMedianMs}{94.443\xspace}
\newcommand{\nWorkCpuReleaseExactMaximumMs}{76.528\xspace}
\newcommand{\nWorkCpuReleaseExactMedianMs}{29.068\xspace}
\newcommand{\nWorkCpuReleaseExactMinimumMs}{20.046\xspace}
\newcommand{\nWorkCpuReleaseExactPeakRssMib}{1803.156\xspace}
\newcommand{\nWorkCpuReleaseFixedCountFourEightMedianMs}{24.144\xspace}
\newcommand{\nWorkCpuReleaseFixedCountOneNineTwoMedianMs}{24.429\xspace}
\newcommand{\nWorkCpuReleaseFixedCountOneSixMedianMs}{25.500\xspace}
\newcommand{\nWorkCpuReleaseFixedCountSevenSixEightMedianMs}{26.112\xspace}
\newcommand{\nWorkCpuReleaseFixedCountThreeEightFourMedianMs}{26.294\xspace}
\newcommand{\nWorkCpuReleaseFixedCrossover}{48\xspace}
\newcommand{\nWorkCpuReleaseFixedMaximumMs}{44.969\xspace}
\newcommand{\nWorkCpuReleaseFixedMedianMs}{24.144\xspace}
\newcommand{\nWorkCpuReleaseFixedMinimumMs}{13.502\xspace}
\newcommand{\nWorkCpuReleaseFixedPeakRssMib}{1860.922\xspace}
\newcommand{\nWorkCpuReleaseIgFiveRerankCountFourEightMedianMs}{71.697\xspace}
\newcommand{\nWorkCpuReleaseIgFiveRerankCountOneNineTwoMedianMs}{209.136\xspace}
\newcommand{\nWorkCpuReleaseIgFiveRerankCountOneSixMedianMs}{44.670\xspace}
\newcommand{\nWorkCpuReleaseIgFiveRerankCountSevenSixEightMedianMs}{762.616\xspace}
\newcommand{\nWorkCpuReleaseIgFiveRerankCountThreeEightFourMedianMs}{391.066\xspace}
\newcommand{\nWorkCpuReleaseIgFiveRerankMaximumMs}{253.972\xspace}
\newcommand{\nWorkCpuReleaseIgFiveRerankMedianMs}{71.697\xspace}
\newcommand{\nWorkCpuReleaseIgFiveRerankMinimumMs}{43.320\xspace}
\newcommand{\nWorkCpuReleaseIgFiveRerankPeakRssMib}{1789.969\xspace}
\newcommand{\nWorkCpuReleaseIgOneZeroRerankCountFourEightMedianMs}{114.994\xspace}
\newcommand{\nWorkCpuReleaseIgOneZeroRerankCountOneNineTwoMedianMs}{365.641\xspace}
\newcommand{\nWorkCpuReleaseIgOneZeroRerankCountOneSixMedianMs}{55.933\xspace}
\newcommand{\nWorkCpuReleaseIgOneZeroRerankCountSevenSixEightMedianMs}{1387.302\xspace}
\newcommand{\nWorkCpuReleaseIgOneZeroRerankCountThreeEightFourMedianMs}{715.368\xspace}
\newcommand{\nWorkCpuReleaseIgOneZeroRerankMaximumMs}{280.671\xspace}
\newcommand{\nWorkCpuReleaseIgOneZeroRerankMedianMs}{114.994\xspace}
\newcommand{\nWorkCpuReleaseIgOneZeroRerankMinimumMs}{64.130\xspace}
\newcommand{\nWorkCpuReleaseIgOneZeroRerankPeakRssMib}{1803.156\xspace}
\newcommand{\nWorkCpuReleaseMarginCountFourEightMedianMs}{33.051\xspace}
\newcommand{\nWorkCpuReleaseMarginCountOneNineTwoMedianMs}{41.116\xspace}
\newcommand{\nWorkCpuReleaseMarginCountOneSixMedianMs}{34.089\xspace}
\newcommand{\nWorkCpuReleaseMarginCountSevenSixEightMedianMs}{76.889\xspace}
\newcommand{\nWorkCpuReleaseMarginCountThreeEightFourMedianMs}{55.817\xspace}
\newcommand{\nWorkCpuReleaseMarginCrossover}{192\xspace}
\newcommand{\nWorkCpuReleaseMarginMaximumMs}{61.216\xspace}
\newcommand{\nWorkCpuReleaseMarginMedianMs}{33.051\xspace}
\newcommand{\nWorkCpuReleaseMarginMinimumMs}{22.716\xspace}
\newcommand{\nWorkCpuReleaseMarginPeakRssMib}{1842.922\xspace}
\newcommand{\nWorkCpuReleaseSoftmaxFixedCountFourEightMedianMs}{29.174\xspace}
\newcommand{\nWorkCpuReleaseSoftmaxFixedCountOneNineTwoMedianMs}{34.716\xspace}
\newcommand{\nWorkCpuReleaseSoftmaxFixedCountOneSixMedianMs}{29.562\xspace}
\newcommand{\nWorkCpuReleaseSoftmaxFixedCountSevenSixEightMedianMs}{59.796\xspace}
\newcommand{\nWorkCpuReleaseSoftmaxFixedCountThreeEightFourMedianMs}{45.475\xspace}
\newcommand{\nWorkCpuReleaseSoftmaxFixedCrossover}{192\xspace}
\newcommand{\nWorkCpuReleaseSoftmaxFixedMaximumMs}{61.896\xspace}
\newcommand{\nWorkCpuReleaseSoftmaxFixedMedianMs}{29.174\xspace}
\newcommand{\nWorkCpuReleaseSoftmaxFixedMinimumMs}{13.998\xspace}
\newcommand{\nWorkCpuReleaseSoftmaxFixedPeakRssMib}{1842.922\xspace}
\newcommand{\nWorkCpuReleaseSoftmaxRerankCountFourEightMedianMs}{40.094\xspace}
\newcommand{\nWorkCpuReleaseSoftmaxRerankCountOneNineTwoMedianMs}{63.308\xspace}
\newcommand{\nWorkCpuReleaseSoftmaxRerankCountOneSixMedianMs}{36.043\xspace}
\newcommand{\nWorkCpuReleaseSoftmaxRerankCountSevenSixEightMedianMs}{163.338\xspace}
\newcommand{\nWorkCpuReleaseSoftmaxRerankCountThreeEightFourMedianMs}{98.257\xspace}
\newcommand{\nWorkCpuReleaseSoftmaxRerankCrossover}{768\xspace}
\newcommand{\nWorkCpuReleaseSoftmaxRerankMaximumMs}{77.349\xspace}
\newcommand{\nWorkCpuReleaseSoftmaxRerankMedianMs}{40.094\xspace}
\newcommand{\nWorkCpuReleaseSoftmaxRerankMinimumMs}{24.844\xspace}
\newcommand{\nWorkCpuReleaseSoftmaxRerankPeakRssMib}{1842.922\xspace}
\newcommand{\nWorkCpuTestbedExactCountNineSixMedianMs}{27.418\xspace}
\newcommand{\nWorkCpuTestbedExactCountOneNineTwoMedianMs}{40.924\xspace}
\newcommand{\nWorkCpuTestbedExactCountOneSixMedianMs}{17.744\xspace}
\newcommand{\nWorkCpuTestbedExactCountSevenSixEightMedianMs}{111.517\xspace}
\newcommand{\nWorkCpuTestbedExactCountThreeEightFourMedianMs}{66.159\xspace}
\newcommand{\nWorkCpuTestbedExactMaximumMs}{60.256\xspace}
\newcommand{\nWorkCpuTestbedExactMedianMs}{27.418\xspace}
\newcommand{\nWorkCpuTestbedExactMinimumMs}{17.294\xspace}
\newcommand{\nWorkCpuTestbedExactPeakRssMib}{2056.969\xspace}
\newcommand{\nWorkCpuTestbedFixedCountNineSixMedianMs}{24.990\xspace}
\newcommand{\nWorkCpuTestbedFixedCountOneNineTwoMedianMs}{25.807\xspace}
\newcommand{\nWorkCpuTestbedFixedCountOneSixMedianMs}{26.562\xspace}
\newcommand{\nWorkCpuTestbedFixedCountSevenSixEightMedianMs}{27.715\xspace}
\newcommand{\nWorkCpuTestbedFixedCountThreeEightFourMedianMs}{27.652\xspace}
\newcommand{\nWorkCpuTestbedFixedCrossover}{96\xspace}
\newcommand{\nWorkCpuTestbedFixedMaximumMs}{41.126\xspace}
\newcommand{\nWorkCpuTestbedFixedMedianMs}{24.990\xspace}
\newcommand{\nWorkCpuTestbedFixedMinimumMs}{13.201\xspace}
\newcommand{\nWorkCpuTestbedFixedPeakRssMib}{2049.766\xspace}
\newcommand{\nWorkCpuTestbedIgFiveRerankCountNineSixMedianMs}{109.292\xspace}
\newcommand{\nWorkCpuTestbedIgFiveRerankCountOneNineTwoMedianMs}{193.481\xspace}
\newcommand{\nWorkCpuTestbedIgFiveRerankCountOneSixMedianMs}{39.361\xspace}
\newcommand{\nWorkCpuTestbedIgFiveRerankCountSevenSixEightMedianMs}{716.912\xspace}
\newcommand{\nWorkCpuTestbedIgFiveRerankCountThreeEightFourMedianMs}{362.279\xspace}
\newcommand{\nWorkCpuTestbedIgFiveRerankMaximumMs}{197.876\xspace}
\newcommand{\nWorkCpuTestbedIgFiveRerankMedianMs}{109.292\xspace}
\newcommand{\nWorkCpuTestbedIgFiveRerankMinimumMs}{59.306\xspace}
\newcommand{\nWorkCpuTestbedIgFiveRerankPeakRssMib}{2039.828\xspace}
\newcommand{\nWorkCpuTestbedIgOneZeroRerankCountNineSixMedianMs}{186.409\xspace}
\newcommand{\nWorkCpuTestbedIgOneZeroRerankCountOneNineTwoMedianMs}{348.648\xspace}
\newcommand{\nWorkCpuTestbedIgOneZeroRerankCountOneSixMedianMs}{50.457\xspace}
\newcommand{\nWorkCpuTestbedIgOneZeroRerankCountSevenSixEightMedianMs}{1348.729\xspace}
\newcommand{\nWorkCpuTestbedIgOneZeroRerankCountThreeEightFourMedianMs}{694.750\xspace}
\newcommand{\nWorkCpuTestbedIgOneZeroRerankMaximumMs}{316.036\xspace}
\newcommand{\nWorkCpuTestbedIgOneZeroRerankMedianMs}{186.409\xspace}
\newcommand{\nWorkCpuTestbedIgOneZeroRerankMinimumMs}{95.705\xspace}
\newcommand{\nWorkCpuTestbedIgOneZeroRerankPeakRssMib}{2184.812\xspace}
\newcommand{\nWorkCpuTestbedMarginCountNineSixMedianMs}{31.456\xspace}
\newcommand{\nWorkCpuTestbedMarginCountOneNineTwoMedianMs}{34.994\xspace}
\newcommand{\nWorkCpuTestbedMarginCountOneSixMedianMs}{31.646\xspace}
\newcommand{\nWorkCpuTestbedMarginCountSevenSixEightMedianMs}{53.211\xspace}
\newcommand{\nWorkCpuTestbedMarginCountThreeEightFourMedianMs}{43.141\xspace}
\newcommand{\nWorkCpuTestbedMarginCrossover}{192\xspace}
\newcommand{\nWorkCpuTestbedMarginMaximumMs}{50.728\xspace}
\newcommand{\nWorkCpuTestbedMarginMedianMs}{31.456\xspace}
\newcommand{\nWorkCpuTestbedMarginMinimumMs}{19.799\xspace}
\newcommand{\nWorkCpuTestbedMarginPeakRssMib}{2031.797\xspace}
\newcommand{\nWorkCpuTestbedSoftmaxFixedCountNineSixMedianMs}{29.592\xspace}
\newcommand{\nWorkCpuTestbedSoftmaxFixedCountOneNineTwoMedianMs}{33.473\xspace}
\newcommand{\nWorkCpuTestbedSoftmaxFixedCountOneSixMedianMs}{27.352\xspace}
\newcommand{\nWorkCpuTestbedSoftmaxFixedCountSevenSixEightMedianMs}{59.079\xspace}
\newcommand{\nWorkCpuTestbedSoftmaxFixedCountThreeEightFourMedianMs}{43.729\xspace}
\newcommand{\nWorkCpuTestbedSoftmaxFixedCrossover}{192\xspace}
\newcommand{\nWorkCpuTestbedSoftmaxFixedMaximumMs}{56.320\xspace}
\newcommand{\nWorkCpuTestbedSoftmaxFixedMedianMs}{29.592\xspace}
\newcommand{\nWorkCpuTestbedSoftmaxFixedMinimumMs}{14.664\xspace}
\newcommand{\nWorkCpuTestbedSoftmaxFixedPeakRssMib}{2031.797\xspace}
\newcommand{\nWorkCpuTestbedSoftmaxRerankCountNineSixMedianMs}{38.297\xspace}
\newcommand{\nWorkCpuTestbedSoftmaxRerankCountOneNineTwoMedianMs}{48.190\xspace}
\newcommand{\nWorkCpuTestbedSoftmaxRerankCountOneSixMedianMs}{32.092\xspace}
\newcommand{\nWorkCpuTestbedSoftmaxRerankCountSevenSixEightMedianMs}{111.039\xspace}
\newcommand{\nWorkCpuTestbedSoftmaxRerankCountThreeEightFourMedianMs}{71.259\xspace}
\newcommand{\nWorkCpuTestbedSoftmaxRerankCrossover}{768\xspace}
\newcommand{\nWorkCpuTestbedSoftmaxRerankMaximumMs}{88.045\xspace}
\newcommand{\nWorkCpuTestbedSoftmaxRerankMedianMs}{38.297\xspace}
\newcommand{\nWorkCpuTestbedSoftmaxRerankMinimumMs}{22.240\xspace}
\newcommand{\nWorkCpuTestbedSoftmaxRerankPeakRssMib}{2031.797\xspace}
\newcommand{\nWorkCudaReleaseExactCountFourEightMedianMs}{2.931\xspace}
\newcommand{\nWorkCudaReleaseExactCountOneNineTwoMedianMs}{3.646\xspace}
\newcommand{\nWorkCudaReleaseExactCountOneSixMedianMs}{2.848\xspace}
\newcommand{\nWorkCudaReleaseExactCountSevenSixEightMedianMs}{6.492\xspace}
\newcommand{\nWorkCudaReleaseExactCountThreeEightFourMedianMs}{4.509\xspace}
\newcommand{\nWorkCudaReleaseExactMaximumMs}{5.535\xspace}
\newcommand{\nWorkCudaReleaseExactMedianMs}{2.931\xspace}
\newcommand{\nWorkCudaReleaseExactMedianShort}{2.93\xspace}
\newcommand{\nWorkCudaReleaseExactMinimumMs}{2.521\xspace}
\newcommand{\nWorkCudaReleaseExactPeakAllocatedMib}{1135.421\xspace}
\newcommand{\nWorkCudaReleaseExactPeakRssMib}{4958.871\xspace}
\newcommand{\nWorkCudaReleaseFixedCountFourEightMedianMs}{10.236\xspace}
\newcommand{\nWorkCudaReleaseFixedCountOneNineTwoMedianMs}{10.195\xspace}
\newcommand{\nWorkCudaReleaseFixedCountOneSixMedianMs}{10.244\xspace}
\newcommand{\nWorkCudaReleaseFixedCountSevenSixEightMedianMs}{10.276\xspace}
\newcommand{\nWorkCudaReleaseFixedCountThreeEightFourMedianMs}{10.242\xspace}
\newcommand{\nWorkCudaReleaseFixedMaximumMs}{16.742\xspace}
\newcommand{\nWorkCudaReleaseFixedMedianMs}{10.236\xspace}
\newcommand{\nWorkCudaReleaseFixedMinimumMs}{8.897\xspace}
\newcommand{\nWorkCudaReleaseFixedPeakAllocatedMib}{1120.841\xspace}
\newcommand{\nWorkCudaReleaseFixedPeakRssMib}{4958.871\xspace}
\newcommand{\nWorkCudaReleaseIgFiveRerankCountFourEightMedianMs}{10.665\xspace}
\newcommand{\nWorkCudaReleaseIgFiveRerankCountOneNineTwoMedianMs}{11.608\xspace}
\newcommand{\nWorkCudaReleaseIgFiveRerankCountOneSixMedianMs}{10.445\xspace}
\newcommand{\nWorkCudaReleaseIgFiveRerankCountSevenSixEightMedianMs}{21.521\xspace}
\newcommand{\nWorkCudaReleaseIgFiveRerankCountThreeEightFourMedianMs}{14.426\xspace}
\newcommand{\nWorkCudaReleaseIgFiveRerankMaximumMs}{16.814\xspace}
\newcommand{\nWorkCudaReleaseIgFiveRerankMedianMs}{10.665\xspace}
\newcommand{\nWorkCudaReleaseIgFiveRerankMinimumMs}{9.422\xspace}
\newcommand{\nWorkCudaReleaseIgFiveRerankPeakAllocatedMib}{1322.166\xspace}
\newcommand{\nWorkCudaReleaseIgFiveRerankPeakRssMib}{4958.871\xspace}
\newcommand{\nWorkCudaReleaseIgOneZeroRerankCountFourEightMedianMs}{10.540\xspace}
\newcommand{\nWorkCudaReleaseIgOneZeroRerankCountOneNineTwoMedianMs}{13.538\xspace}
\newcommand{\nWorkCudaReleaseIgOneZeroRerankCountOneSixMedianMs}{10.389\xspace}
\newcommand{\nWorkCudaReleaseIgOneZeroRerankCountSevenSixEightMedianMs}{31.889\xspace}
\newcommand{\nWorkCudaReleaseIgOneZeroRerankCountThreeEightFourMedianMs}{19.685\xspace}
\newcommand{\nWorkCudaReleaseIgOneZeroRerankMaximumMs}{15.731\xspace}
\newcommand{\nWorkCudaReleaseIgOneZeroRerankMedianMs}{10.540\xspace}
\newcommand{\nWorkCudaReleaseIgOneZeroRerankMinimumMs}{9.430\xspace}
\newcommand{\nWorkCudaReleaseIgOneZeroRerankPeakAllocatedMib}{1546.251\xspace}
\newcommand{\nWorkCudaReleaseIgOneZeroRerankPeakRssMib}{4958.871\xspace}
\newcommand{\nWorkCudaReleaseMarginCountFourEightMedianMs}{10.269\xspace}
\newcommand{\nWorkCudaReleaseMarginCountOneNineTwoMedianMs}{10.836\xspace}
\newcommand{\nWorkCudaReleaseMarginCountOneSixMedianMs}{10.282\xspace}
\newcommand{\nWorkCudaReleaseMarginCountSevenSixEightMedianMs}{12.405\xspace}
\newcommand{\nWorkCudaReleaseMarginCountThreeEightFourMedianMs}{11.403\xspace}
\newcommand{\nWorkCudaReleaseMarginMaximumMs}{17.196\xspace}
\newcommand{\nWorkCudaReleaseMarginMedianMs}{10.269\xspace}
\newcommand{\nWorkCudaReleaseMarginMedianShort}{10.3\xspace}
\newcommand{\nWorkCudaReleaseMarginMinimumMs}{9.082\xspace}
\newcommand{\nWorkCudaReleaseMarginPeakAllocatedMib}{1120.841\xspace}
\newcommand{\nWorkCudaReleaseMarginPeakRssMib}{4958.871\xspace}
\newcommand{\nWorkCudaReleaseSoftmaxFixedCountFourEightMedianMs}{6.514\xspace}
\newcommand{\nWorkCudaReleaseSoftmaxFixedCountOneNineTwoMedianMs}{6.497\xspace}
\newcommand{\nWorkCudaReleaseSoftmaxFixedCountOneSixMedianMs}{6.459\xspace}
\newcommand{\nWorkCudaReleaseSoftmaxFixedCountSevenSixEightMedianMs}{6.692\xspace}
\newcommand{\nWorkCudaReleaseSoftmaxFixedCountThreeEightFourMedianMs}{6.516\xspace}
\newcommand{\nWorkCudaReleaseSoftmaxFixedMaximumMs}{9.803\xspace}
\newcommand{\nWorkCudaReleaseSoftmaxFixedMedianMs}{6.514\xspace}
\newcommand{\nWorkCudaReleaseSoftmaxFixedMinimumMs}{5.660\xspace}
\newcommand{\nWorkCudaReleaseSoftmaxFixedPeakAllocatedMib}{1150.724\xspace}
\newcommand{\nWorkCudaReleaseSoftmaxFixedPeakRssMib}{4958.871\xspace}
\newcommand{\nWorkCudaReleaseSoftmaxRerankCountFourEightMedianMs}{6.680\xspace}
\newcommand{\nWorkCudaReleaseSoftmaxRerankCountOneNineTwoMedianMs}{7.475\xspace}
\newcommand{\nWorkCudaReleaseSoftmaxRerankCountOneSixMedianMs}{6.571\xspace}
\newcommand{\nWorkCudaReleaseSoftmaxRerankCountSevenSixEightMedianMs}{10.106\xspace}
\newcommand{\nWorkCudaReleaseSoftmaxRerankCountThreeEightFourMedianMs}{8.239\xspace}
\newcommand{\nWorkCudaReleaseSoftmaxRerankMaximumMs}{10.160\xspace}
\newcommand{\nWorkCudaReleaseSoftmaxRerankMedianMs}{6.680\xspace}
\newcommand{\nWorkCudaReleaseSoftmaxRerankMinimumMs}{5.928\xspace}
\newcommand{\nWorkCudaReleaseSoftmaxRerankPeakAllocatedMib}{1180.440\xspace}
\newcommand{\nWorkCudaReleaseSoftmaxRerankPeakRssMib}{4958.871\xspace}
\newcommand{\nWorkCudaTestbedExactCountNineSixMedianMs}{2.879\xspace}
\newcommand{\nWorkCudaTestbedExactCountOneNineTwoMedianMs}{3.113\xspace}
\newcommand{\nWorkCudaTestbedExactCountOneSixMedianMs}{2.700\xspace}
\newcommand{\nWorkCudaTestbedExactCountSevenSixEightMedianMs}{4.837\xspace}
\newcommand{\nWorkCudaTestbedExactCountThreeEightFourMedianMs}{3.735\xspace}
\newcommand{\nWorkCudaTestbedExactMaximumMs}{5.339\xspace}
\newcommand{\nWorkCudaTestbedExactMedianMs}{2.879\xspace}
\newcommand{\nWorkCudaTestbedExactMinimumMs}{2.422\xspace}
\newcommand{\nWorkCudaTestbedExactPeakAllocatedMib}{886.558\xspace}
\newcommand{\nWorkCudaTestbedExactPeakRssMib}{4952.086\xspace}
\newcommand{\nWorkCudaTestbedFixedCountNineSixMedianMs}{10.528\xspace}
\newcommand{\nWorkCudaTestbedFixedCountOneNineTwoMedianMs}{10.507\xspace}
\newcommand{\nWorkCudaTestbedFixedCountOneSixMedianMs}{10.608\xspace}
\newcommand{\nWorkCudaTestbedFixedCountSevenSixEightMedianMs}{10.587\xspace}
\newcommand{\nWorkCudaTestbedFixedCountThreeEightFourMedianMs}{10.479\xspace}
\newcommand{\nWorkCudaTestbedFixedMaximumMs}{16.331\xspace}
\newcommand{\nWorkCudaTestbedFixedMedianMs}{10.528\xspace}
\newcommand{\nWorkCudaTestbedFixedMinimumMs}{9.017\xspace}
\newcommand{\nWorkCudaTestbedFixedPeakAllocatedMib}{878.220\xspace}
\newcommand{\nWorkCudaTestbedFixedPeakRssMib}{4952.086\xspace}
\newcommand{\nWorkCudaTestbedIgFiveRerankCountNineSixMedianMs}{10.817\xspace}
\newcommand{\nWorkCudaTestbedIgFiveRerankCountOneNineTwoMedianMs}{11.310\xspace}
\newcommand{\nWorkCudaTestbedIgFiveRerankCountOneSixMedianMs}{10.659\xspace}
\newcommand{\nWorkCudaTestbedIgFiveRerankCountSevenSixEightMedianMs}{20.261\xspace}
\newcommand{\nWorkCudaTestbedIgFiveRerankCountThreeEightFourMedianMs}{14.004\xspace}
\newcommand{\nWorkCudaTestbedIgFiveRerankMaximumMs}{20.559\xspace}
\newcommand{\nWorkCudaTestbedIgFiveRerankMedianMs}{10.817\xspace}
\newcommand{\nWorkCudaTestbedIgFiveRerankMinimumMs}{9.258\xspace}
\newcommand{\nWorkCudaTestbedIgFiveRerankPeakAllocatedMib}{1290.004\xspace}
\newcommand{\nWorkCudaTestbedIgFiveRerankPeakRssMib}{4952.086\xspace}
\newcommand{\nWorkCudaTestbedIgOneZeroRerankCountNineSixMedianMs}{10.984\xspace}
\newcommand{\nWorkCudaTestbedIgOneZeroRerankCountOneNineTwoMedianMs}{13.184\xspace}
\newcommand{\nWorkCudaTestbedIgOneZeroRerankCountOneSixMedianMs}{10.552\xspace}
\newcommand{\nWorkCudaTestbedIgOneZeroRerankCountSevenSixEightMedianMs}{31.051\xspace}
\newcommand{\nWorkCudaTestbedIgOneZeroRerankCountThreeEightFourMedianMs}{19.306\xspace}
\newcommand{\nWorkCudaTestbedIgOneZeroRerankMaximumMs}{18.761\xspace}
\newcommand{\nWorkCudaTestbedIgOneZeroRerankMedianMs}{10.984\xspace}
\newcommand{\nWorkCudaTestbedIgOneZeroRerankMinimumMs}{9.313\xspace}
\newcommand{\nWorkCudaTestbedIgOneZeroRerankPeakAllocatedMib}{1736.402\xspace}
\newcommand{\nWorkCudaTestbedIgOneZeroRerankPeakRssMib}{4952.086\xspace}
\newcommand{\nWorkCudaTestbedMarginCountNineSixMedianMs}{10.534\xspace}
\newcommand{\nWorkCudaTestbedMarginCountOneNineTwoMedianMs}{10.667\xspace}
\newcommand{\nWorkCudaTestbedMarginCountOneSixMedianMs}{10.504\xspace}
\newcommand{\nWorkCudaTestbedMarginCountSevenSixEightMedianMs}{11.607\xspace}
\newcommand{\nWorkCudaTestbedMarginCountThreeEightFourMedianMs}{11.076\xspace}
\newcommand{\nWorkCudaTestbedMarginMaximumMs}{17.009\xspace}
\newcommand{\nWorkCudaTestbedMarginMedianMs}{10.534\xspace}
\newcommand{\nWorkCudaTestbedMarginMinimumMs}{9.096\xspace}
\newcommand{\nWorkCudaTestbedMarginPeakAllocatedMib}{878.220\xspace}
\newcommand{\nWorkCudaTestbedMarginPeakRssMib}{4952.086\xspace}
\newcommand{\nWorkCudaTestbedSoftmaxFixedCountNineSixMedianMs}{6.583\xspace}
\newcommand{\nWorkCudaTestbedSoftmaxFixedCountOneNineTwoMedianMs}{6.561\xspace}
\newcommand{\nWorkCudaTestbedSoftmaxFixedCountOneSixMedianMs}{6.529\xspace}
\newcommand{\nWorkCudaTestbedSoftmaxFixedCountSevenSixEightMedianMs}{6.750\xspace}
\newcommand{\nWorkCudaTestbedSoftmaxFixedCountThreeEightFourMedianMs}{6.558\xspace}
\newcommand{\nWorkCudaTestbedSoftmaxFixedMaximumMs}{11.568\xspace}
\newcommand{\nWorkCudaTestbedSoftmaxFixedMedianMs}{6.583\xspace}
\newcommand{\nWorkCudaTestbedSoftmaxFixedMinimumMs}{5.519\xspace}
\newcommand{\nWorkCudaTestbedSoftmaxFixedPeakAllocatedMib}{915.078\xspace}
\newcommand{\nWorkCudaTestbedSoftmaxFixedPeakRssMib}{4952.086\xspace}
\newcommand{\nWorkCudaTestbedSoftmaxRerankCountNineSixMedianMs}{6.724\xspace}
\newcommand{\nWorkCudaTestbedSoftmaxRerankCountOneNineTwoMedianMs}{6.997\xspace}
\newcommand{\nWorkCudaTestbedSoftmaxRerankCountOneSixMedianMs}{6.563\xspace}
\newcommand{\nWorkCudaTestbedSoftmaxRerankCountSevenSixEightMedianMs}{8.533\xspace}
\newcommand{\nWorkCudaTestbedSoftmaxRerankCountThreeEightFourMedianMs}{7.564\xspace}
\newcommand{\nWorkCudaTestbedSoftmaxRerankMaximumMs}{10.329\xspace}
\newcommand{\nWorkCudaTestbedSoftmaxRerankMedianMs}{6.724\xspace}
\newcommand{\nWorkCudaTestbedSoftmaxRerankMinimumMs}{5.815\xspace}
\newcommand{\nWorkCudaTestbedSoftmaxRerankPeakAllocatedMib}{925.433\xspace}
\newcommand{\nWorkCudaTestbedSoftmaxRerankPeakRssMib}{4952.086\xspace}
\newcommand{\nWorkReleaseFixedError}{0.564\xspace}
\newcommand{\nWorkReleaseMarginError}{0.033\xspace}
\newcommand{\nWorkReleaseQueries}{48\xspace}
\newcommand{\nWorkReleaseSoftmaxFixedError}{0.552\xspace}
\newcommand{\nWorkReleaseSoftmaxRerankError}{0.024\xspace}
\newcommand{\nWorkTestbedFixedError}{0.790\xspace}
\newcommand{\nWorkTestbedIgFiveRerankError}{$9.0\times 10^{-6}$\xspace}
\newcommand{\nWorkTestbedIgOneZeroRerankError}{$2.2\times 10^{-6}$\xspace}
\newcommand{\nWorkTestbedMarginError}{0.017\xspace}
\newcommand{\nWorkTestbedQueries}{96\xspace}
\newcommand{\nWorkTestbedSoftmaxFixedError}{0.788\xspace}
\newcommand{\nWorkTestbedSoftmaxRerankError}{0.013\xspace}
\newcommand{\nLicenceNameA}{MIT\xspace}
\newcommand{\nLicenceNameB}{MIT\xspace}
\newcommand{\nLicenceNameC}{MIT\xspace}
\newcommand{\nLicenceNameD}{MIT\xspace}
\newcommand{\nLicenceNameE}{none stated\xspace}
\newcommand{\nLicenceNameR}{MIT\xspace}
\newcommand{\nModelNameA}{Pythia-70M\xspace}
\newcommand{\nModelNameB}{Pythia-70M\xspace}
\newcommand{\nModelNameC}{Pythia-70M\xspace}
\newcommand{\nModelNameD}{Pythia-160M\xspace}
\newcommand{\nModelNameE}{Pythia-70M-deduped\xspace}
\newcommand{\nModelNameF}{Pythia-70M\xspace}
\newcommand{\nModelNameR}{Pythia-70M-deduped\xspace}
\newcommand{\nRuleNameA}{TopK\xspace}
\newcommand{\nRuleNameB}{TopK\xspace}
\newcommand{\nRuleNameC}{TopK\xspace}
\newcommand{\nRuleNameD}{TopK\xspace}
\newcommand{\nRuleNameE}{TopK\xspace}
\newcommand{\nRuleNameR}{ReLU\xspace}
\newcommand{\nRunsEachConfiguration}{8\xspace}
\newcommand{\nPathErrPublicIgFive}{$2.8\times 10^{-5}$\xspace}
\newcommand{\nPathErrPublicIgTen}{$7.0\times 10^{-6}$\xspace}
\newcommand{\nPathKnownBound}{0.002\xspace}
\newcommand{\nPathKnownShareMax}{0.0028\xspace}
\newcommand{\nWorkMarginExactRatio}{3.5\xspace}
\newcommand{\nAbstractErrorReduction}{84\xspace}
\newcommand{\nPathFailedComparisonsWords}{four\xspace}
\newcommand{\nPathAllowedShare}{1\xspace}
\newcommand{\nDiscussionTenthWords}{tenth\xspace}
\newcommand{\nDiscussionPublicPairsWords}{five\xspace}

\newlength{\declA}\AtBeginDocument{\setlength{\declA}{0.27\textwidth}}
\newlength{\declB}\AtBeginDocument{\setlength{\declB}{0.3\textwidth}}
\newlength{\declC}\AtBeginDocument{\setlength{\declC}{0.2\textwidth}}
\newlength{\declD}\AtBeginDocument{\setlength{\declD}{0.62\textwidth}}

\newlength{\proofSeed}\AtBeginDocument{\setlength{\proofSeed}{0.10\textwidth}}
\newlength{\proofConstruction}\AtBeginDocument{\setlength{\proofConstruction}{0.29\textwidth}}
\newlength{\proofOutput}\AtBeginDocument{\setlength{\proofOutput}{0.52\textwidth}}

\newtheorem{theorem}{Theorem}
\newtheorem{proposition}{Proposition}
\newtheorem{corollary}{Corollary}

\title{Sparse Autoencoder Attribution Graphs Are Decided at the Margin}

\newcommand{\authoremail}{kl3685@columbia.edu}

\author{%
  Karl Li \\
  Columbia University \\
\texttt{\authoremail}
}

\hypersetup{pdfauthor={Karl Li}}
\begin{document}

\maketitle

\begin{abstract}
Interpretability research writes a language model's activation as a weighted sum of learned directions, called features. A sparse autoencoder (SAE) learns the features and gives each a coefficient, its weight in that sum. The SAE's encoder first gives every feature a score. A TopK SAE selects the $k$ features with the largest positive scores and zeroes the rest. A feature's \emph{margin} is how far its score must move before the selection changes. An intervention on one feature's coefficient changes the encoder scores of the SAE at the next layer. An attribution graph predicts that change from the scores' derivatives with respect to the coefficient, taken before the intervention. The graph keeps the original selection, so it misses every feature that enters or leaves. We propose the \emph{margin graph}, which adds the predicted change to every score and selects again. When one feature is zeroed in a public pair of \nModelNameA SAEs, the median error of the predicted coefficient change, relative to the true change, falls from \nFixedErrZeroA for the attribution graph to \nMarginErrZeroA for the margin graph. Written back into the model, the margin graph's coefficients predict the change in the model's output on subject verb agreement less accurately than the attribution graph's. Selecting again better predicts which features change, but on the task and pair we tested, not the model's output.
\end{abstract}

\section{Introduction}
\label{sec:intro}

Sparse autoencoders (SAEs) are a main tool for reading the internal activations of a language model \citep{bricken2023towards,marks2025sparse,ameisen2025circuit}. An SAE writes the activation vector at one layer and one token position as a weighted sum of a few learned directions, called features, and the weight of each feature is its coefficient. Its encoder first gives every feature a score, and a selection rule then decides which features are selected and receive a nonzero coefficient. A TopK SAE selects the $k$ features with the largest positive encoder scores, sets the coefficient of each to its score and sets every other coefficient to zero \citep{gao2025scaling}. The public TopK SAEs on \nModelNameA \citep{biderman2023pythia}, for example, select \nKA of their \nWidthA features at each position \citep{eleutherai2024sae70m}. The $k$th largest score is therefore a cutoff, and an unselected feature keeps a coefficient of zero however close its score comes to the cutoff.

\begin{figure}[t]
\centering
\includegraphics{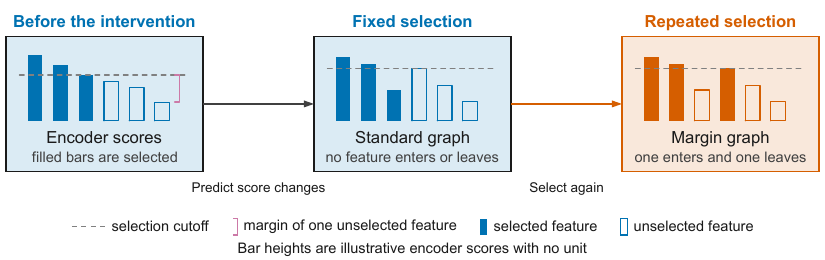}
\caption{\textbf{A small change in encoder scores can alter which features a TopK sparse autoencoder selects.} Bars plot encoder scores before an intervention and as two graphs predict them. A feature's margin is how far its score is from the cutoff that decides its selection. The \emph{standard graph} keeps the original selection, and the margin graph selects again.}
\label{fig:overview}
\end{figure}

To learn what a feature does, a researcher intervenes on it, for example by setting its coefficient to zero, and asks how the coefficients of the SAE at the next layer change \citep{marks2025sparse,ameisen2025circuit}. Running the model again answers this question exactly for one intervention. An attribution graph answers it to first order for every intervention at once \citep{ameisen2025circuit,farnik2025jacobian}. Each entry of the graph is the derivative of one encoder score at the next layer with respect to one coefficient at the earlier layer, taken before the intervention. The graph's entries for the intervened coefficient, multiplied by the change the intervention makes to it, predict how every encoder score at the next layer changes. The graph then adds these predicted changes to the coefficients of the features that were selected before the intervention \citep{ameisen2025circuit,marks2025sparse}.

That last step ignores the cutoff. A feature whose score the intervention lifts past the cutoff is selected afterwards, and its coefficient jumps from zero to near the cutoff, yet the graph predicts no change for it (Figure~\ref{fig:overview}). A selected feature whose score falls below the cutoff leaves the selection in the same way, and its coefficient drops to zero. Such jumps are large even when the intervention is small. We therefore ask how often interventions change the selection of real SAEs, and whether a graph that selects features again predicts the coefficients after an intervention better.

We call the distance a feature's encoder score must move before the feature enters or leaves the selection its \emph{margin}. An intervention that moves a feature's score farther than its margin changes the selection and makes that feature's coefficient jump, so the intervention's effect is decided at the margin. We show that an intervention's effect on a TopK SAE is decided at each feature's margin. We propose the margin graph, which adds the predicted score change to every encoder score at the next layer, selected or not, and then applies the TopK rule again, so it predicts which features enter and which leave. We call the usual prediction, which adds the same changes but keeps the original selection, the standard graph.

We judge each prediction by its relative coefficient error, the norm of its error divided by the norm of the exact coefficient change, so that zero is exact and predicting no change scores one. Our findings follow.
\begin{enumerate}
\item Zeroing one feature changes the next layer's selection in \nCrossZeroMin to \nCrossZeroMax of cases on \nDiscussionPublicPairsWords public pairs of TopK SAEs, each an SAE and the SAE at the next layer. The entering and leaving features account for most of the coefficient change (Section~\ref{sec:crossing}).
\item The margin graph is exact wherever the model's block from one layer to the next is affine. On the public \nModelNameA pair at layers \nLayerUpA and \nLayerDownA, its median error after zeroing one feature is \nAbstractErrorReduction{}\% lower than the standard graph's (Section~\ref{sec:margin}).
\item Attribution patching estimates an intervention's effect to first order, as an attribution graph does. Selecting again also lowers the error of two published corrections of it, integrated gradients, which average derivatives along the intervention, and the softmax correction, which recomputes the attention weights before the first order step (Section~\ref{sec:margin}).
\item Exact recomputation is still faster at the intervention counts we timed, and on a graphics processing unit the margin graph takes \nWorkMarginExactRatio times as long (Section~\ref{sec:margin}).
\item Written back into the model, the margin graph's more accurate coefficients predict how an intervention changes the model's preference for the verb form that agrees with its subject less accurately than the standard graph's coefficients do. The cause is that the entering and leaving features, decoded into the model's activation, give a larger change than the one the model itself makes (Section~\ref{sec:readout}).
\item Graphs of different training seeds are often compared \citep{paulo2026different,gerasimov2026unstable}. Transferred between seeds, a graph loses much of itself when each seed keeps only its own selected features, and the lost part lies on features just below the cutoff (Section~\ref{sec:transport}).
\end{enumerate}

\section{Setup}
\label{sec:setup}

We measure how scaling one upstream feature changes the coefficients of the SAE one layer downstream, on \nDiscussionPublicPairsWords public TopK SAE pairs, one public ReLU pair and the pairs we trained. Every later section uses the SAEs, the margin, the two graphs and the error that this section defines.

\textbf{SAEs and selection rules.} An SAE of width $m$ encodes the residual stream, of width $d$, at one layer of a language model. The residual stream is the activation vector passed between the model's blocks. The encoder, with rows $e_j^\top$, maps an activation $x$ to the encoder scores $p = E(x - b_{\mathrm{pre}}) + b_{\mathrm{enc}}$, which the literature calls preactivations. The vectors $b_{\mathrm{pre}}$ and $b_{\mathrm{enc}}$ are the bias subtracted before encoding and the bias added to the encoder scores. A selection rule turns the encoder scores into the coefficient vector $z$, and the decoder has one unit column for each feature. A TopK rule keeps the $k$ largest positive encoder scores as coefficients and sets every other coefficient to zero \citep{gao2025scaling}. We call the features it keeps selected, as the field calls them active, and we write $S$ for the selected set and $\theta_k$ for the cutoff, the $k$th largest encoder score. A ReLU rule selects every positive encoder score \citep{bricken2023towards}, and a JumpReLU rule selects each encoder score above its feature's threshold \citep{rajamanoharan2024jumping}. A feature enters when an intervention makes it selected, and it leaves when the intervention makes it unselected.

\textbf{The SAE pairs.} We measure the pairs of Table~\ref{tab:releases}, each an upstream SAE and a downstream SAE one layer later. Most comparisons below report the public pair on \nModelNameA at layers \nLayerUpA and \nLayerDownA, whose SAEs have width \nWidthA and select \nKA features \citep{eleutherai2024sae70m}. The same release adds pairs at layers \nLayerUpB and \nLayerDownB and at the last of its \nBlocksA blocks. The release on \nModelNameD adds the middle of its \nBlocksD blocks \citep{eleutherai2024sae160m}, and a deduplicated TopK release shares its model and layers with the ReLU SAEs \citep{eleutherai2024sae70mdeduped,marks2024pythiasaes}. We encode and decode each SAE as the library that trained it does \citep{eleutherai2025sparsify,marks2024dictionary}.

We also trained TopK SAEs on \nModelNameF at layers \nLayerUpA and \nLayerDownA in \nRunsF runs, one pair per run, and we call them our SAEs. Six runs start from different initialization seeds. The other two share an initialization and see the training data in different orders, and we call them the data order pair. All \nRunsF runs share one auxiliary loss coefficient, which Appendix~\ref{app:releases} gives. We trained them because no public release offers a second training seed, and Section~\ref{sec:transport} compares their seeds.

\begin{table}[t]
\caption{\textbf{Our sparse autoencoders (SAEs) reconstruct their layers about as closely as the public SAEs.} Each row is a pair of SAEs at one layer and the next. The fraction of variance unexplained (FVU) divides summed squared reconstruction error by summed squared deviation from the mean, so lower is closer.}
\label{tab:releases}
\centering
\footnotesize
\setlength{\tabcolsep}{2.5pt}
\begin{tabular}{llllrrlrr}
\toprule
source & model & layers & rule & width & k & budget & FVU up & FVU down \\
\midrule
public & \nModelNameA & \nLayerUpA, \nLayerDownA & \nRuleNameA & \nWidthA & \nKA & \nBudgetTokensA B & \nFVUUpA & \nFVUDownA \\
public & \nModelNameB & \nLayerUpB, \nLayerDownB & \nRuleNameB & \nWidthB & \nKB & \nBudgetTokensA B & \nFVUUpB & \nFVUDownB \\
public & \nModelNameC & \nLayerUpC, \nLayerDownC & \nRuleNameC & \nWidthC & \nKC & \nBudgetTokensA B & \nFVUUpC & \nFVUDownC \\
public & \nModelNameD & \nLayerUpD, \nLayerDownD & \nRuleNameD & \nWidthD & \nKD & \nBudgetTokensA B & \nFVUUpD & \nFVUDownD \\
public & \nModelNameE & \nLayerUpE, \nLayerDownE & \nRuleNameE & \nWidthE & \nKE & none stated & \nFVUUpE & \nFVUDownE \\
public & \nModelNameR & \nLayerUpR, \nLayerDownR & \nRuleNameR & \nWidthR & \nKR & \nBudgetTokensR B & \nFVUUpR & \nFVUDownR \\
ours & \nModelNameF & \nLayerUpA, \nLayerDownA & TopK & \nWidthF & \nKF & \nActivationsF & \begin{tabular}[c]{@{}c@{}}\nFVUThreeMinF to\\\nFVUThreeMaxF\end{tabular} & \begin{tabular}[c]{@{}c@{}}\nFVUFourMinF to\\\nFVUFourMaxF\end{tabular} \\
\bottomrule
\end{tabular}

\par\noindent Width counts features, and $k$ is the number that TopK selects. TopK selects the largest positive encoder scores, and ReLU, the rectified linear unit, selects every positive score. A budget in B counts billions of training tokens, and the budget of our SAEs counts activations. Each FVU is measured at \nTokens held out Pile validation positions, and the FVU cells of our SAEs give the range over their \nRunsF runs.
\end{table}

Our SAEs reconstruct their layers about as closely as the public \nModelNameA pair at layers \nLayerUpA and \nLayerDownA, which leaves \nFVUUpA and \nFVUDownA of the variance unexplained at its two layers. We list this fraction of variance unexplained (FVU) for every SAE in Table~\ref{tab:releases}, beside the training budget its release states. Its column FVU up measures the upstream SAE, and FVU down measures the downstream SAE. The FVU is a share of variance. The reconstruction residual is a vector, the activation less its reconstruction, and Section~\ref{sec:readout} measures how that vector changes.

\textbf{The margin.} Intuitively, a feature's margin says how far its own encoder score must move, with every other score held fixed, before the selection rule changes its decision about that feature. An unselected feature enters once its score rises above both the lowest selected score and zero. A selected feature leaves once its score falls below the larger of zero and the highest unselected score.

Precisely, write $\theta_k$ and $\theta_{k+1}$ for the $k$th and the $(k+1)$th largest encoder scores at a position. The margin of feature $j$ is
\begin{equation}
\label{eq:margin}
\mu_j = p_j - \max(\theta_{k+1}, 0) \quad \text{for } j \in S , \qquad \mu_j = p_j - \max(\theta_k, 0) \quad \text{for } j \notin S .
\end{equation}
Every statement below assumes no ties, meaning that no two encoder scores it uses are equal and none is zero. Under that assumption a feature is selected exactly when its margin is positive. Under a ReLU rule the margin is $p_j$ itself, and under a JumpReLU rule it is $p_j$ less the feature's threshold.

\textbf{The two graphs.} Both graphs start from the same first order prediction of how the downstream encoder scores move. They differ only in how they turn the predicted scores into coefficients, since the standard graph keeps the original selection and the margin graph selects again.

Write $D$ for the upstream SAE's decoder, whose unit columns $d_i$ are the upstream features' directions, and $E$ and $G$ for the downstream SAE's encoder and decoder. The block $f$ maps the upstream residual stream at a position to the downstream one, with every other position held fixed, and $J$ is its Jacobian before the intervention. The attribution graph is the matrix $W = EJD$. Its entry $W_{ji}$ is the derivative of downstream encoder score $j$ with respect to upstream coefficient $i$, the virtual weight of \citet{ameisen2025circuit}, whose graph nodes ``represent active features'', the features selected before the intervention.

An intervention scales the coefficient $u_i$ of one selected upstream feature by a factor $\alpha$, which adds $v = (\alpha - 1) u_i d_i$ to the upstream activation. The factor zero is the zero ablation. To first order the downstream encoder scores change by the predicted score change $\widehat{\Delta p} = (\alpha - 1)\, u_i\, W_{:i}$, which is column $i$ of the graph scaled. Let $z$ be the downstream coefficient vector before the intervention and $P$ the mask of its selected set. The two graphs predict the coefficient vector after the intervention as
\begin{equation}
\label{eq:graphs}
\hat z^{\,\mathrm{standard}} = z + P\,\widehat{\Delta p} , \qquad \hat z^{\,\mathrm{margin}} = \mathrm{TopK}\big(p + \widehat{\Delta p}\big) ,
\end{equation}
where TopK keeps the $k$ largest positive entries of its argument and sets the rest to zero. The mask $P$ keeps the coordinates of the features selected before the intervention and sets the other coordinates to zero. Subtracting $z$ from either prediction gives its predicted coefficient change $\widehat{\Delta z}$. The standard graph puts the whole predicted change on the features selected before the intervention. The margin graph selects again among the predicted encoder scores, so a feature enters or leaves when its predicted margin changes sign. Under a JumpReLU rule the same prediction keeps each predicted encoder score above its feature's threshold.

Both graphs read each column of $W$ from one Jacobian vector product (JVP) before the intervention, a single forward derivative of the block along one direction. One JVP for each selected upstream feature at a position therefore serves every intervention there, at every scaling factor and on every subset of features, whose prediction sums its members' columns. The standard graph keeps only the rows of the selected downstream features, and the margin graph keeps every row, unselected features included.

\textbf{The error.} A prediction is good when its coefficient change is close to the exact one, measured against the size of the exact change. A prediction $\widehat{\Delta z}$ of the exact change $\Delta z$ of the downstream coefficients has the relative coefficient error
\begin{equation}
\label{eq:error}
\|\widehat{\Delta z} - \Delta z\| \, / \, \|\Delta z\|
\end{equation}
over every coordinate of the downstream SAE, with the Euclidean norm. Zero is exact, and predicting no change gives one. A graph's error is the median of \eqref{eq:error} over interventions. The symbols used for the prediction and for transfer are in Table~\ref{tab:discussion-terms}.

\textbf{Units and positions.} We measure every SAE in the residual stream's units, each decoder column at unit norm, so that coefficients and cutoffs compare across pairs. Our SAEs normalize each activation by a mean and a scale computed on a sample held out before training, and we convert their coefficients to the stream's units (Appendix~\ref{app:releases}). Each measurement takes one position from each sequence of \nSeqLen tokens of the Pile's validation file \citep{gao2020pile}, on which our SAEs did not train. We intervene on every selected upstream feature at the first \nTokens positions, at the zero ablation and at the scaling factors \nDoseHalf and \nDoseNine.

We fixed the pairs, the positions, the three scaling factors, the pooled median as the summary of a graph's error and every criterion of Appendix~\ref{app:declarations} before we evaluated any predictor. We mark with $^\dagger$ each analysis added after the main results, and we fixed each such analysis before seeing its own result.

\begin{table}[t]
\caption{\textbf{The symbols separate encoder scores, computed before selection, from coefficients, which only selected features keep.} Each graph predicts the downstream coefficients from the same upstream intervention. The subscripts $A$ and $B$ name the source and target runs of the transfer in Section~\ref{sec:transport}.}
\label{tab:discussion-terms}
\centering
\small
\begin{tabular}{p{0.23\textwidth}p{0.71\textwidth}}
\toprule
symbol & meaning \\
\midrule
$p$, $z$ & Downstream encoder scores before selection, and coefficients after selection. \\
$S$, $P$ & Features selected before the intervention, and the mask that keeps their coordinates. \\
$k$, $\theta_k$ & Largest selected count, and the cutoff, the encoder score at rank $k$. \\
$\mu_j$ & Margin, the signed distance of feature $j$'s score from a change of its selection state. \\
$u_i$, $d_i$, $v$ & Upstream feature coefficient, its decoder direction, and the activation change the intervention applies. \\
$D$, $E$, $G$ & Upstream decoder, downstream encoder and downstream decoder. \\
$J$, $W_{:i}$ & Jacobian of the block, and column $i$ of the attribution graph $W=EJD$. \\
$\widehat{\Delta p}$, $\Delta z$ & Predicted score change, and exact change in the coefficients. \\
$A$, $B$ & Source and target SAE runs of the transfer. \\
$M$, $N$ & Sparse maps, with $D_A\approx D_BM$ and $E_A\approx NE_B$. \\
$P_A$, $Q_A$ & Masks of the source's downstream and upstream selected features. \\
$P_B$, $Q_B$ & The same masks for the target. \\
\bottomrule
\end{tabular}

\par\noindent A sparse autoencoder (SAE) represents a model activation by sparse feature coefficients. A selected feature has a nonzero coefficient, and a mask keeps the selected features' coordinates and sets the others to zero.
\end{table}

\section{Selection changes}
\label{sec:crossing}

Zeroing one upstream feature changes the downstream selection in \nCrossZeroMin to \nCrossZeroMax of cases on the \nDiscussionPublicPairsWords public TopK pairs, \nCrossZeroA on the \nModelNameA pair at layers \nLayerUpA and \nLayerDownA, and \nCrossZeroF on our SAEs. The standard graph assumes that the selection before an intervention survives it. This section measures how often that assumption fails and whether the closeness of unselected features to the cutoff explains the failures. Before any intervention, the unselected feature nearest the cutoff on the public TopK pairs has a median margin over positions between $\nNearestInactiveMin$ and $\nNearestInactiveMax$ in the residual stream's units, while the cutoff is \nCutoffOverLargestMin to \nCutoffOverLargestMax of the largest encoder score at the same position. A small change can therefore push such a feature across the cutoff.

\begin{figure}[t]
\centering
\includegraphics[width=\textwidth]{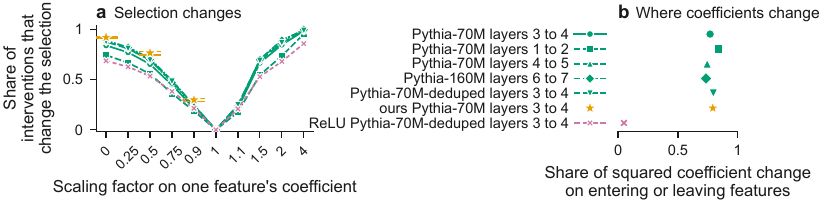}
\caption{\textbf{Most zero ablations change which features the next layer's TopK sparse autoencoder selects, and the entering and leaving features carry most of the coefficient change.} Panel~a plots how often an intervention changes that selection against the scaling factor on one feature's coefficient, and zero removes the feature. Panel~b gives the mean share of the squared coefficient change on entering or leaving features, over zero ablations that change the selection. Ours marks the mean of our \nRunsF runs, measured at zero, \nDoseHalf and \nDoseNine, with bars for their range. ReLU marks the pair that keeps every positive score.}
\label{fig:crossing}
\end{figure}

\textbf{Milder interventions change the selection less often.} As the scaling factor nears one, the share of interventions that change the selection falls to \nCrossNineB to \nCrossNineE at \nDoseNine, and it is zero at one, where the input is unchanged. Past one it rises again, to \nCrossQuadB to \nCrossQuadC at \nDoseQuad (Figure~\ref{fig:crossing}).

\textbf{Most of the change lies on the features that enter or leave.} Over the zero ablations that change the selection, the entering and leaving features account for a mean \nEnergyZeroMin to \nEnergyZeroMax of the squared norm of the coefficient change on the public pairs and \nEnergyZeroF on our SAEs. The entering features alone account for \nEnterShareMin to \nEnterShareMax. At the median, the coefficient of such a feature moves \nDzDpMin to \nDzDpMax times as far as its encoder score, since an entering coefficient jumps from zero to near the cutoff and a leaving coefficient drops to zero. On the ReLU pair, whose coefficients are continuous at zero, the features that cross zero account for only \nEnergyZeroR, although a zero ablation moves some feature across zero in \nCrossZeroR of cases (Appendix~\ref{app:releases}).

\section{The margin graph}
\label{sec:margin}

On the public \nModelNameA pair at layers \nLayerUpA and \nLayerDownA, the margin graph lowers the median relative coefficient error at the zero ablation from \nFixedErrZeroA to \nMarginErrZeroA. This section first shows that the margin graph is exact where the block is affine. It then measures the gain on real SAEs, isolates the part of the gain due to selecting again, applies the same step to two published corrections and compares the cost with exact recomputation.

Intuitively, if the block changes the downstream activation exactly linearly along the intervention, then the predicted score change is the exact score change, and selecting again from exact scores gives the exact coefficients.

\begin{theorem}[exactness on an affine segment]
\label{thm:exact}
Assume no ties and that the block is affine along the intervention, $f(x + tv) = f(x) + tJv$ for $t \in [0,1]$. Then the margin graph's prediction equals the exact downstream coefficient vector after the intervention, entering and leaving features included.
\end{theorem}

By Theorem~\ref{thm:exact}, all of the margin graph's error comes from the curvature of the block. On an affine segment the two graphs differ only on the entering and leaving features. The standard graph's relative error is then at least the square root of the entering features' share of the squared norm of the coefficient change. An entering feature's coefficient jumps from zero to at least the cutoff less the largest predicted change of any downstream encoder score, which is why such a feature moves the coefficients so far. We prove both statements in Corollary~\ref{cor:fixed} of Appendix~\ref{app:proofs}.

\textbf{The margin graph removes most of the error.} Real blocks are not affine, so this comparison measures how much of the standard graph's error selecting again removes in practice. On our SAEs the margin graph errs by \nMarginErrDoseFZero at the zero ablation, against \nFixedErrDoseFZero for the standard graph. Both graphs' median errors on every pair, at the zero ablation and at the scaling factors \nDoseHalf and \nDoseNine, are in Table~\ref{tab:factors}. At \nDoseNine few features enter or leave, and both graphs err little.

Pooled over the three factors, the margin graph errs by \nMarginErrMin to \nMarginErrMax on the public TopK pairs, against \nFixedErrMin to \nFixedErrMax for the standard graph. In all \nBootDraws resamples of the positions$^\dagger$ the difference stays above zero on every pair.

\begin{table}[t]
\caption{\textbf{The margin graph errs less than the standard graph on every pair at every scaling factor.} Cells give the median relative coefficient error, the norm of the prediction's error over the norm of the exact coefficient change, so smaller is better. Column heads name a graph and the scaling factor on one upstream feature's coefficient, where zero removes it.}
\label{tab:factors}
\centering
\scriptsize
\setlength{\tabcolsep}{2.4pt}
\begin{tabular}{lrrrrrr}
\toprule
pair & \begin{tabular}[c]{@{}c@{}}standard\\at zero\end{tabular} & \begin{tabular}[c]{@{}c@{}}margin\\at zero\end{tabular} & \begin{tabular}[c]{@{}c@{}}standard\\at \nDoseHalf\end{tabular} & \begin{tabular}[c]{@{}c@{}}margin\\at \nDoseHalf\end{tabular} & \begin{tabular}[c]{@{}c@{}}standard\\at \nDoseNine\end{tabular} & \begin{tabular}[c]{@{}c@{}}margin\\at \nDoseNine\end{tabular} \\
\midrule
\nModelNameA L\nLayerUpA to L\nLayerDownA & \nFixedErrZeroA & \nMarginErrZeroA & \nFixedErrHalfA & \nMarginErrHalfA & \nFixedErrNineA & \nMarginErrNineA \\
\nModelNameB L\nLayerUpB to L\nLayerDownB & \nFixedErrZeroB & \nMarginErrZeroB & \nFixedErrHalfB & \nMarginErrHalfB & \nFixedErrNineB & \nMarginErrNineB \\
\nModelNameC L\nLayerUpC to L\nLayerDownC & \nFixedErrZeroC & \nMarginErrZeroC & \nFixedErrHalfC & \nMarginErrHalfC & \nFixedErrNineC & \nMarginErrNineC \\
\nModelNameD L\nLayerUpD to L\nLayerDownD & \nFixedErrZeroD & \nMarginErrZeroD & \nFixedErrHalfD & \nMarginErrHalfD & \nFixedErrNineD & \nMarginErrNineD \\
\nModelNameE L\nLayerUpE to L\nLayerDownE & \nFixedErrZeroE & \nMarginErrZeroE & \nFixedErrHalfE & \nMarginErrHalfE & \nFixedErrNineE & \nMarginErrNineE \\
ReLU \nModelNameR L\nLayerUpR to L\nLayerDownR & \nFixedErrZeroR & \nMarginErrZeroR & \nFixedErrHalfR & \nMarginErrHalfR & \nFixedErrNineR & \nMarginErrNineR \\
ours & \nFixedErrDoseFZero & \nMarginErrDoseFZero & \nFixedErrDoseFHalf & \nMarginErrDoseFHalf & \nFixedErrDoseFNine & \nMarginErrDoseFNine \\
\bottomrule
\end{tabular}

\par\noindent Each pair is an upstream sparse autoencoder (SAE) and the SAE at the next layer, and each cell covers every selected upstream feature at \nTokens held out Pile validation positions. Our SAEs, of width \nWidthF and TopK \nKF, average \nRunsF run medians. The standard graph keeps the features selected before the intervention, and the margin graph selects again from the predicted scores. TopK selects the largest positive encoder scores, ReLU, the rectified linear unit, selects every positive score, and L names a model layer.
\end{table}

\textbf{Selecting again does most of the work.} The margin graph differs from the standard graph in two ways, since it keeps every row of the graph and it selects again. A control isolates the second. The control keeps every row but judges each feature against the cutoff from before the intervention. It errs by \nExtraReleaseErrThreshold on the public \nModelNameA pair at layers \nLayerUpA and \nLayerDownA, where the margin graph errs by \nExtraReleaseErrMargin. On our SAEs the two err by means of run medians of \nNoRerankErrF and \nMarginErrF. These errors pool the three scaling factors of the predictor comparison.

\textbf{The margin graph identifies the entering features.} We compare the features that the margin graph lets enter with those that enter, apart from their predicted size. At its own count it finds them on the public pairs with a precision of \nPrecMin to \nPrecMax and a recall of \nRecMin to \nRecMax. Every other predictor receives the true number of entering features, which leaves only the question of which features enter (Table~\ref{tab:support}). Given that count, the margin graph's ranking reaches a precision of \nPrecTrueCountA on the public \nModelNameA pair, the unselected features smallest in margin reach \nPrecMarginAloneA, and those with the largest predicted score increase almost never enter. A feature therefore enters by its margin and its predicted change together, and Appendix~\ref{app:releases} also scores the leaving features.

\textbf{Selecting again also improves two published corrections.} Two corrections of attribution patching predict the score change better than the first order term, at a higher cost. Integrated gradients (IG) average the derivative along each intervention's path \citep{sundararajan2017axiomatic}. The softmax correction recomputes the attention weights exactly at the intervened input and keeps the rest of the block at first order \citep{kramar2024atp,hanna2024faith}. Both corrections err less once they select again (Figure~\ref{fig:corrections} in Appendix~\ref{app:releases}). Counting derivative evaluations without reusing those from before the intervention, IG that selects again errs by \nDevIgFiveRerank at \nDevIgFiveRerankProducts JVPs per intervention on our SAEs, against the margin graph's \nDevMargin at one JVP. The softmax correction that selects again adds a partial attention pass, and it errs by \nDevSoftmaxRerank on our SAEs and by \nSmRerankPubMin to \nSmRerankPubMax on the public TopK pairs$^\dagger$, less than the margin graph on all \nRunsF runs and all \nSmPairsBelow public pairs.

\textbf{Exact recomputation is faster at our counts.} The margin graph reuses one set of derivatives across a position's interventions, while exact recomputation runs the block again for each one. We time the construction of a predictor and all of a position's interventions together. At these counts the margin graph takes longer than exact recomputation on both machines, \nWorkCudaReleaseMarginMedianShort against \nWorkCudaReleaseExactMedianShort ms for a position's \nWorkReleaseQueries interventions on the \nDeviceName, \nWorkMarginExactRatio times as long. Among the counts we measured, it first runs faster than recomputation at \nWorkCpuReleaseMarginCrossover interventions on the \nCostProcessor, for the public \nModelNameA pair and for our SAEs, and never through \nWorkCountFive on the \nDeviceName. A reader who needs only the changed coefficients should recompute them, and Appendix~\ref{app:cost} gives every predictor's time, memory and coefficient error.

Where the block is not affine the first order prediction errs. A selection state is the selected or unselected state of one feature after one intervention. Intuitively, a predicted state cannot change when the feature's predicted margin exceeds the possible error of its own score plus that of its nearest competitor across the cutoff. Theorem~\ref{thm:band} makes this exact, with $t$ tracing the segment from the activation before the intervention to its intervened value and $\|\cdot\|_2$ the spectral norm.

\begin{theorem}[the first order error]
\label{thm:band}
Assume no ties and $\|J(x + tv) - J(x)\|_2 \le L_v\, t\, \|v\|$ for $t \in [0,1]$. Then each downstream encoder score's first order error is at most $\varepsilon_j = \|e_j\| L_v \|v\|^2/2$, and a feature whose predicted margin exceeds in size its own $\varepsilon_j$ plus the largest $\varepsilon_l$ on the other side of the predicted cutoff keeps its predicted selection state after the intervention. The largest $\varepsilon_l$ over a side with no feature is zero.
\end{theorem}

We call the range of margins too small for Theorem~\ref{thm:band} to decide the error band. It widens with the square of the intervention's norm. Under Theorem~\ref{thm:band}, every selection state outside the band keeps its predicted value wherever the bound on the Jacobian is met along the whole segment. We check the bound after the intervention, estimating $L_v$ at three points, which need not bound the whole segment, so the check is empirical. On positions held out from calibration$^\dagger$ the check accepted \nRuleDosePooledCovMin to \nRuleDosePooledCovMax of predicted selection states, and none was wrong (Appendix~\ref{app:declarations}).

\textbf{The advantage shrinks as more features are ablated.} Ablating several features at once moves the intervention farther from the affine case. The margin graph predicts a subset's ablation by summing the stored columns, with no further JVP. On our SAEs its error rises from \nMarginSizeTwoF at \nSizeTwo features to \nMarginSizeSixteenF at \nSizeSixteen, still under the standard graph's \nFixedSizeSixteenF (Table~\ref{tab:sizes} in Appendix~\ref{app:declarations}).

\textbf{At \nSizeSixteen features the order reverses on the public \nModelNameA pair.} There a subset of \nSizeSixteen features is the pair's whole selected set at every position, and the margin graph errs by \nMarginSizeSixteenRel against \nFixedSizeSixteenRel for the standard graph. IG that selects again, at \nPathProducts JVPs on each subset, errs by at most \nPathFiveRerankMaxF on our SAEs and \nPathFiveRerankMaxRel on the public pair at every size. A practitioner who ablates many features at once should therefore pay for those JVPs.

\section{The decoded change}
\label{sec:readout}

Decoded, the entering and leaving features give a change \nSwapRatioMin to \nSwapRatioMax times as large in norm as the model's own activation change, at the median on the public TopK pairs, and the change of the reconstruction residual cancels the excess. A more accurate coefficient prediction therefore need not predict the model better, and this section tests whether it does.

The downstream SAE decodes its coefficient change as $G\,\Delta z$, and we call the decoded change on the entering and leaving features the entering and leaving part. The rest of the activation change, $\Delta x - G\,\Delta z$, is the change of the reconstruction residual. On the public ReLU pair the entering and leaving part has \nSwapRatioR times the norm of the activation change. Appendix~\ref{app:releases} gives each part for every pair.

\textbf{The entering and leaving part exceeds the activation change.} Intuitively, when one feature enters as another leaves, the decoder adds one direction at about the cutoff's size and removes another. The decoded change then has a norm near the cutoff times the distance between the two directions, however small the intervention. Theorem~\ref{thm:readout} in Appendix~\ref{app:proofs} makes this exact. When one feature enters as one leaves, the part's norm is at least the cutoff times the distance between their decoder columns, less a term proportional to the largest exact change of an encoder score.

\textbf{The margin graph's patched coefficients err more.} We patch each graph's predicted downstream coefficients into the model$^\dagger$ and measure how well they predict the intervention's change in the model's output, on one task and the public \nModelNameA pair at layers \nLayerUpA and \nLayerDownA. The task is subject verb agreement, where the model should prefer the verb form that agrees with the subject. The patch replaces the coefficients at layer \nLayerDownA but keeps the reconstruction residual from before the intervention, and the model runs on from that layer. The logit difference subtracts the incorrect verb form's logit from the correct verb form's logit, and we compare its change under the patch with the change the intervention makes. The retained interventions have an absolute true change of at least \nDsFloor in that difference.

On the \nDsAbove test interventions we kept, the margin graph's coefficients err by \nDsErrMargin, \nDsExcess{}\% more than the standard graph's \nDsErrFixed, as mean absolute errors in logit units. They also err more than predicting no change, whose error is \nDsErrNull. The margin decides which features enter, yet the entering features account for \nDsEntrantShare of the intervention's change. Inside the model the residual's change cancels the excess of the entering and leaving part, and the patch leaves that change out. Appendix~\ref{app:downstream} gives the test.

\textbf{Decoded middle coefficients fail across two blocks.} We call a prediction that passes through two blocks in turn a chain. A chain that decodes its predicted middle coefficient change before the second block errs at the third layer's coefficients by \nChainFixed with the standard graph and \nChainMargin with the margin graph, while one JVP through both blocks in the residual stream errs by \nChainResidual (Appendix~\ref{app:releases}). Decoding drops the change of the middle reconstruction residual, so a chain should stay in the residual stream until its last layer.

\section{Transfer across seeds}
\label{sec:transport}

Between two training seeds of our SAEs, writing each feature vector of one seed as a sparse combination of the other seed's vectors reconstructs \nRestricted of a graph's squared norm on one seed's selected features, but \nMasked once the other seed's graph keeps only its own selected features. The part lost lies on features just below the cutoff, which have small margins. Graphs of different seeds are often compared \citep{paulo2026different,gerasimov2026unstable}, so this section measures how much of a graph transfers between seeds and where the rest is lost.

Take a source SAE $A$ and a target SAE $B$ from two seeds of one model at the same pair of layers. We write each decoder column of $A$ as a sparse nonnegative combination of the decoder columns of $B$, $D_A \approx D_B M$, and each encoder row of $A$ likewise, $E_A \approx N E_B$. We call $M$ and $N$ the sparse maps, and a decoder column or an encoder row of $B$ a target vector. Where the maps are exact, one seed's graph can be computed from the other's.

We score a transfer by how much of a reference graph it reconstructs. For an ordered pair $q$ of runs and a position $t$, let $A_{qt}$ be the reference submatrix of the graph and $\widehat A_{qt}$ its transferred prediction. The graph reconstruction score pools squared norms over positions before averaging over pairs,
\begin{equation}
\label{eq:transport-score}
s_q = 1 - \frac{\sum_t\|A_{qt}-\widehat A_{qt}\|_F^2}{\sum_t\|A_{qt}\|_F^2},
\qquad \bar s = \frac{1}{|Q|}\sum_{q\in Q}s_q .
\end{equation}
The set $Q$ contains the ordered pairs of source and target runs. The squared Frobenius norm sums the square of every matrix entry. A score of one is exact, and a negative score means the transferred prediction is worse than a zero graph.

Intuitively, a graph is a product of an encoder, a Jacobian and a decoder, so maps that rewrite the encoder and the decoder rewrite the graph. Below, $P$ and $Q$ are the masks of a run's downstream and upstream selected sets.

\begin{theorem}[transfer]
\label{thm:transport}
Let $W_A = E_A J D_A$ and $W_B = E_B J D_B$. If $D_A = D_B M$ and $E_A = N E_B$, then $W_A = N W_B M$ at every position and for any two widths. Also $P_A W_A Q_A = P_A N (P_B W_B Q_B) M Q_A$ wherever the rows of $N$ at the selected downstream features of $A$ and the columns of $M$ at its selected upstream features put no weight on features unselected in $B$.
\end{theorem}

By Theorem~\ref{thm:transport}, sparse maps transfer a graph exactly, with no matching of features. Masking each graph to its own selected set breaks the transfer only through the unselected features of $B$, whose margins are not positive. The mask depends on each margin's sign and not its size, so the part it drops for a feature appears or vanishes at once when that margin changes sign. Corollary~\ref{cor:residual} of Appendix~\ref{app:proofs} bounds the transfer's error under inexact maps.

We fit the sparse maps by nonnegative orthogonal matching pursuit \citep{nguyen2019nonnegative} over the \nPairsBetween ordered pairs of runs with different seeds and both directions of the data order pair (Appendix~\ref{app:anatomy}), and we score each transfer on \nHeldout held out positions. We make two comparisons, both in Table~\ref{tab:transport} of Appendix~\ref{app:anatomy}. The first takes the whole source graph as reference and compares sparse maps with a one to one matching, which pairs each source vector with at most one target vector (Appendix~\ref{app:anatomy}). The second takes the source graph on the source's selected features as reference. It compares the target graph unmasked, keeping every feature, with the target graph masked to its own selected features, and the drop from masking is the unmasked score less the masked score.

\textbf{Sparse maps reconstruct most of the graph.} Between seeds they reach a graph reconstruction score of \nSparseUnmasked on the whole graph, against \nOneToOne for a one to one matching. They beat every random map of matched sparsity and every permutation null on every pair. On the source's selected features, with the target unmasked, they reach \nRestricted.

\textbf{The maps recombine features.} Only \nAnatBetweenDecOne of decoder columns between seeds are fitted by a single target vector. Appendix~\ref{app:anatomy} gives the sizes of the combinations and the share of columns whose fit does not meet the tolerance.

\textbf{Masking each graph loses much of it.} With the target graph masked, the sparse maps reach \nMasked between seeds, a drop of \nGap from the unmasked target, and the data order pair drops by \nGapRetrain. By Theorem~\ref{thm:transport} the drop comes from the maps' weight on features selected in one seed and unselected in the other. Appendix~\ref{app:anatomy} splits the drop exactly into the squared norm of the part that masking removes and a signed cross term (Table~\ref{tab:transport-full}).

\textbf{The drop lies on features just below the cutoff.} If the drop comes from features near the cutoff, adding back the unselected target features nearest the cutoff should recover it first. We add unselected target features to the target's mask, nearest the cutoff first, counting the share added over all unselected target features. The nearest \nDiscussionTenthWords recovers \nRecoveredTenth of the drop between seeds and \nRecoveredTenthRetrain on the data order pair (Figure~\ref{fig:widening} in Appendix~\ref{app:anatomy}). A \nDiscussionTenthWords drawn at random$^\dagger$, with \nControlDraws draws per pair, recovers \nControlTenth, in proportion to the features added. Between seeds, as within one model, the graph is decided at the margin.

\section{Related work}
\label{sec:related}

\textbf{First order graphs and their corrections.} Attribution patching and its corrections \citep{nanda2023attribution,syed2024attribution,kramar2024atp,hanna2024faith}, integrated gradients \citep{sundararajan2017axiomatic}, circuit searches \citep{conmy2023towards,bhaskar2024finding,mueller2025mib}, replacement models and transcoders \citep{ge2024automatically,dunefsky2024transcoders,ameisen2025circuit}, crosscoders \citep{lindsey2024sparse} and feature circuits \citep{marks2025sparse} build graphs to first order on the features selected before the intervention. For a multilayer perceptron, \citet{kramar2024atp} predict to first order up to the input of the next nonlinearity and then apply the nonlinearity exactly. We apply the same order of operations to the TopK selection of an SAE, where the selection changes it captures account for most of the coefficient change.

\textbf{Interventions on features.} Feature interventions run at high strengths \citep{templeton2024scaling,lindsey2025biology}, their side effects are forecast or checked by ablation \citep{duan2026preintervention,chalnev2024improving,droste2026where}, and benchmarks score features without intervening on them \citep{karvonen2025saebench,chanin2025a}. JumpReLU and batch TopK SAEs select features by thresholds, where the margin is defined as it is here \citep{rajamanoharan2024jumping,lieberum2024gemma,bussmann2024batchtopk}, and a network of ReLU blocks is a max affine spline \citep{balestriero2018spline,hanin2019complexity,hanin2019deep}. We measure how an intervention changes the selection and predict the change from the margins.

\textbf{Comparing SAEs across seeds.} Feature matching scores one feature at a time, by a cosine rule or by how often a feature recurs over many runs \citep{paulo2026different,gerasimov2026unstable}. Meta SAEs write one SAE's decoder directions over another's \citep{leask2025sparse,bussmann2024showing,rubinstein2010double,li2016convergent}, features are matched across layers \citep{balagansky2025mechanistic,cao2026semantic,chen2025transferring,sweeney2026signed}, and SAEs are trained jointly across seeds or models \citep{thasarathan2025universal,varadi2026procrustes,wang2025enforcing}. Model merging maps weights through a permutation or a transport plan \citep{ainsworth2023git,singh2020model,blondel2018smooth,liu2023sparsity,cuturi2023monge}, and identifiability results make two SAEs coincide up to a permutation with scaling or signs \citep{aharon2006on,nelson2026toward,song2025pwmcc,geiger2025causal}. We transfer the graph between two layers through sparse maps, which contain a permutation as the case of one target vector per source vector, and we locate the drop from masking on the features just below the cutoff.

\section{Conclusion}
\label{sec:conclusion}

An attribution graph between TopK SAEs predicts the coefficients after an intervention far better once it selects features again after its predicted score change. A practitioner who reads a graph to learn which features an intervention turns on or off should therefore keep every downstream feature's predicted score change and select again. A practitioner who needs only the changed coefficients should recompute them, which is faster at the intervention counts we timed. When many features are ablated at once, integrated gradients that select again err less than the margin graph. Between seeds, the part of a graph that masking loses lies on the features just below the cutoff. In our patching test the more accurate coefficients predicted the model's behaviour less accurately. Whether better coefficient predictions select more faithful circuits is therefore open.

\bibliography{references}
\bibliographystyle{plainnat}

\clearpage

\appendix
\raggedbottom

\section{Proofs}
\label{app:proofs}

This appendix proves the three theorems of the main text and the results they rely on, the two corollaries, Theorem~\ref{thm:readout} and two propositions. It closes with the checks we ran on constructed instances. Intuitively, every result here rests on one fact. Between changes of rank, a TopK coefficient vector is a linear function of the encoder scores, and at a change of rank it jumps by about the cutoff.

Throughout, $x$ is the upstream activation at the position, $v$ the intervention, $f(x)$ the downstream activation and $p = E(f(x) - b_{\mathrm{pre}}) + b_{\mathrm{enc}}$ the downstream encoder scores, which are affine in the downstream activation. The exact encoder scores after the intervention are $p'$ and the predicted ones are $\hat p = p + \widehat{\Delta p}$, with $\widehat{\Delta p} = EJv$. The selected set before the intervention is $S$, the exact selected set after it is $S'$ and the predicted selected set is $\hat S$. We assume no ties throughout, meaning that no two coordinates of an encoder score vector that a statement uses are equal, none is zero and none equals its feature's threshold under a JumpReLU rule. A largest bound or a largest change taken over an empty set of features is zero.

Intuitively, on an affine block the standard graph is right on every feature except the entering and leaving ones, so its error is at least the part of the exact change that falls on entering features.

\begin{corollary}[the standard graph on an affine segment]
\label{cor:fixed}
Under the hypotheses of Theorem~\ref{thm:exact}, the standard graph and the margin graph differ only on the entering and leaving features, and the standard graph's relative error is at least the square root of the entering features' share of the squared norm of the coefficient change. Where the position has at least $k$ positive encoder scores, an entering feature's coefficient jumps from zero to at least $\theta_k - r$ and a leaving feature's drops from at least $\theta_k$ to zero, with $r$ the largest first order change of a downstream encoder score.
\end{corollary}

Intuitively, an inexact map adds an error in proportion to how badly it fits the decoder and the encoder.

\begin{corollary}[transfer under inexact maps]
\label{cor:residual}
Write $\Delta D = D_A - D_B M$ and $\Delta E = E_A - N E_B$. Then $W_A - N W_B M = \Delta E\, J D_A + N E_B J\, \Delta D$ exactly, and its Frobenius norm is at most $\|J\|_2 (\|\Delta E\|_F \|D_A\|_2 + \|N E_B\|_2 \|\Delta D\|_F)$. Without the conditions of Theorem~\ref{thm:transport} the masked identity gains the term $P_A N (W_B - P_B W_B Q_B) M Q_A$, which depends on the margins of $B$ only through their signs.
\end{corollary}

Intuitively, an entering feature and a leaving feature each move by about the cutoff, so their decoded change is large even when the intervention is small, and only the change of the reconstruction residual can cancel it.

\begin{theorem}[the decoded change when features enter and leave, and the activation derivative chain]
\label{thm:readout}
Write $\Delta x = f(x + v) - f(x)$ for the block's image of the intervention, $R = \max_i |p'_i - p_i|$ for the intervention's exact shift, the largest absolute change of a downstream encoder score, and $\gamma$ for the largest downstream decoder column norm. Assume that the position has at least $k$ positive encoder scores before and after the intervention, and that $s$ features enter the selected set and $s$ leave it, with $I$ the set of entering features and $O$ the set of leaving features. Then the entering and leaving part of the decoded coefficient change and the whole decoded change satisfy
\begin{equation*}
\|G\,\Delta z_{I \cup O}\| \ge \theta_k \Big\|\sum_{l \in I} g_l - \sum_{j \in O} g_j\Big\| - 3 s R \gamma , \qquad \|G\,\Delta z\| \ge \theta_k \Big\|\sum_{l \in I} g_l - \sum_{j \in O} g_j\Big\| - (k + 2s) R \gamma ,
\end{equation*}
and dividing each bound by $\|\Delta x\|$ bounds each ratio to the activation change. On a block affine along the intervention, $\Delta x = Jv$ and $R \le \|EJ\|\,\|v\|$. Where $\theta_k > 0$ and the gap $\theta_k - \theta_{k+1}$ at the cutoff is positive, an intervention whose shift is under $\min(\theta_k, (\theta_k - \theta_{k+1})/2)$ leaves the selected set unchanged. The reconstruction residual's change satisfies $\|\Delta x - G\,\Delta z\| \ge \|G\,\Delta z\| - \|\Delta x\|$. Now let the intervention cross two blocks with Jacobians $J_1$ and $J_2$, the middle SAE's decoder $G_2$ and the third layer's encoder $E_3$, and assume the first block is affine along the intervention and the second along the middle change $\Delta y$ it induces. Then $p_3 + E_3 J_2 J_1 v$ are the exact third layer encoder scores, and the coefficient vector of the activation derivative chain under the third layer's rule is the exact one. A chain that passes a decoded middle coefficient change $G_2\,\Delta \hat z$ through the second block forms third layer encoder scores that differ from those of the activation derivative chain by $E_3 J_2 (\Delta y - G_2\,\Delta \hat z)$, which for the exact middle coefficients is the second block's image of the change of the middle reconstruction residual.
\end{theorem}

Intuitively, no second sparse representation of a vector exists when no few target vectors are linearly dependent. Two good fits on well conditioned supports also lie close together.

\begin{proposition}[unique and stable maps]
\label{thm:maps}
Let $y$ be a feature vector of $A$ and $c$ a column of $M$ that represents it with a support of $s$ target vectors. If every set of at most $2s$ columns of $D_B$ is linearly independent, which is the case when the spark of $D_B$ exceeds $2s$, then no other representation of $y$ uses at most $s$ target vectors, and likewise for each row of $N$ over the rows of $E_B$. If two representations of one feature vector $y$ meet a relative fit residual $\tau$ on supports $S$ and $S'$, they differ by at most $2\tau\|y\|/\sigma$, where $\sigma$ is the smallest singular value of the columns of $D_B$ indexed by $S \cup S'$.
\end{proposition}

Intuitively, on an affine block a feature changes its selection state only when its own move, together with the move of a competitor across the cutoff, covers its margin.

\begin{proposition}[the error band per feature]
\label{thm:mediation}
Assume an affine block, and call the size of a feature's own first order encoder score change its own shift, with the intervention's shift $r$ the largest of them. A selected feature whose margin exceeds its own shift plus the largest shift among the unselected features keeps its selection state, and an unselected feature the size of whose negative margin exceeds its own shift plus the largest shift among the selected features stays out. A selected feature whose own move is under its margin leaves only where an unselected partner's move and its own together cover its margin and the size of the partner's. The decoded first order change of a selected feature that keeps its selection state is the rank one operator $g_j e_j^\top$ applied to the activation change.
\end{proposition}

\begin{proof}[Proof of Theorem~\ref{thm:exact}]
On each cell of the partition cut by the hyperplanes $p_j = p_l$ and $p_j = 0$ the order of the encoder scores and their signs are fixed, so the selected set is a fixed set $S_0$ and the coefficient vector is $P_{S_0}\,p$, which is linear. Each cell is an intersection of halfspaces and so convex. The cells are finitely many, so the coefficient vector is piecewise linear. Across $p_j = 0$ a coordinate enters or leaves at the value zero, so the coefficient vector is continuous there. Across a hyperplane $p_j = p_l$ that separates a selected feature from an unselected one, one coordinate goes from zero to the shared value and the other from that value to zero. The shared value is positive, so the coefficient vector jumps there. These cells are the activation regions of \citet{hanin2019deep} for one SAE, and the spline view of \citet{balestriero2018spline} states the partition into cells and the linearity on each for a network of ReLU blocks.

Suppose $f(x + tv) = f(x) + tJv$ for $t \in [0,1]$. Then $f(x + v) = f(x) + Jv$, and since $p$ is affine in the downstream activation, $p' = p + EJv = \hat p$. The margin graph's prediction is the coefficient vector of $\hat p$, which is the coefficient vector of $p'$, the exact one, on every coordinate, the entering and leaving ones included. The ranking plays no part in the argument, since the prediction is exact before any ranking is computed.
\end{proof}

\begin{proof}[Proof of Corollary~\ref{cor:fixed}]
On a feature $j$ in $S \cap \hat S$ the margin graph predicts $\hat p_j$, and the standard graph predicts $z_j + \widehat{\Delta p}_j = p_j + \widehat{\Delta p}_j = \hat p_j$, since the coefficient equals the encoder score on the selected set. On a predicted leaving feature, in $S$ and not in $\hat S$, the standard graph predicts $\hat p_j$ and the margin graph zero. On a predicted entering feature $l$, in $\hat S$ and not in $S$, the standard graph predicts zero and the margin graph $\hat p_l$, which is positive. Everywhere else both predict zero. So the set where the two differ lies inside the entering and the leaving coordinates, it contains every entering feature, and it contains a leaving feature exactly when $\hat p_j$ is not zero. On an affine segment the exact change on an entering feature $l$ is $\hat p_l$ and the standard graph's change there is zero, so the squared norm of its error is at least the squared norm of the exact coefficient change on the entering set. Dividing by the squared norm of the whole coefficient change gives the bound on its relative error. The leaving coordinates enter no such bound, since the standard graph predicts a leaving feature's change up to its jump.

Where the position has at least $k$ positive encoder scores the cutoff $\theta_k$ is positive and every unselected feature lies outside the top $k$. On an affine segment every encoder score moves by at most $r$, so the $k$th largest encoder score after the intervention is at least $\theta_k - r$. An entering feature is among the $k$ largest after the intervention, so its coefficient rises from zero to at least $\theta_k - r$. A leaving feature was among the $k$ largest before it, so its coefficient was at least $\theta_k$ and drops to zero.
\end{proof}

\begin{proof}[Proof of Theorem~\ref{thm:band}]
Along the segment,
\begin{equation*}
\|f(x + v) - f(x) - Jv\| = \Big\| \int_0^1 \big(J(x + tv) - J(x)\big) v \, dt \Big\| \le \int_0^1 L_v\, t\, \|v\|^2 \, dt = \frac{L_v}{2}\|v\|^2 ,
\end{equation*}
by integrating the derivative along the segment and taking the norm under the integral, which is the mean value form of Taylor's theorem on one segment. Taking the remainder through the encoder row $e_j$ by the Cauchy and Schwarz inequality gives $|p'_j - \hat p_j| \le \|e_j\| (L_v/2)\|v\|^2 = \varepsilon_j(v)$.

Take the margins on $\hat p$, whose $k$th and $(k+1)$th largest values are $\hat\theta_k$ and $\hat\theta_{k+1}$. Write $\bar\varepsilon_{\mathrm{out}}$ for the largest $\varepsilon_l(v)$ over the features outside $\hat S$ and $\bar\varepsilon_{\mathrm{in}}$ for the largest over the features inside it, each zero where its set is empty. Take $j$ in $\hat S$ with $\hat p_j - \max(\hat\theta_{k+1}, 0) > \varepsilon_j(v) + \bar\varepsilon_{\mathrm{out}}$. Every $l$ outside $\hat S$ has $\hat p_l \le \max(\hat\theta_{k+1}, 0)$, because $l$ either lies outside the top $k$ of $\hat p$, where $\hat p_l \le \hat\theta_{k+1}$, or lies inside it with $\hat p_l \le 0$. So $p'_j \ge \hat p_j - \varepsilon_j(v) > \max(\hat\theta_{k+1}, 0) + \bar\varepsilon_{\mathrm{out}} \ge \hat p_l + \varepsilon_l(v) \ge p'_l$, and $p'_j > 0$. Then $j$ outranks every feature outside $\hat S$ in $p'$ and is positive, at most $k - 1$ features can outrank it, and it lies in $S'$.

Take $l$ outside $\hat S$ with $\max(\hat\theta_k, 0) - \hat p_l > \varepsilon_l(v) + \bar\varepsilon_{\mathrm{in}}$. Where $\hat\theta_k > 0$, every one of the $k$ largest predicted encoder scores is positive, so $\hat S$ contains exactly $k$ features each with $\hat p_j \ge \hat\theta_k$, and $p'_l \le \hat p_l + \varepsilon_l(v) < \hat\theta_k - \bar\varepsilon_{\mathrm{in}} \le \hat p_j - \varepsilon_j(v) \le p'_j$ for each of them. So $k$ features outrank $l$ in $p'$ and $l$ lies outside $S'$. Where $\hat\theta_k \le 0$, the condition becomes $-\hat p_l > \varepsilon_l(v) + \bar\varepsilon_{\mathrm{in}} \ge \varepsilon_l(v)$, so $p'_l \le \hat p_l + \varepsilon_l(v) < 0$ and $l$ is not positive, so it lies outside $S'$. Where one constant $L$ bounds $L_v$ for every intervention, $\varepsilon_j(v) \le \|e_j\|(L/2)\|v\|^2$, so every selection state outside the error band under $L$ is outside it under the intervention's own constant as well, and the uniform statement follows.
\end{proof}

\begin{proof}[Proof of Theorem~\ref{thm:transport}]
If $D_A = D_B M$ and $E_A = N E_B$, then $W_A = E_A J D_A = (N E_B) J (D_B M) = N (E_B J D_B) M = N W_B M$. Nothing about $J$, the masks or the widths enters, so the identity is true at every position and for any two widths, and it asks nothing of a matching.

The two conditions of the masked identity are $P_A N (I - P_B) = 0$ and $(I - Q_B) M Q_A = 0$, so $P_A N = P_A N P_B$ and $M Q_A = Q_B M Q_A$. Under both, $P_A W_A Q_A = P_A N W_B M Q_A = (P_A N P_B) W_B (Q_B M Q_A) = P_A N (P_B W_B Q_B) M Q_A$.

Without the two conditions and under exact representation, $P_A W_A Q_A - P_A N (P_B W_B Q_B) M Q_A = P_A N (W_B - P_B W_B Q_B) M Q_A$. In general the masked difference has this term beside $P_A (W_A - N W_B M) Q_A$. Since $W_B - P_B W_B Q_B = (I - P_B) W_B + P_B W_B (I - Q_B)$, the term has two parts. One part comes in through the weight of $N$ on the downstream target vectors of $B$ that $I - P_B$ keeps. The other comes in through the weight of $M$ on the upstream target vectors of $B$ that $I - Q_B$ keeps, through the selected downstream features of $B$. The mask $I - P_B$ keeps exactly the target vectors that are unselected at the position, and a feature is unselected exactly when its margin is not positive. The term therefore depends on the margins only through their signs, which is the last statement of Corollary~\ref{cor:residual}.
\end{proof}

\begin{proof}[Proof of Corollary~\ref{cor:residual}]
For the residual, $\Delta E\, J D_A + N E_B J\, \Delta D = (E_A - N E_B) J D_A + N E_B J (D_A - D_B M) = E_A J D_A - N E_B J D_B M = W_A - N W_B M$. Taking norms by the triangle inequality and $\|XYZ\|_F \le \|X\|_F \|Y\|_2 \|Z\|_2$ with its counterpart $\|XYZ\|_F \le \|X\|_2 \|Y\|_2 \|Z\|_F$ gives the bound.
\end{proof}

\begin{proof}[Proof of Theorem~\ref{thm:readout}]
Since $p$ is affine in the downstream activation, $p' - p = E\,\Delta x$ exactly on any block, so $R = \|E\,\Delta x\|_\infty$. Every order statistic of the encoder scores moves by at most $R$. At least $i$ coordinates of $p$ lie at or above its $i$th largest value $\theta_i$ and each moves by at most $R$, so $\theta'_i \ge \theta_i - R$. At least $m - i + 1$ coordinates lie at or below $\theta_i$ and end at or below $\theta_i + R$, so $\theta'_i \le \theta_i + R$.

Under the assumption both selected sets contain $k$ features, so as many leave as enter. An entering feature $l$ was outside the top $k$, so $p'_l \le p_l + R < \theta_k + R$, and it is inside after the intervention, so $p'_l \ge \theta'_k \ge \theta_k - R$ and $|p'_l - \theta_k| \le R$. A leaving feature $j$ had $p_j \ge \theta_k$ and ends with $p'_j \le \theta'_k \le \theta_k + R$, so $0 \le p_j - \theta_k \le 2R$. The entering feature's coefficient rises from zero to $p'_l$, which is positive, the leaving feature's drops by $p_j$, and each of the $k - s$ features selected before and after changes by at most $R$, so
\begin{equation*}
G\,\Delta z = \theta_k \Big(\sum_{l \in I} g_l - \sum_{j \in O} g_j\Big) + \sum_{l \in I} (p'_l - \theta_k) g_l - \sum_{j \in O} (p_j - \theta_k) g_j + \sum_{i \in S \cap S'} (p'_i - p_i) g_i ,
\end{equation*}
whose first three terms are $G\,\Delta z_{I \cup O}$. The second and third terms have norm at most $3sR\gamma$ together and the fourth at most $(k - s)R\gamma$, and the triangle inequality gives both bounds. A change of the selected set needs $R > 0$, so $\|\Delta x\| > 0$, and dividing gives the ratios. On a block affine along the intervention, $\Delta x = Jv$ and $R = \|EJv\|_\infty \le \|EJ\|\,\|v\|$.

Let $R < \min(\theta_k, (\theta_k - \theta_{k+1})/2)$. Every member of $S$ ends at or above $\theta_k - R$, which is positive and above $\theta_{k+1} + R$, and every other feature ends at or below $\theta_{k+1} + R$. So the $k$ largest encoder scores after the intervention are the members of $S$, all positive, and $S' = S$. Since $\Delta x - G\,\Delta z$ is the change of the reconstruction residual, $\|\Delta x - G\,\Delta z\| \ge \|G\,\Delta z\| - \|\Delta x\|$.

Across two blocks the first block's affinity along the intervention gives the middle change $\Delta y = J_1 v$ exactly, and the second block's affinity along $\Delta y$ gives the third layer's activation change $J_2 J_1 v$. The third layer encoder scores are therefore $p_3 + E_3 J_2 J_1 v$ exactly, which is what one JVP through both blocks forms. The coefficient vector of those encoder scores under the third layer's rule is the exact one by the argument of Theorem~\ref{thm:exact}. A chain that passes $G_2\,\Delta\hat z$ on forms $p_3 + E_3 J_2 G_2\,\Delta\hat z$, which differs from the activation derivative chain's $p_3 + E_3 J_2 \Delta y$ by $E_3 J_2 (\Delta y - G_2\,\Delta\hat z)$. For the exact middle coefficients, $\Delta y - G_2\,\Delta z_2$ is the change of the middle reconstruction residual. The identity is on the third layer's encoder scores, and the third layer's rule applied to the two vectors need not differ by that vector, which is why we compare the chains at the third layer's coefficients.
\end{proof}

\begin{proof}[Proof of Proposition~\ref{thm:maps}]
If $D_B c = D_B c'$ for two representations of $y$ on supports of at most $s$ target vectors each, then $D_B (c - c') = 0$ with $c - c'$ supported on at most $2s$ columns, which are linearly independent, so $c = c'$. That is the uniqueness theorem of \citet{donoho2003optimally} and \citet{gribonval2003sparse} applied column by column, and its restricted isometry form is that of \citet{candes2005decoding}. The same argument on the rows of $E_B$ gives each row of $N$. For stability, $\|D_B (c - c')\| \le \|D_B c - y\| + \|D_B c' - y\| \le 2\tau \|y\|$. Since $c - c'$ is supported on $S \cup S'$, $\sigma \|c - c'\| \le \|D_B (c - c')\| \le 2\tau\|y\|$. The classical forms of the same bound are those of \citet{donoho2006stable} under coherence and \citet{candes2008the} under a restricted isometry.
\end{proof}

\begin{proof}[Proof of Proposition~\ref{thm:mediation}]
On an affine block the first order change of each encoder score is its exact change, so a feature's own shift is the size of its move. The argument of Theorem~\ref{thm:band}, with each feature's own shift in place of its error bound, then gives the first two conclusions. Take a selected feature $j$ whose own move is under its margin, $-\Delta p_j < p_j - \max(\theta_{k+1}, 0)$. Then $p'_j > \max(\theta_{k+1}, 0) \ge 0$, so $j$ is positive after the intervention. If it leaves, it leaves by rank, which needs $k$ features above it in $p'$ and so an unselected feature $l$ with $p'_l \ge p'_j$. Then $\Delta p_l - \Delta p_j \ge p_j - p_l \ge p_j - \max(\theta_{k+1}, 0)$, the margin of $j$, because an unselected $l$ has $p_l \le \max(\theta_{k+1}, 0)$. By the same inequality taken from the other side, $\Delta p_l - \Delta p_j \ge \max(\theta_k, 0) - p_l$, the size of $l$'s margin, since a selected $j$ has $p_j \ge \max(\theta_k, 0)$. A selected feature whose own move is not under its margin can leave with no partner, by dropping to zero. A selected feature that keeps its selection state changes by $\Delta p_j = e_j^\top J v = e_j^\top \Delta y$, so its decoded change is $g_j e_j^\top \Delta y$, the rank one operator $g_j e_j^\top$ applied to the activation change.
\end{proof}

\textbf{Constructed checks.} We also checked Theorems~\ref{thm:exact} to~\ref{thm:readout}, both corollaries and Proposition~\ref{thm:mediation} in double precision on constructed instances from fixed seeds. We list each construction with its saved output in Table~\ref{tab:constructed-checks}. The checks from seed \nTheorySeedMargin confirm Theorem~\ref{thm:exact}'s exact prediction on affine blocks and that the two graphs differ on the entering and leaving coordinates alone. They also confirm the linearity of the coefficient vector between rank changes, Theorem~\ref{thm:band}'s bound and error band, and the mask term of Theorem~\ref{thm:transport} with the intervention that flips it, and all \nTheoryChecksMargin pass. The checks from seed \nTheorySeedTransport confirm Theorem~\ref{thm:transport}'s two identities with the target vector that breaks the masked one, and the decomposition of Corollary~\ref{cor:residual} with its bound. They also confirm that a representation's class, the count of target vectors at the first step that meets the tolerance as Appendix~\ref{app:anatomy} sorts it, depends on the tolerance, and all \nTheoryChecksTransport pass.

The checks from seed \nTheorySeedReadout confirm the bound of Theorem~\ref{thm:readout} for one entering and one leaving feature on affine blocks, and the growth of the ratio across positions whose gap at the cutoff shrinks. They also confirm that the residual absorbs the excess, the identity of the activation derivative chain, the conditions of Proposition~\ref{thm:mediation} and the first order remainder, and all \nTheoryChecksReadout pass. The checks from seed \nTheorySeedRestated confirm the set where the graphs differ, the entering set's share, the error band at a cutoff at or below zero, the clipped margin, the entering feature's jump against the intervention's shift, the tie and the chain identity on encoder scores, and all \nTheoryChecksRestated pass.

The checks of Theorem~\ref{thm:readout} as stated here run from seed \nTheorySeedOvershoot on residual streams of width \nTheoryD, SAEs of width \nTheoryM at a TopK of \nTheoryK and instances built by hand, at a relative tolerance of \nTheoryTol, and all \nTheoryChecksOvershoot pass. They confirm both bounds on interventions through a nonlinear block, the affine form and the floor on the shift of an intervention that changes the selected set. With coincident decoder columns the ratio stays at one half, and with a cutoff that shrinks with the gap it stays at the square root of five over two. At one position with a fixed cutoff and gap, no intervention whose shift is under half the gap changes the selected set.

\begingroup
\begin{longtable}{p{\proofSeed}p{\proofConstruction}p{\proofOutput}}
\caption{\textbf{Every constructed check agrees with its algebraic prediction.} Rows give each check's fixed seed, its construction and its saved output, and the proofs of Appendix~\ref{app:proofs} state the hypotheses.}\label{tab:constructed-checks}\\
\toprule
seed & construction & measured output \\
\midrule
\endfirsthead
\toprule
seed & construction & measured output \\
\midrule
\endhead
\bottomrule
\endfoot
\nTheorySeedMargin & Affine block, selecting again & exact \nProofEight, of \nProofNine, selection changes \nProofSeven \\
\nTheorySeedMargin & Segments between rank changes & linear \nProofOneZero, of \nProofOneOne \\
\nTheorySeedMargin & Nonlinear block with known constant & within bound \nProofOneTwo, of \nProofOneFour, constant \nProofOneThree \\
\nTheorySeedMargin & Selection states outside the error band & outside the band \nProofOneFive, unchanged \nProofOneSeven, changes inside the band \nProofOneSix \\
\nTheorySeedMargin & Affine selection changes & changes differing only on entering and leaving features \nProofOneNine, unchanged selections agreeing \nProofTwoZero \\
\nTheorySeedMargin & Mask identity with unselected target vectors & relative gap \nProofTwoTwo \\
\nTheorySeedMargin & Maps give unselected target vectors no weight & mask norm \nProofTwoThree \\
\nTheorySeedMargin & Flip the smallest negative margin & margin size \nProofThreeOne, mask norm before \nProofThreeZero, after \nProofTwoNine \\
\nTheorySeedTransport & Exact sparse representations & relative identity gap \nProofThreeNine \\
\nTheorySeedTransport & Maps restricted to selected target vectors & relative identity gap \nProofFourZero \\
\nTheorySeedTransport & One unselected target vector & relative identity gap \nProofFourOne \\
\nTheorySeedTransport & Inexact maps and residual decomposition & relative identity gap \nProofSixSix \\
\nTheorySeedTransport & Two target vectors and a nearby third & perturbation \nProofFourFour gives residual \nProofFourFive, perturbation \nProofFiveTwo gives residual \nProofFiveThree, perturbation \nProofSixZero gives residual \nProofSixOne \\
\nTheorySeedTransport & Perturbed sparse maps & residual norm \nProofSixEight, bound \nProofSixSeven \\
\nTheorySeedReadout & One entering and one leaving feature, affine & norm of their decoded change \nProofOneFiveFour, lower bound \nProofOneFourNine \\
\nTheorySeedReadout & Fixed scale with shrinking gap & ratio growth \nProofNineZero \\
\nTheorySeedReadout & Residual for the same change & residual norm \nProofOneFiveThree, activation change \nProofOneFourThree \\
\nTheorySeedReadout & Two affine blocks & largest coefficient gap \nProofOneSixSix, interventions \nProofOneSixNine \\
\nTheorySeedReadout & Decoded middle coefficients omit their residual & identity gap \nProofOneSixFour, median coefficient error \nProofOneSixFive \\
\nTheorySeedReadout & Margins beyond the error band & positions with no such change \nProofOneSixThree \\
\nTheorySeedReadout & A change driven by the opposite side & changed features each explained by a partner \nProofOneSixTwo \\
\nTheorySeedReadout & Selected features whose selection state stays fixed & largest share identity gap \nProofOneSixZero \\
\nTheorySeedReadout & Features near the cutoff & changed features \nProofOneFiveEight, median jump over first order change \nProofOneFiveSix \\
\nTheorySeedReadout & Nonlinear block and local remainder & interventions all within the bound \nProofEightSeven \\
\nTheorySeedReadout & Local and global constants & selection states accepted locally \nProofSevenSix, global \nProofSevenFive \\
\nTheorySeedReadout & Selection states accepted by each band & local changes \nProofEightZero, global changes \nProofSevenNine \\
\nTheorySeedRestated & Coincident decoder columns & ratio at first gap \nProofOneEightFour, last gap \nProofTwoTwoFour \\
\nTheorySeedRestated & Cutoff proportional to the gap & ratio at first gap \nProofTwoThreeFour, last gap \nProofTwoSevenFour \\
\nTheorySeedRestated & Fixed cutoff and shrinking gap & ratio at first gap \nProofTwoEightFive, last gap \nProofThreeTwoNine \\
\nTheorySeedRestated & Graphs differ on a proper subset of the changed features & second fixed coordinate \nProofThreeThreeThree, margin coordinate \nProofThreeThreeFive \\
\nTheorySeedRestated & Entering share and entering and leaving share & relative error \nProofThreeThreeEight, entering share \nProofThreeThreeSeven, entering and leaving share \nProofThreeThreeSix \\
\nTheorySeedRestated & Nonpositive predicted cutoff & cutoff \nProofThreeThreeNine \\
\nTheorySeedRestated & Clipped margin at zero & clipped margin \nProofThreeFourOne, unclipped value \nProofThreeFourTwo \\
\nTheorySeedRestated & Entering feature's jump and largest change & jump \nProofThreeFourFive, bound from largest change \nProofThreeFourFour, bound from own change \nProofThreeFourThree \\
\nTheorySeedRestated & A tie at the cutoff & kept at tie \nProofThreeFourSix, kept feature margin \nProofThreeFourSeven \\
\nTheorySeedRestated & Chain identity on encoder scores & largest relative identity gap \nProofThreeFourNine \\
\nTheorySeedRestated & One target vector per column without a permutation & identity gap \nProofThreeFiveZero \\
\nTheorySeedRestated & Negative target vector with nonnegative fit & coefficient \nProofThreeFiveOne, relative residual \nProofThreeFiveTwo \\
\nTheorySeedOvershoot & Cutoff proportional to the gap & ratio \nProofThreeSixTwo, lower bound \nProofThreeSixZero \\
\nTheorySeedOvershoot & Interventions that change the selection & at or above the floor on the change \nProofThreeSixNine, of \nProofThreeSevenZero \\
\nTheorySeedOvershoot & Coincident decoder columns & ratio \nProofThreeSevenThree, lower bound \nProofThreeSevenOne \\
\nTheorySeedOvershoot & One position with fixed cutoff and gap & half gap \nProofThreeEightZero, smallest shift that changes the selection \nProofThreeNineNine \\
\nTheorySeedOvershoot & Affine form of the decoded bound & largest image gap \nProofFourZeroZero, largest shift gap \nProofFourZeroOne \\
\nTheorySeedOvershoot & Nonlinear block with the theorem's hypotheses & interventions \nProofFourZeroFour, positive part bounds \nProofFourZeroTwo, violations \nProofFourZeroThree \\
\end{longtable}
\par\noindent A selection state is a feature's selected or unselected state, and its margin is the encoder score's signed distance from a change of that state. The error band holds the margins whose predicted state the first order error bound cannot decide. A mask keeps the selected features and sets the other coordinates to zero, and the reconstruction residual is the activation less its decoded coefficients. Identity gaps compare the two sides of the named identity, and a relative gap divides their norm difference by the reference norm. Identity gaps are relative unless the row calls them absolute coefficient or image gaps. Rows with a first and a last gap give the ends of the constructed sequence, and the examples with one entering and one leaving feature use its last gap. Absolute gaps, margins and changes are in the constructed activation's units. The decoded ratio divides the norm of the decoded change on entering and leaving features by the norm of the activation change, and a lower bound is the theorem's bound on the ratio or norm the row names. A count gives constructed instances unless the cell names features, selection states or interventions.

\endgroup

\section{SAEs and measurements}
\label{app:releases}

This appendix gives the choices behind every comparison of Sections~\ref{sec:crossing} to~\ref{sec:readout}. It describes each SAE pair, the corpus and how we sample positions, the training of our SAEs, the checks of the Jacobian, and the measurements that the main text summarizes. We load each public SAE from its repository, on the model its card names, and apply the arithmetic of the library that trained it, which subtracts the decoder bias before the encoder and adds it after the decoder \citep{eleutherai2025sparsify,marks2024dictionary}. Table~\ref{tab:releases-full} gives each SAE's release, model, layers, rule, budget and licence. The EleutherAI releases on \nModelNameA and \nModelNameD state a budget of \nBudgetTokensA billion tokens of the Pile at a context of \nContextA, and the deduplicated TopK release states none. The ReLU release's library states \nBudgetTokensR billion tokens, and its configuration states \nStepsR steps at \nBatchR activations per batch and a context of \nCtxR, with an $\ell_1$ penalty of \nSparsityPenaltyR.

{\raggedright In the order the table first lists them, the releases are hosted at the Hugging Face Hub as \texttt{\nReleaseA}, \texttt{\nReleaseD}, \texttt{\nReleaseE} and \texttt{\nReleaseR}. The released code pins the revision of every model, SAE, tokenizer, dataset and library at which these experiments ran.\par}

\begin{table}[H]
\caption{\textbf{The configurations cover the public releases and the sparse autoencoders (SAEs) we trained.} Rows give each configuration's release, model, layers, selection rule, count of selected features, training budget and stated licence. Our SAEs have width \nWidthF and the narrower configuration width \nWidthFiled.}
\label{tab:releases-full}
\centering
\scriptsize
\setlength{\tabcolsep}{1.2pt}
\begin{tabular}{lllllll}
\toprule
release & model & layers & rule & selected count & budget & licence \\
\midrule
\citet{eleutherai2024sae70m} & \nModelNameA & \nLayerUpA, \nLayerDownA & \nRuleNameA & \nKA & \nBudgetTokensA B & \nLicenceNameA \\
\citet{eleutherai2024sae70m} & \nModelNameB & \nLayerUpB, \nLayerDownB & \nRuleNameB & \nKB & \nBudgetTokensA B & \nLicenceNameB \\
\citet{eleutherai2024sae70m} & \nModelNameC & \nLayerUpC, \nLayerDownC & \nRuleNameC & \nKC & \nBudgetTokensA B & \nLicenceNameC \\
\citet{eleutherai2024sae160m} & \nModelNameD & \nLayerUpD, \nLayerDownD & \nRuleNameD & \nKD & \nBudgetTokensA B & \nLicenceNameD \\
\citet{eleutherai2024sae70mdeduped} & \nModelNameE & \nLayerUpE, \nLayerDownE & \nRuleNameE & \nKE & none stated & \nLicenceNameE \\
\citet{marks2024pythiasaes} & \nModelNameR & \nLayerUpR, \nLayerDownR & \nRuleNameR & \nKR & \nBudgetTokensR B & \nLicenceNameR \\
ours & \nModelNameF & \nLayerUpA, \nLayerDownA & TopK & \nKF & \nActivationsF & none \\
narrower, ours & \nModelNameF & \nLayerUpA, \nLayerDownA & TopK & \nKFiled & \nBudgetFiledEnd & none \\
small sample, ours & \nModelNameF & \nLayerUpA & TopK & \nKF, \nKFiled & \nSmallBlockDistinct distinct & none \\
\bottomrule
\end{tabular}
\par\vspace{2pt}\footnotesize A budget in B is in billions of tokens, and a budget of an SAE we trained is in activations. The narrower configuration has width \nWidthFiled with \nKFiled features kept, and the small sample configurations of both widths fit their distinct activations over \nSmallBlockPresentations presentations, at the upstream layer alone, and appear in Appendices~\ref{app:anatomy} and~\ref{app:budget}. Each configuration we trained has \nRunsEachConfiguration runs, six with different initialization seeds and one data order pair. These training budgets are separate from the held out validation positions of every evaluation.

\par\noindent A sparse autoencoder (SAE) represents model activations by sparse feature coefficients. TopK selects the largest positive encoder scores, and ReLU, the rectified linear unit, every positive score. Width counts an SAE's features, and the selected count $k$ is the most features the rule selects. MIT names the MIT License, and none means that no licence was stated.
\end{table}

The experiments use the validation file of Pile Uncopyrighted, a copy of the Pile hosted at the Hugging Face Hub as \texttt{\nCorpusRepo}. Its card names the subsets removed from the collection, \nCorpusRemoved. Each model's own tokenizer tokenizes the documents in file order. We skip a document under \nMinChars characters and follow each document with the end of text token. The same filter packs the stream that trained our SAEs and every evaluation stream. It skipped \nPackValSkipped of the \nPackValDocs documents the experiments used, which contained \nPackValSkippedTokens tokens, and \nPackTrainSkipped of the \nPackTrainDocs documents that the training of our SAEs and its held out sample used, which contained \nPackTrainSkippedTokens tokens. The stream is cut into sequences of \nSeqLen tokens, and each sequence gives one position, drawn uniformly from seed \nPositionSeed among the positions after the first up to position \nPositionLast of the sequence indexed from zero. We fixed the filter, the cut into sequences and the draw before any experiment. The first \nMeasureSeqs sequences are the measurement sample, which chose the sparsity of the transfer's fit and nothing else, and the next \nHeldout are the held out sample. We intervene on every selected upstream feature at the first \nTokens positions of the held out sample, at \nCrossingDoses scaling factors for the share of interventions that change the selection and at \nPredictorDoses for the predictors. The transfer across seeds uses all \nHeldout.

We trained our SAEs on \nModelNameF at the layers of its first public pair, with a TopK rule selecting \nKF features and an expansion of \nExpansionF over the residual stream's width of \nCfDModel, for \nActivationsF activations in each of \nRunsF runs. The trainer is our own and follows the recipe of \citet{gao2025scaling}. It streams the same repository's training split in file order under the packing rules above and drops the first position of every packed sequence, the one position that attends to nothing but itself, so a sequence of \nSeqLen tokens yields \nPositionLast activations. The first \nCfHeldout activations form the sample held out before training, on which we compute the mean $\mu$ and scale $s$ of Section~\ref{sec:setup}. For that sample $H$, we set $\mu=|H|^{-1}\sum_{x\in H}x$ and $s=\max\{\text{\nExtraScaleFloor},\ |H|^{-1}\sum_{x\in H}\|x-\mu\|_2\}$.

Every run shares $\mu$ and $s$, with $s$ at \nScaleUpF at layer \nLayerUpA and \nScaleDownF at layer \nLayerDownA. In the units of Section~\ref{sec:setup}, every coefficient, encoder score and cutoff is therefore $s$ times its stored value, and an intervention's norm over the cutoff is \nGaugeFactorF times its stored ratio. The fraction of variance unexplained is measured on the first \nCfFvuSample activations of that sample at each checkpoint. On a second sample of \nCfHeldout activations drawn after training, it ranges from \nFvuTailThreeMin to \nFvuTailThreeMax at layer \nLayerUpA and from \nFvuTailFourMin to \nFvuTailFourMax at layer \nLayerDownA over the \nRunsF runs.

The loss is the squared reconstruction error summed over coordinates and averaged over the batch, plus one over \nCfAuxAlphaDen times an auxiliary term. A feature is dead when it was selected on none of the last \nCfDeadTokens activations, and no feature counts as dead before that many were trained. The auxiliary term decodes the positive parts of the \nCfKAux largest encoder scores among the dead features, or of every dead feature where fewer are dead. It fits them to the residual of the reconstruction from the selected features,
\begin{equation*}
\mathcal{L}_{\mathrm{aux}} = c\, \frac{\sum_b \|G\,a_b - \operatorname{sg}(x_b - \hat x_b)\|^2}{\sum_b \|x_b - \bar x\|^2} ,
\end{equation*}
where $\operatorname{sg}$ keeps its argument's forward value and has zero gradient. We detach this residual target, so the term differentiates through the selected dead features' encoder and decoder parameters and each run's preencoder bias, but not through the reconstruction from the selected features. In the display, $a_b$ keeps those encoder scores at row $b$, $\hat x_b$ is the reconstruction from the selected features, $\bar x$ is its run's batch mean, and both sums run over the rows of the \nRunsTogether runs trained together.

The coefficient $c$ is shared across all \nRunsTogether runs. It is their summed count of dead features divided by \nRunsTogether times the smaller of \nCfKAux and the largest count among them, capped at one, and the term is off where no run has a dead feature.

Table~\ref{tab:aux} gives $c$ and every run's count of dead features at the checkpoints where any run had one. The term was off at the first \nAuxOffCount of the \nAuxCheckpoints checkpoints of each layer, through \nAuxOffThrough activations. At the last two, $c$ ranged from \nAuxCoefMin to \nAuxCoefMax, above \nAuxThreshold at \nAuxOverThreshold of the \nAuxCheckpoints.

At the recorded checkpoints, at most \nAuxDeadMax of a run's \nWidthF features were dead. Their direct auxiliary gradients reach only the selected dead features' parameters and the run's bias $b_{\mathrm{pre}}$, whose updates can affect every feature's later encoder score. The checkpoint counts do not bound the updates between checkpoints.

Each run uses Adam at a learning rate of \nCfLr with PyTorch's default betas of \nAdamBetaOne and \nAdamBetaTwo and epsilon of \nAdamEps, on batches of \nCfBatch activations. The model's activations are taken in \nActDtype and cast to single precision before centring. Every parameter and optimizer state is in single precision, under PyTorch \nTorchPin.

Initialization, shuffling and position sampling use PyTorch's uniform draws, random permutations and random integers, each from a generator on the processor that is seeded explicitly. Each run permutes each chunk of \nCfChunk collected activations by its own generator and consumes it in batches of \nCfBatch. That generator is seeded at the run's shuffle seed times \nShuffleRunMult plus the layer times \nShuffleLayerMult plus the chunk's index. The learning rate warms up linearly over \nCfWarmup steps as the sparsify library does, and it stays constant after the warmup where that library decays \citep{eleutherai2025sparsify}. Decoder columns are renormalized to unit norm after every step.

The encoder starts from a uniform draw, from a generator seeded at the run's initialization seed plus \nInitLayerMult times the layer, bounded by the inverse square root of the stream's width, and the decoder starts from its normalized transpose, with both biases at zero.

Each run saves a checkpoint at each of \nCfCheckpointCount targets, one, two and five times each power of ten from \nCfCheckpointFirst to \nCfCheckpointLast activations, and Appendix~\ref{app:budget} uses them as the \nBudgetCount budgets the training reached.

The six runs with different initialization seeds take initialization seeds \nCfSeedInitFirst to \nCfSeedInitLast and shuffle seeds \nCfShuffleFirst to \nCfShuffleLastIndependent. The data order pair shares the initialization seed \nCfSeedInitRetrain and takes the shuffle seeds \nCfShuffleRetrainFirst and \nCfShuffleRetrainSecond.

The narrower configuration at width \nWidthFiled with \nKFiled features selected trains on the same stream to \nBudgetFiledEnd activations. The small sample configurations of both widths fit \nSmallBlockDistinct distinct activations over \nSmallBlockPresentations presentations. Each configuration has \nRunsEachConfiguration runs, six with different initialization seeds and one data order pair, as our SAEs do.

Our remote computations took \nDeviceHours accelerator hours, summed over calls that each ran on one \nDeviceName. Of these, \nHoursTraining trained our SAEs, \nHoursPublic measured the public releases, \nHoursTransport fitted and scored the transfer, and \nHoursReadout measured the decoded change and checked it.

\begin{table}[H]
\caption{\textbf{Only the last two recorded checkpoints have a nonzero shared auxiliary coefficient.} Cells count the dead features of each run of our sparse autoencoders (SAEs) at each layer and checkpoint, those selected on none of the last \nCfDeadTokens activations, and the last row gives the shared coefficient.}
\label{tab:aux}
\centering
\scriptsize
\setlength{\tabcolsep}{2.5pt}
\begin{tabular}{lrrrr}
\toprule
run & \begin{tabular}[c]{@{}c@{}}layer \nLayerUpA at\\\nAuxAtThreeMid\end{tabular} & \begin{tabular}[c]{@{}c@{}}layer \nLayerUpA at\\\nAuxAtThreeEnd\end{tabular} & \begin{tabular}[c]{@{}c@{}}layer \nLayerDownA at\\\nAuxAtFourMid\end{tabular} & \begin{tabular}[c]{@{}c@{}}layer \nLayerDownA at\\\nAuxAtFourEnd\end{tabular} \\
\midrule
run \nRunIndexZero & \nAuxDeadThreeMidZero & \nAuxDeadThreeEndZero & \nAuxDeadFourMidZero & \nAuxDeadFourEndZero \\
run \nRunIndexOne & \nAuxDeadThreeMidOne & \nAuxDeadThreeEndOne & \nAuxDeadFourMidOne & \nAuxDeadFourEndOne \\
run \nRunIndexTwo & \nAuxDeadThreeMidTwo & \nAuxDeadThreeEndTwo & \nAuxDeadFourMidTwo & \nAuxDeadFourEndTwo \\
run \nRunIndexThree & \nAuxDeadThreeMidThree & \nAuxDeadThreeEndThree & \nAuxDeadFourMidThree & \nAuxDeadFourEndThree \\
run \nRunIndexFour & \nAuxDeadThreeMidFour & \nAuxDeadThreeEndFour & \nAuxDeadFourMidFour & \nAuxDeadFourEndFour \\
run \nRunIndexFive & \nAuxDeadThreeMidFive & \nAuxDeadThreeEndFive & \nAuxDeadFourMidFive & \nAuxDeadFourEndFive \\
run \nRunIndexSix & \nAuxDeadThreeMidSix & \nAuxDeadThreeEndSix & \nAuxDeadFourMidSix & \nAuxDeadFourEndSix \\
run \nRunIndexSeven & \nAuxDeadThreeMidSeven & \nAuxDeadThreeEndSeven & \nAuxDeadFourMidSeven & \nAuxDeadFourEndSeven \\
coefficient & \nAuxCoefThreeMid & \nAuxCoefThreeEnd & \nAuxCoefFourMid & \nAuxCoefFourEnd \\
\bottomrule
\end{tabular}

\par\noindent The \nRunsTogether runs are on \nModelNameF at layers \nLayerUpA and \nLayerDownA, of width \nWidthF and TopK \nKF. The shared coefficient divides their summed count of dead features by \nRunsTogether times the smaller of \nCfKAux and the largest count of a run, capped at one, and it is zero when every count is zero. Column heads give the layer and the cumulative count of training activations at the checkpoint. The coefficient weights the auxiliary reconstruction loss for dead features and has no unit. TopK selects the largest positive encoder scores.
\end{table}

The block between two layers is the model's next transformer block, from the residual stream after the upstream layer to the residual stream after the downstream one. Its Jacobian is taken at one position with every earlier position's keys and values fixed. We take every exact coefficient change and every Jacobian in double precision, on a model loaded in double precision. Before any experiment uses the Jacobian, we check it four ways on the first \nCheckTokensCount positions of the held out sample, with \nCheckDirs random directions each. The check at the model's single precision softmax uses seed \nCheckSeedRelease, and the three in double precision use seed \nCheckSeedDouble. We compare the block function with the model's own forward pass at the model's softmax and in double precision, the Jacobian with the JVP through the whole layer, and the Jacobian with a central difference at a step of \nCheckStep. Each check's value is the largest over positions and directions of the Euclidean norm of the difference over the reference's norm. Every value on every pair is within its tolerance, as Table~\ref{tab:checks} shows for each check. On our SAEs we also compared two ways to compute one edge. The first multiplies the encoder row by the forward mode JVP of the decoder column, and the second pulls the encoder row back in reverse mode and multiplies it by the column. The note under the table gives their difference, which the block's single precision softmax puts just over its tolerance, and none of our results uses either computation.

\begin{table}[H]
\caption{\textbf{The four tabulated checks of the Jacobian meet their tolerances.} Cells give each check's largest relative norm difference, beside the tolerance in the first row. The checks compare the block function with the model at each softmax precision, the Jacobian with a Jacobian vector product (JVP) through the whole layer, and the Jacobian with a central difference.}
\label{tab:checks}
\centering
\scriptsize
\setlength{\tabcolsep}{2.5pt}
\begin{tabular}{lllrrrr}
\toprule
SAEs & model & layers & \begin{tabular}[c]{@{}c@{}}the function at the\\model's softmax\end{tabular} & \begin{tabular}[c]{@{}c@{}}the function in\\double precision\end{tabular} & \begin{tabular}[c]{@{}c@{}}the Jacobian against\\a JVP\end{tabular} & \begin{tabular}[c]{@{}c@{}}the Jacobian against\\a central difference\end{tabular} \\
\midrule
tolerance &  &  & \nCheckSoftmaxTol & \nCheckOutputTol & \nCheckJvpTol & \nCheckCentralTol \\
public TopK & \nModelNameA & \nLayerUpA, \nLayerDownA & \nCheckSoftmaxA & \nCheckOutputA & \nCheckJvpA & \nCheckCentralA \\
public TopK & \nModelNameB & \nLayerUpB, \nLayerDownB & \nCheckSoftmaxB & \nCheckOutputB & \nCheckJvpB & \nCheckCentralB \\
public TopK & \nModelNameC & \nLayerUpC, \nLayerDownC & \nCheckSoftmaxC & \nCheckOutputC & \nCheckJvpC & \nCheckCentralC \\
public TopK & \nModelNameD & \nLayerUpD, \nLayerDownD & \nCheckSoftmaxD & \nCheckOutputD & \nCheckJvpD & \nCheckCentralD \\
public TopK & \nModelNameE & \nLayerUpE, \nLayerDownE & \nCheckSoftmaxE & \nCheckOutputE & \nCheckJvpE & \nCheckCentralE \\
public ReLU & \nModelNameR & \nLayerUpR, \nLayerDownR & \nCheckSoftmaxR & \nCheckOutputR & \nCheckJvpR & \nCheckCentralR \\
ours & \nModelNameF & \nLayerUpA, \nLayerDownA & \nCheckSoftmaxF & \nCheckOutputF & \nCheckJvpF & \nCheckCentralF \\
\bottomrule
\end{tabular}
\par\vspace{2pt}\footnotesize A sparse autoencoder (SAE) represents model activations by sparse feature coefficients. Each value is the largest over the first \nCheckTokensCount held out Pile validation positions and \nCheckDirs random directions per position. A separate comparison of two ways to compute one edge did not meet its tolerance, as follows. The Jacobian computation of our SAEs on run \nRunIndexZero takes three further checks. The one position function agrees with the block in double precision at all \nTjSequences positions of the measurement and held out samples, at \nCheckEveryTokenF against a tolerance of \nCheckEveryTokenTolF. On the first \nCheckTokensCount held out positions one edge, from the first selected upstream feature to the first selected downstream one, computed forward and in reverse through the model's single precision softmax, differs by \nCheckPairingsF of the Cauchy and Schwarz bound on that edge, above its tolerance of \nCheckPairingsTolF, and none of our measurements uses either computation. The edge the stored double precision Jacobian gives agrees with both at \nCheckAgainstPairingsF of the bound against \nCheckAgainstPairingsTolF.

\par\noindent TopK selects the largest positive encoder scores, and ReLU, the rectified linear unit, every positive score. A JVP gives the first order activation change along one direction. Each relative difference divides the norm of the difference by the norm of the reference computation that the column head names, and smaller is better.
\end{table}

Table~\ref{tab:counts} counts every pair's interventions and selection changes per scaling factor, and no intervention at any scaling factor leaves the exact coefficients unchanged. The margin graph errs less in every cell of Table~\ref{tab:factors}, which gives both graphs' coefficient errors for each pair, and of Table~\ref{tab:errors-runs}, which gives each of our runs. Each share of squared norm in Section~\ref{sec:crossing} averages over the zero ablations that change the selection. For a coefficient change $\Delta z$, with entering set $I$ and leaving set $O$, the entering and leaving share is $\|\Delta z_{I\cup O}\|^2/\|\Delta z\|^2$ and the entering share is $\|\Delta z_I\|^2/\|\Delta z\|^2$, where $\Delta z_U$ keeps the change on $U$.

The ReLU pair shows the contrast with a continuous rule. There a zero ablation moves a feature across zero in \nCrossZeroR of cases and changes the count of selected features in \nCountChangeZeroR. Those features account for \nEnergyZeroR of the squared norm of the coefficient change, since a ReLU coefficient is continuous at zero. The ReLU and deduplicated TopK releases also differ in library, budget and context, so the contrast is one instance of a fact that Appendix~\ref{app:proofs} proves, that a TopK coefficient jumps where a ReLU coefficient bends.

\begin{table}[H]
\caption{\textbf{Selection changes are common, and every intervention changes the coefficients.} Cells count each pair's interventions and selection changes per scaling factor, and the last two columns give the mean share of the squared coefficient change on the entering features or on the entering and leaving features.}
\label{tab:counts}
\centering
\scriptsize
\setlength{\tabcolsep}{2.5pt}
\begin{tabular}{lllrrrrrrr}
\toprule
SAEs & model & layers & \begin{tabular}[c]{@{}c@{}}interventions\\per scaling factor\end{tabular} & \begin{tabular}[c]{@{}c@{}}selection\\changes\\at zero\end{tabular} & \begin{tabular}[c]{@{}c@{}}selection\\changes\\at \nDoseHalf\end{tabular} & \begin{tabular}[c]{@{}c@{}}selection\\changes\\at \nDoseNine\end{tabular} & \begin{tabular}[c]{@{}c@{}}unchanged\\coefficients\\at every factor\end{tabular} & \begin{tabular}[c]{@{}c@{}}entering\\share\end{tabular} & \begin{tabular}[c]{@{}c@{}}entering and\\leaving share\end{tabular} \\
\midrule
public TopK & \nModelNameA & \nLayerUpA, \nLayerDownA & \nAuditPerDoseZeroA & \nCrossZeroCountA & \nCrossHalfCountA & \nCrossNineCountA & \nZeroResponseA & \nEnterShareA & \nEnergyZeroA \\
public TopK & \nModelNameB & \nLayerUpB, \nLayerDownB & \nAuditPerDoseZeroB & \nCrossZeroCountB & \nCrossHalfCountB & \nCrossNineCountB & \nZeroResponseB & \nEnterShareB & \nEnergyZeroB \\
public TopK & \nModelNameC & \nLayerUpC, \nLayerDownC & \nAuditPerDoseZeroC & \nCrossZeroCountC & \nCrossHalfCountC & \nCrossNineCountC & \nZeroResponseC & \nEnterShareC & \nEnergyZeroC \\
public TopK & \nModelNameD & \nLayerUpD, \nLayerDownD & \nAuditPerDoseZeroD & \nCrossZeroCountD & \nCrossHalfCountD & \nCrossNineCountD & \nZeroResponseD & \nEnterShareD & \nEnergyZeroD \\
public TopK & \nModelNameE & \nLayerUpE, \nLayerDownE & \nAuditPerDoseZeroE & \nCrossZeroCountE & \nCrossHalfCountE & \nCrossNineCountE & \nZeroResponseE & \nEnterShareE & \nEnergyZeroE \\
public ReLU & \nModelNameR & \nLayerUpR, \nLayerDownR & \nAuditPerDoseZeroR & \nCrossZeroCountR & \nCrossHalfCountR & \nCrossNineCountR & \nZeroResponseR & \nEnterShareR & \nEnergyZeroR \\
ours & \nModelNameF & \nLayerUpA, \nLayerDownA & \nAuditPerDoseZeroF & \nCrossZeroCountF & \nCrossHalfCountF & \nCrossNineCountF & \nZeroResponseF & \nEnterShareF & \nEnergyZeroF \\
\bottomrule
\end{tabular}
\par\vspace{2pt}\footnotesize A sparse autoencoder (SAE) represents model activations by sparse feature coefficients. Counts cover every selected feature at \nTokens held out Pile validation positions per pair, summed over the \nRunsF runs of our SAEs. A selection change is an intervention that changes at least one selected downstream feature. The two shares average, over the zero ablations that change the selection, the squared coefficient change on the named features over the whole change's squared norm, and our SAEs average their runs. Every position of every pair had a selected upstream feature, so no position was excluded, and no other exclusion applies.

\par\noindent TopK selects the largest positive encoder scores, and ReLU, the rectified linear unit, every positive score. The scaling factor multiplies a selected upstream feature's coefficient, and zero removes that feature. The shares are fractions with no unit.
\end{table}

\begin{table}[H]
\caption{\textbf{Every run of our sparse autoencoders (SAEs) orders the two graphs alike.} Cells give each run's median relative coefficient error for the standard and the margin graph, at each scaling factor and pooled over zero, \nDoseHalf and \nDoseNine.}
\label{tab:errors-runs}
\centering
\scriptsize
\setlength{\tabcolsep}{2.5pt}
\begin{tabular}{lrrrrrrrr}
\toprule
run & \begin{tabular}[c]{@{}c@{}}standard\\at zero\end{tabular} & \begin{tabular}[c]{@{}c@{}}margin\\at zero\end{tabular} & \begin{tabular}[c]{@{}c@{}}standard\\at \nDoseHalf\end{tabular} & \begin{tabular}[c]{@{}c@{}}margin\\at \nDoseHalf\end{tabular} & \begin{tabular}[c]{@{}c@{}}standard\\at \nDoseNine\end{tabular} & \begin{tabular}[c]{@{}c@{}}margin\\at \nDoseNine\end{tabular} & \begin{tabular}[c]{@{}c@{}}standard\\pooled\end{tabular} & \begin{tabular}[c]{@{}c@{}}margin\\pooled\end{tabular} \\
\midrule
run \nRunIndexZero & \nFixedErrRunZeroZero & \nMarginErrRunZeroZero & \nFixedErrRunZeroHalf & \nMarginErrRunZeroHalf & \nFixedErrRunZeroNine & \nMarginErrRunZeroNine & \nFixedErrRunZero & \nMarginErrRunZero \\
run \nRunIndexOne & \nFixedErrRunOneZero & \nMarginErrRunOneZero & \nFixedErrRunOneHalf & \nMarginErrRunOneHalf & \nFixedErrRunOneNine & \nMarginErrRunOneNine & \nFixedErrRunOne & \nMarginErrRunOne \\
run \nRunIndexTwo & \nFixedErrRunTwoZero & \nMarginErrRunTwoZero & \nFixedErrRunTwoHalf & \nMarginErrRunTwoHalf & \nFixedErrRunTwoNine & \nMarginErrRunTwoNine & \nFixedErrRunTwo & \nMarginErrRunTwo \\
run \nRunIndexThree & \nFixedErrRunThreeZero & \nMarginErrRunThreeZero & \nFixedErrRunThreeHalf & \nMarginErrRunThreeHalf & \nFixedErrRunThreeNine & \nMarginErrRunThreeNine & \nFixedErrRunThree & \nMarginErrRunThree \\
run \nRunIndexFour & \nFixedErrRunFourZero & \nMarginErrRunFourZero & \nFixedErrRunFourHalf & \nMarginErrRunFourHalf & \nFixedErrRunFourNine & \nMarginErrRunFourNine & \nFixedErrRunFour & \nMarginErrRunFour \\
run \nRunIndexFive & \nFixedErrRunFiveZero & \nMarginErrRunFiveZero & \nFixedErrRunFiveHalf & \nMarginErrRunFiveHalf & \nFixedErrRunFiveNine & \nMarginErrRunFiveNine & \nFixedErrRunFive & \nMarginErrRunFive \\
run \nRunIndexSix & \nFixedErrRunSixZero & \nMarginErrRunSixZero & \nFixedErrRunSixHalf & \nMarginErrRunSixHalf & \nFixedErrRunSixNine & \nMarginErrRunSixNine & \nFixedErrRunSix & \nMarginErrRunSix \\
run \nRunIndexSeven & \nFixedErrRunSevenZero & \nMarginErrRunSevenZero & \nFixedErrRunSevenHalf & \nMarginErrRunSevenHalf & \nFixedErrRunSevenNine & \nMarginErrRunSevenNine & \nFixedErrRunSeven & \nMarginErrRunSeven \\
\bottomrule
\end{tabular}

\par\noindent Our SAEs are on \nModelNameF at layers \nLayerUpA and \nLayerDownA, of width \nWidthF and TopK \nKF. Each factor uses \nInterventionsPerRunF interventions at \nTokens held out Pile validation positions, and the six seed runs and the data order pair share the auxiliary coefficient of Table~\ref{tab:aux}. The standard graph keeps the features selected before the intervention, and the margin graph selects again from the predicted scores. The relative coefficient error divides the norm of a prediction's error by the norm of the exact coefficient change, and smaller is better. The scaling factor multiplies a selected upstream feature's coefficient, and zero removes that feature. TopK selects the largest positive encoder scores, and each run label names one trained pair.
\end{table}

We fixed a bootstrap over the positions before running it, to bound the two graphs' medians and their difference. It draws the \nTokens positions with replacement \nBootDraws times from seed \nBootSeed and keeps every intervention at a drawn position once per draw. No resample on any pair puts the difference at or under zero, and Table~\ref{tab:bootstrap} gives the interval of each statistic.

\begin{table}[H]
\caption{\textbf{Every bootstrap comparison favours the margin graph.} Intervals give \nBootInterval of \nBootDraws resamples of positions for each graph's median relative coefficient error and for their difference, the standard graph's error less the margin graph's. The last column gives the fraction of resamples whose difference is zero or less.}
\label{tab:bootstrap}
\centering
\scriptsize
\setlength{\tabcolsep}{2.5pt}
\begin{tabular}{llllllr}
\toprule
SAEs & model & layers & standard graph & margin graph & difference & \begin{tabular}[c]{@{}c@{}}resample fraction at\\or under zero\end{tabular} \\
\midrule
public TopK & \nModelNameA & \nLayerUpA, \nLayerDownA & \nBootFixedLowA to \nBootFixedHighA & \nBootMarginLowA to \nBootMarginHighA & \nBootDiffLowA to \nBootDiffHighA & \nBootAtOrUnderZeroA \\
public TopK & \nModelNameB & \nLayerUpB, \nLayerDownB & \nBootFixedLowB to \nBootFixedHighB & \nBootMarginLowB to \nBootMarginHighB & \nBootDiffLowB to \nBootDiffHighB & \nBootAtOrUnderZeroB \\
public TopK & \nModelNameC & \nLayerUpC, \nLayerDownC & \nBootFixedLowC to \nBootFixedHighC & \nBootMarginLowC to \nBootMarginHighC & \nBootDiffLowC to \nBootDiffHighC & \nBootAtOrUnderZeroC \\
public TopK & \nModelNameD & \nLayerUpD, \nLayerDownD & \nBootFixedLowD to \nBootFixedHighD & \nBootMarginLowD to \nBootMarginHighD & \nBootDiffLowD to \nBootDiffHighD & \nBootAtOrUnderZeroD \\
public TopK & \nModelNameE & \nLayerUpE, \nLayerDownE & \nBootFixedLowE to \nBootFixedHighE & \nBootMarginLowE to \nBootMarginHighE & \nBootDiffLowE to \nBootDiffHighE & \nBootAtOrUnderZeroE \\
public ReLU & \nModelNameR & \nLayerUpR, \nLayerDownR & \nBootFixedLowR to \nBootFixedHighR & \nBootMarginLowR to \nBootMarginHighR & \nBootDiffLowR to \nBootDiffHighR & \nBootAtOrUnderZeroR \\
ours & \nModelNameF & \nLayerUpA, \nLayerDownA & \nBootFixedLowF to \nBootFixedHighF & \nBootMarginLowF to \nBootMarginHighF & \nBootDiffLowF to \nBootDiffHighF & \nBootAtOrUnderZeroF \\
\bottomrule
\end{tabular}

\par\noindent A sparse autoencoder (SAE) represents model activations by sparse feature coefficients. Each resample draws the \nTokens held out Pile validation positions with replacement and keeps every feature and the factors zero, \nDoseHalf and \nDoseNine at each drawn position. Each draw for our SAEs averages the medians of their \nRunsF runs at the same drawn positions. TopK selects the largest positive encoder scores, and ReLU, the rectified linear unit, every positive score. The standard graph keeps the features selected before the intervention, and the margin graph selects again from the predicted scores. The relative coefficient error divides the norm of a prediction's error by the norm of the exact coefficient change, and smaller is better.
\end{table}

Table~\ref{tab:graphs} gives both graphs' errors pooled over the three scaling factors, on the coefficients and on their decoded change, with the margin graph's precision and recall on the entering features.

\begin{table}[H]
\caption{\textbf{Pooling the scaling factors keeps the margin graph's advantage on the TopK pairs.} Cells give median errors of the predicted coefficients and of their decoded change, each relative to the exact change's norm in its own space, and the margin graph's precision and recall on the entering features.}
\label{tab:graphs}
\centering
\scriptsize
\setlength{\tabcolsep}{2.5pt}
\begin{tabular}{lllrrrrrr}
\toprule
SAEs & model & layers & \begin{tabular}[c]{@{}c@{}}standard graph\\coefficients\end{tabular} & \begin{tabular}[c]{@{}c@{}}margin graph\\coefficients\end{tabular} & \begin{tabular}[c]{@{}c@{}}standard graph\\decoded\end{tabular} & \begin{tabular}[c]{@{}c@{}}margin graph\\decoded\end{tabular} & precision & recall \\
\midrule
public TopK & \nModelNameA & \nLayerUpA, \nLayerDownA & \nFixedErrA & \nMarginErrA & \nFixedErrDecA & \nMarginErrDecA & \nPrecA & \nRecA \\
public TopK & \nModelNameB & \nLayerUpB, \nLayerDownB & \nFixedErrB & \nMarginErrB & \nFixedErrDecB & \nMarginErrDecB & \nPrecB & \nRecB \\
public TopK & \nModelNameC & \nLayerUpC, \nLayerDownC & \nFixedErrC & \nMarginErrC & \nFixedErrDecC & \nMarginErrDecC & \nPrecC & \nRecC \\
public TopK & \nModelNameD & \nLayerUpD, \nLayerDownD & \nFixedErrD & \nMarginErrD & \nFixedErrDecD & \nMarginErrDecD & \nPrecD & \nRecD \\
public TopK & \nModelNameE & \nLayerUpE, \nLayerDownE & \nFixedErrE & \nMarginErrE & \nFixedErrDecE & \nMarginErrDecE & \nPrecE & \nRecE \\
public ReLU & \nModelNameR & \nLayerUpR, \nLayerDownR & \nFixedErrR & \nMarginErrR & \nFixedErrDecR & \nMarginErrDecR & \nPrecR & \nRecR \\
ours & \nModelNameF & \nLayerUpA, \nLayerDownA & \nFixedErrF & \nMarginErrF & \nFixedErrDecF & \nMarginErrDecF & \nPrecF & \nRecF \\
\bottomrule
\end{tabular}

\par\noindent A sparse autoencoder (SAE) represents model activations by sparse feature coefficients. Cells pool every selected feature at \nTokens held out Pile validation positions and the factors zero, \nDoseHalf and \nDoseNine. Precision and recall pool the confusion counts of the entering features, cells for our SAEs average \nRunsF runs, and Table~\ref{tab:counts} gives the counts by pair. TopK selects the largest positive encoder scores, and ReLU, the rectified linear unit, every positive score. The standard graph keeps the features selected before the intervention, and the margin graph selects again from the predicted scores. Decoding maps the coefficient vector back to model activations. Precision is the share of predicted entering features that enter, and recall the share of entering features predicted. Every value has no unit.
\end{table}

\begin{table}[H]
\caption{\textbf{The margin and the predicted score change together predict the entering features better than either alone.} Cells give each predictor's precision or recall on the entering features, on the public \nModelNameA pair and on our sparse autoencoders (SAEs). The last five predictors receive the true number of entering features, so their precision equals their recall.}
\label{tab:support}
\centering
\footnotesize
\setlength{\tabcolsep}{6pt}
\begin{tabular}{lrr}
\toprule
predictor of the entering features & public \nModelNameA & ours \\
\midrule
margin graph, precision & \nPrecA & \nPrecF \\
margin graph, recall & \nRecA & \nRecF \\
margin graph ranking, true count & \nPrecTrueCountA & \nPrecTrueCountF \\
margin alone, true count & \nPrecMarginAloneA & \nPrecMarginAloneF \\
predicted change alone, true count & \nPrecEdgeAloneA & \nPrecEdgeAloneF \\
uniform draw, true count & \nPrecNullA & \nPrecNullF \\
exact oracle, true count & \nOraclePrecAll & \nOraclePrecAll \\
\bottomrule
\end{tabular}

\par\noindent Precision and recall pool confusion counts over the factors zero, \nDoseHalf and \nDoseNine at \nTokens held out Pile validation positions, on the public pair at layers \nLayerUpA and \nLayerDownA and on \nRunsF runs of our SAEs on the same model and layers, at width \nWidthF and TopK \nKF. A sparse autoencoder (SAE) represents model activations by sparse feature coefficients. TopK selects the largest positive encoder scores, and ReLU every positive score. The standard graph keeps the features selected before the intervention, and the margin graph selects again from the predicted scores. A margin is a score's signed distance from a change of selection, and the exact oracle selects from the recomputed scores. Precision divides correct predictions by predicted entering features, and recall divides them by actual entering features. Both are fractions with no unit, and higher is better.
\end{table}

The predictors of entering features in Section~\ref{sec:margin} rank the features unselected before the intervention. The margin graph is scored at its own count, and its ordering at the true count $n$ of entering features ranks those features by their predicted encoder scores. Each predictor below takes in the true count of entering features and removes the true count of leaving features, which differ under the ReLU rule, whose selected count moves. The margin alone lets enter the unselected features whose margins are smallest in size and lets leave the selected features whose margins are smallest. The predicted change alone lets enter the unselected features with the largest predicted score increase and lets leave the selected features with the largest predicted decrease. The uniform draw moves unselected features drawn uniformly at random into the selected set and selected features out of it, from seed \nNullSeed plus the run on our SAEs and from seed \nNullSeed on the public pairs. The exact oracle selects again from the exact encoder scores at the true count, and it gives the entering set at a precision and a recall of \nOraclePrecAll on every pair.

We treat a feature as selected where its encoder score is positive and at or above the $k$th largest, and no exact encoder score was tied at that boundary on any of the \nAuditRowsAll interventions the oracle scored. The results for the margin alone, the predicted change alone and the uniform draw list each one's hits beside its predicted and true counts. Its true positives are the hits, and its false positives and false negatives are the two counts less the hits. Table~\ref{tab:confusion} gives the confusion counts of all six predictors on the entering set, where only the exact oracle reaches perfect precision and recall and the uniform draw stays near zero.

\begingroup
\scriptsize
\setlength{\tabcolsep}{2.5pt}
\begin{longtable}{lllrrrrrr}
\caption{\normalsize{}\textbf{The margin graph finds most entering features.} Cells count each predictor's true positives, entering features predicted to enter, its false positives, predicted entering features that do not enter, and its false negatives, entering features not predicted.}
\label{tab:confusion}\\
\toprule
SAEs & model & layers & predictor & \begin{tabular}[c]{@{}c@{}}true\\positives\end{tabular} & \begin{tabular}[c]{@{}c@{}}false\\positives\end{tabular} & \begin{tabular}[c]{@{}c@{}}false\\negatives\end{tabular} & precision & recall \\
\midrule
\endfirsthead
\toprule
SAEs & model & layers & predictor & \begin{tabular}[c]{@{}c@{}}true\\positives\end{tabular} & \begin{tabular}[c]{@{}c@{}}false\\positives\end{tabular} & \begin{tabular}[c]{@{}c@{}}false\\negatives\end{tabular} & precision & recall \\
\midrule
\endhead
public TopK & \nModelNameA & \nLayerUpA, \nLayerDownA & exact oracle & \nOracleTPA & \nOracleFPA & \nOracleFNA & \nOraclePrecAuditA & \nOracleRecAuditA \\
 &  &  & margin graph & \nMgTPA & \nMgFPA & \nMgFNA & \nMgPrecAuditA & \nMgRecAuditA \\
 &  &  & at the true count & \nTcTPA & \nTcFPA & \nTcFNA & \nTcPrecAuditA & \nTcRecAuditA \\
 &  &  & margin alone & \nMaEnterTPA & \nMaEnterFPA & \nMaEnterFNA & \nMaEnterPrecA & \nMaEnterRecA \\
 &  &  & predicted change alone & \nEaEnterTPA & \nEaEnterFPA & \nEaEnterFNA & \nEaEnterPrecA & \nEaEnterRecA \\
 &  &  & uniform draw & \nNlEnterTPA & \nNlEnterFPA & \nNlEnterFNA & \nNlEnterPrecA & \nNlEnterRecA \\
public TopK & \nModelNameB & \nLayerUpB, \nLayerDownB & exact oracle & \nOracleTPB & \nOracleFPB & \nOracleFNB & \nOraclePrecAuditB & \nOracleRecAuditB \\
 &  &  & margin graph & \nMgTPB & \nMgFPB & \nMgFNB & \nMgPrecAuditB & \nMgRecAuditB \\
 &  &  & at the true count & \nTcTPB & \nTcFPB & \nTcFNB & \nTcPrecAuditB & \nTcRecAuditB \\
 &  &  & margin alone & \nMaEnterTPB & \nMaEnterFPB & \nMaEnterFNB & \nMaEnterPrecB & \nMaEnterRecB \\
 &  &  & predicted change alone & \nEaEnterTPB & \nEaEnterFPB & \nEaEnterFNB & \nEaEnterPrecB & \nEaEnterRecB \\
 &  &  & uniform draw & \nNlEnterTPB & \nNlEnterFPB & \nNlEnterFNB & \nNlEnterPrecB & \nNlEnterRecB \\
public TopK & \nModelNameC & \nLayerUpC, \nLayerDownC & exact oracle & \nOracleTPC & \nOracleFPC & \nOracleFNC & \nOraclePrecAuditC & \nOracleRecAuditC \\
 &  &  & margin graph & \nMgTPC & \nMgFPC & \nMgFNC & \nMgPrecAuditC & \nMgRecAuditC \\
 &  &  & at the true count & \nTcTPC & \nTcFPC & \nTcFNC & \nTcPrecAuditC & \nTcRecAuditC \\
 &  &  & margin alone & \nMaEnterTPC & \nMaEnterFPC & \nMaEnterFNC & \nMaEnterPrecC & \nMaEnterRecC \\
 &  &  & predicted change alone & \nEaEnterTPC & \nEaEnterFPC & \nEaEnterFNC & \nEaEnterPrecC & \nEaEnterRecC \\
 &  &  & uniform draw & \nNlEnterTPC & \nNlEnterFPC & \nNlEnterFNC & \nNlEnterPrecC & \nNlEnterRecC \\
public TopK & \nModelNameD & \nLayerUpD, \nLayerDownD & exact oracle & \nOracleTPD & \nOracleFPD & \nOracleFND & \nOraclePrecAuditD & \nOracleRecAuditD \\
 &  &  & margin graph & \nMgTPD & \nMgFPD & \nMgFND & \nMgPrecAuditD & \nMgRecAuditD \\
 &  &  & at the true count & \nTcTPD & \nTcFPD & \nTcFND & \nTcPrecAuditD & \nTcRecAuditD \\
 &  &  & margin alone & \nMaEnterTPD & \nMaEnterFPD & \nMaEnterFND & \nMaEnterPrecD & \nMaEnterRecD \\
 &  &  & predicted change alone & \nEaEnterTPD & \nEaEnterFPD & \nEaEnterFND & \nEaEnterPrecD & \nEaEnterRecD \\
 &  &  & uniform draw & \nNlEnterTPD & \nNlEnterFPD & \nNlEnterFND & \nNlEnterPrecD & \nNlEnterRecD \\
public TopK & \nModelNameE & \nLayerUpE, \nLayerDownE & exact oracle & \nOracleTPE & \nOracleFPE & \nOracleFNE & \nOraclePrecAuditE & \nOracleRecAuditE \\
 &  &  & margin graph & \nMgTPE & \nMgFPE & \nMgFNE & \nMgPrecAuditE & \nMgRecAuditE \\
 &  &  & at the true count & \nTcTPE & \nTcFPE & \nTcFNE & \nTcPrecAuditE & \nTcRecAuditE \\
 &  &  & margin alone & \nMaEnterTPE & \nMaEnterFPE & \nMaEnterFNE & \nMaEnterPrecE & \nMaEnterRecE \\
 &  &  & predicted change alone & \nEaEnterTPE & \nEaEnterFPE & \nEaEnterFNE & \nEaEnterPrecE & \nEaEnterRecE \\
 &  &  & uniform draw & \nNlEnterTPE & \nNlEnterFPE & \nNlEnterFNE & \nNlEnterPrecE & \nNlEnterRecE \\
public ReLU & \nModelNameR & \nLayerUpR, \nLayerDownR & exact oracle & \nOracleTPR & \nOracleFPR & \nOracleFNR & \nOraclePrecAuditR & \nOracleRecAuditR \\
 &  &  & margin graph & \nMgTPR & \nMgFPR & \nMgFNR & \nMgPrecAuditR & \nMgRecAuditR \\
 &  &  & at the true count & \nTcTPR & \nTcFPR & \nTcFNR & \nTcPrecAuditR & \nTcRecAuditR \\
 &  &  & margin alone & \nMaEnterTPR & \nMaEnterFPR & \nMaEnterFNR & \nMaEnterPrecR & \nMaEnterRecR \\
 &  &  & predicted change alone & \nEaEnterTPR & \nEaEnterFPR & \nEaEnterFNR & \nEaEnterPrecR & \nEaEnterRecR \\
 &  &  & uniform draw & \nNlEnterTPR & \nNlEnterFPR & \nNlEnterFNR & \nNlEnterPrecR & \nNlEnterRecR \\
ours & \nModelNameF & \nLayerUpA, \nLayerDownA & exact oracle & \nOracleTPF & \nOracleFPF & \nOracleFNF & \nOraclePrecAuditF & \nOracleRecAuditF \\
 &  &  & margin graph & \nMgTPF & \nMgFPF & \nMgFNF & \nMgPrecAuditF & \nMgRecAuditF \\
 &  &  & at the true count & \nTcTPF & \nTcFPF & \nTcFNF & \nTcPrecAuditF & \nTcRecAuditF \\
 &  &  & margin alone & \nMaEnterTPF & \nMaEnterFPF & \nMaEnterFNF & \nMaEnterPrecF & \nMaEnterRecF \\
 &  &  & predicted change alone & \nEaEnterTPF & \nEaEnterFPF & \nEaEnterFNF & \nEaEnterPrecF & \nEaEnterRecF \\
 &  &  & uniform draw & \nNlEnterTPF & \nNlEnterFPF & \nNlEnterFNF & \nNlEnterPrecF & \nNlEnterRecF \\
\bottomrule
\end{longtable}
\par\vspace{2pt}\footnotesize Counts pool every intervention on a selected feature at \nTokens held out Pile validation positions and the factors zero, \nDoseHalf and \nDoseNine, summed over the \nRunsF runs of our sparse autoencoders (SAEs). The exact oracle selects from the exact encoder scores, the margin graph chooses its own count, and the other predictors receive the true number of entering features. The precision is the true positives over the predicted count and the recall the true positives over the true count, each micro averaged over the interventions, and higher is better.

\par\noindent A sparse autoencoder (SAE) represents model activations by sparse feature coefficients. TopK selects the largest positive encoder scores, and ReLU, the rectified linear unit, every positive score. The standard graph keeps the features selected before the intervention, and the margin graph selects again from the predicted scores. At the true count ranks the margin graph's predicted scores. Margin alone ranks scores by their distance from a change of selection, and predicted change alone ranks the predicted score changes. Precision and recall are fractions with no unit.
\endgroup

Table~\ref{tab:confusion-leave} gives the same counts on the leaving set, where the margin graph's precision runs from \nLeavePrecMin to \nLeavePrecMax and its recall from \nLeaveRecMin to \nLeaveRecMax on the \nDiscussionPublicPairsWords public TopK pairs.

\begingroup
\scriptsize
\setlength{\tabcolsep}{2.5pt}
\begin{longtable}{lllrrrrrr}
\caption{\normalsize{}\textbf{The margin graph finds most leaving features.} Cells count each predictor's true positives, leaving features predicted to leave, its false positives, predicted leaving features that stay, and its false negatives, leaving features not predicted.}
\label{tab:confusion-leave}\\
\toprule
SAEs & model & layers & predictor & \begin{tabular}[c]{@{}c@{}}true\\positives\end{tabular} & \begin{tabular}[c]{@{}c@{}}false\\positives\end{tabular} & \begin{tabular}[c]{@{}c@{}}false\\negatives\end{tabular} & precision & recall \\
\midrule
\endfirsthead
\toprule
SAEs & model & layers & predictor & \begin{tabular}[c]{@{}c@{}}true\\positives\end{tabular} & \begin{tabular}[c]{@{}c@{}}false\\positives\end{tabular} & \begin{tabular}[c]{@{}c@{}}false\\negatives\end{tabular} & precision & recall \\
\midrule
\endhead
public TopK & \nModelNameA & \nLayerUpA, \nLayerDownA & exact oracle & \nOracleLeaveTPA & \nOracleLeaveFPA & \nOracleLeaveFNA & \nOracleLeavePrecA & \nOracleLeaveRecA \\
 &  &  & margin graph & \nMgLeaveTPA & \nMgLeaveFPA & \nMgLeaveFNA & \nMgLeavePrecA & \nMgLeaveRecA \\
 &  &  & at the true count & \nTcLeaveTPA & \nTcLeaveFPA & \nTcLeaveFNA & \nTcLeavePrecA & \nTcLeaveRecA \\
 &  &  & margin alone & \nMaLeaveTPA & \nMaLeaveFPA & \nMaLeaveFNA & \nMaLeavePrecA & \nMaLeaveRecA \\
 &  &  & predicted change alone & \nEaLeaveTPA & \nEaLeaveFPA & \nEaLeaveFNA & \nEaLeavePrecA & \nEaLeaveRecA \\
 &  &  & uniform draw & \nNlLeaveTPA & \nNlLeaveFPA & \nNlLeaveFNA & \nNlLeavePrecA & \nNlLeaveRecA \\
public TopK & \nModelNameB & \nLayerUpB, \nLayerDownB & exact oracle & \nOracleLeaveTPB & \nOracleLeaveFPB & \nOracleLeaveFNB & \nOracleLeavePrecB & \nOracleLeaveRecB \\
 &  &  & margin graph & \nMgLeaveTPB & \nMgLeaveFPB & \nMgLeaveFNB & \nMgLeavePrecB & \nMgLeaveRecB \\
 &  &  & at the true count & \nTcLeaveTPB & \nTcLeaveFPB & \nTcLeaveFNB & \nTcLeavePrecB & \nTcLeaveRecB \\
 &  &  & margin alone & \nMaLeaveTPB & \nMaLeaveFPB & \nMaLeaveFNB & \nMaLeavePrecB & \nMaLeaveRecB \\
 &  &  & predicted change alone & \nEaLeaveTPB & \nEaLeaveFPB & \nEaLeaveFNB & \nEaLeavePrecB & \nEaLeaveRecB \\
 &  &  & uniform draw & \nNlLeaveTPB & \nNlLeaveFPB & \nNlLeaveFNB & \nNlLeavePrecB & \nNlLeaveRecB \\
public TopK & \nModelNameC & \nLayerUpC, \nLayerDownC & exact oracle & \nOracleLeaveTPC & \nOracleLeaveFPC & \nOracleLeaveFNC & \nOracleLeavePrecC & \nOracleLeaveRecC \\
 &  &  & margin graph & \nMgLeaveTPC & \nMgLeaveFPC & \nMgLeaveFNC & \nMgLeavePrecC & \nMgLeaveRecC \\
 &  &  & at the true count & \nTcLeaveTPC & \nTcLeaveFPC & \nTcLeaveFNC & \nTcLeavePrecC & \nTcLeaveRecC \\
 &  &  & margin alone & \nMaLeaveTPC & \nMaLeaveFPC & \nMaLeaveFNC & \nMaLeavePrecC & \nMaLeaveRecC \\
 &  &  & predicted change alone & \nEaLeaveTPC & \nEaLeaveFPC & \nEaLeaveFNC & \nEaLeavePrecC & \nEaLeaveRecC \\
 &  &  & uniform draw & \nNlLeaveTPC & \nNlLeaveFPC & \nNlLeaveFNC & \nNlLeavePrecC & \nNlLeaveRecC \\
public TopK & \nModelNameD & \nLayerUpD, \nLayerDownD & exact oracle & \nOracleLeaveTPD & \nOracleLeaveFPD & \nOracleLeaveFND & \nOracleLeavePrecD & \nOracleLeaveRecD \\
 &  &  & margin graph & \nMgLeaveTPD & \nMgLeaveFPD & \nMgLeaveFND & \nMgLeavePrecD & \nMgLeaveRecD \\
 &  &  & at the true count & \nTcLeaveTPD & \nTcLeaveFPD & \nTcLeaveFND & \nTcLeavePrecD & \nTcLeaveRecD \\
 &  &  & margin alone & \nMaLeaveTPD & \nMaLeaveFPD & \nMaLeaveFND & \nMaLeavePrecD & \nMaLeaveRecD \\
 &  &  & predicted change alone & \nEaLeaveTPD & \nEaLeaveFPD & \nEaLeaveFND & \nEaLeavePrecD & \nEaLeaveRecD \\
 &  &  & uniform draw & \nNlLeaveTPD & \nNlLeaveFPD & \nNlLeaveFND & \nNlLeavePrecD & \nNlLeaveRecD \\
public TopK & \nModelNameE & \nLayerUpE, \nLayerDownE & exact oracle & \nOracleLeaveTPE & \nOracleLeaveFPE & \nOracleLeaveFNE & \nOracleLeavePrecE & \nOracleLeaveRecE \\
 &  &  & margin graph & \nMgLeaveTPE & \nMgLeaveFPE & \nMgLeaveFNE & \nMgLeavePrecE & \nMgLeaveRecE \\
 &  &  & at the true count & \nTcLeaveTPE & \nTcLeaveFPE & \nTcLeaveFNE & \nTcLeavePrecE & \nTcLeaveRecE \\
 &  &  & margin alone & \nMaLeaveTPE & \nMaLeaveFPE & \nMaLeaveFNE & \nMaLeavePrecE & \nMaLeaveRecE \\
 &  &  & predicted change alone & \nEaLeaveTPE & \nEaLeaveFPE & \nEaLeaveFNE & \nEaLeavePrecE & \nEaLeaveRecE \\
 &  &  & uniform draw & \nNlLeaveTPE & \nNlLeaveFPE & \nNlLeaveFNE & \nNlLeavePrecE & \nNlLeaveRecE \\
public ReLU & \nModelNameR & \nLayerUpR, \nLayerDownR & exact oracle & \nOracleLeaveTPR & \nOracleLeaveFPR & \nOracleLeaveFNR & \nOracleLeavePrecR & \nOracleLeaveRecR \\
 &  &  & margin graph & \nMgLeaveTPR & \nMgLeaveFPR & \nMgLeaveFNR & \nMgLeavePrecR & \nMgLeaveRecR \\
 &  &  & at the true count & \nTcLeaveTPR & \nTcLeaveFPR & \nTcLeaveFNR & \nTcLeavePrecR & \nTcLeaveRecR \\
 &  &  & margin alone & \nMaLeaveTPR & \nMaLeaveFPR & \nMaLeaveFNR & \nMaLeavePrecR & \nMaLeaveRecR \\
 &  &  & predicted change alone & \nEaLeaveTPR & \nEaLeaveFPR & \nEaLeaveFNR & \nEaLeavePrecR & \nEaLeaveRecR \\
 &  &  & uniform draw & \nNlLeaveTPR & \nNlLeaveFPR & \nNlLeaveFNR & \nNlLeavePrecR & \nNlLeaveRecR \\
ours & \nModelNameF & \nLayerUpA, \nLayerDownA & exact oracle & \nOracleLeaveTPF & \nOracleLeaveFPF & \nOracleLeaveFNF & \nOracleLeavePrecF & \nOracleLeaveRecF \\
 &  &  & margin graph & \nMgLeaveTPF & \nMgLeaveFPF & \nMgLeaveFNF & \nMgLeavePrecF & \nMgLeaveRecF \\
 &  &  & at the true count & \nTcLeaveTPF & \nTcLeaveFPF & \nTcLeaveFNF & \nTcLeavePrecF & \nTcLeaveRecF \\
 &  &  & margin alone & \nMaLeaveTPF & \nMaLeaveFPF & \nMaLeaveFNF & \nMaLeavePrecF & \nMaLeaveRecF \\
 &  &  & predicted change alone & \nEaLeaveTPF & \nEaLeaveFPF & \nEaLeaveFNF & \nEaLeavePrecF & \nEaLeaveRecF \\
 &  &  & uniform draw & \nNlLeaveTPF & \nNlLeaveFPF & \nNlLeaveFNF & \nNlLeavePrecF & \nNlLeaveRecF \\
\bottomrule
\end{longtable}
\par\vspace{2pt}\footnotesize Counts pool every intervention on a selected feature at \nTokens held out Pile validation positions and the factors zero, \nDoseHalf and \nDoseNine, summed over the \nRunsF runs of our sparse autoencoders (SAEs). The exact oracle selects from the exact encoder scores, the margin graph chooses its own count, and the other predictors receive the true number of leaving features. The precision is the true positives over the predicted count and the recall the true positives over the true count, each micro averaged over the interventions, and higher is better.

\par\noindent A sparse autoencoder (SAE) represents model activations by sparse feature coefficients. TopK selects the largest positive encoder scores, and ReLU, the rectified linear unit, every positive score. The standard graph keeps the features selected before the intervention, and the margin graph selects again from the predicted scores. At the true count ranks the margin graph's predicted scores. Margin alone ranks selected scores by their distance from leaving the selection, and predicted change alone ranks the predicted score decreases. Precision and recall are fractions with no unit.
\endgroup

\begin{figure}[H]
\centering
\includegraphics[width=\textwidth]{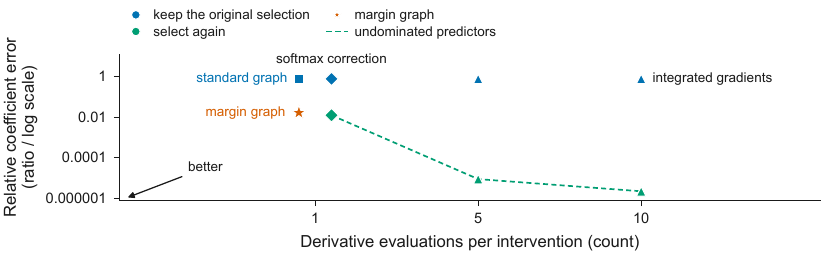}
\caption{\textbf{Selecting features again lowers the coefficient error of both corrections.} Each mark plots a predictor's median relative coefficient error on our sparse autoencoders against the derivative evaluations, each one Jacobian vector product, that one intervention needs. The margin graph, in orange, is the standard graph that selects again. The dashed line joins the predictors that no other beats on both axes, and marks at one evaluation are set apart so that none hides another.}
\label{fig:corrections}
\end{figure}

Of the corrections of Section~\ref{sec:margin}, integrated gradients average the first order term at $n$ points along the intervention, at the fractions $(s - \tfrac{1}{2})/n$ for $s$ from one to $n$, at \nDevIgFiveFixedProducts and \nDevIgTenFixedProducts JVPs. The softmax correction recomputes the attention weights exactly at the intervened input, through the first layer norm, the query and key projections of the intervened position with the earlier positions' keys fixed, the rotary rotation, the scores and the softmax. The exact weights then combine the earlier positions' fixed values with the position's own value taken at first order. The multilayer perceptron, which takes the block's input through the second layer norm, is taken at first order by the same JVP along the intervention, and the output projection applies exactly. We evaluate each correction twice, keeping the features selected before the intervention and selecting again, and Figure~\ref{fig:corrections} compares them on our SAEs. The two softmax corrections were also evaluated on the \nDiscussionPublicPairsWords public TopK pairs, over one intervention on every selected upstream feature at each of the \nTokens positions at the three scaling factors. Table~\ref{tab:devices-pairs} gives their errors beside both graphs', with each correction's JVPs and retained rows.

\begin{table}[H]
\caption{\textbf{The softmax correction that selects again lowers the error on every public TopK pair.} Cells give median relative coefficient errors at each scaling factor and pooled. The last two columns give the median of the correction's error less the margin graph's on each intervention, negative where the correction errs less, and the share of interventions where it does.}
\label{tab:devices-pairs}
\centering
\scriptsize
\setlength{\tabcolsep}{2.5pt}
\begin{tabular}{lllrrrrrr}
\toprule
SAEs & layers & correction & at zero & at \nDoseHalf & at \nDoseNine & pooled & \begin{tabular}[c]{@{}c@{}}error difference\\median\end{tabular} & \begin{tabular}[c]{@{}c@{}}fraction of interventions\\with lower error\end{tabular} \\
\midrule
public \nModelNameA & \nLayerUpA, \nLayerDownA & standard graph & \nDevTabFxZeroA & \nDevTabFxHalfA & \nDevTabFxNineA & \nDevTabFxPooledA &  &  \\
 &  & margin graph & \nDevTabMgZeroA & \nDevTabMgHalfA & \nDevTabMgNineA & \nDevTabMgPooledA &  &  \\
 &  & softmax, original selection & \nDevTabSfZeroA & \nDevTabSfHalfA & \nDevTabSfNineA & \nDevTabSfPooledA &  &  \\
 &  & softmax, selects again & \nDevTabSrZeroA & \nDevTabSrHalfA & \nDevTabSrNineA & \nDevTabSrPooledA & $\nDevTabPairedA$ & \nDevTabBelowA \\
public \nModelNameB & \nLayerUpB, \nLayerDownB & standard graph & \nDevTabFxZeroB & \nDevTabFxHalfB & \nDevTabFxNineB & \nDevTabFxPooledB &  &  \\
 &  & margin graph & \nDevTabMgZeroB & \nDevTabMgHalfB & \nDevTabMgNineB & \nDevTabMgPooledB &  &  \\
 &  & softmax, original selection & \nDevTabSfZeroB & \nDevTabSfHalfB & \nDevTabSfNineB & \nDevTabSfPooledB &  &  \\
 &  & softmax, selects again & \nDevTabSrZeroB & \nDevTabSrHalfB & \nDevTabSrNineB & \nDevTabSrPooledB & $\nDevTabPairedB$ & \nDevTabBelowB \\
public \nModelNameC & \nLayerUpC, \nLayerDownC & standard graph & \nDevTabFxZeroC & \nDevTabFxHalfC & \nDevTabFxNineC & \nDevTabFxPooledC &  &  \\
 &  & margin graph & \nDevTabMgZeroC & \nDevTabMgHalfC & \nDevTabMgNineC & \nDevTabMgPooledC &  &  \\
 &  & softmax, original selection & \nDevTabSfZeroC & \nDevTabSfHalfC & \nDevTabSfNineC & \nDevTabSfPooledC &  &  \\
 &  & softmax, selects again & \nDevTabSrZeroC & \nDevTabSrHalfC & \nDevTabSrNineC & \nDevTabSrPooledC & $\nDevTabPairedC$ & \nDevTabBelowC \\
public \nModelNameD & \nLayerUpD, \nLayerDownD & standard graph & \nDevTabFxZeroD & \nDevTabFxHalfD & \nDevTabFxNineD & \nDevTabFxPooledD &  &  \\
 &  & margin graph & \nDevTabMgZeroD & \nDevTabMgHalfD & \nDevTabMgNineD & \nDevTabMgPooledD &  &  \\
 &  & softmax, original selection & \nDevTabSfZeroD & \nDevTabSfHalfD & \nDevTabSfNineD & \nDevTabSfPooledD &  &  \\
 &  & softmax, selects again & \nDevTabSrZeroD & \nDevTabSrHalfD & \nDevTabSrNineD & \nDevTabSrPooledD & $\nDevTabPairedD$ & \nDevTabBelowD \\
public \nModelNameE & \nLayerUpE, \nLayerDownE & standard graph & \nDevTabFxZeroE & \nDevTabFxHalfE & \nDevTabFxNineE & \nDevTabFxPooledE &  &  \\
 &  & margin graph & \nDevTabMgZeroE & \nDevTabMgHalfE & \nDevTabMgNineE & \nDevTabMgPooledE &  &  \\
 &  & softmax, original selection & \nDevTabSfZeroE & \nDevTabSfHalfE & \nDevTabSfNineE & \nDevTabSfPooledE &  &  \\
 &  & softmax, selects again & \nDevTabSrZeroE & \nDevTabSrHalfE & \nDevTabSrNineE & \nDevTabSrPooledE & $\nDevTabPairedE$ & \nDevTabBelowE \\
ours & \nLayerUpA, \nLayerDownA & standard graph & \nDevTabFxZeroF & \nDevTabFxHalfF & \nDevTabFxNineF & \nDevTabFxPooledF &  &  \\
 &  & margin graph & \nDevTabMgZeroF & \nDevTabMgHalfF & \nDevTabMgNineF & \nDevTabMgPooledF &  &  \\
 &  & softmax, original selection & \nDevTabSfZeroF & \nDevTabSfHalfF & \nDevTabSfNineF & \nDevTabSfPooledF &  &  \\
 &  & softmax, selects again & \nDevTabSrZeroF & \nDevTabSrHalfF & \nDevTabSrNineF & \nDevTabSrPooledF & $\nDevTabPairedF$ & \nDevTabBelowF \\
\bottomrule
\end{tabular}
\par\vspace{2pt}\footnotesize A sparse autoencoder (SAE) represents model activations by sparse feature coefficients. A Jacobian vector product (JVP) gives the first order activation change along one direction. Cells cover every selected feature at \nTokens held out Pile validation positions and the factors zero, \nDoseHalf and \nDoseNine. Errors for our SAEs, on \nModelNameF with width \nWidthF and TopK \nKF, average the medians of their \nRunsF runs, and so do their paired medians and shares. The standard graph and the margin graph take one JVP per selected feature at a position, which every scaling factor shares. The softmax correction reuses the derivatives from before the intervention across interventions and adds a partial forward pass from the first layer norm to the attention softmax for each intervention. The standard graph and the softmax correction with the original selection keep the selected rows of each column, the $k$ of Table~\ref{tab:releases}, and the margin graph and the correction that selects again keep every row, the width there.

\par\noindent TopK selects the largest positive encoder scores, and ReLU every positive score. The standard graph keeps the features selected before the intervention, and the margin graph selects again from the predicted scores. The relative coefficient error divides the norm of a prediction's error by the norm of the exact coefficient change, and smaller is better. The softmax correction recomputes the attention weights before the rest of the first order prediction. The share in the last column is a fraction of interventions with no unit.
\end{table}

\begin{figure}[H]
\centering
\includegraphics[width=\textwidth]{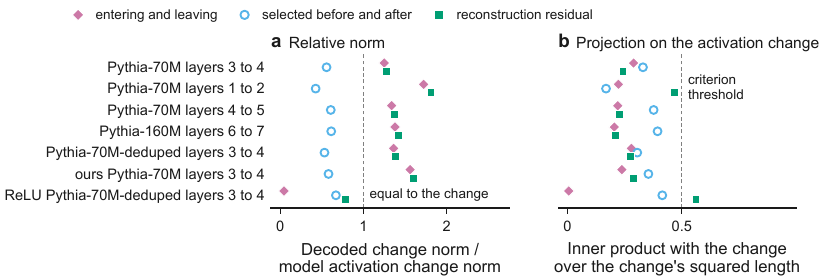}
\caption{\textbf{The entering and leaving features of a TopK sparse autoencoder (SAE) decode to more than the activation change, and those of a ReLU SAE do not.} The activation change splits into the decoded change on entering and leaving features, the decoded change on features selected before and after, and the change of the reconstruction residual, the part the SAE leaves out. Each mark is one pair's median over the interventions that change the selection. Dashed lines mark a norm equal to the change's own and the projection \nSwapShareCeiling of a criterion fixed in advance, and ours marks our SAEs.}
\label{fig:readout}
\end{figure}

\textbf{The decoded change.} Section~\ref{sec:readout} splits the activation change $\Delta x$ into three parts. They are the decoded change on features selected before and after the intervention, the decoded change on the entering and leaving features, and the change of the reconstruction residual, which sum to $\Delta x$ to within \nDecompGapMax on every intervention measured. Over the interventions that change the selection, the entering and leaving part's projection on the activation change, its inner product with the change over the change's squared norm, has a median of \nSwapShareMin to \nSwapShareMax on the public TopK pairs and \nSwapShareF on our SAEs (Figure~\ref{fig:readout}). On our SAEs the entering and leaving part has \nSwapRatioF times the activation change's norm at the median. The whole decoded change has \nDecodedRatioMin to \nDecodedRatioMax times the activation change's norm, at a median cosine with it of \nDecodedCosMin to \nDecodedCosMax and \nDecodedCosF on our SAEs.

\begin{figure}[H]
\centering
\includegraphics[width=\textwidth]{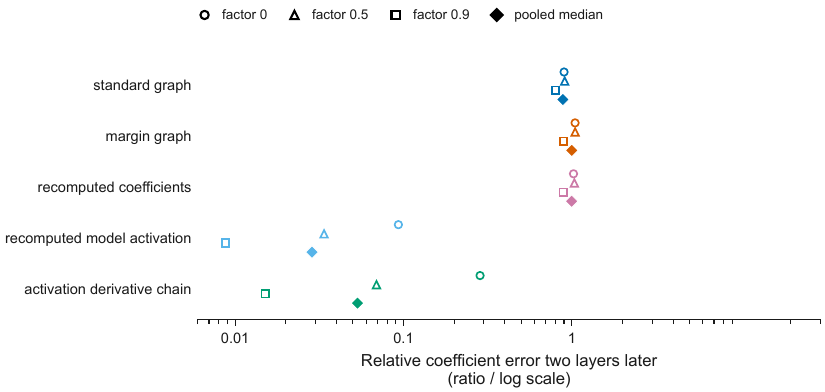}
\caption{\textbf{Recomputing the middle model activation lowers the error two layers downstream, while decoded middle coefficients do not.} Each row gives one predictor's median relative coefficient error two layers downstream, at each scaling factor on one upstream feature's coefficient and pooled. Marks take the hue of their row's predictor, and the legend gives only each factor's shape.}
\label{fig:exhibit-chains}
\end{figure}

\textbf{Chains across two blocks.}\label{par:chains} A chain predicts the coefficients two layers downstream by passing a prediction through both blocks in turn, and the question is whether the prediction should pass between the blocks as decoded coefficients or as a model activation. The standard graph's chain decodes its predicted middle coefficient change and keeps the features selected before the intervention again at the third layer. The margin graph's chain decodes its predicted middle coefficient change and selects again at the third layer. The chain with recomputed coefficients substitutes the exact middle coefficient change before selecting again. The chain with the recomputed model activation instead applies the second block's Jacobian $J_2$, taken before the intervention, to the exact middle activation change, projects with $E_3$ and selects again. The activation derivative chain applies $J_2J_1$ to the intervention before projecting and selecting again, and every chain's error divides by the norm of the exact change of the third layer's coefficients. In Figure~\ref{fig:exhibit-chains}, the first three rows decode a predicted or exact middle coefficient change, and the last two pass on a model activation change, recomputed or predicted through both blocks' derivatives.

On \nModelNameA at layers \nLayerUpA, \nLayerDownA and \nLayerThird, over \nRowsTwoHop interventions at each scaling factor, a chain that decodes its middle coefficient change errs at the third layer's coefficients by \nChainFixed with the standard graph's middle coefficients, by \nChainMargin with the margin graph's and by \nChainOracle with recomputed middle coefficients (Figure~\ref{fig:exhibit-chains}). A chain can instead stay in the residual stream. The activation derivative chain errs by \nChainResidual, and the chain with the recomputed model activation errs by \nChainExactInput. A decoded middle change drops the change of the middle reconstruction residual, which the third layer receives by Theorem~\ref{thm:readout}. A chain across blocks should therefore stay in the residual stream and decode at its last layer.

Theorem~\ref{thm:readout} bounds the entering and leaving part from below when one feature enters and another leaves, by the cutoff times the separation of their decoder columns less three largest shifts for each such pair. On all \nOvershootCrossingWrites single feature interventions that change the selection, at every position and scaling factor of the \nDiscussionPublicPairsWords public TopK pairs and our SAEs, the measured part is at or above the bound. On the interventions where one feature enters and one leaves, the bound is positive on \nOvershootPartPositivePublic of the public pairs' interventions and \nOvershootPartPositiveF of our SAEs'. Where it is positive, its median ratio to the activation change is \nOvershootBoundMedianPublic and \nOvershootBoundMedianF, against a measured median of \nOvershootMeasuredMedianPublic and \nOvershootMeasuredMedianF.

Over every intervention, the ReLU pair's decoded change has a median projection share of \nDecodedShareR on the activation change, a median norm ratio of \nDecodedRatioR and a median cosine of \nDecodedCosR. A prespecified criterion counts a decoded change as following the activation change where its share is at least \nReluShareFloor and its norm ratio lies between \nReluRatioLow and \nReluRatioHigh. The ReLU pair's share is under that floor, so the criterion was met (Table~\ref{tab:rules}).

\section{Fitting the maps}
\label{app:anatomy}

This appendix explains how we fit the sparse maps of Section~\ref{sec:transport} and what the fitted maps look like. The maps write each feature vector of a source run, a decoder column or an encoder row, as a sparse nonnegative combination of the target run's vectors. The greedy search below chooses those target vectors, and its stopping rule fixes how many each source vector may use.

\begin{table}[H]
\caption{\textbf{Masking loses much of what transfers across seeds.} Cells give the graph reconstruction score, the share of a reference graph's squared norm that a transferred graph reproduces, between seeds and on the data order pair. The first two columns take the whole source graph as reference, and the last three its part on the source's selected features.}
\label{tab:transport}
\centering
\footnotesize
\setlength{\tabcolsep}{4pt}
\begin{tabular}{lrrrrr}
\toprule
pair & \begin{tabular}[c]{@{}c@{}}matching\\whole graph\end{tabular} & \begin{tabular}[c]{@{}c@{}}sparse maps\\whole graph\end{tabular} & \begin{tabular}[c]{@{}c@{}}sparse maps\\target unmasked\end{tabular} & \begin{tabular}[c]{@{}c@{}}sparse maps\\target masked\end{tabular} & drop from masking \\
\midrule
between seeds & \nOneToOne & \nSparseUnmasked & \nRestricted & \nMasked & \nGap \\
\quad low end & \nIvLowMatchUnmasked & \nIvLowSparseUnmasked & \nIvLowWholeTarget & \nIvLowSparseMasked & \nIvLowGap \\
\quad high end & \nIvHighMatchUnmasked & \nIvHighSparseUnmasked & \nIvHighWholeTarget & \nIvHighSparseMasked & \nIvHighGap \\
data order pair & \nOneToOneRetrain & \nSparseRetrain & \nRestrictedRetrain & \nMaskedRetrain & \nGapRetrain \\
\quad low end & \nIvLowMatchUnmaskedRetrain & \nIvLowSparseUnmaskedRetrain & \nIvLowWholeTargetRetrain & \nIvLowSparseMaskedRetrain & \nIvLowGapRetrain \\
\quad high end & \nIvHighMatchUnmaskedRetrain & \nIvHighSparseUnmaskedRetrain & \nIvHighWholeTargetRetrain & \nIvHighSparseMaskedRetrain & \nIvHighGapRetrain \\
\bottomrule
\end{tabular}

\par\noindent A sparse autoencoder (SAE) represents model activations by sparse feature coefficients. Scores average \nPairsBetween ordered pairs of runs of our SAEs on \nModelNameF at layers \nLayerUpA and \nLayerDownA, width \nWidthF and TopK \nKF, or both directions of the data order pair, at \nHeldout held out Pile validation positions. Intervals give the \nIvPctLow and \nIvPctHigh percentiles of \nIvDraws resamples of positions, leaving one seed run out between seeds and none for the data order pair. The graph reconstruction score is one minus the summed squared prediction error over the summed squared reference entries, pooled over positions before averaging pairs. Between seeds compares runs with different initialization seeds, and the data order pair shares an initialization and sees the training examples in different orders. A mask keeps the selected features and sets the other coordinates to zero. Matching pairs single features, while sparse maps combine target features to represent each source feature. Target unmasked keeps every target feature, target masked keeps only the target's selected features, and the drop from masking is the unmasked score less the masked score. Every value has no unit.
\end{table}

We fit each sparse map one source vector at a time by a greedy search, which adds target vectors until the source vector's fit residual meets a tolerance or its target vectors reach the fit limit. Each source vector and each target vector is set to unit norm, so the tolerance is a relative fit residual and a coefficient is measured against a unit target vector. The tolerance of \nTolerance is the residual that a unit vector leaves on one unit target vector at a cosine of \nToleranceCosine. At each step the search takes the target vector with the largest positive inner product with the current residual and refits the coefficients on every chosen target vector by nonnegative least squares. The refit is an accelerated proximal gradient in double precision. Its step is the inverse of \nNnlsStepMultiple times the exact largest eigenvalue of the chosen vectors' Gram matrix, and its momentum follows $t' = (1 + \sqrt{1 + 4t^2})/2$ and $y = x' + ((t - 1)/t')(x' - x)$, restarted at $y = x'$ and $t = 1$ whenever the objective rises.

Each step's refit starts from the previous step's coefficients, with the new target vector at its inner product with the residual floored at zero. The refit checks the optimality violation every \nNnlsCheckEvery iterations and stops once it drops to \nNnlsStop or at \nNnlsIters iterations, where the violation is $\max_i \nu_i / \|b\|$. Here $G$ is the chosen vectors' Gram matrix and $b$ their inner products with the source vector. The entry $\nu_i$ is $|(Gc - b)_i|$ where $c_i > 0$ and $\max(b_i - (Gc)_i, 0)$ where $c_i = 0$. A source vector stops at the first step that meets the tolerance, at the fit limit of \nCap target vectors, or sooner where no target vector lowers its residual. A source vector that never meets the tolerance keeps its closest fit. A coefficient above \nNonzero counts as a target vector, and the count deletes no coefficient.

The fit limit was evaluated at one, two, four, eight, sixteen and thirty-two target vectors on the measurement sample, where the mean graph reconstruction score of the unmasked transfer between seeds is \nLevelOne, \nLevelTwo, \nLevelFour, \nLevelEight, \nLevelSixteen and \nLevelThirtyTwo. The rule fixed in advance chose the smallest limit within \nLevelWithin of the best, which is \nLevelChosen. At a limit of one, each source vector keeps the target vector of largest positive inner product, with no constraint that two source vectors take different target vectors. That limit gives the unconstrained nearest vector map, which reaches \nLevelOne against \nOneToOneExact for the one to one matching.

The one to one matching assigns each source vector to at most one target vector so as to maximize the summed positive cosine, with a negative cosine counting as zero. Each matched vector takes its target vector at the nonnegative projection, and an unmatched vector takes none. We convert the fitted coefficients back to each SAE's rows in the units of Section~\ref{sec:setup}. The decoder columns have unit norm in those units, so $M$ is the fitted coefficients themselves. A source encoder row is rebuilt as its own norm times the combination of unit target rows, so $N_{jl} = C_{jl}\,\|e^A_j\| / \|e^B_l\|$, with $C$ the coefficients fitted on unit rows.

An ordered pair of distinct runs with different initialization seeds, source $a$ and target $b$, takes the index $5a + b$, less one where $b$ exceeds $a$. The random maps of matched sparsity give each source vector the count of target vectors its fitted map uses, on that many target vectors drawn uniformly without replacement, with the coefficients refitted by the same nonnegative least squares, at \nRandomDrawsPerPair draws per pair from seed \nMapSeedRandom plus the pair's index. The permutation null relabels the target's vectors uniformly at random under the fitted maps, independently at each layer, keeping the maps' sparsity and coefficients, at \nNullDrawsPerPair draws per pair from seed \nMapSeedNull plus the pair's index. The rotation control rotates the target's vectors by a Haar random orthogonal matrix of the residual stream's coordinates, which keeps every internal angle of the target and breaks every correspondence with the source. It compares each run $a$ with an initialization seed with run $a + 1$ modulo six rotated, from seed \nRotationSeedDec plus $a$ at the upstream layer and \nRotationSeedEnc plus $a$ at the downstream one.

\begin{figure}[H]
\centering
\includegraphics[width=\textwidth]{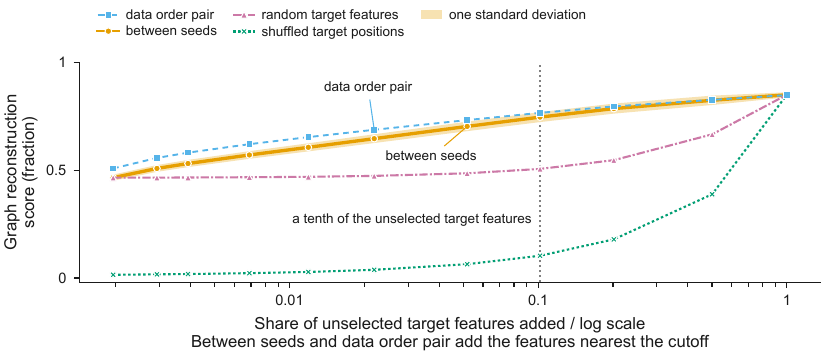}
\caption{\textbf{Adding back the unselected target features nearest the cutoff recovers most of what masking loses.} Curves plot the graph reconstruction score on the source's selected features against the share of unselected target features added to the target's mask. Nearest selection cutoff adds the features nearest the cutoff between seeds, and the data order pair's curve adds them on that pair. Random target features adds them at random, and shuffled target positions takes the target's selection at a permuted position. The dotted line marks a share of \nBandSmall, and shading gives one standard deviation over pairs.}
\label{fig:widening}
\end{figure}

The shuffled target positions of Figure~\ref{fig:widening} permute the target's positions from seed \nPositionSeed, so the target's selected sets and margins are taken at a permuted position. The recovery curve keeps the target's selected features in its mask and adds, at a share of the unselected features, the floor of that share times the count of its unselected features nearest the cutoff, per position and on each side. The random control adds the same count of unselected features drawn uniformly without replacement, at \nControlDraws draws per pair from seed \nControlSeedBase plus the pair's index, on the same \nBandTokens positions.

Equation~\ref{eq:transport-score} defines the graph reconstruction score. The scores \nRestricted and \nMasked of Section~\ref{sec:transport} follow this order of pooling, with $Q$ the ordered pairs of seed runs and the reference the source's selected submatrix. The data order pair averages its two directions by the same rule.

Table~\ref{tab:transport-runs} gives every score of the transfer run by run, each run with a different initialization seed as the source of \nTrPairsPerRun ordered pairs. The interval of each central score takes \nIvDraws draws from seed \nIvSeed. Draw $b$ leaves out run $b$ modulo six, keeps the ordered pairs among the other five runs and resamples the \nHeldout positions with replacement once for every pair, and the data order pair's value takes the same drawn positions with no run left out. The ends are the \nIvPctLow and \nIvPctHigh percentiles of the draws.

\begin{table}[H]
\caption{\textbf{The drop from masking appears for every source run.} Cells give graph reconstruction scores for each source run of our sparse autoencoders (SAEs), averaged over its target runs, and for each direction of the data order pair. The masked rows and the target unmasked row take the source graph on the source's selected features as reference.}
\label{tab:transport-runs}
\centering
\scriptsize
\setlength{\tabcolsep}{2.5pt}
\begin{tabular}{lrrrrrrrr}
\toprule
score & run \nRunIndexZero & run \nRunIndexOne & run \nRunIndexTwo & run \nRunIndexThree & run \nRunIndexFour & run \nRunIndexFive & \begin{tabular}[c]{@{}c@{}}data order,\\\nRunIndexSix from \nRunIndexSeven\end{tabular} & \begin{tabular}[c]{@{}c@{}}data order,\\\nRunIndexSeven from \nRunIndexSix\end{tabular} \\
\midrule
matching, unmasked & \nTrZeroMatchUnmasked & \nTrOneMatchUnmasked & \nTrTwoMatchUnmasked & \nTrThreeMatchUnmasked & \nTrFourMatchUnmasked & \nTrFiveMatchUnmasked & \nTrRetrainSixSevenMatchUnmasked & \nTrRetrainSevenSixMatchUnmasked \\
sparse, unmasked & \nTrZeroSparseUnmasked & \nTrOneSparseUnmasked & \nTrTwoSparseUnmasked & \nTrThreeSparseUnmasked & \nTrFourSparseUnmasked & \nTrFiveSparseUnmasked & \nTrRetrainSixSevenSparseUnmasked & \nTrRetrainSevenSixSparseUnmasked \\
matching, masked & \nTrZeroMatchMasked & \nTrOneMatchMasked & \nTrTwoMatchMasked & \nTrThreeMatchMasked & \nTrFourMatchMasked & \nTrFiveMatchMasked & \nTrRetrainSixSevenMatchMasked & \nTrRetrainSevenSixMatchMasked \\
sparse, masked & \nTrZeroSparseMasked & \nTrOneSparseMasked & \nTrTwoSparseMasked & \nTrThreeSparseMasked & \nTrFourSparseMasked & \nTrFiveSparseMasked & \nTrRetrainSixSevenSparseMasked & \nTrRetrainSevenSixSparseMasked \\
sparse, target unmasked & \nTrZeroWholeTarget & \nTrOneWholeTarget & \nTrTwoWholeTarget & \nTrThreeWholeTarget & \nTrFourWholeTarget & \nTrFiveWholeTarget & \nTrRetrainSixSevenWholeTarget & \nTrRetrainSevenSixWholeTarget \\
drop from masking & \nTrZeroGap & \nTrOneGap & \nTrTwoGap & \nTrThreeGap & \nTrFourGap & \nTrFiveGap & \nTrRetrainSixSevenGap & \nTrRetrainSevenSixGap \\
mask term & \nTrZeroMaskTerm & \nTrOneMaskTerm & \nTrTwoMaskTerm & \nTrThreeMaskTerm & \nTrFourMaskTerm & \nTrFiveMaskTerm & \nTrRetrainSixSevenMaskTerm & \nTrRetrainSevenSixMaskTerm \\
cross term & \nTrZeroCrossTerm & \nTrOneCrossTerm & \nTrTwoCrossTerm & \nTrThreeCrossTerm & \nTrFourCrossTerm & \nTrFiveCrossTerm & \nTrRetrainSixSevenCrossTerm & \nTrRetrainSevenSixCrossTerm \\
unmasked error & \nTrZeroResid & \nTrOneResid & \nTrTwoResid & \nTrThreeResid & \nTrFourResid & \nTrFiveResid & \nTrRetrainSixSevenResid & \nTrRetrainSevenSixResid \\
\bottomrule
\end{tabular}

\par\noindent The runs are on \nModelNameF at layers \nLayerUpA and \nLayerDownA, width \nWidthF and TopK \nKF, and each source run's column averages its \nTrPairsPerRun ordered target pairs over \nHeldout held out Pile validation positions. The last three rows divide by the reference graph's squared norm, and the signed cross term is twice the mask term's inner product with the unmasked error. A sparse autoencoder (SAE) represents model activations by sparse feature coefficients. The graph reconstruction score is one minus the summed squared prediction error over the summed squared reference entries, pooled over positions before averaging pairs. Between seeds compares runs with different initialization seeds, and the data order pair shares an initialization and sees the training examples in different orders. A mask keeps the selected features and sets the other coordinates to zero. Matching pairs single features, and sparse maps use nonnegative combinations of target features. The mask term is the part removed when the target keeps only its selected features, and the unmasked error is the reference graph less the prediction with the target unmasked. The drop from masking is the unmasked score less the masked score. Every value has no unit.
\end{table}

The drop from masking splits exactly into two terms, each divided by the reference's squared norm (Table~\ref{tab:transport-full}). Masking removes part of the target's contribution, which we call the mask term. The first term is the mask term's own squared norm, \nTermMask between seeds. The second is twice the mask term's inner product with the unmasked error, the reference graph less the prediction with the target unmasked, and it is \nTermCross. The unmasked error's own squared norm, \nTermResid, enters both scores and cancels in the drop.

\begin{table}[H]
\caption{\textbf{The drop from masking is the mask term's squared norm plus a signed cross term.} Cells give graph reconstruction scores between seeds of our sparse autoencoders (SAEs) and on the data order pair. The masked and target unmasked columns take the source graph on its selected features as reference, and the last four columns divide by its squared norm.}
\label{tab:transport-full}
\centering
\scriptsize
\setlength{\tabcolsep}{2.5pt}
\begin{tabular}{lrrrrrrrrr}
\toprule
pair & \begin{tabular}[c]{@{}c@{}}matching\\unmasked\end{tabular} & \begin{tabular}[c]{@{}c@{}}sparse\\unmasked\end{tabular} & \begin{tabular}[c]{@{}c@{}}matching\\masked\end{tabular} & \begin{tabular}[c]{@{}c@{}}sparse\\masked\end{tabular} & \begin{tabular}[c]{@{}c@{}}sparse, target\\unmasked\end{tabular} & \begin{tabular}[c]{@{}c@{}}drop\end{tabular} & \begin{tabular}[c]{@{}c@{}}mask\\term\end{tabular} & \begin{tabular}[c]{@{}c@{}}cross\\term\end{tabular} & \begin{tabular}[c]{@{}c@{}}unmasked\\error\end{tabular} \\
\midrule
between seeds & \nOneToOne & \nSparseUnmasked & \nOneToOneMasked & \nMasked & \nRestricted & \nGap & \nTermMask & \nTermCross & \nTermResid \\
\quad low end & \nIvLowMatchUnmasked & \nIvLowSparseUnmasked & \nIvLowMatchMasked & \nIvLowSparseMasked & \nIvLowWholeTarget & \nIvLowGap & \nIvLowMaskTerm & \nIvLowCrossTerm & \nIvLowResid \\
\quad high end & \nIvHighMatchUnmasked & \nIvHighSparseUnmasked & \nIvHighMatchMasked & \nIvHighSparseMasked & \nIvHighWholeTarget & \nIvHighGap & \nIvHighMaskTerm & \nIvHighCrossTerm & \nIvHighResid \\
data order pair & \nOneToOneRetrain & \nSparseRetrain & \nOneToOneMaskedRetrain & \nMaskedRetrain & \nRestrictedRetrain & \nGapRetrain & \nTermMaskRetrain & \nTermCrossRetrain & \nTermResidRetrain \\
\quad low end & \nIvLowMatchUnmaskedRetrain & \nIvLowSparseUnmaskedRetrain & \nIvLowMatchMaskedRetrain & \nIvLowSparseMaskedRetrain & \nIvLowWholeTargetRetrain & \nIvLowGapRetrain & \nIvLowMaskTermRetrain & \nIvLowCrossTermRetrain & \nIvLowResidRetrain \\
\quad high end & \nIvHighMatchUnmaskedRetrain & \nIvHighSparseUnmaskedRetrain & \nIvHighMatchMaskedRetrain & \nIvHighSparseMaskedRetrain & \nIvHighWholeTargetRetrain & \nIvHighGapRetrain & \nIvHighMaskTermRetrain & \nIvHighCrossTermRetrain & \nIvHighResidRetrain \\
\bottomrule
\end{tabular}

\par\noindent Scores average \nPairsBetween ordered pairs of seed runs on \nModelNameF at layers \nLayerUpA and \nLayerDownA, width \nWidthF and TopK \nKF, or the data order pair's two directions, at \nHeldout held out Pile validation positions. Intervals use \nIvDraws resamples of positions at the \nIvPctLow and \nIvPctHigh percentiles, leaving out one seed run between seeds. A sparse autoencoder (SAE) represents model activations by sparse feature coefficients. The graph reconstruction score is one minus the summed squared prediction error over the summed squared reference entries, pooled over positions before averaging pairs. Between seeds compares runs with different initialization seeds, and the data order pair shares an initialization and sees the training examples in different orders. A mask keeps the selected features and sets the other coordinates to zero. Matching pairs single features, and sparse maps use nonnegative combinations. The mask term is the target contribution that masking removes, and the unmasked error is the prediction error with the target unmasked. The mask term and unmasked error columns give their squared norms over the reference's squared norm, and the cross term is twice their inner product over that norm. The drop from masking is the unmasked score less the masked score, and every value has no unit.
\end{table}

\begin{figure}[H]
\centering
\includegraphics[width=\textwidth]{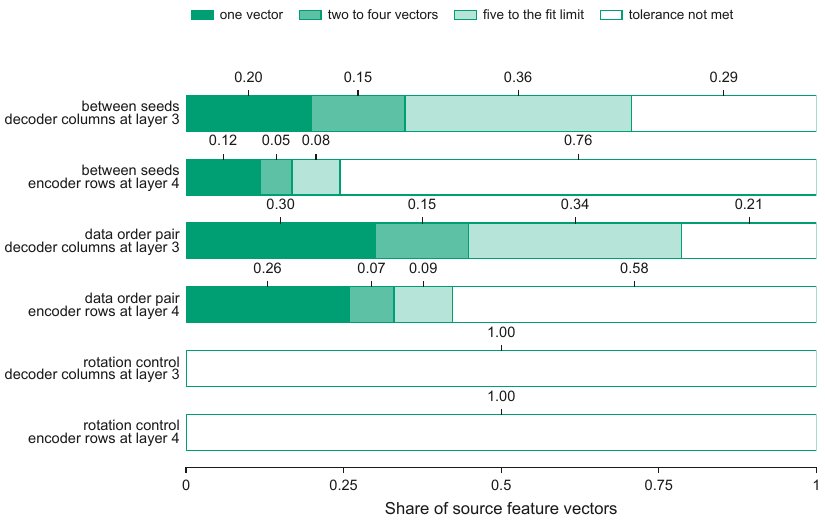}
\caption{\textbf{Between seeds, sparse maps mostly recombine the target vectors they reach, while the data order pair's encoder rows that the fit reaches mostly copy one.} Each bar sorts source feature vectors by how many target vectors the fit used once its relative residual fell to \nTolerance, within the fit limit \nCap. The rotation control rotates the target at random.}
\label{fig:anatomy}
\end{figure}

The number of target vectors at the first step that meets the tolerance sorts a source vector into a class, the column heads of Figure~\ref{fig:anatomy}. One vector means a copy of a single target vector, and the other classes are two to four vectors and five vectors up to the fit limit. Tolerance not met means the vector never meets the tolerance within the fit limit. The first step at which the greedy path meets the tolerance bounds the smallest representation from above, and the class counts that step's coefficients above \nNonzero, which can be fewer. Near a coherent target the class also depends on the tolerance. A target vector equal to the normalized sum of two others is represented exactly by those two and closely by a third vector near their sum, which is the failure of the exact recovery condition of \citet{tropp2004greed}.

Between seeds the upstream decoder columns divide into one vector at \nAnatBetweenDecOne, two to four vectors at \nAnatBetweenDecFew, five to the fit limit at \nAnatBetweenDecMany and tolerance not met at \nAnatBetweenDecNone (Figure~\ref{fig:anatomy}). The downstream encoder rows divide into one vector at \nAnatBetweenEncOne, two to four at \nAnatBetweenEncFew, five to the fit limit at \nAnatBetweenEncMany and tolerance not met at \nAnatBetweenEncNone. On the data order pair the decoder columns take one vector in \nAnatRetrainDecOne of cases and miss the tolerance in \nAnatRetrainDecNone, and the encoder rows in \nAnatRetrainEncOne and \nAnatRetrainEncNone. Over \nControlPairs pairs, a control that rotates the target's vectors represents \nControlRepresented of its source's decoder columns, misses the tolerance on \nAnatControlDecNone of them and reaches a graph reconstruction score of \nControlExplained on the unmasked graph.

The maps recombine target vectors more often than they copy them. They represent \nRepresentedDec of the upstream decoder columns at a median of \nMedianAtomsDec target vectors each and \nRepresentedEnc of the downstream encoder rows at a median of \nMedianAtomsEnc. Feature by feature, a selected downstream feature's decoder partner in the other seed is that seed's decoder column of largest cosine, counted only where the cosine reaches \nPartnerCosine. Over every selected feature with a partner at every position, with each pair taken in both directions, the partner is selected at the same position in \nPartnerActiveDec of cases, against \nPartnerActiveDecNull with shuffled target positions.

Written over our SAEs' vectors, a narrower SAE of width \nWidthFiled trained on the same stream takes two or more target vectors for a mean share of \nFFourAcrossMean of its represented columns over its six pairs, which range from \nFFourAcrossMin to \nFFourAcrossMax. Between two seeds of our SAEs at the same budget the mean share is \nFFourBetweenMean, ranging from \nFFourBetweenMin to \nFFourBetweenMax. The criterion fixed in advance compared the mean across widths with the largest share between seeds, \nMultiAtomBetweenMax. The mean across widths is also under the mean between seeds.

\section{Overlap and training budget}
\label{app:budget}

How often the features of two seeds match depends on how long the SAEs trained, so a match rate measures the training budget as well as the model. The nearest neighbour overlap of two SAEs at a cosine of \nOverlapThreshold is the share of one SAE's decoder columns, among the features selected at least once in training, whose largest absolute cosine with such a column of the other exceeds \nOverlapThreshold. Several columns may share one partner. We measure each pair from each of its two SAEs in turn, and the curve averages the two directions.

\begin{figure}[H]
\centering
\includegraphics[width=\textwidth]{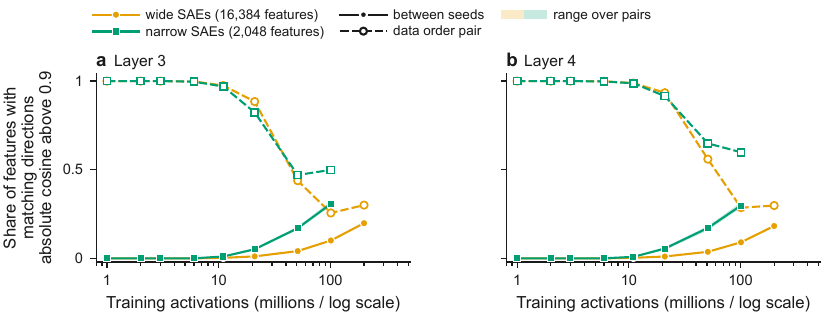}
\caption{\textbf{How often the features of two runs match depends on how long they trained.} Curves plot the share of matching features against training activations, averaged over both directions. Wide sparse autoencoders (SAEs) have \nWidthF features and narrow SAEs \nWidthFiled. Between seeds compares runs from different initialization seeds, with shading for the range over pairs.}
\label{fig:budget}
\end{figure}

Between seeds of our SAEs, averaged over \nBudgetPairs pairs of runs, the overlap rises with the training budget, to \nOverlapTestbedEnd at \nBudgetEnd activations at layer \nLayerUpA, while on the data order pair, which starts from one initialization, it falls to \nWithinTestbedEnd (Figure~\ref{fig:budget}). A narrower configuration at width \nWidthFiled with \nKFiled features selected reaches \nOverlapFiledEnd between seeds at \nBudgetFiledEnd activations. SAEs of the two configurations fitted over many passes to a small sample of \nDistinctActivationsBlock distinct activations reach \nOverlapTestbedConverged and \nOverlapFiledConverged between seeds. We therefore fit the maps of Section~\ref{sec:transport} at the full budget of our SAEs.

A checkpoint target is recorded at the first chunk boundary at or past it, and a chunk trains \nCfChunkTrained activations, its \nCfChunk cut to whole batches. The \nCfTargetsFirstChunk targets below one chunk are therefore recorded together at the first boundary, each larger target at the first boundary past it, and the last at the end of training. The \nCfCheckpointCount targets thus give the \nBudgetCount budgets that the curve plots.

\section{Criteria fixed in advance}
\label{app:declarations}

We fixed each criterion of Table~\ref{tab:rules}, with the quantity it measures and the threshold at which it is met, before the measurement it governs, and the table gives each one beside its result. Theorem~\ref{thm:band} decides a selection state only where its bound on the Jacobian is met along the whole segment, which no measurement can confirm. So we accept a predicted selection state only where it lies outside the error band under a constant sampled at three points and the intervention is smaller than the size at which the median first order error reaches a threshold fixed in advance. We call this check the acceptance rule, and it adds no new guarantee.

\begingroup
\scriptsize
\setlength{\tabcolsep}{3pt}
\begin{longtable}{lp{\declA}p{\declB}p{\declC}l}
\caption{\normalsize{}\textbf{The criteria separate the predictions the data support from their limits.} Each row gives the quantity a criterion measures, the threshold at which it is met, the result and whether it was met. The ReLU contrast meets \nRulesFiredRelu criteria and the TopK sparse autoencoders (SAEs) meet \nRulesFiredOther.}
\label{tab:rules}\\
\toprule
criterion & quantity measured & met when & result & met \\
\midrule
\endfirsthead
\toprule
criterion & quantity measured & met when & result & met \\
\midrule
\endhead
$\mathrm{G1}$ & share of zero ablations that change the selection, each public TopK pair & under \nCrossingFloor & \nCrossZeroMin to \nCrossZeroMax & no \\
$\mathrm{G2}$ & share of squared norm on the entering and leaving coordinates, each public TopK pair & under \nEnergyFloor & \nEnergyZeroMin to \nEnergyZeroMax & no \\
$\mathrm{G2}$ & the same on the ReLU pair & under \nEnergyFloor & \nEnergyZeroR & yes \\
$\mathrm{C2}$ & the ReLU pair's share of squared norm & over \nReluEnergyBound & \nEnergyZeroR & no \\
$\mathrm{G3}$ & the margin graph's error, each public TopK pair & not under \nDesignFactor of the standard graph's, which is \nDesignBoundA, \nDesignBoundB, \nDesignBoundC, \nDesignBoundD and \nDesignBoundE on the pairs in the order of Table~\ref{tab:releases} & \nMarginErrMin to \nMarginErrMax & no \\
$\mathrm{F5}$ & the margin graph's error, our SAEs & not under \nDesignFactor of the standard graph's, which is \nDesignBoundF & \nMarginErrF & no \\
$\mathrm{F6}$ & runs whose regressor on six inputs, the encoder score, its predicted change and the two order statistics at the cutoff before and after the change, has a median error inside the margin graph's interquartile range & \nRegressorRunsToFire runs or more & \nRegressorRunsWithin runs & no \\
$\mathrm{F7}$ & share of squared norm on the entering and leaving coordinates, our SAEs & under \nEnergyFloor & \nEnergyZeroF & no \\
$\mathrm{F8}$ & the margin graph's precision and recall over those of the margin alone and of the predicted change alone, our SAEs & a lead under \nSupportMargin & precision \nPrecF and recall \nRecF against \nPrecMarginAloneF and \nPrecEdgeAloneF for both & no \\
$\mathrm{B1}$ & selection states outside the error band under the run's largest constant that change & any & \nCertFlips & no \\
acceptance rule & accepted selection states that the exact encoder scores contradict, on the scoring half of every population & any & \nRuleWrong & no \\
$\mathrm{M1}$ & the margin graph on subsets against the standard graph & not under \nDesignFactor of the standard graph's error at a size & not under at the largest sizes & yes \\
$\mathrm{P1}$ & integrated gradients that select again, at \nPathProducts JVPs, on subsets & not under \nRemedyFactor of the standard graph's error at a size & at most \nPathFiveRerankMaxF on our SAEs and \nPathFiveRerankMaxRel on the public \nModelNameA pair & no \\
$\mathrm{P2}$ & the ratio of the two norm boundaries at the error threshold \nRegimeLevel & over \nBoundaryFactor & \nBoundaryRatio & no \\
$\mathrm{R1}$ & the entering and leaving part, each TopK pair and our SAEs & norm ratio under \nSwapRatioFloor or share over \nSwapShareCeiling & \nSwapRatioMin to \nSwapRatioMax and \nSwapShareMin to \nSwapShareMax & no \\
$\mathrm{R2}$ & the ReLU pair's decoded change over every intervention & share under \nReluShareFloor or ratio outside \nReluRatioLow to \nReluRatioHigh & \nDecodedShareR and \nDecodedRatioR & yes \\
$\mathrm{H1}$ & the margin graph's decoded chain against the standard graph's & not under \nDesignFactor of the standard graph's chain's error & \nChainMargin against \nChainFixed & yes \\
$\mathrm{H2}$ & the activation derivative chain & over \nChainBoundFixed, which is \nDesignFactor of the standard graph's chain's error, or over \nChainBoundOracle, which is \nChainOracleMultiple times the error of the chain with the recomputed model activation & \nChainResidual & no \\
$\mathrm{H3}$ & the projection of the decoded change on middle features beyond one ablation's first order change of the cutoff, above the part within it & a difference under \nMediationMargin & \nMediationDiff & yes \\
$\mathrm{F1}$ & the sparse maps' gain over one to one matching, unmasked & under \nGainBound, a quarter of what the matching leaves & \nSparseGain & no \\
$\mathrm{F2}$ & pairs whose maps beat every one of \nRandomDrawsPerPair random draws and \nNullDrawsPerPair permutation draws & any pair short & \nPairsBeatingNull of \nPairsBetween & no \\
$\mathrm{F3}$ & maps of at most \nFThreeBound target vectors at the median that represent half the vectors, the decoder columns at layer \nLayerUpA and the encoder rows at layer \nLayerDownA & a median count over \nFThreeBound or a represented share under \nFThreeFloor on either family & a median of \nMedianAtomsDec target vectors on the decoder columns and a represented share of \nRepresentedEnc on the encoder rows & yes \\
$\mathrm{F4}$ & the mean share of columns that use two or more target vectors over the six pairs from the narrower configuration into our SAEs, against the largest such share among the pairs between seeds of our SAEs & at or under & \nMultiAtomAcross against \nMultiAtomBetweenMax & yes \\
$\mathrm{C1}$ & the rotation control's represented share of the decoder columns at layer \nLayerUpA, and its unmasked graph reconstruction score & at or over half the represented share between seeds, or at or over the one to one matching's score & \nControlRepresented and \nControlExplained & no \\
$\mathrm{K1}$ & the recovery curve at a share of \nBandSmall of the unselected target features & under \nBandNeeded of the drop from masking & \nRecoveredTenth & no \\
\bottomrule
\end{longtable}

\par\noindent A met entry answers the adverse condition of its row, so yes is not a general claim of success. The labels in the first column are row identifiers with no numerical meaning, and the preceding sections define each quantity. A sparse autoencoder (SAE) represents model activations by sparse feature coefficients. TopK selects the largest positive encoder scores, and ReLU, the rectified linear unit, every positive score. The standard graph keeps the features selected before the intervention, and the margin graph selects again from the predicted scores. Errors are relative coefficient errors, the norm of the prediction's error over the norm of the exact coefficient change. Integrated gradients average derivatives along the intervention's path, and a Jacobian vector product (JVP) gives the first order activation change in one direction. An accepted selection state lies outside the error band under a sampled constant, for an intervention whose norm lies under the \emph{norm boundary}, the calibrated norm at which the median first order error reaches its threshold. A chain predicts through successive model blocks. Shares, errors and norm ratios have no unit, and counts name their unit in the cells.
\endgroup

On the ReLU pair \nRulesFiredRelu criteria were met, testing the contrast both ways. The other \nRulesFiredOther that were met test the margin graph on the largest subsets, the decoded chains, the middle features through which a path runs across two blocks, and the counts of target vectors in the maps. Section~\ref{sec:margin} states the subset result and Appendix~\ref{app:releases} the chain result. This appendix gives the result on middle features below, and Appendix~\ref{app:anatomy} gives the counts.

Table~\ref{tab:nulls} gives the fitted scores and the largest random and permutation scores for every pair behind the criterion on null maps.

\begin{table}[H]
\caption{\textbf{Every fitted map beats the largest of both null maps.} Cells give the whole graph reconstruction score of each ordered pair's fitted map beside the largest score of its random maps and of its permutations, so larger is better. The rows are the comparisons behind the criterion on null maps in Table~\ref{tab:rules}.}
\label{tab:nulls}
\centering
\small
\begin{tabular}{rrrrr}
\toprule
source run & target run & fitted score & largest random & largest permutation \\
\midrule
\nExtraMapZeroSourceRun & \nExtraMapZeroTargetRun & \nExtraMapZeroFitted & \nExtraMapZeroRandomMax & \nExtraMapZeroNullMax \\
\nExtraMapOneSourceRun & \nExtraMapOneTargetRun & \nExtraMapOneFitted & \nExtraMapOneRandomMax & \nExtraMapOneNullMax \\
\nExtraMapTwoSourceRun & \nExtraMapTwoTargetRun & \nExtraMapTwoFitted & \nExtraMapTwoRandomMax & \nExtraMapTwoNullMax \\
\nExtraMapThreeSourceRun & \nExtraMapThreeTargetRun & \nExtraMapThreeFitted & \nExtraMapThreeRandomMax & \nExtraMapThreeNullMax \\
\nExtraMapFourSourceRun & \nExtraMapFourTargetRun & \nExtraMapFourFitted & \nExtraMapFourRandomMax & \nExtraMapFourNullMax \\
\nExtraMapFiveSourceRun & \nExtraMapFiveTargetRun & \nExtraMapFiveFitted & \nExtraMapFiveRandomMax & \nExtraMapFiveNullMax \\
\nExtraMapSixSourceRun & \nExtraMapSixTargetRun & \nExtraMapSixFitted & \nExtraMapSixRandomMax & \nExtraMapSixNullMax \\
\nExtraMapSevenSourceRun & \nExtraMapSevenTargetRun & \nExtraMapSevenFitted & \nExtraMapSevenRandomMax & \nExtraMapSevenNullMax \\
\nExtraMapEightSourceRun & \nExtraMapEightTargetRun & \nExtraMapEightFitted & \nExtraMapEightRandomMax & \nExtraMapEightNullMax \\
\nExtraMapNineSourceRun & \nExtraMapNineTargetRun & \nExtraMapNineFitted & \nExtraMapNineRandomMax & \nExtraMapNineNullMax \\
\nExtraMapOneZeroSourceRun & \nExtraMapOneZeroTargetRun & \nExtraMapOneZeroFitted & \nExtraMapOneZeroRandomMax & \nExtraMapOneZeroNullMax \\
\nExtraMapOneOneSourceRun & \nExtraMapOneOneTargetRun & \nExtraMapOneOneFitted & \nExtraMapOneOneRandomMax & \nExtraMapOneOneNullMax \\
\nExtraMapOneTwoSourceRun & \nExtraMapOneTwoTargetRun & \nExtraMapOneTwoFitted & \nExtraMapOneTwoRandomMax & \nExtraMapOneTwoNullMax \\
\nExtraMapOneThreeSourceRun & \nExtraMapOneThreeTargetRun & \nExtraMapOneThreeFitted & \nExtraMapOneThreeRandomMax & \nExtraMapOneThreeNullMax \\
\nExtraMapOneFourSourceRun & \nExtraMapOneFourTargetRun & \nExtraMapOneFourFitted & \nExtraMapOneFourRandomMax & \nExtraMapOneFourNullMax \\
\nExtraMapOneFiveSourceRun & \nExtraMapOneFiveTargetRun & \nExtraMapOneFiveFitted & \nExtraMapOneFiveRandomMax & \nExtraMapOneFiveNullMax \\
\nExtraMapOneSixSourceRun & \nExtraMapOneSixTargetRun & \nExtraMapOneSixFitted & \nExtraMapOneSixRandomMax & \nExtraMapOneSixNullMax \\
\nExtraMapOneSevenSourceRun & \nExtraMapOneSevenTargetRun & \nExtraMapOneSevenFitted & \nExtraMapOneSevenRandomMax & \nExtraMapOneSevenNullMax \\
\nExtraMapOneEightSourceRun & \nExtraMapOneEightTargetRun & \nExtraMapOneEightFitted & \nExtraMapOneEightRandomMax & \nExtraMapOneEightNullMax \\
\nExtraMapOneNineSourceRun & \nExtraMapOneNineTargetRun & \nExtraMapOneNineFitted & \nExtraMapOneNineRandomMax & \nExtraMapOneNineNullMax \\
\nExtraMapTwoZeroSourceRun & \nExtraMapTwoZeroTargetRun & \nExtraMapTwoZeroFitted & \nExtraMapTwoZeroRandomMax & \nExtraMapTwoZeroNullMax \\
\nExtraMapTwoOneSourceRun & \nExtraMapTwoOneTargetRun & \nExtraMapTwoOneFitted & \nExtraMapTwoOneRandomMax & \nExtraMapTwoOneNullMax \\
\nExtraMapTwoTwoSourceRun & \nExtraMapTwoTwoTargetRun & \nExtraMapTwoTwoFitted & \nExtraMapTwoTwoRandomMax & \nExtraMapTwoTwoNullMax \\
\nExtraMapTwoThreeSourceRun & \nExtraMapTwoThreeTargetRun & \nExtraMapTwoThreeFitted & \nExtraMapTwoThreeRandomMax & \nExtraMapTwoThreeNullMax \\
\nExtraMapTwoFourSourceRun & \nExtraMapTwoFourTargetRun & \nExtraMapTwoFourFitted & \nExtraMapTwoFourRandomMax & \nExtraMapTwoFourNullMax \\
\nExtraMapTwoFiveSourceRun & \nExtraMapTwoFiveTargetRun & \nExtraMapTwoFiveFitted & \nExtraMapTwoFiveRandomMax & \nExtraMapTwoFiveNullMax \\
\nExtraMapTwoSixSourceRun & \nExtraMapTwoSixTargetRun & \nExtraMapTwoSixFitted & \nExtraMapTwoSixRandomMax & \nExtraMapTwoSixNullMax \\
\nExtraMapTwoSevenSourceRun & \nExtraMapTwoSevenTargetRun & \nExtraMapTwoSevenFitted & \nExtraMapTwoSevenRandomMax & \nExtraMapTwoSevenNullMax \\
\nExtraMapTwoEightSourceRun & \nExtraMapTwoEightTargetRun & \nExtraMapTwoEightFitted & \nExtraMapTwoEightRandomMax & \nExtraMapTwoEightNullMax \\
\nExtraMapTwoNineSourceRun & \nExtraMapTwoNineTargetRun & \nExtraMapTwoNineFitted & \nExtraMapTwoNineRandomMax & \nExtraMapTwoNineNullMax \\
\bottomrule
\end{tabular}

\par\noindent A sparse autoencoder (SAE) represents model activations by sparse feature coefficients. Each pair joins two seed runs of our SAEs on \nModelNameF at layers \nLayerUpA and \nLayerDownA, width \nWidthF and TopK \nKF, and its fitted map is compared with \nRandomDrawsPerPair random maps and \nNullDrawsPerPair permutations at \nHeldout held out Pile validation positions. The graph reconstruction score is one minus the summed squared prediction error over the summed squared reference entries, pooled over positions before averaging pairs. A negative score means a larger squared error than predicting a zero graph. Random maps match the fitted sparsity, and permutations shuffle the target feature assignments. Run numbers identify initialization seeds and are not quantities.
\end{table}

The regressors take the inputs the margin graph takes and nothing more. The six input regressor predicts each coordinate's coefficient change from that coordinate's own six inputs, with one set of weights shared across coordinates. The inputs are the encoder score before the intervention, its predicted change, the $k$th and the $(k+1)$th largest encoder scores before the intervention, and the same two order statistics of the predicted encoder scores. The two input regressor takes the first two of these. The margin graph is one function of the same six inputs, so the criterion tests whether a fitted learner reaches that function within a budget fixed in advance.

Each regressor is a multilayer perceptron of two hidden layers of \nRegWidth units with rectifiers, trained per run on \nRegTrain interventions drawn uniformly from the first \nRegTrainTokens positions at the three scaling factors, with every coordinate of each a training row, \nRegRows rows in all. The inputs are standardized by the training rows' mean and standard deviation. Each regressor minimizes a mean squared error by Adam at a learning rate of \nRegLr and the default betas and epsilon of Appendix~\ref{app:releases}, on batches of \nRegBatch rows for \nRegEpochs passes, \nRegSteps steps. The draw of interventions and the order of the training rows take seed \nRegSeedBase plus the run. The networks' weights start from the global generator seeded at \nRegInitSeed before the first run and drawn from in run order, and each layer takes PyTorch's default initialization, its weights and biases drawn uniformly within the inverse square root of its input width. On the scoring half, the six input regressor's median error runs from \nRegSixMin to \nRegSixMax over the runs and the two input regressor's from \nRegTwoMin to \nRegTwoMax. Both are above one, the error of predicting no change, against a mean over the runs of \nRegMarginScoringMean for the margin graph's median on the same half (Table~\ref{tab:regressor}).

\begin{table}[H]
\caption{\textbf{Fitted regressors err more than predicting no change.} Cells give each run's relative coefficient errors on the scoring half, the margin graph's quartiles and each regressor's median. Predicting no change errs by one, so a regressor above one does worse than no prediction.}
\label{tab:regressor}
\centering
\scriptsize
\setlength{\tabcolsep}{2.5pt}
\begin{tabular}{lrrrrr}
\toprule
run & \begin{tabular}[c]{@{}c@{}}margin graph\\first quartile\end{tabular} & \begin{tabular}[c]{@{}c@{}}margin graph\\median\end{tabular} & \begin{tabular}[c]{@{}c@{}}margin graph\\third quartile\end{tabular} & \begin{tabular}[c]{@{}c@{}}six input\\regressor\end{tabular} & \begin{tabular}[c]{@{}c@{}}two input\\regressor\end{tabular} \\
\midrule
\nRunIndexZero & \nRegMarginQOneZero & \nRegMarginMedZero & \nRegMarginQThreeZero & \nRegSixZero & \nRegTwoZero \\
\nRunIndexOne & \nRegMarginQOneOne & \nRegMarginMedOne & \nRegMarginQThreeOne & \nRegSixOne & \nRegTwoOne \\
\nRunIndexTwo & \nRegMarginQOneTwo & \nRegMarginMedTwo & \nRegMarginQThreeTwo & \nRegSixTwo & \nRegTwoTwo \\
\nRunIndexThree & \nRegMarginQOneThree & \nRegMarginMedThree & \nRegMarginQThreeThree & \nRegSixThree & \nRegTwoThree \\
\nRunIndexFour & \nRegMarginQOneFour & \nRegMarginMedFour & \nRegMarginQThreeFour & \nRegSixFour & \nRegTwoFour \\
\nRunIndexFive & \nRegMarginQOneFive & \nRegMarginMedFive & \nRegMarginQThreeFive & \nRegSixFive & \nRegTwoFive \\
\nRunIndexSix & \nRegMarginQOneSix & \nRegMarginMedSix & \nRegMarginQThreeSix & \nRegSixSix & \nRegTwoSix \\
\nRunIndexSeven & \nRegMarginQOneSeven & \nRegMarginMedSeven & \nRegMarginQThreeSeven & \nRegSixSeven & \nRegTwoSeven \\
\bottomrule
\end{tabular}

\par\noindent A sparse autoencoder (SAE) represents model activations by sparse feature coefficients. Errors cover \nRegScoreTokens held out scoring positions and the factors zero, \nDoseHalf and \nDoseNine, for each run of our SAEs on \nModelNameF at layers \nLayerUpA and \nLayerDownA, width \nWidthF and TopK \nKF, and each regressor is trained on the separate training half. TopK selects the largest positive encoder scores, and ReLU every positive score. The standard graph keeps the features selected before the intervention, and the margin graph selects again from the predicted scores. The smaller regressor takes the encoder score before the intervention and its predicted change, and the larger adds the selected and the next unselected score at the cutoff before and after that change. Each error has no unit, and smaller is better.
\end{table}

The sampled constant uses Jacobians at $x$, $x+v/2$ and $x+v$,
\begin{equation}
\widehat L_v=\max_{s,t\in\{0,1/2,1\},\ s<t}
\frac{\|J(x+tv)-J(x+sv)\|_2}{(t-s)\|v\|},
\end{equation}
with the spectral norm. This sampled maximum is a lower bound on the segment's Lipschitz constant, so it need not satisfy the hypothesis of Theorem~\ref{thm:band}. The selection states reported below are therefore an empirical check under this estimate.

We estimate the constant of Theorem~\ref{thm:band} on \nLipWritesPerRun zero ablations per run, \nLipFeaturesPerToken selected upstream features drawn without replacement at each of the \nTokens positions from seed \nLipSeed plus the run. Its median runs from \nConstMedianMin to \nConstMedianMax over the \nRunsF runs, and its largest value from \nConstGlobalMin to \nConstGlobalMax. Under the run's largest constant, \nCertGlobalActive of \nCertGlobalPredicted predicted selected states lie outside the error band over the runs, and none of them changed, which a criterion of Table~\ref{tab:rules} tests. Every prediction in these checks has as many predicted selected features as its SAE selects, so no largest bound is taken over an empty set.

\begin{figure}[H]
\centering
\includegraphics[width=\textwidth]{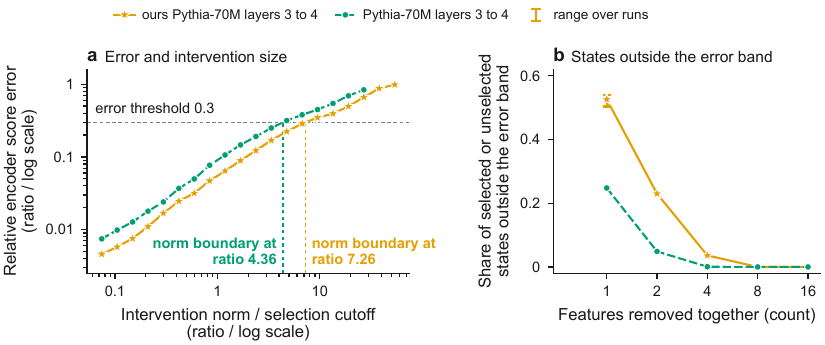}
\caption{\textbf{The first order error grows with the intervention's size, and fewer selection states lie outside the error band as more features are removed together.} Panel~a plots the median relative encoder score error of zero ablations against the intervention's norm over the median selection cutoff. Each vertical line marks a norm boundary, where that configuration's error reaches the threshold \nRegimeLevel. Panel~b plots the share of predicted selection states outside the error band against the count of features removed. Ours marks our sparse autoencoders, with bars for the range over \nRunsF runs at one feature.}
\label{fig:regime}
\end{figure}

The subsets are drawn \nSubsetsPerSize per size at each position from seed \nSubsetSeed plus the run, each of distinct features, with the draws independent. The check of the bound and the acceptance rule use the first two subsets drawn at each size and position, in the order drawn. On the first of our runs that gives \nBandWritesPerSize interventions per size. Under each intervention's own constant, the share of selection states outside the error band drops from \nCertOwnSizeTwoF on the subsets of \nSizeTwo features to \nCertOwnSizeEightFSci on the subsets of \nSizeEight and \nCertOwnSizeSixteenFSci on the subsets of \nSizeSixteen (Figure~\ref{fig:regime}). No selection state outside the error band changes at any size, and every first order remainder lies within its own bound at every size, a share of \nRemainderSizesMin on the first of our runs and on the public \nModelNameA pair.

The first order error grows with the intervention. We measure the relative encoder score error, the norm of the predicted score change less the exact change over the exact change's norm, against the intervention's norm divided by the median downstream cutoff over the positions. We call the value of the intervention's norm over the median cutoff at which the median error first reaches the error threshold \nRegimeLevel the norm boundary. On single feature interventions the norm boundary is \nBoundarySingleF on our SAEs, their \nRunsF runs pooled, and \nBoundarySingleRel on the public \nModelNameA pair. On subset interventions it is \nBoundarySubsetF and \nBoundarySubsetRel. The two populations pooled give norm boundaries of \nBoundaryF and \nBoundaryRel, within a ratio of \nBoundaryRatio, under the factor of \nBoundaryFactor fixed in advance, and the median intervention has a norm of \nNormMedianF and \nNormMedianRel times the median cutoff. On the public pairs' single feature interventions the norm boundaries run from \nBoundaryPubSingleMin to \nBoundaryPubSingleMax.

The acceptance rule accepts a predicted selection state where the state lies outside the error band of Theorem~\ref{thm:band} under the intervention's own constant and the intervention's norm lies under the norm boundary, set on half the \nTokens positions. On the other half it measures \nLipFeaturesPerToken sampled features per position at the three scaling factors, over our SAEs' \nRunsF runs and the \nDiscussionPublicPairsWords public TopK pairs. It accepts no wrong selection state among \nRuleDosePooledAccepted, at \nRuleDosePooledCovMin to \nRuleDosePooledCovMax of a single feature intervention's predicted selection states and \nRuleDosePooledLiveMin to \nRuleDosePooledLiveMax of its predicted selected ones. At the zero ablation it accepts \nRuleCovSubsetMin to \nRuleCovSubsetMax of a subset intervention's predicted selection states, again with none wrong.

We find the norm boundary by binning the intervention's norm geometrically from \nBinLow to \nBinHigh times the median cutoff, at \nBinsPerDoubling bins per doubling. A bin with fewer than \nBinsMinRows interventions is left empty, and the median error is computed in each bin. The norm boundary is interpolated linearly in the logarithm of the bin centres between the last bin below the error threshold and the first at or above it.

The acceptance rule is scored on every population in Table~\ref{tab:rule}. Its constant is the three point constant above. The norm boundary of each run and each pair is set on the calibration half of the positions, one permutation from seed \nSplitSeed putting \nSplitHalf of the \nTokens positions in each half, with the calibration half's median cutoff as the unit. No accepted selection state is wrong on any population, and the error band alone on the same interventions, with no condition on the norm, accepts no wrong selection state either.

\begin{table}[H]
\caption{\textbf{The acceptance rule accepts no wrong selection state on the scoring half.} Rows give zero ablations of one feature or of subsets, scored on half the positions after calibration on the other half. Coverage divides accepted selection states by all predicted ones, or by predicted selected ones in the selected coverage column.}
\label{tab:rule}
\centering
\scriptsize
\setlength{\tabcolsep}{2.5pt}
\begin{tabular}{lllrrrrrrr}
\toprule
SAEs & layers & interventions & norm boundary & \begin{tabular}[c]{@{}c@{}}median error\\inside\end{tabular} & scored & inside & coverage & \begin{tabular}[c]{@{}c@{}}selected\\coverage\end{tabular} & wrong \\
\midrule
ours, run \nRunIndexZero & \nLayerUpA, \nLayerDownA & single & \nRuleBoundFZeroSingle & \nRuleErrFZeroSingle & \nRuleWritesFZeroSingle & \nRuleInsideFZeroSingle & \nRuleCovFZeroSingle & \nRuleLiveFZeroSingle & \nRuleWrongFZeroSingle \\
ours, run \nRunIndexOne & \nLayerUpA, \nLayerDownA & single & \nRuleBoundFOneSingle & \nRuleErrFOneSingle & \nRuleWritesFOneSingle & \nRuleInsideFOneSingle & \nRuleCovFOneSingle & \nRuleLiveFOneSingle & \nRuleWrongFOneSingle \\
ours, run \nRunIndexTwo & \nLayerUpA, \nLayerDownA & single & \nRuleBoundFTwoSingle & \nRuleErrFTwoSingle & \nRuleWritesFTwoSingle & \nRuleInsideFTwoSingle & \nRuleCovFTwoSingle & \nRuleLiveFTwoSingle & \nRuleWrongFTwoSingle \\
ours, run \nRunIndexThree & \nLayerUpA, \nLayerDownA & single & \nRuleBoundFThreeSingle & \nRuleErrFThreeSingle & \nRuleWritesFThreeSingle & \nRuleInsideFThreeSingle & \nRuleCovFThreeSingle & \nRuleLiveFThreeSingle & \nRuleWrongFThreeSingle \\
ours, run \nRunIndexFour & \nLayerUpA, \nLayerDownA & single & \nRuleBoundFFourSingle & \nRuleErrFFourSingle & \nRuleWritesFFourSingle & \nRuleInsideFFourSingle & \nRuleCovFFourSingle & \nRuleLiveFFourSingle & \nRuleWrongFFourSingle \\
ours, run \nRunIndexFive & \nLayerUpA, \nLayerDownA & single & \nRuleBoundFFiveSingle & \nRuleErrFFiveSingle & \nRuleWritesFFiveSingle & \nRuleInsideFFiveSingle & \nRuleCovFFiveSingle & \nRuleLiveFFiveSingle & \nRuleWrongFFiveSingle \\
ours, run \nRunIndexSix & \nLayerUpA, \nLayerDownA & single & \nRuleBoundFSixSingle & \nRuleErrFSixSingle & \nRuleWritesFSixSingle & \nRuleInsideFSixSingle & \nRuleCovFSixSingle & \nRuleLiveFSixSingle & \nRuleWrongFSixSingle \\
ours, run \nRunIndexSeven & \nLayerUpA, \nLayerDownA & single & \nRuleBoundFSevenSingle & \nRuleErrFSevenSingle & \nRuleWritesFSevenSingle & \nRuleInsideFSevenSingle & \nRuleCovFSevenSingle & \nRuleLiveFSevenSingle & \nRuleWrongFSevenSingle \\
ours, run \nRunIndexZero & \nLayerUpA, \nLayerDownA & subsets & \nRuleBoundFZeroSubset & \nRuleErrFZeroSubset & \nRuleWritesFZeroSubset & \nRuleInsideFZeroSubset & \nRuleCovFZeroSubset & \nRuleLiveFZeroSubset & \nRuleWrongFZeroSubset \\
ours, run \nRunIndexOne & \nLayerUpA, \nLayerDownA & subsets & \nRuleBoundFOneSubset & \nRuleErrFOneSubset & \nRuleWritesFOneSubset & \nRuleInsideFOneSubset & \nRuleCovFOneSubset & \nRuleLiveFOneSubset & \nRuleWrongFOneSubset \\
ours, run \nRunIndexTwo & \nLayerUpA, \nLayerDownA & subsets & \nRuleBoundFTwoSubset & \nRuleErrFTwoSubset & \nRuleWritesFTwoSubset & \nRuleInsideFTwoSubset & \nRuleCovFTwoSubset & \nRuleLiveFTwoSubset & \nRuleWrongFTwoSubset \\
ours, run \nRunIndexThree & \nLayerUpA, \nLayerDownA & subsets & \nRuleBoundFThreeSubset & \nRuleErrFThreeSubset & \nRuleWritesFThreeSubset & \nRuleInsideFThreeSubset & \nRuleCovFThreeSubset & \nRuleLiveFThreeSubset & \nRuleWrongFThreeSubset \\
ours, run \nRunIndexFour & \nLayerUpA, \nLayerDownA & subsets & \nRuleBoundFFourSubset & \nRuleErrFFourSubset & \nRuleWritesFFourSubset & \nRuleInsideFFourSubset & \nRuleCovFFourSubset & \nRuleLiveFFourSubset & \nRuleWrongFFourSubset \\
ours, run \nRunIndexFive & \nLayerUpA, \nLayerDownA & subsets & \nRuleBoundFFiveSubset & \nRuleErrFFiveSubset & \nRuleWritesFFiveSubset & \nRuleInsideFFiveSubset & \nRuleCovFFiveSubset & \nRuleLiveFFiveSubset & \nRuleWrongFFiveSubset \\
ours, run \nRunIndexSix & \nLayerUpA, \nLayerDownA & subsets & \nRuleBoundFSixSubset & \nRuleErrFSixSubset & \nRuleWritesFSixSubset & \nRuleInsideFSixSubset & \nRuleCovFSixSubset & \nRuleLiveFSixSubset & \nRuleWrongFSixSubset \\
ours, run \nRunIndexSeven & \nLayerUpA, \nLayerDownA & subsets & \nRuleBoundFSevenSubset & \nRuleErrFSevenSubset & \nRuleWritesFSevenSubset & \nRuleInsideFSevenSubset & \nRuleCovFSevenSubset & \nRuleLiveFSevenSubset & \nRuleWrongFSevenSubset \\
public \nModelNameA & \nLayerUpA, \nLayerDownA & single & \nRuleBoundASingle & \nRuleErrASingle & \nRuleWritesASingle & \nRuleInsideASingle & \nRuleCovASingle & \nRuleLiveASingle & \nRuleWrongASingle \\
public \nModelNameB & \nLayerUpB, \nLayerDownB & single & \nRuleBoundBSingle & \nRuleErrBSingle & \nRuleWritesBSingle & \nRuleInsideBSingle & \nRuleCovBSingle & \nRuleLiveBSingle & \nRuleWrongBSingle \\
public \nModelNameC & \nLayerUpC, \nLayerDownC & single & \nRuleBoundCSingle & \nRuleErrCSingle & \nRuleWritesCSingle & \nRuleInsideCSingle & \nRuleCovCSingle & \nRuleLiveCSingle & \nRuleWrongCSingle \\
public \nModelNameD & \nLayerUpD, \nLayerDownD & single & \nRuleBoundDSingle & \nRuleErrDSingle & \nRuleWritesDSingle & \nRuleInsideDSingle & \nRuleCovDSingle & \nRuleLiveDSingle & \nRuleWrongDSingle \\
public \nModelNameE & \nLayerUpE, \nLayerDownE & single & \nRuleBoundESingle & \nRuleErrESingle & \nRuleWritesESingle & \nRuleInsideESingle & \nRuleCovESingle & \nRuleLiveESingle & \nRuleWrongESingle \\
public \nModelNameA & \nLayerUpA, \nLayerDownA & subsets & \nRuleBoundASubset & \nRuleErrASubset & \nRuleWritesASubset & \nRuleInsideASubset & \nRuleCovASubset & \nRuleLiveASubset & \nRuleWrongASubset \\
\bottomrule
\end{tabular}

\par\noindent A sparse autoencoder (SAE) represents model activations by sparse feature coefficients. Zero ablations are scored on \nRuleScoring held out Pile validation positions. Rows marked ours are our SAEs on \nModelNameF at layers \nLayerUpA and \nLayerDownA, width \nWidthF and TopK \nKF. The norm boundary is the intervention norm, over the calibration positions' median downstream cutoff, at which the relative encoder score error reaches \nRuleLevel, and nothing beyond it is accepted. Median error inside is that error on the interventions inside the boundary. TopK selects the largest positive encoder scores, and ReLU every positive score. A selection state is a feature's selected or unselected state, and the rule accepts one where its predicted margin exceeds the sampled error bound on either side of the cutoff. Scored and inside count interventions, and wrong counts accepted selection states with the wrong value. The relative encoder score error divides the norm of the predicted score change's error by the norm of the exact score change. Coverage and errors have no unit.
\end{table}

Table~\ref{tab:rule-doses} scores the same rule at the scaling factors \nDoseHalf and \nDoseNine on the same \nLipFeaturesPerToken features per position of the scoring half, under the same norm boundaries. No accepted selection state is wrong at any scaling factor, and the coverage rises with the scaling factor, from \nRuleDoseZeroCovMin to \nRuleDoseZeroCovMax at the zero ablation to \nRuleDoseNineCovMin to \nRuleDoseNineCovMax at the scaling factor \nDoseNine.

\begin{table}[H]
\caption{\textbf{The acceptance rule's coverage rises with the scaling factor, and no accepted selection state is wrong.} Cells give the rule's coverage at each factor and pooled, per public pair or per run of our sparse autoencoders (SAEs). Coverage divides accepted selection states by all predicted ones, and selected coverage divides by predicted selected ones.}
\label{tab:rule-doses}
\centering
\scriptsize
\setlength{\tabcolsep}{2.5pt}
\begin{tabular}{llrrrrrrr}
\toprule
SAEs & layers & \begin{tabular}[c]{@{}c@{}}coverage\\at zero\end{tabular} & \begin{tabular}[c]{@{}c@{}}coverage\\at \nDoseHalf\end{tabular} & \begin{tabular}[c]{@{}c@{}}coverage\\at \nDoseNine\end{tabular} & \begin{tabular}[c]{@{}c@{}}coverage\\pooled\end{tabular} & \begin{tabular}[c]{@{}c@{}}selected coverage\\pooled\end{tabular} & \begin{tabular}[c]{@{}c@{}}inside\\pooled\end{tabular} & \begin{tabular}[c]{@{}c@{}}wrong\\pooled\end{tabular} \\
\midrule
ours, run \nRunIndexZero & \nLayerUpA, \nLayerDownA & \nRuleDoseCovFZeroZero & \nRuleDoseCovFZeroHalf & \nRuleDoseCovFZeroNine & \nRuleDoseCovFZeroPooled & \nRuleDoseLiveFZero & \nRuleDoseInsideFZeroPooled & \nRuleDoseWrongFZeroPooled \\
ours, run \nRunIndexOne & \nLayerUpA, \nLayerDownA & \nRuleDoseCovFOneZero & \nRuleDoseCovFOneHalf & \nRuleDoseCovFOneNine & \nRuleDoseCovFOnePooled & \nRuleDoseLiveFOne & \nRuleDoseInsideFOnePooled & \nRuleDoseWrongFOnePooled \\
ours, run \nRunIndexTwo & \nLayerUpA, \nLayerDownA & \nRuleDoseCovFTwoZero & \nRuleDoseCovFTwoHalf & \nRuleDoseCovFTwoNine & \nRuleDoseCovFTwoPooled & \nRuleDoseLiveFTwo & \nRuleDoseInsideFTwoPooled & \nRuleDoseWrongFTwoPooled \\
ours, run \nRunIndexThree & \nLayerUpA, \nLayerDownA & \nRuleDoseCovFThreeZero & \nRuleDoseCovFThreeHalf & \nRuleDoseCovFThreeNine & \nRuleDoseCovFThreePooled & \nRuleDoseLiveFThree & \nRuleDoseInsideFThreePooled & \nRuleDoseWrongFThreePooled \\
ours, run \nRunIndexFour & \nLayerUpA, \nLayerDownA & \nRuleDoseCovFFourZero & \nRuleDoseCovFFourHalf & \nRuleDoseCovFFourNine & \nRuleDoseCovFFourPooled & \nRuleDoseLiveFFour & \nRuleDoseInsideFFourPooled & \nRuleDoseWrongFFourPooled \\
ours, run \nRunIndexFive & \nLayerUpA, \nLayerDownA & \nRuleDoseCovFFiveZero & \nRuleDoseCovFFiveHalf & \nRuleDoseCovFFiveNine & \nRuleDoseCovFFivePooled & \nRuleDoseLiveFFive & \nRuleDoseInsideFFivePooled & \nRuleDoseWrongFFivePooled \\
ours, run \nRunIndexSix & \nLayerUpA, \nLayerDownA & \nRuleDoseCovFSixZero & \nRuleDoseCovFSixHalf & \nRuleDoseCovFSixNine & \nRuleDoseCovFSixPooled & \nRuleDoseLiveFSix & \nRuleDoseInsideFSixPooled & \nRuleDoseWrongFSixPooled \\
ours, run \nRunIndexSeven & \nLayerUpA, \nLayerDownA & \nRuleDoseCovFSevenZero & \nRuleDoseCovFSevenHalf & \nRuleDoseCovFSevenNine & \nRuleDoseCovFSevenPooled & \nRuleDoseLiveFSeven & \nRuleDoseInsideFSevenPooled & \nRuleDoseWrongFSevenPooled \\
public \nModelNameA & \nLayerUpA, \nLayerDownA & \nRuleDoseCovAZero & \nRuleDoseCovAHalf & \nRuleDoseCovANine & \nRuleDoseCovAPooled & \nRuleDoseLiveA & \nRuleDoseInsideAPooled & \nRuleDoseWrongAPooled \\
public \nModelNameB & \nLayerUpB, \nLayerDownB & \nRuleDoseCovBZero & \nRuleDoseCovBHalf & \nRuleDoseCovBNine & \nRuleDoseCovBPooled & \nRuleDoseLiveB & \nRuleDoseInsideBPooled & \nRuleDoseWrongBPooled \\
public \nModelNameC & \nLayerUpC, \nLayerDownC & \nRuleDoseCovCZero & \nRuleDoseCovCHalf & \nRuleDoseCovCNine & \nRuleDoseCovCPooled & \nRuleDoseLiveC & \nRuleDoseInsideCPooled & \nRuleDoseWrongCPooled \\
public \nModelNameD & \nLayerUpD, \nLayerDownD & \nRuleDoseCovDZero & \nRuleDoseCovDHalf & \nRuleDoseCovDNine & \nRuleDoseCovDPooled & \nRuleDoseLiveD & \nRuleDoseInsideDPooled & \nRuleDoseWrongDPooled \\
public \nModelNameE & \nLayerUpE, \nLayerDownE & \nRuleDoseCovEZero & \nRuleDoseCovEHalf & \nRuleDoseCovENine & \nRuleDoseCovEPooled & \nRuleDoseLiveE & \nRuleDoseInsideEPooled & \nRuleDoseWrongEPooled \\
\bottomrule
\end{tabular}

\par\noindent At each factor the rule is scored on \nLipFeaturesPerToken features at each of \nRuleScoring held out Pile validation positions, on our SAEs on \nModelNameF at layers \nLayerUpA and \nLayerDownA, width \nWidthF and TopK \nKF. The last two columns count the interventions inside the calibrated norm boundary and the wrong accepted selection states. A sparse autoencoder (SAE) represents model activations by sparse feature coefficients. TopK selects the largest positive encoder scores, and ReLU every positive score. The scaling factor multiplies a selected upstream feature's coefficient, and zero removes that feature. A selection state is a feature's selected or unselected state, and the rule accepts one where its predicted margin exceeds the sampled error bound and the intervention's norm lies under the calibrated norm boundary. Coverage has no unit, and pooled combines the three factors.
\end{table}

Table~\ref{tab:sizes} gives the errors by subset size behind the subset criteria. On our SAEs the standard graph's error decreases with the subset's size, while the margin graph's increases.

\begin{table}[H]
\caption{\textbf{Integrated gradients that select again predict subset ablations more closely than either graph.} Cells give median relative coefficient errors by the number of features ablated together, on our sparse autoencoders (SAEs) and on the public \nModelNameA pair. Integrated gradients (IG) take \nPathProducts or \nPathTenProducts points for each subset, while both graphs reuse columns built once per position.}
\label{tab:sizes}
\centering
\scriptsize
\setlength{\tabcolsep}{2.5pt}
\begin{tabular}{lrrrrrrrr}
\toprule
\begin{tabular}[c]{@{}c@{}}features\\ablated\end{tabular} & \begin{tabular}[c]{@{}c@{}}ours\\standard graph\end{tabular} & \begin{tabular}[c]{@{}c@{}}ours\\margin graph\end{tabular} & \begin{tabular}[c]{@{}c@{}}ours\\IG at \nPathProducts\end{tabular} & \begin{tabular}[c]{@{}c@{}}ours\\IG at \nPathTenProducts\end{tabular} & \begin{tabular}[c]{@{}c@{}}public\\standard graph\end{tabular} & \begin{tabular}[c]{@{}c@{}}public\\margin graph\end{tabular} & \begin{tabular}[c]{@{}c@{}}public\\IG at \nPathProducts\end{tabular} & \begin{tabular}[c]{@{}c@{}}public\\IG at \nPathTenProducts\end{tabular} \\
\midrule
\nSizeTwo & \nFixedSizeTwoF & \nMarginSizeTwoF & \nPathFiveRerankSizeTwoF & \nPathTenRerankSizeTwoF & \nFixedSizeTwoRel & \nMarginSizeTwoRel & \nPathFiveRerankSizeTwoRel & \nPathTenRerankSizeTwoRel \\
\nSizeFour & \nFixedSizeFourF & \nMarginSizeFourF & \nPathFiveRerankSizeFourF & \nPathTenRerankSizeFourF & \nFixedSizeFourRel & \nMarginSizeFourRel & \nPathFiveRerankSizeFourRel & \nPathTenRerankSizeFourRel \\
\nSizeEight & \nFixedSizeEightF & \nMarginSizeEightF & \nPathFiveRerankSizeEightF & \nPathTenRerankSizeEightF & \nFixedSizeEightRel & \nMarginSizeEightRel & \nPathFiveRerankSizeEightRel & \nPathTenRerankSizeEightRel \\
\nSizeSixteen & \nFixedSizeSixteenF & \nMarginSizeSixteenF & \nPathFiveRerankSizeSixteenF & \nPathTenRerankSizeSixteenF & \nFixedSizeSixteenRel & \nMarginSizeSixteenRel & \nPathFiveRerankSizeSixteenRel & \nPathTenRerankSizeSixteenRel \\
\bottomrule
\end{tabular}

\par\noindent Each size takes \nSubsetsPerSize zero ablated subsets at each of \nTokens held out Pile validation positions. The public columns are the pair at layers \nLayerUpA and \nLayerDownA, whose whole selected set has \nSizeSixteen features, and the columns for our SAEs, on the same model and layers with width \nWidthF and TopK \nKF, average \nRunsF run medians. A sparse autoencoder (SAE) represents model activations by sparse feature coefficients. TopK selects the largest positive encoder scores, and ReLU every positive score. The relative coefficient error divides the norm of a prediction's error by the norm of the exact coefficient change, and smaller is better. The standard graph keeps the features selected before the intervention, and the margin graph selects again from the predicted scores. Integrated gradients average derivatives along the intervention's path, and IG here selects again from its predicted scores.
\end{table}

A further criterion asks through which middle features a path across two blocks runs. Write $v$ for the intervention, $\Delta y$ for the exact middle activation change, $\Delta\hat y = J_1 v$ for its first order change and $\Delta\hat z$ for the margin graph's middle coefficient change. Let $i^\ast$ be the position's selected upstream feature of largest coefficient, and let $\rho_j = |u_{i^\ast}\, e_{2,j}^\top J_1 d_{i^\ast}|$, with $e_{2,j}$ the middle encoder's row, be the first order change that its zero ablation makes at middle feature $j$. A middle feature lies within that change when the size of its margin is at most $\rho_j$, and beyond it otherwise. The parts are
\begin{equation*}
\Delta y_{\mathrm{b}} = G_2 (\Delta\hat z \odot \mathbf{1}_{\mathrm{beyond}}) , \qquad \Delta y_{\mathrm{w}} = G_2 (\Delta\hat z \odot \mathbf{1}_{\mathrm{within}}) , \qquad \Delta\hat y - \Delta y_{\mathrm{b}} - \Delta y_{\mathrm{w}} ,
\end{equation*}
the last the remainder, and each part's share is its inner product with $\Delta y$ over $\|\Delta y\|^2$. Over every intervention measured two blocks downstream at the three scaling factors, the criterion asked the median share of the part beyond to exceed that of the part within by \nMediationMargin. The part beyond projects at \nMedStableShare and the part within at \nMedMarginalShare, a difference of \nMediationDiff, so the criterion was met. Passed alone through the second block, the part beyond errs at the third layer by \nCarryStable, the part within by \nCarryMarginal and the remainder by \nCarryRemainder.

\section{The downstream test}
\label{app:downstream}

Patched into the model, the margin graph's coefficients predict the change in the agreement logit difference with a mean error of \nDsErrMargin, against \nDsErrFixed for the standard graph's, on one SAE pair and one task. The test asks whether a more accurate coefficient prediction also predicts the model's behaviour. We patch each graph's predicted downstream coefficients into the model and compare the change this makes in a task's logit difference with the change the intervention makes. The test is an analysis we added after the main results$^\dagger$, and we fixed its tasks, interventions, criteria and checks before seeing its outcome. Comparing changes in a logit difference follows the faithfulness measures of \citet{wang2022interpretability} and \citet{marks2025sparse}.

\textbf{Tasks.} The test runs two templated tasks on the public \nModelNameA pair at layers \nLayerUpA and \nLayerDownA. Subject verb agreement follows the attractor design of \citet{linzen2016assessing} and \citet{finlayson2021causal}. Each prompt opens with the end of text token and continues with \emph{The}, a subject noun, a preposition, \emph{the} and an attractor noun of the other number, and it ends where the verb would come. Its lists contain \nDsSubjects subject nouns and \nDsAttractors attractor nouns in both numbers, with \nDsPrepositions prepositions. Its metric is the logit of the verb form that agrees with the subject minus that of the form that does not, for \nDsVerbPairs pairs of verb forms. Induction follows \citet{olsson2022context}. Each prompt contains \nDsSequenceLength distinct tokens drawn from the \nDsPoolSize words of the vocabulary that have three to eight letters, and then the first of them again. Its metric is the logit of the token that followed the first occurrence minus the largest logit among the sequence's other tokens. Each task draws \nDsDevPrompts development prompts at seed \nDsSeedDev and \nDsTestPrompts test prompts at seed \nDsSeedTest, and the two sets share no prompt.

The subject noun pairs are author/authors, doctor/doctors, teacher/teachers, farmer/farmers, student/students, manager/managers, officer/officers, pilot/pilots, singer/singers, writer/writers, lawyer/lawyers, driver/drivers, player/players, worker/workers, leader/leaders, artist/artists, nurse/nurses, judge/judges, guard/guards, friend/friends, engineer/engineers, soldier/soldiers, scientist/scientists, member/members, girl/girls, boy/boys, woman/women, man/men, child/children and king/kings.

The attractor noun pairs are cabinet/cabinets, table/tables, window/windows, building/buildings, house/houses, car/cars, book/books, door/doors, box/boxes, garden/gardens, road/roads, school/schools, office/offices, tree/trees, park/parks, room/rooms, chair/chairs, wall/walls, bridge/bridges, river/rivers, station/stations, church/churches, hotel/hotels, farm/farms and shop/shops.

The prepositions are near, behind, beside, with, by, from, above, below, across and under.

The verb pairs are is/are, was/were and has/have.

The random entering feature control, the prompt bootstrap and the constructed patch checks each use seed \nDsSeedTest. The development and test prompt seeds are given above.

\textbf{Which task is included.} We include a task where the model's metric with no intervention is positive on at least \nDsIncludeShare of its development prompts. Subject verb agreement is positive on \nDsAgreeShare of them and is included. Induction is positive on \nDsInductShare and is left out, so every result below is on subject verb agreement.

\textbf{Interventions and errors.} Each test prompt has \nDsActive selected upstream features at its last position, and each is scaled at the zero ablation and the factors \nDoseHalf and \nDoseNine of Section~\ref{sec:setup}, which gives \nDsUnitsTest interventions. The intervention's change is the metric after the intervention minus the metric before it, through the model. Write $y$ for the downstream activation at the last position before the intervention and $z$ for its coefficients. Each graph's predicted coefficients $\hat z$ are patched in at layer \nLayerDownA as $y + G(\hat z - z)$, which keeps the reconstruction residual from before the intervention, and the model runs on to the logits. A graph's error on an intervention is the absolute difference between the metric's change under the patch and the intervention's change. Predicting no change patches $z$ itself, so its error is the absolute change. Interventions whose absolute change lies under \nDsFloor, the median absolute change over the development prompts' \nDsUnitsDev interventions, are left out, which keeps \nDsAbove of the \nDsUnitsTest.

\textbf{Criteria fixed in advance.} The margin graph's mean error over the standard graph's, pooled over the kept interventions, had to lie under \nDsRatioBound, with the upper end of its \nDsIntervalLevel{}\% interval from \nDsDraws bootstrap draws over prompts under one. The margin graph's mean error also had to lie under that of predicting no change and under that of a random control. The control adds to the standard graph's coefficients as many features unselected before the intervention as the margin graph lets enter, drawn at random and set at the margin graph's cutoff after selecting again. A standard graph error under \nDsAttainShare of the mean absolute change would have left the first criterion out of reach, and fewer than \nDsMinUnits kept interventions would have left the test without a result.

\textbf{Checks of the code.} Table~\ref{tab:downstream-checks} lists the checks of the test's code on constructed instances, each run before the outcome, with the two controls. The first run stopped before the test prompts because two checks failed. Its patched continuation ran each block's softmax in double precision while the model runs it in single precision, and it missed the model's metric by \nDsContFirst against a bound of \nDsContBound. Its reproduction of the errors of Section~\ref{sec:margin} differed from them by up to \nDsReproGap against a bound of \nDsCheckBound, since those errors move by up to \nDsReproMove when the earlier blocks' softmax runs in double rather than single precision. The one change made before the second run, and before any outcome, runs every pass at the model's own softmax and sets the reproduction's bound at \nDsReproBound, under half a unit of the third decimal place at which we print errors.

\begin{table}[H]
\caption{\textbf{The checks pass, and both controls err less than the margin graph.} Check rows give each check's largest deviation beside its bound, and control rows give mean absolute errors of the agreement logit difference on the kept test interventions. Both concern subject verb agreement on the public \nModelNameA pair at layers \nLayerUpA and \nLayerDownA.}
\label{tab:downstream-checks}
\centering
\small
\begin{tabular}{p{\declD}ll}
\toprule
check or control & bound & result \\
\midrule
the patched coefficients from before the intervention reproduce the model's metric on every development prompt & \nDsContBound & \nDsContSecond \\
the exact coefficients, patched with the reconstruction residual of the intervened pass, reproduce the intervention's change & \nDsCheckBound & \nDsPlantedLargest \\
predicting no change errs by the absolute change & \nDsCheckBound & \nDsNullLargest \\
a scaling factor of one changes nothing & \nDsCheckBound & \nDsOneLargest \\
the errors of Section~\ref{sec:margin}, reproduced on \nDsReproRows interventions & \nDsReproBound & \nDsReproGap \\
the bootstrap on identical errors returns an interval of zero width & zero width & zero width \\
predicting no change, mean error & & \nDsErrNull \\
the random control, mean error & & \nDsErrControl \\
\bottomrule
\end{tabular}

\par\noindent Check results are the largest deviations on the development populations of the preceding paragraph, except the \nDsReproRows separate validation interventions that reproduce the coefficient errors. The \nDsAbove kept test interventions have a true change of at least \nDsFloor, the floor fixed on the development prompts. The agreement logit difference is the correct verb form's logit less the incorrect form's logit. A patch replaces the decoded feature coefficients and keeps the reconstruction residual from before the intervention. The random control adds randomly chosen unselected features at the margin graph's cutoff. Bounds and results are in logit units, except the reproduction check, which compares relative coefficient errors, the norm of the prediction's error over the norm of the exact coefficient change, with no unit.
\end{table}

\textbf{The standard graph's coefficients err least.} The criteria were not met. Over the \nDsAbove kept interventions the margin graph's coefficients err by \nDsErrMargin and the standard graph's by \nDsErrFixed, a ratio of \nDsRatio with a \nDsIntervalLevel{}\% interval from \nDsRatioLow to \nDsRatioHigh. Predicting no change errs by \nDsErrNull and the random control by \nDsErrControl, so the margin graph's coefficients move the logit difference less faithfully than either. Table~\ref{tab:downstream-factors} gives each error at each scaling factor, and the order is the same at all three. The margin graph's error is the smaller on \nDsShareSmaller of the interventions.

\begin{table}[H]
\caption{\textbf{The standard graph's coefficients err least at every scaling factor.} Cells give the mean absolute error in the agreement logit difference, in logit units, over the test interventions counted in each row, on the public \nModelNameA pair at layers \nLayerUpA and \nLayerDownA.}
\label{tab:downstream-factors}
\centering
\small
\begin{tabular}{lrrrrr}
\toprule
scaling factor & interventions & standard graph & margin graph & no change & random control \\
\midrule
zero ablation & \nDsUnitsZero & \nDsErrFixedZero & \nDsErrMarginZero & \nDsErrNullZero & \nExtraFactorZeroControlError \\
\nDoseHalf & \nDsUnitsHalf & \nDsErrFixedHalf & \nDsErrMarginHalf & \nDsErrNullHalf & \nExtraFactorOneControlError \\
\nDoseNine & \nDsUnitsNine & \nDsErrFixedNine & \nDsErrMarginNine & \nDsErrNullNine & \nExtraFactorTwoControlError \\
\bottomrule
\end{tabular}

\par\noindent Every intervention scales one selected feature at the final prompt position and has an absolute true change of at least \nDsFloor, fixed on the development prompts, and the text reports the pooled ratio's bootstrap interval. The agreement logit difference is the correct verb form's logit less the incorrect form's logit. The standard graph keeps the features selected before the intervention, and the margin graph selects again from the predicted scores. The scaling factor multiplies a selected upstream feature's coefficient, and zero removes that feature. The random control adds as many randomly chosen entering features as the margin graph does, at its cutoff.
\end{table}

\textbf{The entering features raise the error.} Added alone to the standard graph's coefficients, the margin graph's entering features raise the error to \nDsErrEntrants, above the random control's. They account for \nDsEntrantShare of the intervention's change, the sum of their part of the patched change times the intervention's change over the sum of the squared changes.

\textbf{The complete population gives the same ordering.} We fixed the floor on the development prompts before seeing any test outcome. For comparison, Table~\ref{tab:downstream-populations} also gives every test intervention without the floor. The margin graph's mean error is \nReadAllMarginMeanShort against \nReadAllFixedMeanShort for the standard graph, a ratio of \nReadAllRatioShort with a \nDsIntervalLevel{}\% interval from \nReadAllLowerShort to \nReadAllUpperShort under the same prompt bootstrap. These descriptive analyses change neither the test's criteria, fixed in advance, nor its outcome.

\textbf{Exact coefficients still leave residual error.} Patching the exact downstream coefficients with the reconstruction residual from before the intervention gives a mean error of \nReadKeptExactCleanResidualMeanShort on kept interventions and \nReadAllExactCleanResidualMeanShort on all interventions. We give their medians beside both graphs' in Table~\ref{tab:downstream-populations}. This patch removes the error of the coefficient prediction while keeping the omitted change of the residual. It therefore separates the two sources by comparison, though not by an additive decomposition of absolute errors through the nonlinear rest of the model.

\begin{table}[H]
\caption{\textbf{The margin graph errs more on both test populations.} Cells give mean and median absolute errors of the change in the agreement logit difference, in logit units, for predicted and exact coefficients patched with the reconstruction residual from before the intervention. Kept interventions have an absolute true change of at least \nDsFloor, and all interventions drop this floor.}
\label{tab:downstream-populations}
\centering
\small
\begin{tabular}{lrlrr}
\toprule
population & interventions & predictor & mean logit error & median logit error \\
\midrule
kept & \nReadKeptUnits & standard graph & \nReadKeptFixedMeanError & \nReadKeptFixedMedianError \\
kept & \nReadKeptUnits & margin graph & \nReadKeptMarginMeanError & \nReadKeptMarginMedianError \\
kept & \nReadKeptUnits & no change & \nReadKeptNullMeanError & \nReadKeptNullMedianError \\
kept & \nReadKeptUnits & random entering features & \nReadKeptControlMeanError & \nReadKeptControlMedianError \\
kept & \nReadKeptUnits & exact coefficients, earlier residual & \nReadKeptExactCleanResidualMeanError & \nReadKeptExactCleanResidualMedianError \\
all & \nReadAllUnits & standard graph & \nReadAllFixedMeanError & \nReadAllFixedMedianError \\
all & \nReadAllUnits & margin graph & \nReadAllMarginMeanError & \nReadAllMarginMedianError \\
all & \nReadAllUnits & no change & \nReadAllNullMeanError & \nReadAllNullMedianError \\
all & \nReadAllUnits & random entering features & \nReadAllControlMeanError & \nReadAllControlMedianError \\
all & \nReadAllUnits & exact coefficients, earlier residual & \nReadAllExactCleanResidualMeanError & \nReadAllExactCleanResidualMedianError \\
\bottomrule
\end{tabular}

\par\noindent The test runs on the public \nModelNameA pair at layers \nLayerUpA and \nLayerDownA. The floor was fixed on the development prompts, and all interventions keep every selected feature and scaling factor at the final position of \nDsTestPrompts test prompts. The agreement logit difference is the correct verb form's logit less the incorrect form's logit. The standard graph keeps the features selected before the intervention, and the margin graph selects again from the predicted scores. The reconstruction residual is the model activation less its decoded feature coefficients. Random entering features adds as many randomly chosen unselected features as the margin graph lets enter, at its cutoff.
\end{table}

\textbf{What the patch measures.} The test patches each graph's coefficients with the reconstruction residual from before the intervention, so it measures each graph's coefficients as that patch delivers them. The change of the reconstruction residual, which Section~\ref{sec:readout} shows cancels the excess of the entering and leaving part inside the model, is absent from every patch. The exact downstream coefficients patched the same way err by a median of \nDsExactMedian and a mean of \nDsExactMean on \nDsCheckUnits development interventions whose mean absolute change is \nDsExactTrue. The test covers one SAE pair and one task, with single feature interventions at a prompt's last position.

\section{Measured computation}
\label{app:cost}

At the intervention counts of the evaluation, the margin graph takes longer than exact recomputation on both machines, \nWorkCudaReleaseMarginMedianShort against \nWorkCudaReleaseExactMedianShort ms for \nWorkReleaseQueries interventions on the \nDeviceName. This appendix gives the timing protocol and each predictor's time, memory and coefficient error, so that a reader can judge when a graph pays for its construction. A complete workload times the construction and all of a position's interventions together, on the public \nModelNameA pair at layers \nLayerUpA and \nLayerDownA and on the first of our runs. Each uses \nTokens held out Pile validation positions, with \nWorkReleaseQueries interventions for the public pair or \nWorkTestbedQueries for our run, which are the evaluation's three scaling factors applied to every selected feature. Tables~\ref{tab:workload} and~\ref{tab:workload-counts} report medians with full ranges, and medians at further counts. Each workload has \nCostWarmUps uncounted warm up and \nCostRepetitions measured repetitions. The \nCostProcessor uses \nCostThreads numerical threads, and the \nDeviceName is synchronized before and after each clock.

Interventions follow the order of the selected features and then of the scaling factors, cycling through the same interventions when the requested count exceeds them. A shared model pass before the intervention, with the extraction of its context, precedes every workload. Construction and all interventions run together within one call per position. Host memory is sampled every millisecond and at the endpoints, so short peaks can be missed. Accelerator memory is the peak allocated tensor memory, excluding reserved storage, and both memory measurements include the resident model and SAEs.

The two coefficient errors marked $^\dagger$, those of integrated gradients that select again on the public pair, were computed after the timing, on a processor, by the predictors' own code. There, integrated gradients that select again err by \nPathErrPublicIgFive at five points and \nPathErrPublicIgTen at ten, against the margin graph's \nWorkReleaseMarginError. Recomputed the same way, the standard and the margin graph on the public pair and integrated gradients at five and at ten points on our run give the medians that the table prints. Our first check required every recomputed error to lie within \nPathKnownBound of its saved value, and it failed in all \nPathFailedComparisonsWords comparisons. We then allowed \nPathAllowedShare{}\% of the interventions beyond that bound, and in every comparison at most \nPathKnownShareMax of them lie beyond it.

\begingroup
\scriptsize
\setlength{\tabcolsep}{3pt}
\begin{longtable}{lrrrrr}
\caption{\textbf{The margin graph takes longer than exact recomputation at both counts of interventions.} Cells give each predictor's median time and full range in milliseconds, its peak host and accelerator memory in mebibytes (MiB) and its median coefficient error. Errors marked $^\dagger$ were computed after the timing, on a processor.}
\label{tab:workload}\\
\toprule
predictor & median time ms & full time range ms & host MiB & accelerator MiB & coefficient error \\*
\midrule
\endfirsthead
\toprule
predictor & median time ms & full time range ms & host MiB & accelerator MiB & coefficient error \\*
\midrule
\endhead
\bottomrule
\endfoot
\multicolumn{6}{l}{\textbf{\nDeviceName, public pair, \nWorkReleaseQueries interventions}}\\*
\midrule
standard graph & \nWorkCudaReleaseFixedMedianMs & \nWorkCudaReleaseFixedMinimumMs to \nWorkCudaReleaseFixedMaximumMs & \nWorkCudaReleaseFixedPeakRssMib & \nWorkCudaReleaseFixedPeakAllocatedMib & \nWorkReleaseFixedError \\
margin graph & \nWorkCudaReleaseMarginMedianMs & \nWorkCudaReleaseMarginMinimumMs to \nWorkCudaReleaseMarginMaximumMs & \nWorkCudaReleaseMarginPeakRssMib & \nWorkCudaReleaseMarginPeakAllocatedMib & \nWorkReleaseMarginError \\
softmax, original selection & \nWorkCudaReleaseSoftmaxFixedMedianMs & \nWorkCudaReleaseSoftmaxFixedMinimumMs to \nWorkCudaReleaseSoftmaxFixedMaximumMs & \nWorkCudaReleaseSoftmaxFixedPeakRssMib & \nWorkCudaReleaseSoftmaxFixedPeakAllocatedMib & \nWorkReleaseSoftmaxFixedError \\
softmax, selects again & \nWorkCudaReleaseSoftmaxRerankMedianMs & \nWorkCudaReleaseSoftmaxRerankMinimumMs to \nWorkCudaReleaseSoftmaxRerankMaximumMs & \nWorkCudaReleaseSoftmaxRerankPeakRssMib & \nWorkCudaReleaseSoftmaxRerankPeakAllocatedMib & \nWorkReleaseSoftmaxRerankError \\
IG selecting again, five points & \nWorkCudaReleaseIgFiveRerankMedianMs & \nWorkCudaReleaseIgFiveRerankMinimumMs to \nWorkCudaReleaseIgFiveRerankMaximumMs & \nWorkCudaReleaseIgFiveRerankPeakRssMib & \nWorkCudaReleaseIgFiveRerankPeakAllocatedMib & \nPathErrPublicIgFive$^\dagger$ \\
IG selecting again, ten points & \nWorkCudaReleaseIgOneZeroRerankMedianMs & \nWorkCudaReleaseIgOneZeroRerankMinimumMs to \nWorkCudaReleaseIgOneZeroRerankMaximumMs & \nWorkCudaReleaseIgOneZeroRerankPeakRssMib & \nWorkCudaReleaseIgOneZeroRerankPeakAllocatedMib & \nPathErrPublicIgTen$^\dagger$ \\
exact recomputation & \nWorkCudaReleaseExactMedianMs & \nWorkCudaReleaseExactMinimumMs to \nWorkCudaReleaseExactMaximumMs & \nWorkCudaReleaseExactPeakRssMib & \nWorkCudaReleaseExactPeakAllocatedMib & reference \\
\midrule
\multicolumn{6}{l}{\textbf{\nDeviceName, first of our runs, \nWorkTestbedQueries interventions}}\\*
\midrule
standard graph & \nWorkCudaTestbedFixedMedianMs & \nWorkCudaTestbedFixedMinimumMs to \nWorkCudaTestbedFixedMaximumMs & \nWorkCudaTestbedFixedPeakRssMib & \nWorkCudaTestbedFixedPeakAllocatedMib & \nWorkTestbedFixedError \\
margin graph & \nWorkCudaTestbedMarginMedianMs & \nWorkCudaTestbedMarginMinimumMs to \nWorkCudaTestbedMarginMaximumMs & \nWorkCudaTestbedMarginPeakRssMib & \nWorkCudaTestbedMarginPeakAllocatedMib & \nWorkTestbedMarginError \\
softmax, original selection & \nWorkCudaTestbedSoftmaxFixedMedianMs & \nWorkCudaTestbedSoftmaxFixedMinimumMs to \nWorkCudaTestbedSoftmaxFixedMaximumMs & \nWorkCudaTestbedSoftmaxFixedPeakRssMib & \nWorkCudaTestbedSoftmaxFixedPeakAllocatedMib & \nWorkTestbedSoftmaxFixedError \\
softmax, selects again & \nWorkCudaTestbedSoftmaxRerankMedianMs & \nWorkCudaTestbedSoftmaxRerankMinimumMs to \nWorkCudaTestbedSoftmaxRerankMaximumMs & \nWorkCudaTestbedSoftmaxRerankPeakRssMib & \nWorkCudaTestbedSoftmaxRerankPeakAllocatedMib & \nWorkTestbedSoftmaxRerankError \\
IG selecting again, five points & \nWorkCudaTestbedIgFiveRerankMedianMs & \nWorkCudaTestbedIgFiveRerankMinimumMs to \nWorkCudaTestbedIgFiveRerankMaximumMs & \nWorkCudaTestbedIgFiveRerankPeakRssMib & \nWorkCudaTestbedIgFiveRerankPeakAllocatedMib & \nWorkTestbedIgFiveRerankError \\
IG selecting again, ten points & \nWorkCudaTestbedIgOneZeroRerankMedianMs & \nWorkCudaTestbedIgOneZeroRerankMinimumMs to \nWorkCudaTestbedIgOneZeroRerankMaximumMs & \nWorkCudaTestbedIgOneZeroRerankPeakRssMib & \nWorkCudaTestbedIgOneZeroRerankPeakAllocatedMib & \nWorkTestbedIgOneZeroRerankError \\
exact recomputation & \nWorkCudaTestbedExactMedianMs & \nWorkCudaTestbedExactMinimumMs to \nWorkCudaTestbedExactMaximumMs & \nWorkCudaTestbedExactPeakRssMib & \nWorkCudaTestbedExactPeakAllocatedMib & reference \\
\midrule
\multicolumn{6}{l}{\textbf{\nCostProcessor, public pair, \nWorkReleaseQueries interventions}}\\*
\midrule
standard graph & \nWorkCpuReleaseFixedMedianMs & \nWorkCpuReleaseFixedMinimumMs to \nWorkCpuReleaseFixedMaximumMs & \nWorkCpuReleaseFixedPeakRssMib & not applicable & \nWorkReleaseFixedError \\
margin graph & \nWorkCpuReleaseMarginMedianMs & \nWorkCpuReleaseMarginMinimumMs to \nWorkCpuReleaseMarginMaximumMs & \nWorkCpuReleaseMarginPeakRssMib & not applicable & \nWorkReleaseMarginError \\
softmax, original selection & \nWorkCpuReleaseSoftmaxFixedMedianMs & \nWorkCpuReleaseSoftmaxFixedMinimumMs to \nWorkCpuReleaseSoftmaxFixedMaximumMs & \nWorkCpuReleaseSoftmaxFixedPeakRssMib & not applicable & \nWorkReleaseSoftmaxFixedError \\
softmax, selects again & \nWorkCpuReleaseSoftmaxRerankMedianMs & \nWorkCpuReleaseSoftmaxRerankMinimumMs to \nWorkCpuReleaseSoftmaxRerankMaximumMs & \nWorkCpuReleaseSoftmaxRerankPeakRssMib & not applicable & \nWorkReleaseSoftmaxRerankError \\
IG selecting again, five points & \nWorkCpuReleaseIgFiveRerankMedianMs & \nWorkCpuReleaseIgFiveRerankMinimumMs to \nWorkCpuReleaseIgFiveRerankMaximumMs & \nWorkCpuReleaseIgFiveRerankPeakRssMib & not applicable & \nPathErrPublicIgFive$^\dagger$ \\
IG selecting again, ten points & \nWorkCpuReleaseIgOneZeroRerankMedianMs & \nWorkCpuReleaseIgOneZeroRerankMinimumMs to \nWorkCpuReleaseIgOneZeroRerankMaximumMs & \nWorkCpuReleaseIgOneZeroRerankPeakRssMib & not applicable & \nPathErrPublicIgTen$^\dagger$ \\
exact recomputation & \nWorkCpuReleaseExactMedianMs & \nWorkCpuReleaseExactMinimumMs to \nWorkCpuReleaseExactMaximumMs & \nWorkCpuReleaseExactPeakRssMib & not applicable & reference \\
\midrule
\multicolumn{6}{l}{\textbf{\nCostProcessor, first of our runs, \nWorkTestbedQueries interventions}}\\*
\midrule
standard graph & \nWorkCpuTestbedFixedMedianMs & \nWorkCpuTestbedFixedMinimumMs to \nWorkCpuTestbedFixedMaximumMs & \nWorkCpuTestbedFixedPeakRssMib & not applicable & \nWorkTestbedFixedError \\
margin graph & \nWorkCpuTestbedMarginMedianMs & \nWorkCpuTestbedMarginMinimumMs to \nWorkCpuTestbedMarginMaximumMs & \nWorkCpuTestbedMarginPeakRssMib & not applicable & \nWorkTestbedMarginError \\
softmax, original selection & \nWorkCpuTestbedSoftmaxFixedMedianMs & \nWorkCpuTestbedSoftmaxFixedMinimumMs to \nWorkCpuTestbedSoftmaxFixedMaximumMs & \nWorkCpuTestbedSoftmaxFixedPeakRssMib & not applicable & \nWorkTestbedSoftmaxFixedError \\
softmax, selects again & \nWorkCpuTestbedSoftmaxRerankMedianMs & \nWorkCpuTestbedSoftmaxRerankMinimumMs to \nWorkCpuTestbedSoftmaxRerankMaximumMs & \nWorkCpuTestbedSoftmaxRerankPeakRssMib & not applicable & \nWorkTestbedSoftmaxRerankError \\
IG selecting again, five points & \nWorkCpuTestbedIgFiveRerankMedianMs & \nWorkCpuTestbedIgFiveRerankMinimumMs to \nWorkCpuTestbedIgFiveRerankMaximumMs & \nWorkCpuTestbedIgFiveRerankPeakRssMib & not applicable & \nWorkTestbedIgFiveRerankError \\
IG selecting again, ten points & \nWorkCpuTestbedIgOneZeroRerankMedianMs & \nWorkCpuTestbedIgOneZeroRerankMinimumMs to \nWorkCpuTestbedIgOneZeroRerankMaximumMs & \nWorkCpuTestbedIgOneZeroRerankPeakRssMib & not applicable & \nWorkTestbedIgOneZeroRerankError \\
exact recomputation & \nWorkCpuTestbedExactMedianMs & \nWorkCpuTestbedExactMinimumMs to \nWorkCpuTestbedExactMaximumMs & \nWorkCpuTestbedExactPeakRssMib & not applicable & reference \\
\midrule
\end{longtable}
\par\noindent Each setting uses \nTokens held out Pile validation positions, \nCostRepetitions repetitions after \nCostWarmUps warm up, and \nWorkReleaseQueries interventions per position for the public pair or \nWorkTestbedQueries for the first of our runs. Memory is the sampled peak of host resident and accelerator allocated memory, and coefficient errors are medians over the same interventions. A sparse autoencoder (SAE) represents model activations by sparse feature coefficients. TopK selects the largest positive encoder scores, and ReLU every positive score. Public is the \nModelNameA pair at layers \nLayerUpA and \nLayerDownA, of width \nWidthA and TopK \nKA, and ours is the first of our runs, of width \nWidthF and TopK \nKF, on the same model and layers. The standard graph keeps the features selected before the intervention, and the margin graph selects again from the predicted scores. Integrated gradients (IG) average derivatives along the intervention's path. The softmax correction recomputes the attention weights before the rest of the first order prediction. The relative coefficient error divides the norm of a prediction's error by the norm of the exact coefficient change, and smaller is better. Exact recomputation is the reference that each error is measured against.

\endgroup

\begingroup
\small
\setlength{\tabcolsep}{5pt}
\begin{longtable}{lrrrrr}
\caption{\textbf{The margin graph takes no longer than exact recomputation only on the processor.} Cells give median total times in milliseconds, construction included, and each column head counts interventions at one position.}
\label{tab:workload-counts}\\
\toprule
\multicolumn{6}{l}{\textbf{\nDeviceName, public pair}}\\*
predictor & \nWorkCountZero & \nWorkCountOne & \nWorkCountThree & \nWorkCountFour & \nWorkCountFive \\*
\midrule
standard graph & \nWorkCudaReleaseFixedCountOneSixMedianMs & \nWorkCudaReleaseFixedCountFourEightMedianMs & \nWorkCudaReleaseFixedCountOneNineTwoMedianMs & \nWorkCudaReleaseFixedCountThreeEightFourMedianMs & \nWorkCudaReleaseFixedCountSevenSixEightMedianMs \\
margin graph & \nWorkCudaReleaseMarginCountOneSixMedianMs & \nWorkCudaReleaseMarginCountFourEightMedianMs & \nWorkCudaReleaseMarginCountOneNineTwoMedianMs & \nWorkCudaReleaseMarginCountThreeEightFourMedianMs & \nWorkCudaReleaseMarginCountSevenSixEightMedianMs \\
softmax, original selection & \nWorkCudaReleaseSoftmaxFixedCountOneSixMedianMs & \nWorkCudaReleaseSoftmaxFixedCountFourEightMedianMs & \nWorkCudaReleaseSoftmaxFixedCountOneNineTwoMedianMs & \nWorkCudaReleaseSoftmaxFixedCountThreeEightFourMedianMs & \nWorkCudaReleaseSoftmaxFixedCountSevenSixEightMedianMs \\
softmax, selects again & \nWorkCudaReleaseSoftmaxRerankCountOneSixMedianMs & \nWorkCudaReleaseSoftmaxRerankCountFourEightMedianMs & \nWorkCudaReleaseSoftmaxRerankCountOneNineTwoMedianMs & \nWorkCudaReleaseSoftmaxRerankCountThreeEightFourMedianMs & \nWorkCudaReleaseSoftmaxRerankCountSevenSixEightMedianMs \\
IG selecting again, five points & \nWorkCudaReleaseIgFiveRerankCountOneSixMedianMs & \nWorkCudaReleaseIgFiveRerankCountFourEightMedianMs & \nWorkCudaReleaseIgFiveRerankCountOneNineTwoMedianMs & \nWorkCudaReleaseIgFiveRerankCountThreeEightFourMedianMs & \nWorkCudaReleaseIgFiveRerankCountSevenSixEightMedianMs \\
IG selecting again, ten points & \nWorkCudaReleaseIgOneZeroRerankCountOneSixMedianMs & \nWorkCudaReleaseIgOneZeroRerankCountFourEightMedianMs & \nWorkCudaReleaseIgOneZeroRerankCountOneNineTwoMedianMs & \nWorkCudaReleaseIgOneZeroRerankCountThreeEightFourMedianMs & \nWorkCudaReleaseIgOneZeroRerankCountSevenSixEightMedianMs \\
exact recomputation & \nWorkCudaReleaseExactCountOneSixMedianMs & \nWorkCudaReleaseExactCountFourEightMedianMs & \nWorkCudaReleaseExactCountOneNineTwoMedianMs & \nWorkCudaReleaseExactCountThreeEightFourMedianMs & \nWorkCudaReleaseExactCountSevenSixEightMedianMs \\
\toprule
\multicolumn{6}{l}{\textbf{\nDeviceName, first of our runs}}\\*
predictor & \nWorkCountZero & \nWorkCountTwo & \nWorkCountThree & \nWorkCountFour & \nWorkCountFive \\*
\midrule
standard graph & \nWorkCudaTestbedFixedCountOneSixMedianMs & \nWorkCudaTestbedFixedCountNineSixMedianMs & \nWorkCudaTestbedFixedCountOneNineTwoMedianMs & \nWorkCudaTestbedFixedCountThreeEightFourMedianMs & \nWorkCudaTestbedFixedCountSevenSixEightMedianMs \\
margin graph & \nWorkCudaTestbedMarginCountOneSixMedianMs & \nWorkCudaTestbedMarginCountNineSixMedianMs & \nWorkCudaTestbedMarginCountOneNineTwoMedianMs & \nWorkCudaTestbedMarginCountThreeEightFourMedianMs & \nWorkCudaTestbedMarginCountSevenSixEightMedianMs \\
softmax, original selection & \nWorkCudaTestbedSoftmaxFixedCountOneSixMedianMs & \nWorkCudaTestbedSoftmaxFixedCountNineSixMedianMs & \nWorkCudaTestbedSoftmaxFixedCountOneNineTwoMedianMs & \nWorkCudaTestbedSoftmaxFixedCountThreeEightFourMedianMs & \nWorkCudaTestbedSoftmaxFixedCountSevenSixEightMedianMs \\
softmax, selects again & \nWorkCudaTestbedSoftmaxRerankCountOneSixMedianMs & \nWorkCudaTestbedSoftmaxRerankCountNineSixMedianMs & \nWorkCudaTestbedSoftmaxRerankCountOneNineTwoMedianMs & \nWorkCudaTestbedSoftmaxRerankCountThreeEightFourMedianMs & \nWorkCudaTestbedSoftmaxRerankCountSevenSixEightMedianMs \\
IG selecting again, five points & \nWorkCudaTestbedIgFiveRerankCountOneSixMedianMs & \nWorkCudaTestbedIgFiveRerankCountNineSixMedianMs & \nWorkCudaTestbedIgFiveRerankCountOneNineTwoMedianMs & \nWorkCudaTestbedIgFiveRerankCountThreeEightFourMedianMs & \nWorkCudaTestbedIgFiveRerankCountSevenSixEightMedianMs \\
IG selecting again, ten points & \nWorkCudaTestbedIgOneZeroRerankCountOneSixMedianMs & \nWorkCudaTestbedIgOneZeroRerankCountNineSixMedianMs & \nWorkCudaTestbedIgOneZeroRerankCountOneNineTwoMedianMs & \nWorkCudaTestbedIgOneZeroRerankCountThreeEightFourMedianMs & \nWorkCudaTestbedIgOneZeroRerankCountSevenSixEightMedianMs \\
exact recomputation & \nWorkCudaTestbedExactCountOneSixMedianMs & \nWorkCudaTestbedExactCountNineSixMedianMs & \nWorkCudaTestbedExactCountOneNineTwoMedianMs & \nWorkCudaTestbedExactCountThreeEightFourMedianMs & \nWorkCudaTestbedExactCountSevenSixEightMedianMs \\
\toprule
\multicolumn{6}{l}{\textbf{\nCostProcessor, public pair}}\\*
predictor & \nWorkCountZero & \nWorkCountOne & \nWorkCountThree & \nWorkCountFour & \nWorkCountFive \\*
\midrule
standard graph & \nWorkCpuReleaseFixedCountOneSixMedianMs & \nWorkCpuReleaseFixedCountFourEightMedianMs & \nWorkCpuReleaseFixedCountOneNineTwoMedianMs & \nWorkCpuReleaseFixedCountThreeEightFourMedianMs & \nWorkCpuReleaseFixedCountSevenSixEightMedianMs \\
margin graph & \nWorkCpuReleaseMarginCountOneSixMedianMs & \nWorkCpuReleaseMarginCountFourEightMedianMs & \nWorkCpuReleaseMarginCountOneNineTwoMedianMs & \nWorkCpuReleaseMarginCountThreeEightFourMedianMs & \nWorkCpuReleaseMarginCountSevenSixEightMedianMs \\
softmax, original selection & \nWorkCpuReleaseSoftmaxFixedCountOneSixMedianMs & \nWorkCpuReleaseSoftmaxFixedCountFourEightMedianMs & \nWorkCpuReleaseSoftmaxFixedCountOneNineTwoMedianMs & \nWorkCpuReleaseSoftmaxFixedCountThreeEightFourMedianMs & \nWorkCpuReleaseSoftmaxFixedCountSevenSixEightMedianMs \\
softmax, selects again & \nWorkCpuReleaseSoftmaxRerankCountOneSixMedianMs & \nWorkCpuReleaseSoftmaxRerankCountFourEightMedianMs & \nWorkCpuReleaseSoftmaxRerankCountOneNineTwoMedianMs & \nWorkCpuReleaseSoftmaxRerankCountThreeEightFourMedianMs & \nWorkCpuReleaseSoftmaxRerankCountSevenSixEightMedianMs \\
IG selecting again, five points & \nWorkCpuReleaseIgFiveRerankCountOneSixMedianMs & \nWorkCpuReleaseIgFiveRerankCountFourEightMedianMs & \nWorkCpuReleaseIgFiveRerankCountOneNineTwoMedianMs & \nWorkCpuReleaseIgFiveRerankCountThreeEightFourMedianMs & \nWorkCpuReleaseIgFiveRerankCountSevenSixEightMedianMs \\
IG selecting again, ten points & \nWorkCpuReleaseIgOneZeroRerankCountOneSixMedianMs & \nWorkCpuReleaseIgOneZeroRerankCountFourEightMedianMs & \nWorkCpuReleaseIgOneZeroRerankCountOneNineTwoMedianMs & \nWorkCpuReleaseIgOneZeroRerankCountThreeEightFourMedianMs & \nWorkCpuReleaseIgOneZeroRerankCountSevenSixEightMedianMs \\
exact recomputation & \nWorkCpuReleaseExactCountOneSixMedianMs & \nWorkCpuReleaseExactCountFourEightMedianMs & \nWorkCpuReleaseExactCountOneNineTwoMedianMs & \nWorkCpuReleaseExactCountThreeEightFourMedianMs & \nWorkCpuReleaseExactCountSevenSixEightMedianMs \\
\toprule
\multicolumn{6}{l}{\textbf{\nCostProcessor, first of our runs}}\\*
predictor & \nWorkCountZero & \nWorkCountTwo & \nWorkCountThree & \nWorkCountFour & \nWorkCountFive \\*
\midrule
standard graph & \nWorkCpuTestbedFixedCountOneSixMedianMs & \nWorkCpuTestbedFixedCountNineSixMedianMs & \nWorkCpuTestbedFixedCountOneNineTwoMedianMs & \nWorkCpuTestbedFixedCountThreeEightFourMedianMs & \nWorkCpuTestbedFixedCountSevenSixEightMedianMs \\
margin graph & \nWorkCpuTestbedMarginCountOneSixMedianMs & \nWorkCpuTestbedMarginCountNineSixMedianMs & \nWorkCpuTestbedMarginCountOneNineTwoMedianMs & \nWorkCpuTestbedMarginCountThreeEightFourMedianMs & \nWorkCpuTestbedMarginCountSevenSixEightMedianMs \\
softmax, original selection & \nWorkCpuTestbedSoftmaxFixedCountOneSixMedianMs & \nWorkCpuTestbedSoftmaxFixedCountNineSixMedianMs & \nWorkCpuTestbedSoftmaxFixedCountOneNineTwoMedianMs & \nWorkCpuTestbedSoftmaxFixedCountThreeEightFourMedianMs & \nWorkCpuTestbedSoftmaxFixedCountSevenSixEightMedianMs \\
softmax, selects again & \nWorkCpuTestbedSoftmaxRerankCountOneSixMedianMs & \nWorkCpuTestbedSoftmaxRerankCountNineSixMedianMs & \nWorkCpuTestbedSoftmaxRerankCountOneNineTwoMedianMs & \nWorkCpuTestbedSoftmaxRerankCountThreeEightFourMedianMs & \nWorkCpuTestbedSoftmaxRerankCountSevenSixEightMedianMs \\
IG selecting again, five points & \nWorkCpuTestbedIgFiveRerankCountOneSixMedianMs & \nWorkCpuTestbedIgFiveRerankCountNineSixMedianMs & \nWorkCpuTestbedIgFiveRerankCountOneNineTwoMedianMs & \nWorkCpuTestbedIgFiveRerankCountThreeEightFourMedianMs & \nWorkCpuTestbedIgFiveRerankCountSevenSixEightMedianMs \\
IG selecting again, ten points & \nWorkCpuTestbedIgOneZeroRerankCountOneSixMedianMs & \nWorkCpuTestbedIgOneZeroRerankCountNineSixMedianMs & \nWorkCpuTestbedIgOneZeroRerankCountOneNineTwoMedianMs & \nWorkCpuTestbedIgOneZeroRerankCountThreeEightFourMedianMs & \nWorkCpuTestbedIgOneZeroRerankCountSevenSixEightMedianMs \\
exact recomputation & \nWorkCpuTestbedExactCountOneSixMedianMs & \nWorkCpuTestbedExactCountNineSixMedianMs & \nWorkCpuTestbedExactCountOneNineTwoMedianMs & \nWorkCpuTestbedExactCountThreeEightFourMedianMs & \nWorkCpuTestbedExactCountSevenSixEightMedianMs \\
\bottomrule
\end{longtable}
\par\noindent Times cover \nTokens held out Pile validation positions with \nCostRepetitions repetitions after \nCostWarmUps warm up, for the public pair and the first of our runs, and the interventions repeat in order when the count exceeds the position's interventions. A sparse autoencoder (SAE) represents model activations by sparse feature coefficients. TopK selects the largest positive encoder scores, and ReLU every positive score. Public is the \nModelNameA pair at layers \nLayerUpA and \nLayerDownA, of width \nWidthA and TopK \nKA, and ours is the first of our runs, of width \nWidthF and TopK \nKF, on the same model and layers. The standard graph keeps the features selected before the intervention, and the margin graph selects again from the predicted scores. Integrated gradients (IG) average derivatives along the intervention's path. The softmax correction recomputes the attention weights before the rest of the first order prediction.

\endgroup

Table~\ref{tab:workload-crossovers} gives, for each predictor, the first measured count at which it runs no slower than exact recomputation.

\begin{table}[H]
\caption{\textbf{A predictor takes no longer than exact recomputation only on the processor.} Each entry is the first measured count of interventions at one position at which a predictor's median total time is at or below that of exact recomputation, and none means that no measured count meets it.}
\label{tab:workload-crossovers}
\centering
\small
\begin{tabular}{lrrrr}
\toprule
 & \multicolumn{2}{c}{\nDeviceName} & \multicolumn{2}{c}{\nCostProcessor} \\
predictor & public & ours & public & ours \\
\midrule
standard graph & none & none & \nWorkCpuReleaseFixedCrossover & \nWorkCpuTestbedFixedCrossover \\
margin graph & none & none & \nWorkCpuReleaseMarginCrossover & \nWorkCpuTestbedMarginCrossover \\
softmax, original selection & none & none & \nWorkCpuReleaseSoftmaxFixedCrossover & \nWorkCpuTestbedSoftmaxFixedCrossover \\
softmax, selects again & none & none & \nWorkCpuReleaseSoftmaxRerankCrossover & \nWorkCpuTestbedSoftmaxRerankCrossover \\
IG selecting again, five points & none & none & none & none \\
IG selecting again, ten points & none & none & none & none \\
\bottomrule
\end{tabular}

\par\noindent Times cover the \nTokens positions and \nCostRepetitions repetitions of Table~\ref{tab:workload}, and equality meets the condition and shows no saving. A sparse autoencoder (SAE) represents model activations by sparse feature coefficients. TopK selects the largest positive encoder scores, and ReLU every positive score. Public is the \nModelNameA pair at layers \nLayerUpA and \nLayerDownA, of width \nWidthA and TopK \nKA, and ours is the first of our runs, of width \nWidthF and TopK \nKF, on the same model and layers. The standard graph keeps the features selected before the intervention, and the margin graph selects again from the predicted scores. Integrated gradients (IG) average derivatives along the intervention's path. The softmax correction recomputes the attention weights before the rest of the first order prediction.
\end{table}

We compare the two phases on the processor in Table~\ref{tab:cost}, timing construction and the amortized interventions separately, with accelerator memory unmeasured. Construction keeps the context, coefficients and projected derivative columns from before the intervention, with the selected downstream rows for the standard graph and all rows for the predictors that select again. Intervention time divides a batch over all selected features and scaling factors by its count of interventions. It includes the intervention's selection, encoder projection and partial attention work where these are performed. The softmax correction stores the derivatives from before the intervention, while integrated gradients evaluate derivatives along each intervention's path. Exact recomputation keeps the context and runs the intervened block for each intervention. The shared model pass precedes construction for every method.

\begin{table}[H]
\caption{\textbf{The margin graph takes more construction and intervention time than the standard graph.} Cells give processor times as medians with full ranges in brackets, construction per position in milliseconds and intervention time per intervention in microseconds. Accelerator memory, the peak bytes column, was not measured.}
\label{tab:cost}
\centering
\scriptsize
\setlength{\tabcolsep}{2pt}
\begin{tabular}{llrrl}
\toprule
setting & predictor & construction ms & intervention $\mu$s & peak bytes \\
\midrule
public & standard graph & \nCostReleaseFixedConstructionMedianMs [\nCostReleaseFixedConstructionMinimumMs, \nCostReleaseFixedConstructionMaximumMs] & \nCostReleaseFixedQueryMedianUs [\nCostReleaseFixedQueryMinimumUs, \nCostReleaseFixedQueryMaximumUs] & not measured \\
public & margin graph & \nCostReleaseMarginConstructionMedianMs [\nCostReleaseMarginConstructionMinimumMs, \nCostReleaseMarginConstructionMaximumMs] & \nCostReleaseMarginQueryMedianUs [\nCostReleaseMarginQueryMinimumUs, \nCostReleaseMarginQueryMaximumUs] & not measured \\
public & softmax, original selection & \nCostReleaseSoftmaxFixedConstructionMedianMs [\nCostReleaseSoftmaxFixedConstructionMinimumMs, \nCostReleaseSoftmaxFixedConstructionMaximumMs] & \nCostReleaseSoftmaxFixedQueryMedianUs [\nCostReleaseSoftmaxFixedQueryMinimumUs, \nCostReleaseSoftmaxFixedQueryMaximumUs] & not measured \\
public & softmax, selects again & \nCostReleaseSoftmaxRerankConstructionMedianMs [\nCostReleaseSoftmaxRerankConstructionMinimumMs, \nCostReleaseSoftmaxRerankConstructionMaximumMs] & \nCostReleaseSoftmaxRerankQueryMedianUs [\nCostReleaseSoftmaxRerankQueryMinimumUs, \nCostReleaseSoftmaxRerankQueryMaximumUs] & not measured \\
public & IG selecting again, five points & \nCostReleaseIgFiveRerankConstructionMedianMs [\nCostReleaseIgFiveRerankConstructionMinimumMs, \nCostReleaseIgFiveRerankConstructionMaximumMs] & \nCostReleaseIgFiveRerankQueryMedianUs [\nCostReleaseIgFiveRerankQueryMinimumUs, \nCostReleaseIgFiveRerankQueryMaximumUs] & not measured \\
public & IG selecting again, ten points & \nCostReleaseIgOneZeroRerankConstructionMedianMs [\nCostReleaseIgOneZeroRerankConstructionMinimumMs, \nCostReleaseIgOneZeroRerankConstructionMaximumMs] & \nCostReleaseIgOneZeroRerankQueryMedianUs [\nCostReleaseIgOneZeroRerankQueryMinimumUs, \nCostReleaseIgOneZeroRerankQueryMaximumUs] & not measured \\
public & exact recomputation & \nCostReleaseExactConstructionMedianMs [\nCostReleaseExactConstructionMinimumMs, \nCostReleaseExactConstructionMaximumMs] & \nCostReleaseExactQueryMedianUs [\nCostReleaseExactQueryMinimumUs, \nCostReleaseExactQueryMaximumUs] & not measured \\
ours & standard graph & \nCostTestbedFixedConstructionMedianMs [\nCostTestbedFixedConstructionMinimumMs, \nCostTestbedFixedConstructionMaximumMs] & \nCostTestbedFixedQueryMedianUs [\nCostTestbedFixedQueryMinimumUs, \nCostTestbedFixedQueryMaximumUs] & not measured \\
ours & margin graph & \nCostTestbedMarginConstructionMedianMs [\nCostTestbedMarginConstructionMinimumMs, \nCostTestbedMarginConstructionMaximumMs] & \nCostTestbedMarginQueryMedianUs [\nCostTestbedMarginQueryMinimumUs, \nCostTestbedMarginQueryMaximumUs] & not measured \\
ours & softmax, original selection & \nCostTestbedSoftmaxFixedConstructionMedianMs [\nCostTestbedSoftmaxFixedConstructionMinimumMs, \nCostTestbedSoftmaxFixedConstructionMaximumMs] & \nCostTestbedSoftmaxFixedQueryMedianUs [\nCostTestbedSoftmaxFixedQueryMinimumUs, \nCostTestbedSoftmaxFixedQueryMaximumUs] & not measured \\
ours & softmax, selects again & \nCostTestbedSoftmaxRerankConstructionMedianMs [\nCostTestbedSoftmaxRerankConstructionMinimumMs, \nCostTestbedSoftmaxRerankConstructionMaximumMs] & \nCostTestbedSoftmaxRerankQueryMedianUs [\nCostTestbedSoftmaxRerankQueryMinimumUs, \nCostTestbedSoftmaxRerankQueryMaximumUs] & not measured \\
ours & IG selecting again, five points & \nCostTestbedIgFiveRerankConstructionMedianMs [\nCostTestbedIgFiveRerankConstructionMinimumMs, \nCostTestbedIgFiveRerankConstructionMaximumMs] & \nCostTestbedIgFiveRerankQueryMedianUs [\nCostTestbedIgFiveRerankQueryMinimumUs, \nCostTestbedIgFiveRerankQueryMaximumUs] & not measured \\
ours & IG selecting again, ten points & \nCostTestbedIgOneZeroRerankConstructionMedianMs [\nCostTestbedIgOneZeroRerankConstructionMinimumMs, \nCostTestbedIgOneZeroRerankConstructionMaximumMs] & \nCostTestbedIgOneZeroRerankQueryMedianUs [\nCostTestbedIgOneZeroRerankQueryMinimumUs, \nCostTestbedIgOneZeroRerankQueryMaximumUs] & not measured \\
ours & exact recomputation & \nCostTestbedExactConstructionMedianMs [\nCostTestbedExactConstructionMinimumMs, \nCostTestbedExactConstructionMaximumMs] & \nCostTestbedExactQueryMedianUs [\nCostTestbedExactQueryMinimumUs, \nCostTestbedExactQueryMaximumUs] & not measured \\
\bottomrule
\end{tabular}

\par\noindent Times on the \nCostProcessor cover \nTokens held out Pile validation positions and \nCostRepetitions repetitions after \nCostWarmUps warm up, and each intervention batch runs over every selected feature and the factors zero, \nDoseHalf and \nDoseNine. The preceding paragraph defines each phase. A sparse autoencoder (SAE) represents model activations by sparse feature coefficients. TopK selects the largest positive encoder scores, and ReLU every positive score. Public is the \nModelNameA pair at layers \nLayerUpA and \nLayerDownA, of width \nWidthA and TopK \nKA, and ours is the first of our runs, of width \nWidthF and TopK \nKF, on the same model and layers. The standard graph keeps the features selected before the intervention, and the margin graph selects again from the predicted scores. Integrated gradients (IG) average derivatives along the intervention's path. The softmax correction recomputes the attention weights before the rest of the first order prediction. Construction stores the context and the derivative columns, and the intervention time applies one prediction using them.
\end{table}

A control differs from the margin graph only in judging each feature against the cutoff from before the intervention, where the margin graph selects again. Table~\ref{tab:old-cutoff} gives the median of the public pair and the medians of our runs on the held out positions, reproducing the saved predictions.

\begin{table}[H]
\caption{\textbf{Selecting again lowers the error on the public pair and on every run of our sparse autoencoders (SAEs).} Cells give median relative coefficient errors when each feature is judged against the cutoff from before the intervention or selected again from the predicted scores.}
\label{tab:old-cutoff}
\centering
\small
\begin{tabular}{lrr}
\toprule
configuration or run & earlier cutoff & selecting again \\
\midrule
public \nModelNameA pair & \nExtraReleaseErrThreshold & \nExtraReleaseErrMargin \\
\midrule
\nCutoffZeroRun & \nCutoffZeroOldCutoff & \nCutoffZeroMargin \\
\nCutoffOneRun & \nCutoffOneOldCutoff & \nCutoffOneMargin \\
\nCutoffTwoRun & \nCutoffTwoOldCutoff & \nCutoffTwoMargin \\
\nCutoffThreeRun & \nCutoffThreeOldCutoff & \nCutoffThreeMargin \\
\nCutoffFourRun & \nCutoffFourOldCutoff & \nCutoffFourMargin \\
\nCutoffFiveRun & \nCutoffFiveOldCutoff & \nCutoffFiveMargin \\
\nCutoffSixRun & \nCutoffSixOldCutoff & \nCutoffSixMargin \\
\nCutoffSevenRun & \nCutoffSevenOldCutoff & \nCutoffSevenMargin \\
\bottomrule
\end{tabular}

\par\noindent Errors are on \nModelNameA at layers \nLayerUpA and \nLayerDownA, and each row pools every selected feature at \nTokens held out Pile validation positions and the factors zero, \nDoseHalf and \nDoseNine. The public pair has width \nWidthA and TopK \nKA, and our runs width \nWidthF and TopK \nKF, sharing one auxiliary coefficient. A sparse autoencoder (SAE) represents model activations by sparse feature coefficients. TopK selects the largest positive encoder scores, and ReLU every positive score. The relative coefficient error divides the norm of a prediction's error by the norm of the exact coefficient change, and smaller is better. Numbered rows identify our training runs.
\end{table}

\end{document}